UNIVERSITAT POLITÈCNICA DE CATALUNYA

Programa de doctorat:

AUTOMATITZACIÓ AVANÇADA I ROBÒTICA

Tesi doctoral

# FUNCTION-DESCRIBED GRAPHS FOR STRUCTURAL PATTERN RECOGNITION

Francesc Serratosa i Casanelles

Director: Alberto Sanfeliu Cortés

Institut d'Organització i Control de Sistemes Industrials

Juny de 2000

To my wife, Fina.

# Preface

We propose a new representation of an ensemble of attributed graphs for structural pattern recognition called *Function-Described Graphs* (FDGs), where not only the semantic information is preserved, but also part of the structure of the AGs. We also describe some optimal and efficient algorithms for computing the distance and a sub-optimal distance, respectively, between an unknown object and a class. The clustering of attributed graphs and the synthesis of FDGs is also presented.

FDGs are applied in a 3D-object recognition problem due to the importance of considering or not the structural relations. FDGs are generated by a set of attributed graphs representing some views of the model. Results show that FDGs preserves the semantic and structural information of the objects.

The main ideas, methods and algorithms have been previously published in various congresses, workshops, technical reports and publications.
FDGs were first presented in (Serratosa and Sanfeliu, 1997(a); Serratosa and Sanfeliu, 1997(b)) and compared with random graphs in (Serratosa, Sanfeliu and Alquézar (a), 1999; Serratosa, Sanfeliu and Alquézar (b), 1999). Some ideas about the synthesis process were reported in (Alquézar *et al.*, 1998) and the unsupervised clustering of AGs using FDGs was proposed in (Riaño and Serratosa, 1999; Sanfeliu *et al.*, 2000). The distances between AGs and FDGs were presented in (Serratosa *et al.*, 1998; Alquézar *et al.*, 2000) and the graph matching algorithms that compute these distances in (Serratosa, Alquézar and Sanfeliu (c), 1999; Serratosa, Alquézar and Sanfeliu, 2000; Serratosa, Alquézar and Sanfeliu (a), 1999; Serratosa, Alquézar and Sanfeliu (b), 1999). Finally, the methods and algorithms in these papers were tested on random graphs, face recognition (Vergés *et al.*, 1999) or 3D object recognition.

Chapter 1 summarises the state of the art in structural pattern recognition. Nevertheless, the approaches that deserve special attention are discussed in separate sections or in the introduction to the chapters. Chapter 2 and 3 deal with attributed graphs and random graphs (Wong *et al.*), respectively. FDGs are presented in detail in chapters 4 to 8. An experimental validation and a real application of FDGs are presented in chapters 9 and 10. Finally, conclusions and further work are outlined in chapter 11. There is an appendix which shows some tables and figures obtained from the test in chapters 9 and 10.

Most of this work has been partially funded by the Spanish CICYT under the projects TAP96-0629-C04-02 and TAP98-0473.

# Abstract

A fundamental problem in pattern recognition is selecting suitable representations for objects and classes. In the decision-theoretic approach to pattern recognition, a pattern is represented by a set of numerical values, which forms a feature vector. Although, in many tasks, objects can be recognised successfully using only global features such as size and compactness, in some applications it is helpful to describe an object in terms of its basic parts and the relations between them.

Nevertheless, there are two major problems that practical applications using structural pattern recognition are confronted with. The first problem is the computational complexity of comparing two structures. The time required by any of the optimal algorithms may in the worst case become exponential in the size of the graphs. The approximate algorithms, on the other hand, have only polynomial time complexity, but do not guarantee to find the optimal solution. For some of the applications, this may not be acceptable. The second problem is the fact that there is more than one model graph that must be matched with an input graph, then the conventional graph matching algorithms must be applied to each model-input pair sequentially. As a consequence, the performance is linearly dependent on the size of the database of model graphs. For applications dealing with large database, this may be prohibitive.

Function-described graphs (FDGs) are a compact representation of a set of attributed graphs. They have borrowed from "random graphs" proposed by Wong *et al*. the ability to probabilistically model structural attribute information, while improving the capacity to record structural relationships that consistently appear throughout the data. They do this by incorporating qualitative knowledge of the second-order probabilities of the elements that are expressed as relations (Boolean functions) between pairs of vertices and pairs of arcs in the FDGs. Four approaches and algorithms for building FDGs from an ensemble of attributed graphs are presented. The first synthesises an FDG in a supervised manner. The other three use the supervised clustering algorithms: dynamic, complete and single clustering.

The problem of matching attributed graphs (AGs) to FDGs for recognition or classification purposes is studied from a Bayesian perspective. A distance measure between AGs and FDGs is derived from the principle of maximum likelihood, but robustness is enforced by considering only locally the effects of extraneous and missing elements. A second measure is also given in which the second-order constraints are incorporated as additional costs. A branch-and-bound algorithm is proposed to compute these distance measures together with their corresponding optimal labelling. Because of the exponential cost of this algorithm, three efficient algorithms are also proposed and compared to compute sub-optimal distances between AGs and FDGs. Two of them are based on a probabilistic relaxation approach, and the other does not have an iterative technique.

Some experimental tests are presented in random graphs and a 3D-object recognition problem. In the 3D-object recognition application, an FDG model is synthesised (in a supervised and an unsupervised method) for each object from a set of views (AGs). The second-order information in FDGs is shown so that the recognition ration is better than when the first-order probability distributions are only used. Results of efficient algorithms show that there is an important decrease in the run time although there is only a slight decrease in effectiveness.

# Acknowledgements

Many people have helped me either directly or indirectly in this study. I am grateful to all of them.

I would like to thank Professor Alberto Sanfeliu and Professor René Alquézar. I would never have finished this thesis without their support and guidance. The novel ideas presented here and many more that could not be pursued due to lack of time had their origins in the talks and meetings we had.

I should also acknowledge my colleagues Maria Ferré and David Riaño for their scientific contribution. The former helped me to generate the 3D objects. The latter introduced me to clustering algorithms. I also thank Albert Larré for his unconditional technical support and for helping me to prepare the laboratory sessions of my students when time flew faster than I would have liked. Furthermore, I Acknowledge Jaume Vergés-Lahí, he helped me in the segmentation module in the application tests.

Last but not least, I would like to acknowledge my parents, brothers and sisters, for encouraging me to study and for giving me everything I need to be able to work in the profession that I have always wanted.

Thank you, Fina, for giving me self-confidence, hope and unconditional support.

# Contents

# List of Figures

# List of Tables

# 1. Introduction

This chapter presents the state of the art of the structural pattern recognition as well as the main methods and algorithms proposed from the beginnings in the early seventies until our days. We introduce our new method with the aim of overcoming as much as possible the problems that involves working with graphs. Finally, we present the organisation of this document.

## *1.1. Structural pattern recognition*

A fundamental problem in pattern recognition is selecting suitable representations for objects and classes. In the decision-theoretic approach to pattern recognition, a pattern is represented by a set of numerical values, which forms a feature vector (Meisel, 1972). Although, in many tasks, objects can be recognised successfully using only global features such as size and compactness, in some applications it is helpful to describe an object in terms of its basic parts and the relations between them (Bunke and Sanfeliu, 1990). Hence, in the syntactic and structural approach to pattern recognition, an object is decomposed into a set of pattern primitives and a structure relating these primitives. In the syntactic methods, the structure of an object is described as a syntactic pattern (a sentence in some formal language) whereas the classes of objects are represented by grammars. However, a more powerful way of representing pattern structure, which has been given considerable attention in the literature, is the use of graphs.

A graph consists of a set of vertices representing pattern primitives and a set of edges representing relations between the primitives. In order to incorporate more semantic information about the properties of both the parts and the relations, *Attributed Graphs* (AGs) were proposed by Tsai and Fu (Tsai and Fu, 1979). AGs have been widely used in the literature of pattern recognition ever since. There are numerous applications in which the comparison between graphs plays a relevant role. In fact, many of the algorithms described below have been developed with a particular application in mind. One of the earliest applications was in the field of chemical documentation and the analysis of chemical structures (Bouvray and Balaban, 1979). More recently, graph matching has also been proposed for the retrieval of cases in case-based reasoning (Poole, 1993) and for the analysis of semantic networks in combination with graph grammars (Ehrig, 1992). In machine learning, graph matching is used for the learning common sub-structures of different concepts (Bhanu and Ming, 1988; Fisher, 1987; Cook and Holder, 1994). However, most applications of graph matching have been documented in the fields of pattern recognition and computer vision. For example, the sub-graph detection was successfully applied to Chinese character recognition (Lu, Ren and Suen, 1991), the interpretation of schematic diagrams (Bunke and Allerman, 1983; Lee, Kim and Groen, 1990; Messmer and Bunke, 1996), seal verification (Lee and Kim, 1989) or it was combined with evident based systems for shape analysis (Pearce, Caelli and Bischof, 1994). In computer vision, it was mainly used for the localisation and identification of three-dimensional objects (Bmuer and Bunke, 1990; Horaud and Skordas, 1988; Wong, 1990; Cheng and Huang, 1981; Wong, Lu and Rioux, 1989; cho and Kim, 1992; Wong, 1992). In (Wallace, 1987) there is a comparison of methods of high-level interpretation of two-dimensional segmented images. The reviews by Bunke and Messmer (Bunke, 1993; Bunke and Messmer, 1997) reported these applications and additional ones are described in (Walischewski, 1997; Shearer, 1998; Lourens, 1998).

A *hypergraph* is a type of graph that was introduced by (Berge, 1970) and it has been considered as a useful tool to analyse the structure of a system and to represent a partition (Lee-Kwang and C.H. Cho, 1996). Recently, the concept of hypergraphs has been extended to the fuzzy hypergraphs to represent fuzzy partitions (Lee-Kwang and K.M Lee, 1995; S.M. Chen, 1997).

Typically, graphs are used to represent known models from a database and unknown input patterns. The recognition task, therefore, turns into a graph-matching problem. That is, the database is searched for models that are similar to the unknown input graph. Standard algorithms for graph matching include graph isomorphism, sub-graph isomorphism, and maximum common graph search (Corneil and Gotlieb, 1970; Levi, 1972; Berztiss 1973; Ullman, 1976; Barrow, 1976; Schmidt and Druffel, 1976; McGregor, 1982). When matching two graphs $G^i$ and $G^j$ by means of graph isomorphism, we are looking for a bijective mapping between the vertices and arcs of $G^i$ and $G^j$ such that the structure of the graphs is preserved by the mapping function. When such a mapping function can be found then $G^i$ and $G^j$ are isomorphic. If one of the graphs involved in the matching process is larger than the other, i.e. $G^j$ contains more vertices than $G^i$, then we are looking for a sub-graph isomorphism from $G^i$ to $G^j$. That is, we are interested in finding a sub-graph $S$ of $G^j$ such that $G^i$ and $S$ are isomorphic. However, in real world applications we cannot always expect a perfect match between the input graph and one of the graphs in the database. Therefore, what is needed is an algorithm for error-tolerant graph matching (*etgm*), which computes a measure of similarity between two given graphs. Notice that the definition of an error model is strongly application dependent.

One of the drawbacks of graph matching is its computational complexity. For the graph isomorphism problem it is up to this days an open question whether it belongs to the complexity class P or NP (Garey and Johnson, 1979; Booth and Colbourn, 1979). All algorithms that have been developed so far for the general graph isomorphism problem require in the worst case exponential time. For the sub-graph isomorphism problem and also the error-correcting sub-graph isomorphism problem it is well known that it is NP-complete (Garey and Johnson, 1979). Consequently, no algorithm could be constructed that guarantees to find error-correcting sub-graph isomorphisms in polynomial time. However, research in the past twenty years has shown that there are methods for graph matching that behave reasonably well on the average in terms of performance and become computationally intractable only in few cases. Moreover, if the constraints of graph matching are loosened, then it is possible to find solutions in polynomial time by using approximate methods. In the following, an overview of the methods for graph, sub-graph and error-correcting sub-graph isomorphism detection that have been proposed by various authors in the past is given.

### 1.1.1. Graph and sub-graph isomorphism

The graph and sub-graph isomorphism problem has been the focus of intensive research since the seventies (Read and Corneil, 1977; Gati, 1979). There are basically two approaches that have been taken to solve the graph isomorphism problem: One is based on the **backtracking search** and the other is based on the idea of building a so-called **association graph**. Due to the fact that it is not yet known whether the graph isomorphism problem is P or NP, there is another approach to solve the graph isomorphism problem in addition to the two above commented. This approach is based on the **group-theoretic** concepts. All the algorithms described so far are optimal algorithms. That is, they are guaranteed to find all graph and sub-graph isomorphisms from a given graph $G^i$ to another graph $G^j$. The main problem of these algorithms, however, is that for large graphs they may require exponential time. To solve this problem the **discrete relaxation** methods are proposed, which compute sub-optimal distances in polynomial time.

The **backtracking search** approach is practically oriented. It aims directly at constructing graph or sub-graphs isomorphisms in a procedural manner. One of the best known methods for graph and sub-graph isomorphism detection is based on depth-first backtracking search, first described in (Corneil and Gotlieb, 1970). Informally speaking, the method works as follows. Given two graphs $G^i$ and $G^j$, the vertices of $G^i$ are mapped one after the other onto the vertices of $G^j$ and after each mapping, it is checked whether the arc structure of $G^i$ is preserved in $G^j$ by the mapping. If all the vertices of $G^i$ are successfully mapped onto vertices of $G^j$ and, $G^i$ and $G^j$ are of equal size then a graph isomorphism is found. If $G^i$ is smaller than $G^j$ then a sub-graph isomorphism from $G^i$ to $G^j$ is found. Although this method performs well for small graphs, the number of required steps explodes combinatorially when the graphs grows. Hence, Ullman proposed in (Ullman, 1976) to combine backtracking with forward cheking procedure which greatly reduces the number of backtracking steps. A comprehensive analysis of the performance of different forward-checking and looking-ahead procedures for backtracking is given in (Haralick and Elliot, 1980).

The **association graph** approach described in (Falkenhainer, Forbus and Gentner, 1990; Myaeng and Lopez-Lopez, 1992; Horaud and Skordas, 1989) is based on the idea of building a so-called association graph in which each consistent vertex to vertex mapping is represented by a vertex in the association graph and each locally consistent pair of vertex to vertex mappings is represented by an edge between the corresponding vertices in the association graph. Graph or sub-graph isomorphisms are found by searching for maximal cliques in the association graph.

The **group-theoretic** concepts aims at classifying the adjacency matrices of graphs into permutation groups. With this, it was possible to prove that there exists a moderately exponential bound for the general graph isomorphism problem (Babi, 1981). Notice that the group-theoretic methods are only applicable for graph isomorphism but not sub-graph isomorphism. Furthermore, by imposing certain restrictions on the graphs, it was possible to derive algorithms that have polynomially bounded complexity. For example, (Hoffman, 1982) describe a polynomially bounded method for the isomorphism detection of graph with bounded valence. For the special case of the trivalent graph isomorphism, it was shown in (Hoffman, 1982; Luks, 1982) that algorithms with a computational complexity of $O(n^4)$ exist. In (Hopcroft and Wong, 1974) a method for the computation of the isomorphism of planar graphs is proposed that requires time that is only linear in the size of the graphs. Although these methods are very interesting from a theoretical point of view, they are usually not applicable in practice due to a large constant overhead.

The **discrete relaxation** methods were first presented in (Rosenfeld et al., 1976; Kitchen and Rosenfeld, 1979; Zucker et al., 1977; Peleg and Rosenfeld, 1978; O'learly and Peleg, 1983; Hummel and Zucker, 1983). These methods do not always find the optimal solution but they have the advantage that they require only polynomial time and that they are easily parallelised. However, because only local consistency is checked, ambiguities must be resolved in the end by applying again a backtracking procedure. Discrete relaxation methods aim to gradually reduce the number of possible mappings for each vertex of $G^i$ onto vertices of $G^j$ by only allowing vertex to vertex mappings that are locally consistent. More recent work is presented in (Hancock and Kittler, 1990; Kim and Kak, 1991; Wilson and Hancock, 1997; Cross and Hancock, 1998).

### 1.1.2. Error-correcting sub-graph isomorphism

One of the requirements of error-correcting sub-graph isomorphism is the definition of the errors that are to be taken into account. Most of the **optimal algorithms** proposed so far are based on the A* algorithm (Nilsson, 1980). An optimal algorithm guarantees to find all graph an sub-graph isomorphisms from a given graph $G^i$ to another graph $G^j$. The main problem of these algorithms, however, is that for large graphs they may require exponential time. **Approximate or continuous optimisation algorithms**, on the other hand, do not always find the optimal solution but require only polynomial time. These methods generate solutions that are as close as possible to graph or sub-graph isomorphism. Hence, these approximate methods are usually a first step towards the optimal algorithms.

**Optimal approaches**

The A* algorithms compute the error-correcting sub-graph isomorphism in the following manner. Given a graph $G^i$ and a possibly distorted graph $G^j$ a search tree is expanded such that each state in the tree corresponds to a partial mapping of the vertices of $G^i$ onto vertices in $G^j$. At the top of the search tree, the first vertex of $G^i$ is mapped onto every vertex of $G^j$. Each such mapping and its corresponding cost is a state in the search tree. The generation of successor states is then guided by the cost of the mappings. That is, the vertex mapping with the least cost is extended by mapping a new vertex of $G^i$ onto every vertex of $G^j$ that has not yet been used. Eventually, all vertices of $G^i$ are mapped onto vertices of $G^j$ and an error-correcting sub-graph isomorphism is found. The performance of such an algorithm strongly depends on the number of states that are expanded in the search tree. By introducing a heuristic future cost estimation function, the size of the search tree can be greatly reduced. In (Wong, You and Chan, 1990; Tsai and Fu, 1979; Eshera and Fu, 1984; Sanfeliu and Fu, 1983; Shapiro and Haralick, 1981) various heuristic functions have been proposed for error-correcting sub-graph isomorphism detection based on the A* algorithm. The most common used is based on **graph edit operations** (Sanfeliu and Fu, 1983).

**Graph edit operations**

Probably the best known error correction model for graph matching is similar to the model used in string edit distance (Wagner and Fischer, 1974). It is based on the idea of introducing graph edit operations (Sanfeliu and Fu, 1983). For each possible error type a corresponding graph edit operation is defined. In order to model the fact that certain error types are more likely than others, cost functions are assigned to the edit operations. The definition of the cost functions is strongly application dependent. The graph edit operations are then used to correct errors in the graphs. Thus, informally speaking, an error-correcting sub-graph isomorphism is defined as a sequence of edit operations with minimal cost that must be applied to one of the graphs such that a sub-graph isomorphism exists. This approach is described in section 2.2 and proposed as a distance between attributed graph in the 3D object application in chapter 10.

**Approximate approaches**

Various approximate methods have been proposed for the error-correcting graph isomorphism.

In (Kittler, Christmas and Petrou, 1992; Christmas, Kittler and Petrou; 1995) a method based on **probabilistic relaxation** is described. The basic idea is that each vertex to vertex mapping is assigned a certain probability that reflects the cost of the mapping and the local consistency of the mapping. Similar to discrete relaxation, the probabilities of each mapping are then corrected until a maximum probability for a set of vertex to vertex mapping results. The method was tested on fairly large graphs and interesting results were obtained. However, as expected, the optimal solution was missed in some cases.

Another continuous optimisation approach is based on **neural networks**. In (Feng, Laumy and Dhome, 1994; Metha and Fulop, 1990) it was proposed to solve the error-correcting problem by representing each vertex to vertex mapping by a neurone in a Hopfield network and optimise the output of the network. Another method using Kohonen network was also presented in (Xu and Oja, 1990).

In (Herault, Horaud, Veillon and Niez, 1990) an approximate algorithm based on relaxation and **simulated annealing** is presented.

In (De Jong and Spears, 1989; Brown, Huntley and Spillane, 1989; Ford and Zhang, 1991), it is proposed to encode sequences of vertex to vertex mappings as chromosomes. A **genetic** algorithm is then used to optimise the pool of chromosomes such that the encoded vertex to vertex mappings represent perfect or close to perfect graph or sub-graph isomorphisms.

A special case of error-correcting graph matching is the weighted graph matching problem in which two completely connected graphs of equal size with weights assigned to the edges must be matched onto each other such that the weight differences in the edges are minimised. A **linear programming** method providing an approximate solution to this problem is presented in (Almohamad and Duffuaa, 1993). The method was originally designed for problems in the domain of operations research.

Finally, in (Umeyama, 1988) an algorithm based on the **eigen-decomposition** of the adjacency matrices of the weighted graphs is proposed. While this method is very fast, it can only be applied to graphs with completely different eigen-values. Furthermore, only small distortions in the input graph can be handled.

### 1.1.3. Pre-processing structures

In addition, some attempts have been made to try to reduce the computational time of matching the unknown input patterns to the whole set of models from the database. The basic assumption is that the models in the database are not completely dissimilar. Two different approaches for reducing computational time are mentioned below.

In the approach by Messmer and Bunke (Messmer and Bunke, 1994; Bunke, 1998), the model graphs are pre-processed and generate a symbolic data structure called *network of models*. This network is a compact representation of the models in the sense that multiple occurrences of the same sub-graph are represented only once. Consequently, such sub-graphs will be matched only once with the input and the computational effort will be reduced.

In the other approach, AGs are extended in several ways to include either probabilistic or fuzzy information. Thus, *random graphs* were defined by Wong *et al.* for modelling classes of patterns described by AGs through a joint probability space of random variables, but due to the computational intractability of general random graphs, f*irst-* *order random graphs* (FORGs) were proposed for real applications (Wong and You, 1985; Wong, Constant and You, 1990). Likewise, Chan proposed the f*uzzy attributed graphs* (FAGs) and the *hard FAGs* to represent objects and templates, respectively (Chan 1996).

Several studies have been carried out to solve the three-dimensional object recognition problem by the use of structures. For instance (Bamieh and Figueiredo, 1986; Chen and Kak, 1989) presented some structures based on attributed graphs to represent the objects. Likewise, *aspect graphs* is one of the approaches to representing a 3D shape for the purposes of object recognition. In this approach, the viewing space of an object is partitioned into regions such that in each region the topology of the line drawing of the object does not change. Vertices represent regions and arcs path from one region to another. Several papers have been presented to compute and represent aspect graphs and others to match one view to the 3D object (Bowyer and Dyer, 1990; Gigus and Malik, 1990; Gigus *et al.*, 1991; Laurentini, 1995).

## *1.2. New approaches to structural pattern recognition*

There are two major problems that practical applications using exact or error-correcting graph matching are confronted with. The first problem is the computational complexity of graph matching. As mentioned before, the time required by any of the optimal algorithms listed above may in the worst case become exponential in the size of the graphs. The approximate algorithms, on the other hand, have only polynomial time complexity, but do not guarantee to find the optimal solution. For some of the applications described in section 1.1, this may not be acceptable. The second problem is the fact that there is more than one model graph that must be matched with an input graph, then the conventional graph matching algorithms must be applied to each model-input pair sequentially. As a consequence, the performance is linearly dependent on the size of the database of model graphs. For applications dealing with large database, this may be prohibitive. The preprocessing structures such as Random graphs or Fuzzy attributed graphs represent the cluster of graphs with only one “representing structure” and so the performance is linearly dependent on the number of clusters and not on the number of model graphs. Nevertheless, these structures do not guarantee to keep the structural knowledge of the ensemble of graphs and this also may not be acceptable for some applications.

In this thesis, we propose a new approach, called Function-Described Graphs (FDGs) to error-tolerant graph matching that is particularly efficient with regard to the problems mentioned at the end of the previous paragraph. There are two common assumptions and a common idea behind our approach. The first assumption is that the graph-matching problem always involves one or several graphs, so-called model graphs, that are a priori known model graphs to the input graph. This approach is especially designed for applications dealing with large databases of model graphs. The other assumption is that model graphs that belong to the same cluster have a certain degree of similarity between each other. That is, they share part of their sub-graphs. The common idea of our approach is to compute a special representation of the ensemble of the model graphs that belong to the same cluster in the learning process without loosening, as much as possible, the structural information of the ensemble of attributed graphs. This representation is then used in the recognition process in order to efficiently detect the distances between the input graph and the clusters and reduce, as far as possible the computational time required to recognise the input graph.

Function-Described Graphs are computed in the learning process and used in the recognition process to represent a cluster of model graphs. An FDG is a prototype structure that contains, on one hand, probabilistic functions which represent the semantic information of the local parts of the patterns and, on the other, binary functions to maintain, as much as possible, the local features of the attributed graphs that belong to the class and also to reject the graphs that do not belong to it.

In general, an FDG can be derived by synthesising a cluster or set of individual attributed graphs. For the generation of FDGs, we present three different methods which build the FDGs from the ensemble of AGs and minimises the distances between them. In the first two methods, we know a priori in which clusters the model graphs belong to. (supervised clustering). In the first, there is a given common labelling between the vertices and arcs of the attributed graphs (supervised synthesis) and in the second, the labelling has to be computed (non-supervised synthesis). In the last method, the cluster of graph models and the labellings between the graph elements of these graphs have to be computed (non-supervised synthesis and non-suprevised clustering).

Likewise, we present two well-founded distance measures, which use the Bayesian theory framework, between unknown patterns (AGs) and classes (FDGs). The first is tested to be useful in a non-noisy environment, the second is more practically oriented. These distances are computed by an error tolerant graph-matching algorithm based on the A* algorithm with a future cost estimation function. Although the estimation function reduces considerably the expansion of the search tree, the computational complexity still remains exponential. To shorten the time required, we also propose an efficient method for reducing the search space or our branch-and-bound algorithm. This method initially discards mappings between vertices and is based on the distance between the sub-units of the graphs.

### *1.3. Organisation*

This thesis is organised in the following manner.

**Chapter 2** (Attributed Graphs) gives some basic definitions of attributed graphs and summarises the fundamental information about the distance measure between AGs proposed by (Sanfeliu and Fu, 1983). This information will be useful in chapter 5 where the properties of the distance measures between AGs and FDGs are discussed. Moreover, one of the parameters of the clustering process is a distance measure between AGs. In our application we have chosen the Sanfeliu distance.

**Chapter 3** (Random Graphs) describes general random graphs and first order random graphs (Wong *et al.*) as an example of a representation of an ensemble of AGs. The distance measure between first order random graphs is presented and a simple example is given, which shows that the statistical independence between vertices or arcs leads to an excessive generalisation of the sample graphs.

**Chapter 4** (Function-Described Graphs) presents and studies FDGs in depths. First, FDGs are compared with Random Graphs as prototypes of a set of AGs and then they are formally defined before the properties of the information represented are studied. Finally, FDGs are built from a set of AGs or a set of FDGs using a supervised synthesis.

These approaches are used in chapter 8 in the clustering algorithms. Some experiments are added in order to examine the behaviour of the new representation.

**Chapter 5** (Distance measures for matching AGs to FDGs) explains two robust and well-founded distance measures between AGs and FDGs. In one, the structural information ($2^{ond}$ order relations) of FDGs are used to constraint the allowed labellings between graph elements of both graphs. Nevertheless, the spurious elements that appear in real applications would lead to a coarse measure. Thus, in the other distance, second order constraints are relaxed and used to apply a cost on the distance value. An example is included.

**Chapter 6** (Algorithms for computing the distance measures) presents an A* algorithm for computing the distance measures and their associated optimal morphisms. The search space is reduced by a *branch and bound* technique, which uses semantic and structural knowledge of both graphs and also the second order constraints. Finally, the complexity of the distance computation is outlined. Some experimental results with random graphs are added in order to examine the behaviour of the matching algorithms.

**Chapter 7** (Efficient algorithms for computing sub-optimal distance measures) presents three different efficient algorithms. One of the drawbacks of optimal distance measures between graphs is that they are computationally expensive. We propose some methods that reduce further the search space of our branch and bound algorithm by initially discarding mappings between vertices. Some experimental results with random graphs are added in order to examine the relation between the efficiency and effectiveness of the efficient matching algorithms.

**Chapter 8** (Clustering of AGs using FDGs) presents two algorithms for the non-supervised clustering. The incremental clustering and the hierarchical clustering. The advantages and disadvantages of both methods are discussed. Some experimental results with random graphs are added in order to examine the behaviour of the clustering and synthesis algorithms (learning process).

**Chapter 9** (Experimental validation of FDGs using artificial 3D objects) reports some experimental tests on artificial data for assessing how capable FDGs are of representing an ensemble of AGs, how efficient the sub-optimal distance algorithm is and how good the proposed approach is at classification.

**Chapter 10** (Application of FDGs to recognise office objects) presents an application on real tree-dimensional objects. Some 2D images are taken from some office objects and segmented using a neural-network method. From each view, an attributed graph is extracted and from all the views of an object, an FDGs is automatically synthesised.

**Chapter 11** (Conclusions) summarises the advantages of the new structure and its related algorithms and it draws the relevant conclusions. Possibilities for future research are also discussed.

**Appendix 1** and **Appendix 2** provide some data from the experimental validation (chapter 9) and the application (chapter 10). They show different views of the objects, the structure and attributes of the AGs and the structure of the FDGs.

## 2. Attributed Graphs (AGs)

This section gives some basic definitions about attributed graphs and recalls the distance measure between AGs proposed by Sanfeliu and Fu (Sanfeliu and Fu, 1983). AGs need to be defined to explain the supervised synthesis of FDGs (sections 4.4 and 4.5) and the distance between AGs and FDGs (section 5). Moreover, a comparison to the Sanfeliu and Fu measure will be useful so that the properties of the dissimilarity measures between AGs and FDGs presented in Section 5 can be discussed. The clustering algorithms (section 8) are parameterised by any distance measure between AGs, but in our application (chapter 9) we use the Sanfeliu distance.

### *2.1. Basic definitions about AGs*

Let $H=(\Sigma_v,\Sigma_e)$ be a directed graph structure of order *n* where $\Sigma_v=\{v_k \mid k=1,...,n\}$ is a set of vertices (or nodes) that represents basic parts of an object and $\Sigma_e=\{e_{ij} \mid i,j\in\{1,...,n\}, i\neq j\}$ is a set of edges (or arcs), which represents relationships between parts, where the arc $e_{ij}$ connects vertices $v_i$ and $v_j$ from $v_i$ to $v_j$. We use the term *graph element* to refer to either a vertex or an edge.

Let $Z_v=\{z_i \mid i=1,...,t\}$ be a nonempty finite set of names for the attributes in a vertex, and for each $z_i$ in $Z_v$ let $D_{vi}$ denote the corresponding domain of attribute values. Similarly, let $Z_e=\{z'_i \mid i=1,...,s\}$ be a nonempty finite set of names for the attributes in an arc, and let $D_{ei}$ denote the domain of attribute values for $z'_i$. In this way, a set of attribute-value pairs $\{(z_i,a_i) \mid z_i\in Z_v, a_i\in D_{vi}\}$ in which each attribute appears at most once can be associated with a vertex, and a set $\{(z'_i,b_i) \mid z'_i\in Z_e, b_i\in D_{ei}\}$ satisfying the same restriction can be associated with an arc. However, a set of attribute-value pairs can be listed always in the same order and the attribute names suppressed, provided that a generic null value $\phi$ is given to any attribute not appearing in the set. Hence, a set of attribute-value pairs can be transformed into a t-tuple of values for a vertex or an s-tuple of values for an arc, where $\Delta_v = \{(a_1,...,a_i,...a_t) \mid a_i \in D_{vi} \cup \{\phi\}, 1 \le i \le t, \exists i : a_i \ne \phi\}$ is the global domain of possible values for non-null attributed vertices and $\Delta_e = \{(b_1,...,b_i,...b_s) \mid b_i \in D_{ei} \cup \{\phi\}, 1 \le i \le s, \exists i : b_i \ne \phi\}$ is the global domain of possible values for non-null attributed arcs.

A *null graph element* is a vertex or an arc in which all the attributes are instantiated to the null value $\phi$. The value of these elements is represented by $\Phi$, where $\Phi = (\phi,...,\phi)$.

***Definition 1*** (Attributed graph): An *attributed graph* $G$ over $(\Delta_v, \Delta_e)$ with an underlying graph structure $H = (\Sigma_v, \Sigma_e)$ is defined as a pair $(V, A)$ in which $V = (\Sigma_v, \gamma_v)$ is an *attributed vertex set* and $A = (\Sigma_e, \gamma_e)$ is called an *attributed arc set.* The mappings $\gamma_v : \Sigma_v \to \Delta_\omega$ and $\gamma_e : \Sigma_e \to \Delta_\varepsilon$, which assign attribute values to graph elements, are called *vertex interpreter* and *arc interpreter*, respectively, where $\Delta_\varepsilon = \Delta_e \cup \{\Phi\}$ and $\Delta_\omega = \Delta_v \cup \{\Phi\}$.

A *complete AG* is an AG with a complete graph structure $H$ (vertices are totally connected by arcs, but may include null elements).

***Definition 2*** (*k-extension* of an AG): An attributed graph $G = (V, A)$ of order $n$ can be *extended* to form a complete AG $G' = (V', A')$ of order $k$, $k \ge n$, by adding vertices and arcs with null attribute values $\Phi$. We call $G'$ the *k-extension* of $G$.

The matching scheme discussed in Sections 5 and 6 formally needs the two graphs to be structurally isomorphic and complete. To this end, an *extending process* is carried out on both graphs, which may be of different orders, to make them complete and with the same order. The AG before the extending process is called *initial graph* and the AG obtained after the process is called *extended graph* (Figure 1). However, both the initial and the extended graph provide exactly the same information about an object.

Note that two different domains have been defined for the vertices $(\Delta_\omega, \Delta_v)$ and for the arcs $(\Delta_\varepsilon, \Delta_e)$. This is because it is sometimes helpful to distinguish whether an element is allowed to take the null value $\Phi$ or not. It is supposed that initial AGs do not contain any null element, and therefore, their attribute assignment mappings of vertices and edges could range over $\Delta_v$ and $\Delta_e$, respectively. However, there are null elements in extended AGs and, therefore, their attribute mappings range necessarily over the domains $\Delta_\omega$ and $\Delta_\varepsilon$.

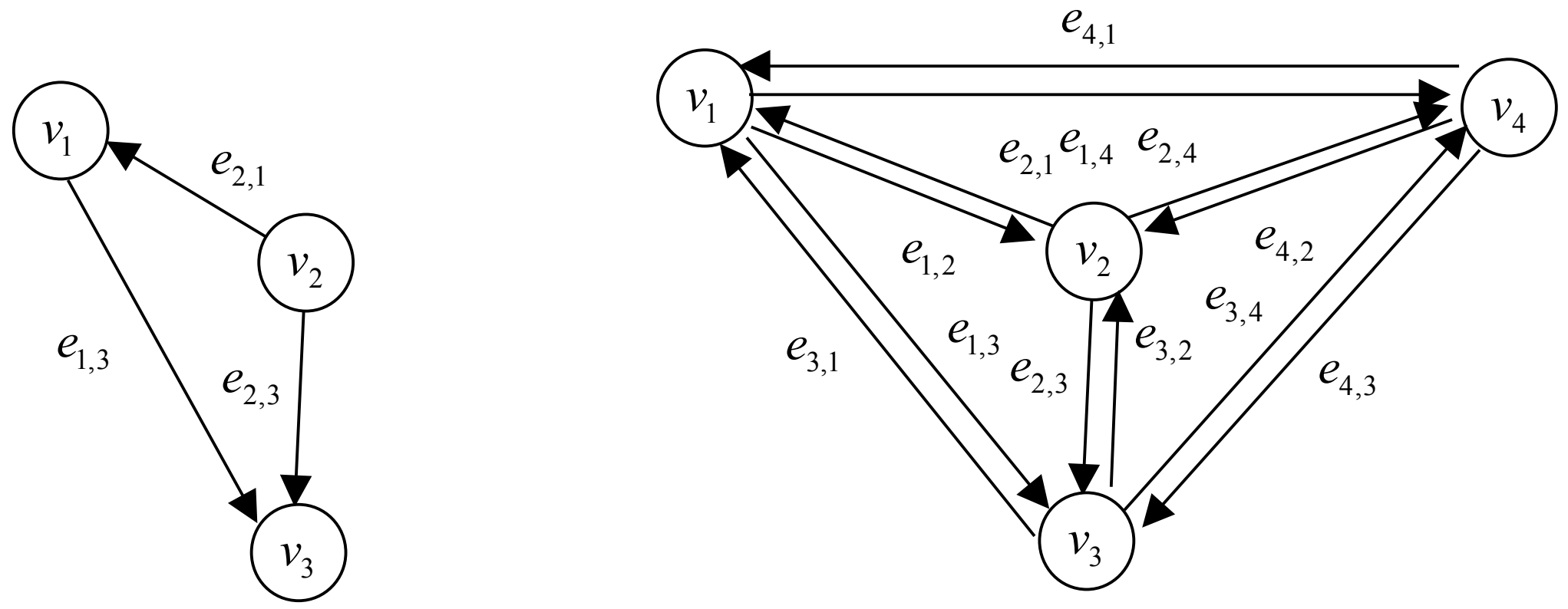

Figure 1. On the left, an AG with 3 vertices and 3 arcs. On the right, its 4-extension.

### 2.2. Distance measure between AGs based on edit operations

Distance measures between structures based on edit operations are in the heart of transforming one structure into another. The three basic structures are strings (Levenshtein, 1966; Tanaka and Kasai, 1976), trees (Fu and Bhargava, 1973; Tai, 1979; Lu, 1984) and graphs (Tanaka, 1977; Sanfeliu and Fu, 1983).

The distance measure between AGs proposed by Sanfeliu and Fu (Sanfeliu and Fu, 1983) requires computing the minimum number of transformations needed to convert an input AG $G_1$ into another $G_2$ using six common edit operations: 1) vertex insertion and 2) arc insertion (a new vertex or arc is inserted into $G_1$ with a given attribute value); 3) vertex deletion and 4) arc deletion (a vertex or arc is deleted from $G_1$); 5) vertex substitution and 6) arc substitution (an attribute value in $G_1$ is substituted by an attribute value in $G_2$).

There is a fixed cost associated with each edit operation and, thus, the global cost of a transformation is the sum of the costs of the edit operations involved. Let $N_{vi}, N_{ei}, N_{vd}, N_{ed}, N_{vs}$ and $N_{es}$ be the number of insertions of vertices and arcs, deletions of vertices and arcs and substitutions of vertices and arcs, respectively. Let $C_{vi}, C_{ei}, C_{vd}, C_{ed}, C_{vs}$ and $C_{es}$ be the individual costs or weights of the corresponding edit operations; these values are generally found heuristically.

Then, the Sanfeliu's distance measure between AGs is given as

$$d(G_1, G_2) = \min_{configurations} \{N_{vi} * C_{vi} + N_{ei} * C_{ei} + N_{vd} * C_{vd} + N_{ed} * C_{ed} + N_{vs} * C_{vs} + N_{es} * C_{es}\} \tag{1}$$

where *configurations* represent the set of allowed chains of edit operations that transform one AG into another. Note that the exact computation of this distance measure is a combinatorial problem for which there is no known solution of polynomial complexity.

## 3. Random graphs (RGs)

A *random graph*, defined by Wong *et al.* (Wong and You, 1985; Wong, Constant and You, 1990), is a graph structure with randomly varying vertex and arc attribute values. Put in another way, it is a graph, together with a set of jointly distributed random variables, some (one for each vertex) ranging over pattern primitives and others (one for each arc) ranging over relations. Any AG obtained by instantiating all random vertices and random arcs is called an *outcome graph* of the random graph. Hence, a random graph represents the set of all possible AGs that can be outcome graphs of it, according to an associated probability distribution.

Below we review the definitions of general random graphs (section 3.1) and first-order random graphs (FORGs) (section 3.2) proposed in (Wong and You, 1985). We have adapted the notation to our convenience. Then we define the distance between two FORGs (section 3.3) and give a simple example (section 3.4) which shows that this probabilistic approach is not enough to represent an ensemble of AGs and that structural information has to be incorporated.

### 3.1. General random graphs

***Definition 3*** (General random graphs): A *random graph R* over $(\Delta_v, \Delta_e)$ with an underlying graph structure $H = (\Sigma_\omega, \Sigma_\varepsilon)$ is defined as a tuple $(W, B, P)$ where $W = (\Sigma_\omega, \gamma_\omega)$, $B = (\Sigma_\varepsilon, \gamma_\varepsilon)$, $\gamma_\omega : \Sigma_\omega \to \Omega_\omega$, $\gamma_\varepsilon : \Sigma_\varepsilon \to \Omega_\varepsilon$. $\Omega_\omega$ is a set of random variables with values in $\Delta_v \cup \{\Phi\}$ (random vertices), $\Omega_\varepsilon$ is a set of random variables with values in $\Delta_e \cup \{\Phi\}$ (random arcs) and, finally, $P$ is a joint probability distribution $P(\alpha_1, \ldots, \alpha_n, \beta_1, \ldots, \beta_m)$ of all the random vertices $\{\alpha_i \mid \alpha_i = \gamma_\omega(\omega_i), 1 \le i \le n\}$ and all the random arcs $\{\beta_j \mid \beta_j = \gamma_\varepsilon(\varepsilon_{kl}), 1 \le j \le m\}$.

For each outcome graph *G* of a random graph *R*, a probability measure $P_R([G])$ of obtaining any AG completely isomorphic to *G,* $[G]$, is given by the sum of the joint probabilities of random vertices and arcs over all instantiations that produce *G*, and any such instantiation is associated with a structural isomorphism $\mu : G' \rightarrow R$, where $G'$ is the extension of *G* to the order of *R*. Let *G* be oriented with respect to *R* by isomorphism $\mu$; for each vertex $\omega_i$ in *R*, let $\mathbf{a}_i = \gamma_v\left(\mu^{-1}(\omega_i)\right)$ be the corresponding attribute value in *G'*, and similarly, for each arc $\varepsilon_{kl}$ in *R* (associated with random variable $\beta_j$) let $\mathbf{b}_j = \gamma_e\left(\mu^{-1}(\varepsilon_{kl})\right)$ be the corresponding attribute value in *G'*. Then the *probability of G according to* (or given by) *the orientation* $\mu$, denoted by $P_R(G|\mu)$, is defined as

$$P_R(G|\mu) = \Pr\left( \wedge_{i=1}^{n} (\alpha_i = \mathbf{a}_i) \wedge \wedge_{j=1}^{m} (\beta_j = \mathbf{b}_j) \right) = P(\mathbf{a}_1, \ldots, \mathbf{a}_n, \mathbf{b}_1, \ldots, \mathbf{b}_m) \tag{2}$$

It is easy to show that the probability $P_R([G])$ can be expressed as the following sum,

$$P_R([G]) = \sum_{[\mu]} P_R(G|\mu) \tag{3}$$

where $[\mu]$ denotes the class of isomorphisms that are equivalent to $\mu$ (two such isomorphisms are considered equivalent whenever one composed with the inverse of the other yields a complete automorphism of *G'*) and $P_R(G|\mu)$ is then the probability of *G* according to any member $\mu$ of the class.

### 3.2. First-order random graphs (FORGs)

General random graphs are absolutely impractical due to the difficulty of estimating and handling the high-order joint probability distribution *P*. Consequently, a strong simplification must be made so that random graphs can be used in practical cases. This is done by introducing suppositions about the probabilistic independence between vertices and/or arcs. Wong and You (Wong and You, 1985) proposed the class of *First Order Random Graphs* (FORGs) for real applications. They assumed the following:

1) The random vertices are mutually independent;

2) The random arcs are independent given values for the random vertices;

3) The arcs are independent of the vertices except for the vertices that they connect.

***Definition 4*** (First order random graph or FORG): A *first order random graph R* over $(\Delta_v, \Delta_e)$ with an underlying graph structure $H = (\Sigma_\omega, \Sigma_\varepsilon)$ is a random graph that satisfies the assumptions 1, 2 and 3 shown above.

Based on these assumptions, for a FORG *R*, the probability $P_R(G|\mu)$ becomes

$$P_R(G|\mu) = \prod_{i=1}^{n} p_i(\mathbf{a}_i) \prod_{j=1}^{m} q_j(\mathbf{b}_j \mid \mathbf{a}_{j1}, \mathbf{a}_{j2}) \tag{4}$$

where $p_i(\mathbf{a}) \hat{=} \Pr(\alpha_i = \mathbf{a})$, $1 \le i \le n$, and $q_j(\mathbf{b} \mid \mathbf{a}_{j1}, \mathbf{a}_{j2}) \hat{=} \Pr(\beta_j = \mathbf{b} \mid \alpha_{j1} = \mathbf{a}_{j1}, \alpha_{j2} = \mathbf{a}_{j2})$, $1 \le j \le m$, are independent probability density functions for vertices and arcs, respectively, and $\alpha_{j1}, \alpha_{j2}$ refer to the random vertices for the endpoints of the random arc $\beta_j$. Note, on the other hand, that a *general random graph* is defined through a joint probability *P* of all the graph elements without any supposition about probabilistic independence.

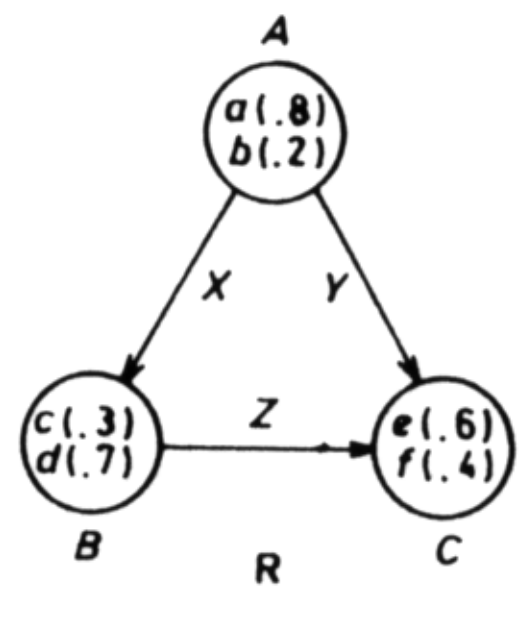

(a)

| X | p | Y | p | Z | p |
|---|---|---|---|---|---|
| u\|a, c | 0.8 | w\|a, e | 0.1 | y\|c, e | 0 |
| v\|a, c | 0.2 | z\|a, e | 0.9 | z\|c, e | 1.0 |
| u\|a, d | 0.7 | w\|a, f | 0.8 | y\|c, f | 0.1 |
| v\|a, d | 0.3 | z\|a, f | 0.2 | z\|c, f | 0.9 |
| u\|b, c | 0.1 | w\|b, e | 0.4 | y\|d, e | 0 |
| v\|b, c | 0.9 | z\|b, e | 0.6 | z\|d, e | 1.0 |
| u\|b, d | 0.2 | w\|b, f | 0 | y\|d, f | 0.4 |
| v\|b, d | 0.8 | z\|b, f | 1.0 | z\|d, f | 0.6 |

(b)

Figure 2. Example of first order random graph taken from (Wong and You, 1985).

Figure 2 shows a simple example of a first order random graph taken from (Wong and You, 1985). On the left there is the structure with three vertices and three arcs. It can be deduced that the attributed graphs used to synthesise it have one attribute in the vertices in the domain $\Delta_v = \{a, b, c, d, e, f\}$ and one attribute in the arcs in the domain $\Delta_e = \{u, v, w, x, y, z\}$. The probabilities related to the vertices are shown inside them, but the probabilities related to the arcs are shown in the table on the right because they depend on the values of the extreme vertices.
First order random graphs were applied to represent and recognise hand written characters in (Wong and You, 1985).

### 3.3. Distance measure between first-order random graphs

A FORG can be derived by synthesising a set of individual attributed graphs or a set of previous FORGs once a common labelling is established. The (weighted) increment of entropy in the synthesis of two FORGs has properties that render it a distance measure between the random graphs involved. The entropy of a FORG reflects the variability in the structural and attribute values of its outcome AGs, and it can be calculated as the sum of the entropy of its random elements (vertices and arcs) from their probability density functions.
Let $R = +(R_1, R_2, \mu)$ denote the synthesis of the FORGs $R_1$ and $R_2$ under a common labelling established by a given morphism $\mu$, and let $r_1$ and $r_2$ be *a priori* probabilities associated with $R_1$ and $R_2$ respectively. So the weighted increment of entropy $H''(+(R_1, R_2, \mu))$ in the synthesis of random graph *R* from $R_1$ and $R_2$ according to $\mu$ is defined as the sum of the weighted increments of entropy in the synthesis of all the random elements of *R*, i.e.

$$H''(+(R_1, R_2, \mu)) = \sum_{\gamma \in R} d(\gamma_1, \gamma_2)(H(\gamma) - (r_1 H(\gamma_1) + r_2 H(\gamma_2))) \tag{5}$$

where $\gamma = +(\gamma_1, \gamma_2, \mu)$ denotes the synthesis of the random elements $\gamma_1$ and $\gamma_2$ in $R_1$ and $R_2$, respectively, according to $\mu$; $d(\gamma_1, \gamma_2)$ is a distance between the random elements $\gamma_1$ and $\gamma_2$ which acts as a weight and which is based on the difference between their probability density functions; and $H(\gamma)$ is the entropy of the random element $\gamma$ defined as,

$$H(\gamma) = -\sum_{a \in \text{"range of } \gamma\text{"}} \Pr(\gamma = a) * \log(\Pr(\gamma = a)) \tag{6}$$

A distance measure between the two FORGs $R_1$ and $R_2$ is defined as the minimum weighted increment of entropy in their synthesis over the set of possible morphisms $\mu$, i.e.

$$d(R_1, R_2) = \min_{\mu} (H''(+(R_1, R_2, \mu))) \tag{7}$$

Since an AG can be treated as a special case of a FORG, the above distance can be applied to two AGs or one AG and one FORG or two FORGs. These distance measures can be used to perform various tasks in structural pattern recognition such as learning and classification.

### 3.4. Example of the distance between first-order random graphs

Figure 3 shows the structure of random graphs $A_1$ and $A_2$, which have been synthesised from a single AG, and the structure of $G$, which has been synthesised from $A_1$ and $A_2$ and an optimal labelling.

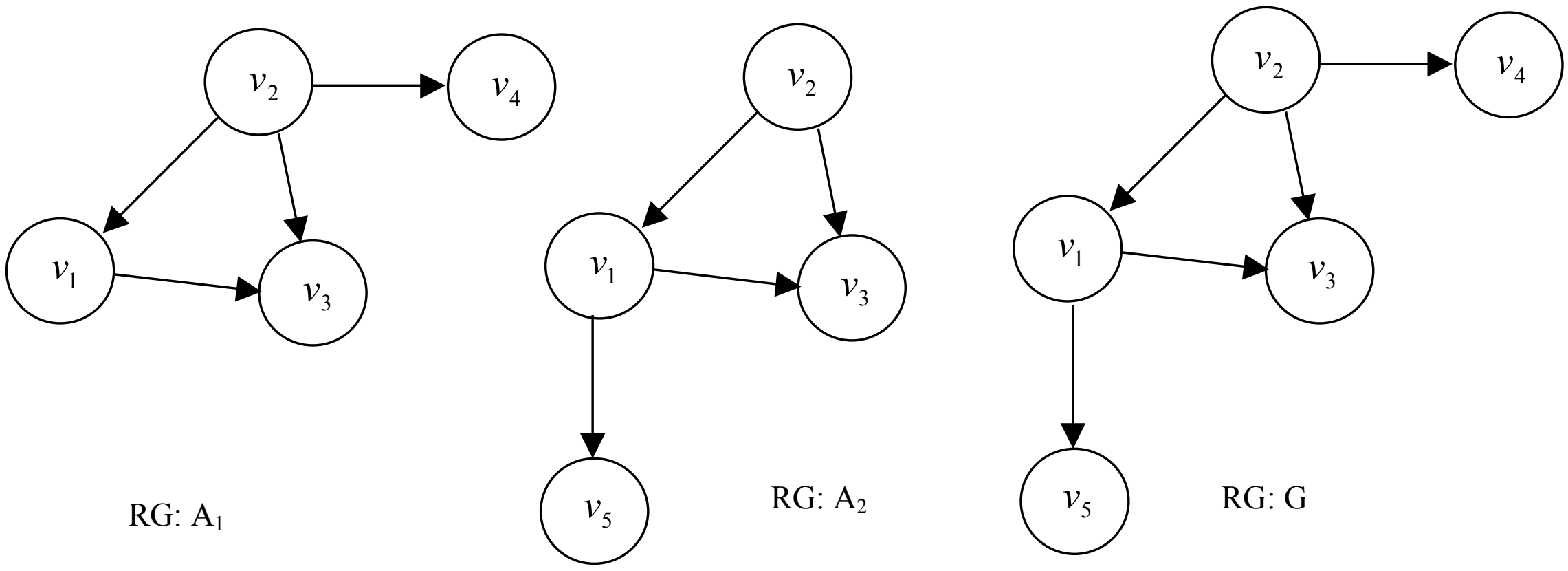

(a) (b) (c)

Figure 3. Structure of random graphs $A_1$, $A_2$ and $G$.

Table 1 shows the attribute values of $A_1$, $A_2$ and $G$. The domain of the attribute in the vertices is $\{a, b, c, d, e\}$ and in the arcs it is $\{X, Y, Z, L, K\}$. The non-existence of graph elements (vertices or arcs) is represented by $\Phi$. The entropy of a random graph synthesised from only one AG is always zero. Thus, $H(A_1)=0$ and $H(A_2)=0$. On the other hand, the entropy of $G$ is $H(G)=-4*0.5*\log_2(0.5)=2$.

| Graph element | $A_1$ | $A_2$ | $G=+(A_1,A_2,\mu^{opt})$ |
|---|---|---|---|
| $v_1$ | $\Pr(b)=1$ | $\Pr(b)=1$ | $\Pr(b)=1$ |
| $v_2$ | $\Pr(a)=1$ | $\Pr(a)=1$ | $\Pr(a)=1$ |
| $v_3$ | $\Pr(c)=1$ | $\Pr(c)=1$ | $\Pr(c)=1$ |
| $v_4$ | $\Pr(d)=1$ | $\Pr(\Phi)=1$ | $\Pr(d)=0.5$ & $\Pr(\Phi)=0.5$ |
| $v_5$ | $\Pr(\Phi)=1$ | $\Pr(e)=1$ | $\Pr(e)=0.5$ & $\Pr(\Phi)=0.5$ |
| $e_{21}$ | $\Pr(X\|a,b)=1$ | $\Pr(X\|a,b)=1$ | $\Pr(X\|a,b)=1$ |
| $e_{23}$ | $\Pr(Y\|a,c)=1$ | $\Pr(Y\|a,c)=1$ | $\Pr(Y\|a,c)=1$ |
| $e_{13}$ | $\Pr(Z\|b,c)=1$ | $\Pr(Z\|b,c)=1$ | $\Pr(Z\|b,c)=1$ |
| $e_{15}$ | $\Pr(\Phi)=1$ | $\Pr(L\|b,e)=1$ | $\Pr(L\|b,e)=1$ & $\Pr(\Phi)=0.5$ |
| $e_{24}$ | $\Pr(K\|a,d)=1$ | $\Pr(\Phi)=1$ | $\Pr(K\|a,d)=1$ & $\Pr(\Phi)=0.5$ |

Table 1. Attribute values of $A_1$, $A_2$ and $G$.

Figure 4 shows the structure of three other random graphs $G_1$, $G_2$ and $G_3$, and Table 2 shows the attribute values of these graphs. Note that $G_1$ is a sub-graph of $A_1$ or $A_2$, $G_2$ is exactly $A_1$, and $G_3$ is the union of both $A_1$ and $A_2$.

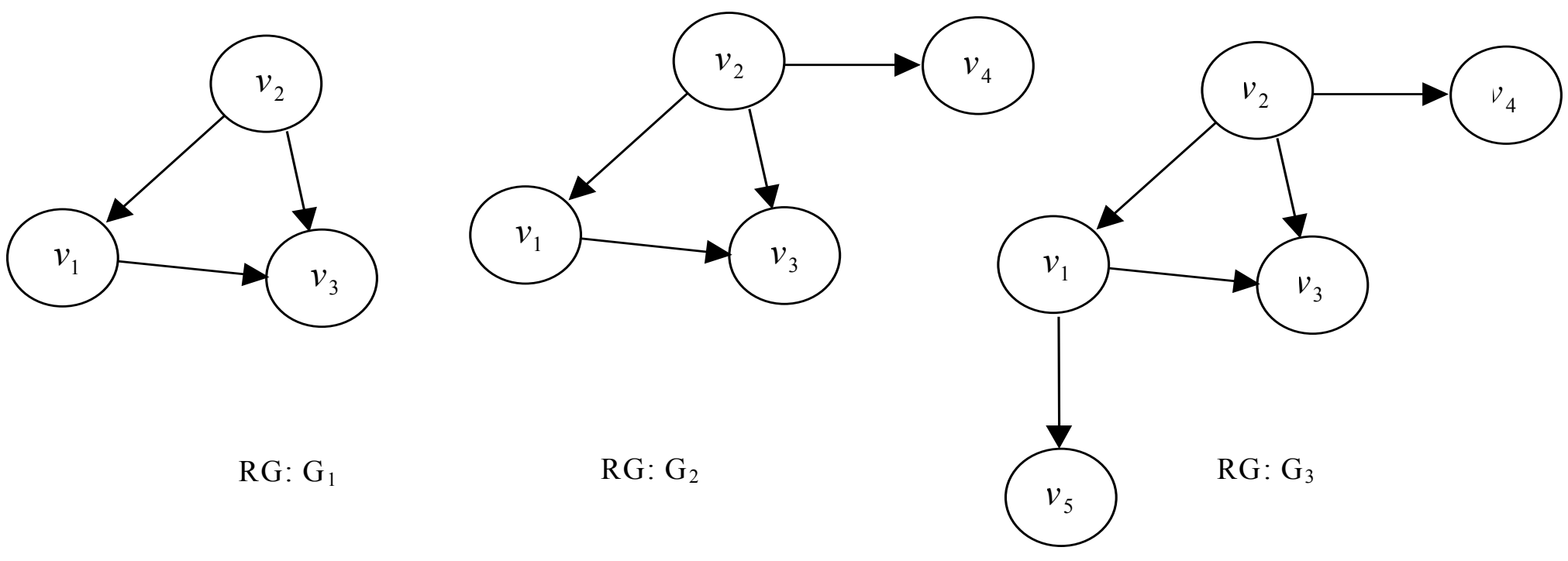

Figure 4. Random graphs $G_1$, $G_2$ and $G_3$.

| Graph element | $G_1$ | $G_2$ | $G_3$ |
|---|---|---|---|
| $v_1$ | $\Pr(b)=1$ | $\Pr(b)=1$ | $\Pr(b)=1$ |
| $v_2$ | $\Pr(a)=1$ | $\Pr(a)=1$ | $\Pr(a)=1$ |
| $v_3$ | $\Pr(c)=1$ | $\Pr(c)=1$ | $\Pr(c)=1$ |
| $v_4$ | $\Pr(\Phi)=1$ | $\Pr(d)=1$ | $\Pr(d)=1$ |
| $v_5$ | $\Pr(\Phi)=1$ | $\Pr(\Phi)=1$ | $\Pr(e)=1$ |
| $e_{21}$ | $\Pr(X \vert a,b)=1$ | $\Pr(X \vert a,b)=1$ | $\Pr(X \vert a,b)=1$ |
| $e_{23}$ | $\Pr(Y \vert a,c)=1$ | $\Pr(Y \vert a,c)=1$ | $\Pr(Y \vert a,c)=1$ |
| $e_{13}$ | $\Pr(Z \vert b,c)=1$ | $\Pr(Z \vert b,c)=1$ | $\Pr(Z \vert b,c)=1$ |
| $e_{15}$ | $\Pr(\Phi)=1$ | $\Pr(\Phi)=1$ | $\Pr(L \vert b,e)=1$ |
| $e_{24}$ | $\Pr(\Phi)=1$ | $\Pr(K \vert a,d)=1$ | $\Pr(K \vert a,d)=1$ |

Table 2. Attribute values of RGs $G_1$, $G_2$ and $G_3$.

We wish to obtain and compare the distances between $G$ and $G_1$, $G_2$ and $G_3$. First, the RGs $G'_1 = +\left(G_1, G, \mu_1^{opt}\right)$, $G'_2 = +\left(G_2, G, \mu_2^{opt}\right)$ and $G'_3 = +\left(G_3, G, \mu_3^{opt}\right)$ have to be synthesised. They have the same structure as $G$ and the attribute values presented in Table 3.

From the probabilities in Table 3, it can be deduced that the three random graphs have the same entropy, which is

$$H(G'_i) = -2 * 0.666 * \log_2(0.666) - 2 * 0.333 * \log_2(0.333) = 1.838$$

and therefore, the distance measure between each of the three $G_i$ and $G$ is

$$d(G_i, G) = H(G'_i) - \left(\frac{2}{3}H(G) + \frac{1}{3}H(G_i)\right) = 1.838 - \frac{2}{3}2.0 = 0.505$$

To conclude, the distance is the same although the structure of $G_1$, $G_2$ and $G_3$ is different. This is because first-order random graphs are probabilistic structures that do not keep any information about the global structure of the graphs used in the synthesis.

For this reason the distance measure does not distinguish whether $G_1$, $G_2$ and $G_3$ are a sub-graph of $A_1$ or $A_2$ or the union of them.

| | $G'_1$ | $G'_2$ | $G'_3$ |
|---|---|---|---|
| $v_1$ | $\Pr(b)=1$ | $\Pr(b)=1$ | $\Pr(b)=1$ |
| $v_2$ | $\Pr(a)=1$ | $\Pr(a)=1$ | $\Pr(a)=1$ |
| $v_3$ | $\Pr(c)=1$ | $\Pr(c)=1$ | $\Pr(c)=1$ |
| $v_4$ | $\Pr(d)=0.333$<br>$\Pr(\Phi)=0.666$ | $\Pr(d)=0.666$<br>$\Pr(\Phi)=0.333$ | $\Pr(d)=0.666$<br>$\Pr(\Phi)=0.333$ |
| $v_5$ | $\Pr(e)=0.333$<br>$\Pr(\Phi)=0.666$ | $\Pr(e)=0.333$<br>$\Pr(\Phi)=0.666$ | $\Pr(e)=0.666$<br>$\Pr(\Phi)=0.333$ |
| $e_{21}$ | $\Pr(X\|a,b)=1$ | $\Pr(X\|a,b)=1$ | $\Pr(X\|a,b)=1$ |
| $e_{23}$ | $\Pr(Y\|a,c)=1$ | $\Pr(Y\|a,c)=1$ | $\Pr(Y\|a,c)=1$ |
| $e_{13}$ | $\Pr(Z\|b,c)=1$ | $\Pr(Z\|b,c)=1$ | $\Pr(Z\|b,c)=1$ |
| $e_{15}$ | $\Pr(L\|b,e)=1$<br>$\Pr(\Phi)=0.666$ | $\Pr(L\|b,e)=1$<br>$\Pr(\Phi)=0.666$ | $\Pr(L\|b,e)=1$<br>$\Pr(\Phi)=0.333$ |
| $e_{24}$ | $\Pr(K\|a,d)=1$<br>$\Pr(\Phi)=0.666$ | $\Pr(K\|a,d)=1$<br>$\Pr(\Phi)=0.333$ | $\Pr(K\|a,d)=1$<br>$\Pr(\Phi)=0.333$ |

Table 3. Attribute values of $G'_1$, $G'_2$ and $G'_3$.

# 4. Function-described graphs (FDGs)

*Function-described graphs* or FDGs are proposed here for modelling classes of patterns described by attributed graphs. An FDG is a prototype structure that contains, on one hand, probabilistic functions which represent the semantic information of the local parts of the patterns and, on the other, binary functions to maintain, as much as possible, the local features of the AGs that belong to the class and also to reject the AGs that do not belong to it. In general, an FDG can be derived, like random graphs, by synthesising a cluster or set of individual AGs.

Section 4.1 compares FDGs and FORGs and describes the novel features of FDGs that represent a set of AGs. Then section 4.2 gives a formal definition of an FDG and section 4.3 presents some technical details about FDG functions. Finally, sections 4.4 and 4.5 discuss the synthesis of an FDG from a set of AGs and from a set of FDGs with given common labellings of their vertices and arcs. Some experiments with random graphs are added in section 4.6.

## 4.1. FDGs versus FORGs as prototypes of a set of AGs

In order to attain a compact representation of a set of AGs by means of a prototype, the ensemble needs to be described probabilistically so that the variations in the structural patterns of the reference set or sample can be accounted for. As has been mentioned in Section 3, random graphs (RGs) provide such a representation. Nevertheless, when estimating the probability distribution of the structural patterns from an ensemble of AGs, it is impractical to consider the high order probability distribution where all the graph elements are taken jointly. The creators of random graphs believed -and so do we- that general RGs cannot be used in real applications, and so, they proposed first-order random graphs (FORGs).

Although the FORG approach simplifies the representation considerably, it is still difficult to apply in real problems in which AGs have a large number of vertices and attributes with an extensive domain. The main cause of this problem is that the attributes of the arc depend on the attributes of the vertices that the arc connects (assumption 3 in Section 3.2). Although this supposition is useful to constrain the generalisation of the given set of AGs, a huge amount of data is required to estimate the probability density functions and the computational cost is high.

On the other hand, because of the probability independence assumptions 1 and 2 in section 3.2, FORGs have the considerable drawback that the structural information in a sample of AGs is not well preserved in the FORG which is synthesised from them. That is to say, a FORG represents an over-generalised prototype that may cover graph structures quite different from those in the sample. For example, if $C$ is a set of AGs describing different perspective views of an object $O$, many of the outcome graphs of the FORG synthesised from $C$ will represent impossible views of $O$ (just from the topological point of view, without further consideration of the attributes of primitives and relations).

Function-described graphs (FDGs) defined below, aim to offer a more practical approach and can be seen as a different type of simplification of general random graphs. They also adopt a different approximation of the joint probability $P$ of the random elements. On one hand, some independence assumptions are considered, but on the other hand, some useful functions are included that permit to constrain the generalisation of the structure.

We decided not to maintain the conditional probabilities of the arcs in the FDGs due to space and time restrictions. This means that the third assumption in the FORGs is replaced by the assumption that the arcs are independent except for the existence of the extreme vertices which is mandatory for structural coherence. Hence, the arc conditional probability density functions $q_j(\mathbf{b} \mid \mathbf{a}_{j1}, \mathbf{a}_{j2}) \hat{=} \Pr(\beta_j = \mathbf{b} \mid \alpha_{j1} = \mathbf{a}_{j1}, \alpha_{j2} = \mathbf{a}_{j2})$, $1 \le j \le m$, in FORGs are converted into marginal probability density functions $q_j(\mathbf{b}) \hat{=} \Pr(\beta_j = \mathbf{b} \mid \alpha_{j1} \ne \Phi, \alpha_{j2} \ne \Phi)$, $1 \le j \le m$, in FDGs. The underlying hypothesis is that the probability of any outcome of a random arc is the same regardless of the actual non-null outcomes of the endpoints.

In order to tackle the problem of the over-generalisation of the sample, we introduce the *antagonism*, *occurrence* and *existence* relations into FDGs, which apply to pairs of graph elements. In this way, both random vertices and arcs are not assumed to be mutually independent, at least with regards to the structural information. These second-order relations, that involve a small increase in the amount of data to be stored in the prototype, are useful for two reasons: they constrain the set of outcome graphs covered by the prototype, thus tending to cut down the structural over-generalisation quite considerably, and, they reduce the size of the search space of the AG-to-FDG matching algorithm, thus decreasing the overall time of the recognition process (see Section 5.1).

Let us now explain in more detail the kind of simplification made in FDGs with respect to General RGs. Let us begin with the independence assumptions:

1) The attributes in the vertices are independent of the other vertices and also of the arcs.
2) The attributes in the arcs are independent of the other arcs and also of the vertices. However, it is mandatory that all non-null arcs be linked to a non-null vertex at each extreme in every AG covered by an FDG. In other words, any outcome AG of the FDG has to be structurally consistent.

With these assumptions, the probability density functions are themselves independent since the attributes in the arcs do not depend on the attributes in the vertices that they connect, but only on the existence of the extreme vertices. Consequently, associated with each graph element in an FDG, there is a *random variable* that represents the distribution of the semantic information of the corresponding graph elements in the set of outcome AGs. A random variable has a probability density function defined over the same attribute domains of the AGs, including the null value $\Phi$, that denotes the non-instantiation of an FDG graph element in an outcome AG.

It is interesting to emphasise that the attribute domain of each AG element is in general a tuple. For this reason, the random variables are associated with joint probability density functions which depend on the whole set of attributes in the tuple. However, in real applications, the tuple elements are usually considered mutually independent which avoids the spatial cost of representing a joint density function in each graph element. Hence, the joint probability of an AG element can be estimated as the product of the probabilities of all the attributes in the tuple.

The probability density functions are not represented analytically in practice, but non-parametrically. Thus, any type of function can be defined, whether discrete or continuous. If the attributes are discrete, the representation is simple, just by storing the frequencies of each possible value. When the values are continuous, we use a discretisation process to represent the density function computationally, although a parametric model (e.g. mixture of Gaussians) could be estimated alternatively.

The supposition of independence between graph elements can involve an excessive generalisation of the set of patterns used to build the FDG, because objects that do not belong to the target class will be covered. To improve the representation capability, second order probability information (i.e. the joint probabilities of two graph elements) could be added. But since it is extremely difficult in practice to estimate and handle these joint probabilities, only qualitative information of the second order probability functions is added. Hence, in FDGs, the marginal (first order) probability functions are complemented by second order Boolean functions (relations), which provide this qualitative second order information.

Suppose two vertices in the FDG, $\omega_1$ and $\omega_2$, the attributes of which are two random elements $\alpha_1$ and $\alpha_2$ in the domain $\Delta_w = \{\Delta_v \cup \Phi\}$, where $\Delta_v$ is the domain of the attributes of the AGs that the FDG represents and $\Phi$ represents the non-existence of vertices in the AG. By definition, the sum of the joint probabilities of these two elements of an FDG over their attribute domain $\Delta_\omega$ equals 1,

$$\sum_{x\in\Delta_\omega}\sum_{y\in\Delta_\omega}\Pr(\alpha_1 = x \wedge \alpha_2 = y) = 1 \quad (8)$$

Moreover, the domain $\Delta_\omega \times \Delta_\omega$ can be split into four regions (Figure 5 shows these four regions for the vertices). The first one is composed of the points that belong to the Cartesian product of the sets of actual attributes of the two elements $\Delta_v$, corresponding to the cases in which both elements are defined in the initial non-extended AG and so their value is not null. The second and third regions are both straight lines in which only one of the elements has the null value. This covers the cases when one of the two elements does not belong to the initial AG and has been added in the extending process. Finally, the fourth region is the single point where both elements have the null value ( $x = \Phi \wedge y = \Phi$ ) and, therefore, this includes the cases in which none of them appear in the initial AG.

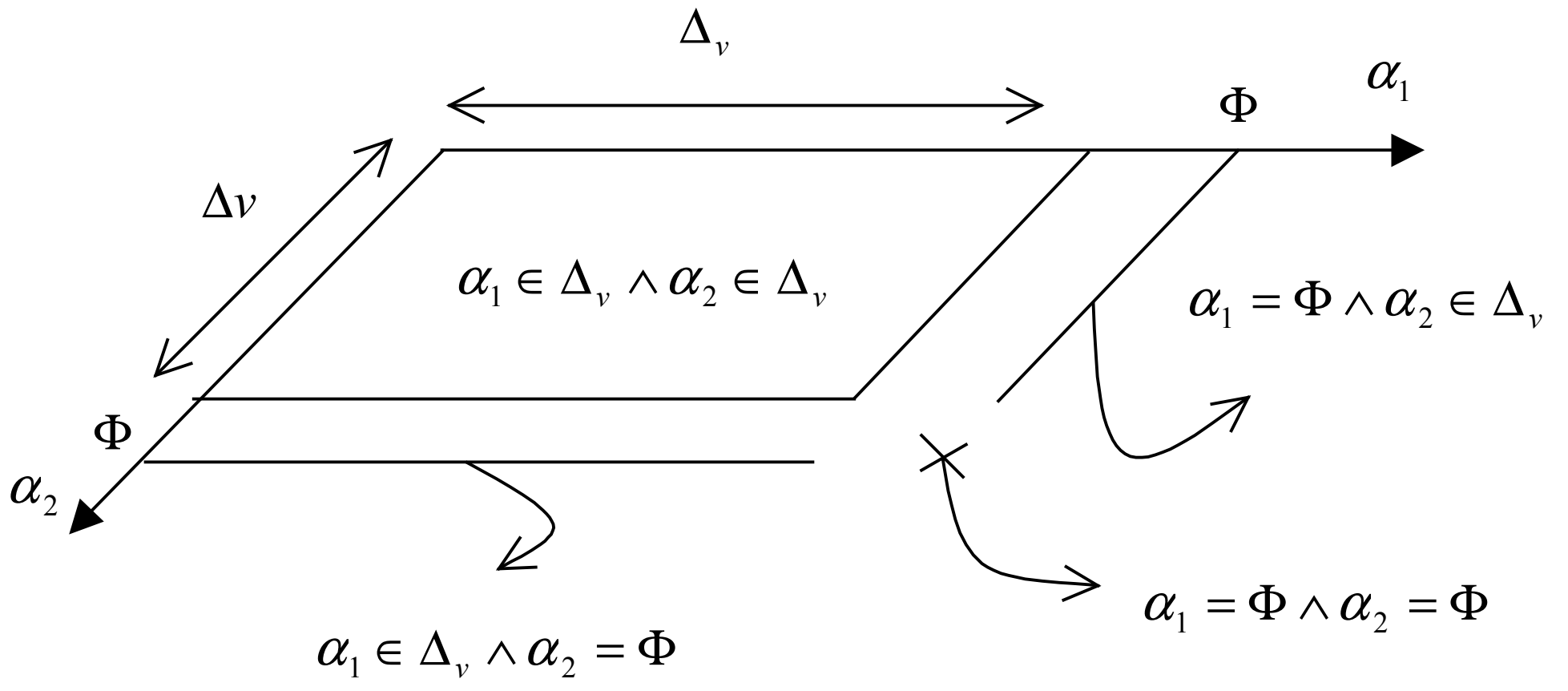

Figure 5. Joint probability of two vertices split into four regions.

The sum of joint probabilities is composed of four terms according to these four regions as follows,

$$\begin{pmatrix} \Pr(\alpha_1 \neq \Phi \wedge \alpha_2 \neq \Phi) + \Pr(\alpha_1 = \Phi \wedge \alpha_2 \neq \Phi) + \\ \Pr(\alpha_1 \neq \Phi \wedge \alpha_2 = \Phi) + \Pr(\alpha_1 = \Phi \wedge \alpha_2 = \Phi) \end{pmatrix} = 1 \tag{9}$$

This last equation is used to implement a *qualitative approximation* of the joint probability function of two elements represented through the *antagonism*, *occurrence* and *existence functions* in the FDGs.

In the case that the probabilities in the first region are all zero, $\Pr(\alpha_1 \neq \Phi \wedge \alpha_2 \neq \Phi) = 0$, we say the two graph elements of the FDG are *antagonistic*, which means that, although they are included in the prototype as different elementary parts of the covered patterns, they have never taken place together in any AG of the reference set used to synthesise the FDG. On the other hand, if the joint probability function equals zero in the second region, $\Pr(\alpha_1 \neq \Phi \wedge \alpha_2 = \Phi) = 0$, it is possible to assure that if the element $\alpha_1$ appears in any AG of the reference set then the element $\alpha_2$ must also appear. So, there is a structural dependence of the element $\alpha_2$ on the element $\alpha_1$. That is called an *occurrence* relation. The case of the third region is analogous to the second one, with the only difference that the elements are swapped. Finally, if $\Pr(\alpha_1 = \Phi \wedge \alpha_2 = \Phi) = 0$, all the objects in the class described by the FDG have at least one of the two elements, and we say that there is an *existence* relation between them.

Figure 6 shows three possible joint probabilities of the vertices $\omega_i$ and $\omega_j$. In case (a), they are defined as antagonistic, whereas in case (b), there is an occurrence relation from $\omega_i$ to $\omega_j$, and in case (c), there is an existence relation between $\omega_i$ and $\omega_j$.

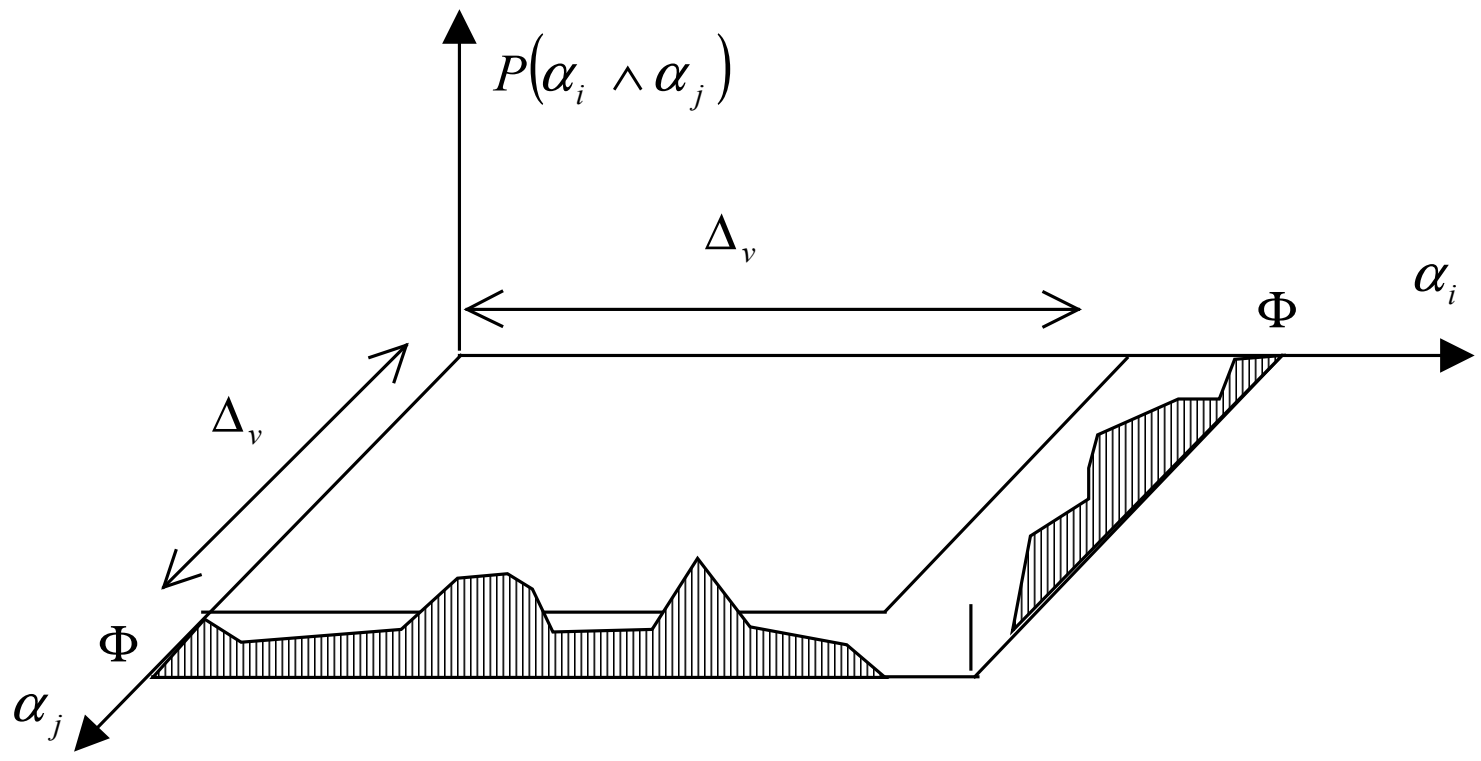

Figure 6.a. Example of joint probability that defines an antagonistic relation.

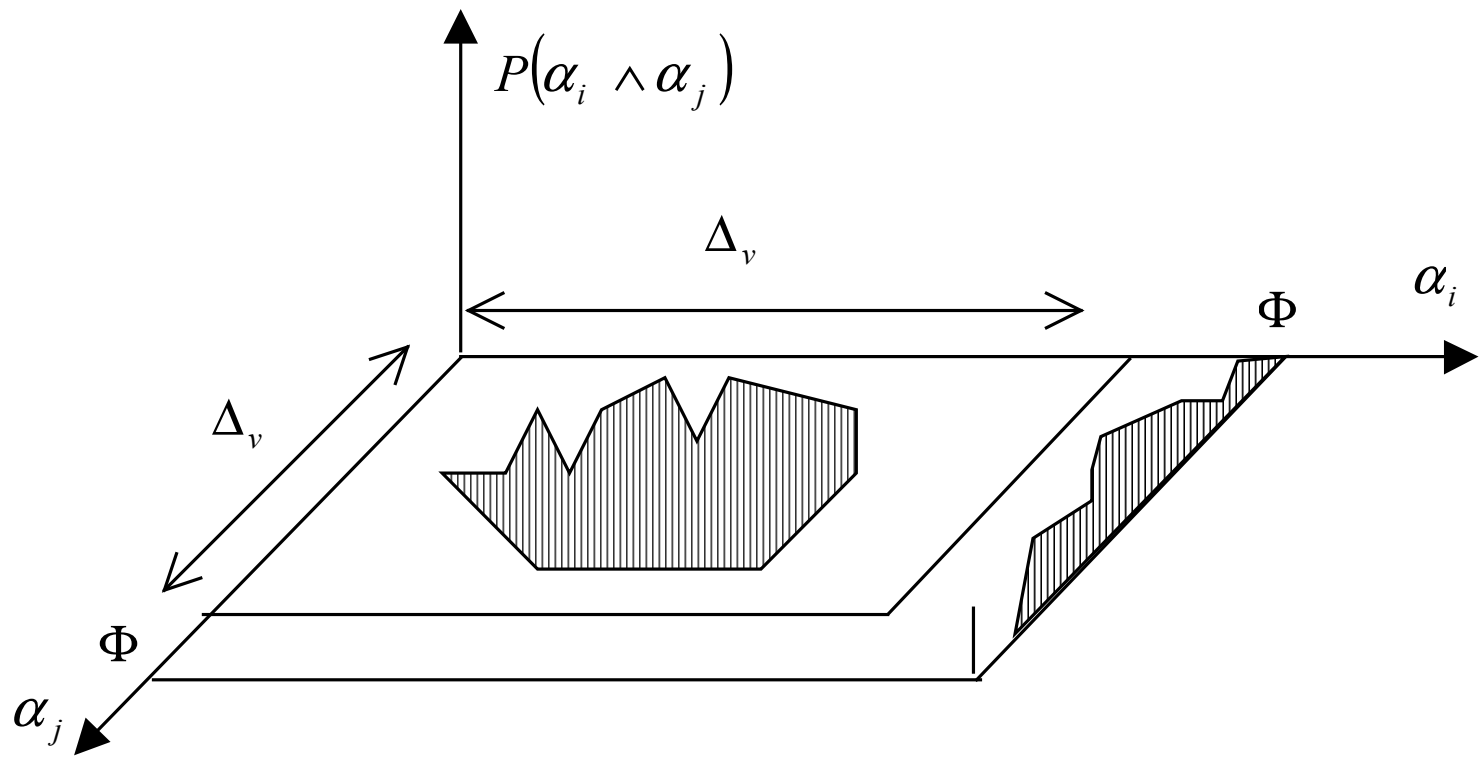

Figure 6.b. Example of joint probability that defines an occurrence relation.

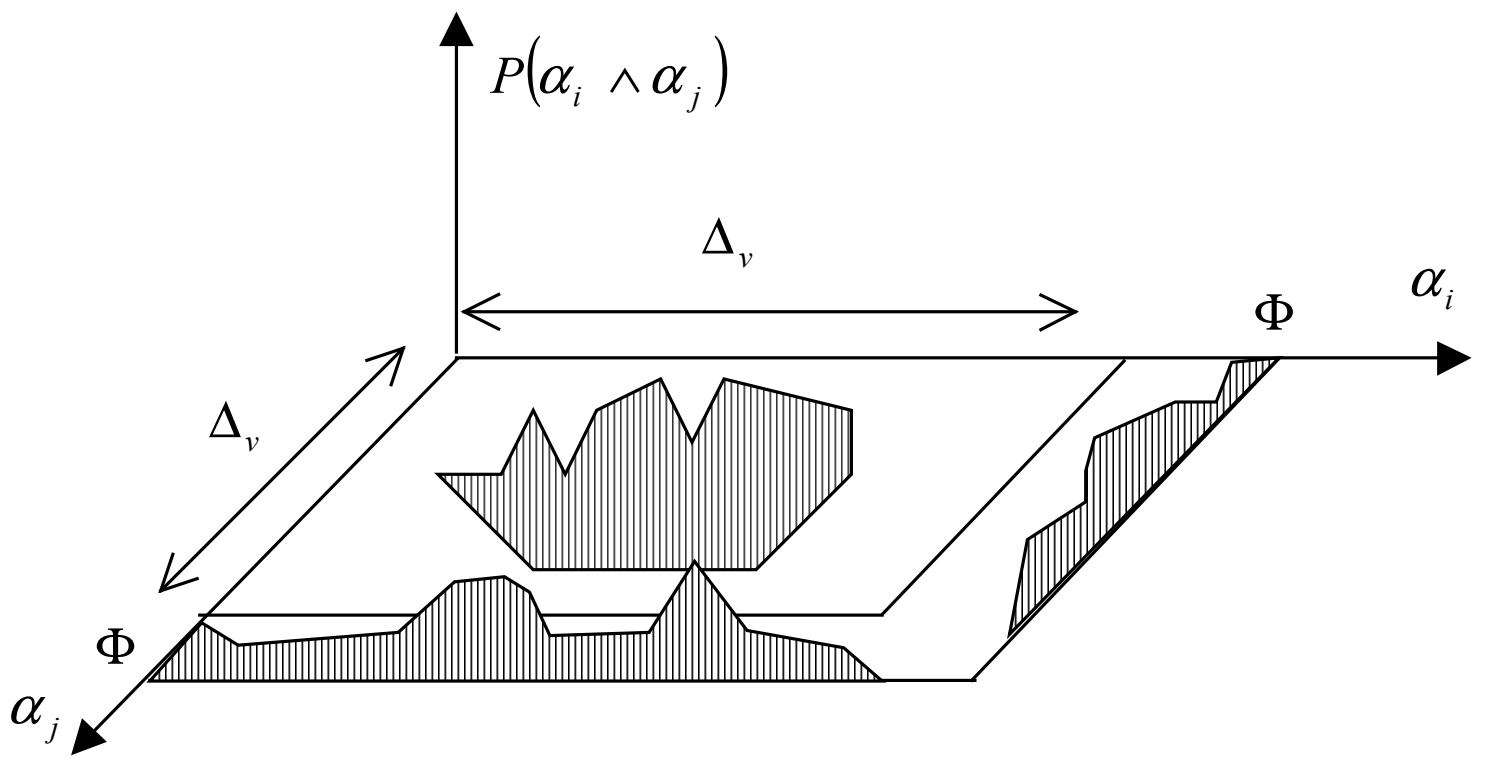

Figure 6.c. Example of joint probability that defines an existence relation.

### 4.2. Formal definition of FDGs

***Definition 5*** (Function-described graph or FDG): A *function-described graph F* over $(\Delta_v, \Delta_e)$ with an underlying graph structure $H = (\Sigma_\omega, \Sigma_\varepsilon)$ is defined as a tuple $(W, B, P, R)$ such that

1. $W = (\Sigma_\omega, \gamma_\omega)$ is a *random vertex set* and $\gamma_\omega : \Sigma_\omega \to \Omega_\omega$ is a mapping that associates each vertex $\omega_i \in \Sigma_\omega$ with a random variable $\alpha_i = \gamma_\omega(\omega_i)$ with values in $\Delta_\omega$.
2. $B = (\Sigma_\varepsilon, \gamma_\varepsilon)$ is a *random arc set* and $\gamma_\varepsilon : \Sigma_\varepsilon \to \Omega_\varepsilon$ is a mapping that associates each arc $\varepsilon_{kl} \in \Sigma_\varepsilon$ with a random variable $\beta_j = \gamma_\varepsilon(\varepsilon_{kl})$ with values in $\Delta_\varepsilon$.
3. $P = (P_\omega, P_\varepsilon)$ are two sets of marginal (or first-order) probability density functions for random vertices and edges, respectively. That is, $P_\omega = \{p_i(\mathbf{a}), 1 \le i \le n\}$ and $P_\varepsilon = \{q_j(\mathbf{b}), 1 \le j \le m\}$ (being *m* the number of edges), where $p_i(\mathbf{a}) \equiv \Pr(\alpha_i = \mathbf{a})$ for all $\mathbf{a} \in \Delta_\omega$ and $q_j(\mathbf{b}) \equiv \Pr(\beta_j = \mathbf{b} | \alpha_{j1} \ne \Phi \wedge \alpha_{j2} \ne \Phi)$ for all $\mathbf{b} \in \Delta_\varepsilon$ such that $\alpha_{j1}, \alpha_{j2}$ refer to the random variables for the endpoints of the random arc associated with $\beta_j$.
4. $R = (A_\omega, A_\varepsilon, O_\omega, O_\varepsilon, E_\omega, E_\varepsilon)$ is a collection of Boolean functions defined over pairs of graph elements (i.e. relations on the sets of vertices and arcs) that allow qualitative second-order probability information to be incorporated. $A_\omega$ and $A_\varepsilon$ are the so-called *vertex antagonism* and *arc antagonism functions*, respectively, where $A_\omega : \Sigma_\omega \times \Sigma_\omega \to \{0,1\}$ is defined by $A_\omega(\omega_i, \omega_j) = 1 \Leftrightarrow \Pr(\alpha_i \neq \Phi \wedge \alpha_j \neq \Phi) = 0$, and similarly, $A_\varepsilon : \Sigma_\varepsilon \times \Sigma_\varepsilon \to \{0,1\}$ is defined by $A_\varepsilon(\varepsilon_{kl}, \varepsilon_{pq}) = 1 \Leftrightarrow \Pr(\beta_i \neq \Phi \wedge \beta_j \neq \Phi) = 0$, where $\beta_i = \gamma_\varepsilon(\varepsilon_{kl})$ and $\beta_j = \gamma_\varepsilon(\varepsilon_{pq})$. The above functions can be seen alternatively as symmetric binary relations on the sets $\Sigma_\omega$ and $\Sigma_\varepsilon$, respectively. In addition, $O_\omega$ and $O_\varepsilon$ are the so-called *vertex occurrence* and *arc occurrence functions*, where $O_\omega : \Sigma_\omega \times \Sigma_\omega \to \{0,1\}$ is defined by $O_\omega(\omega_i, \omega_j) = 1 \Leftrightarrow \Pr(\alpha_i \neq \Phi \wedge \alpha_j = \Phi) = 0$, and $O_\varepsilon : \Sigma_\varepsilon \times \Sigma_\varepsilon \to \{0,1\}$ is defined by $O_\varepsilon(\varepsilon_{kl}, \varepsilon_{pq}) = 1 \Leftrightarrow \Pr(\beta_i \neq \Phi \wedge \beta_j = \Phi) = 0$, where $\beta_i = \gamma_\varepsilon(\varepsilon_{kl})$ and $\beta_j = \gamma_\varepsilon(\varepsilon_{pq})$. These last functions can be seen alternatively as reflexive and transitive relations (partial orders) on the sets $\Sigma_\omega$ and $\Sigma_\varepsilon$, respectively. Finally, $E_\omega$ and $E_\varepsilon$ are the so-called *vertex existence* and *arc existence functions*, where $E_\omega : \Sigma_\omega \times \Sigma_\omega \to \{0,1\}$ is defined by $E_\omega(\omega_i, \omega_j) = 1 \Leftrightarrow \Pr(\alpha_i = \Phi \wedge \alpha_j = \Phi) = 0$, and $E_\varepsilon : \Sigma_\varepsilon \times \Sigma_\varepsilon \to \{0,1\}$ is defined by $E_\varepsilon(\varepsilon_{kl}, \varepsilon_{pq}) = 1 \Leftrightarrow \Pr(\beta_i = \Phi \wedge \beta_j = \Phi) = 0$, where $\beta_i = \gamma_\varepsilon(\varepsilon_{kl})$ and $\beta_j = \gamma_\varepsilon(\varepsilon_{pq})$. These two last functions can be seen alternatively as symmetric binary relations on the sets $\Sigma_\omega$ and $\Sigma_\varepsilon$, respectively.

Figure 7 shows a simple FDG with an underlying graph structure composed by 4 vertices and 6 arcs. There is also an antagonistic relation between vertices $\omega_1$ and $\omega_3$ and an occurrence relation from arc $\varepsilon_{4,2}$ to arc $\varepsilon_{3,2}$.

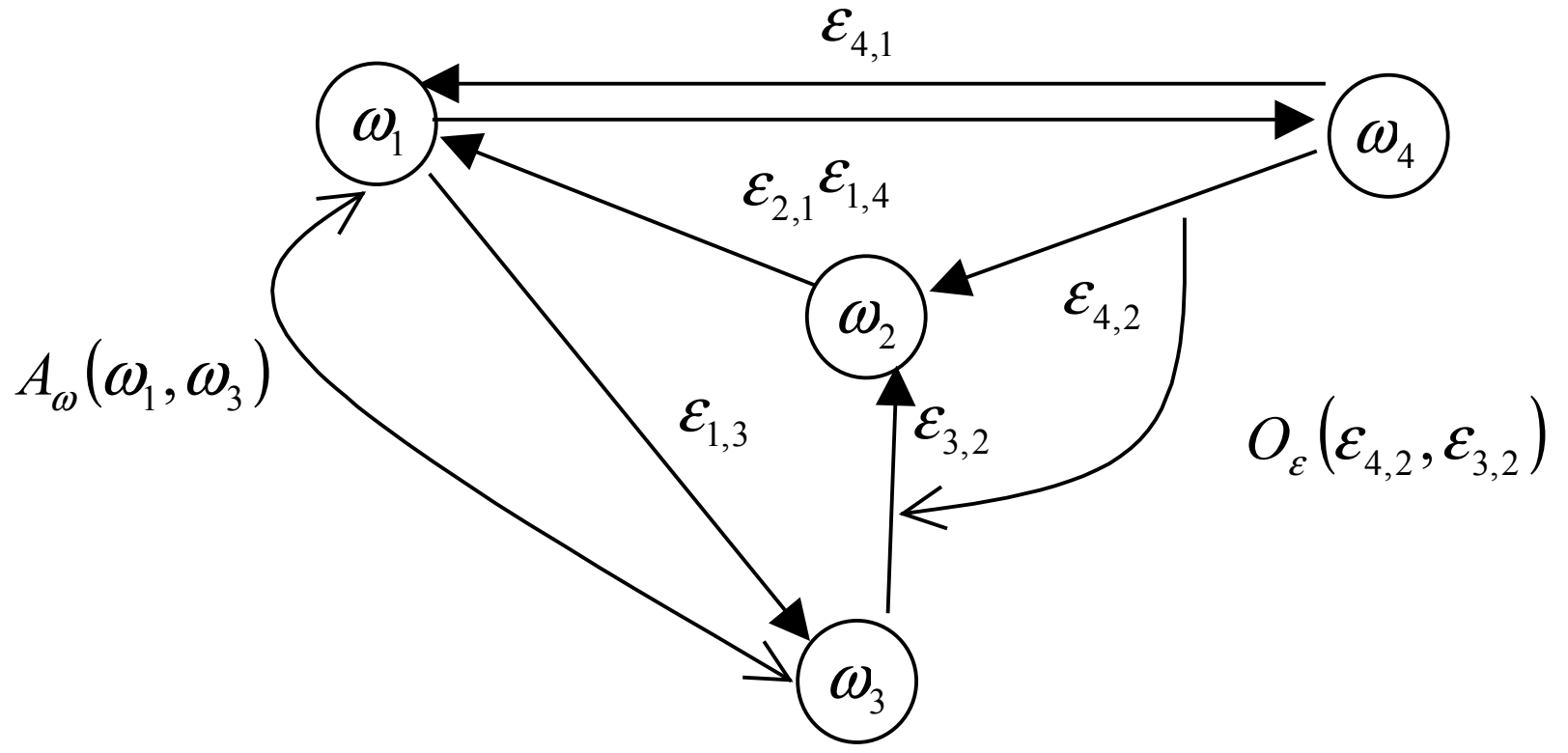

Figure 7. An example of FDG.

Because of the structural consistency requirements, there is no need to store the conditional probabilities $\Pr(\beta_j = \mathbf{b} | \alpha_{j1} = \Phi \vee \alpha_{j2} = \Phi)$ in the structure of the FDGs, since, by definition, an arc cannot exist or has to be defined as null if one of the connecting vertices does not exist or is null, that is, $\Pr(\beta_j = \Phi | \alpha_{j1} = \Phi \vee \alpha_{j2} = \Phi) = 1$.

Furthermore, two graph elements (of the same type) are *co-occurrent* if and only if the occurrence relation applies to them in both directions. Figure 8 shows a joint probability of two vertices that defines a co-occurrence relation. $C_\omega$ and $C_\varepsilon$ are the so-called *vertex co-occurrence* and *arc co-occurrence functions*, where $C_\omega : \Sigma_\omega \times \Sigma_\omega \rightarrow \{0,1\}$ is defined by $C_\omega(\omega_i, \omega_j) = 1 \Leftrightarrow \Pr(\alpha_i \neq \Phi \wedge \alpha_j = \Phi) + \Pr(\alpha_i = \Phi \wedge \alpha_j \neq \Phi) = 0$, and $C_\varepsilon : \Sigma_\varepsilon \times \Sigma_\varepsilon \rightarrow \{0,1\}$ is defined by $C_\varepsilon(\varepsilon_{kl}, \varepsilon_{pq}) = 1 \Leftrightarrow \Pr(\beta_i \neq \Phi \wedge \beta_j = \Phi) + \Pr(\beta_i = \Phi \wedge \beta_j \neq \Phi) = 0$, where $\beta_i = \gamma_\varepsilon(\varepsilon_{kl})$ and $\beta_j = \gamma_\varepsilon(\varepsilon_{pq})$. It follows that *co-occurrence* of vertices or arcs are symmetric binary relations on the sets $\Sigma_\omega$ and $\Sigma_\varepsilon$, respectively. The *co-occurrence* relations are not stored in the FDGs since they can be easily deduced from the occurrence relations.

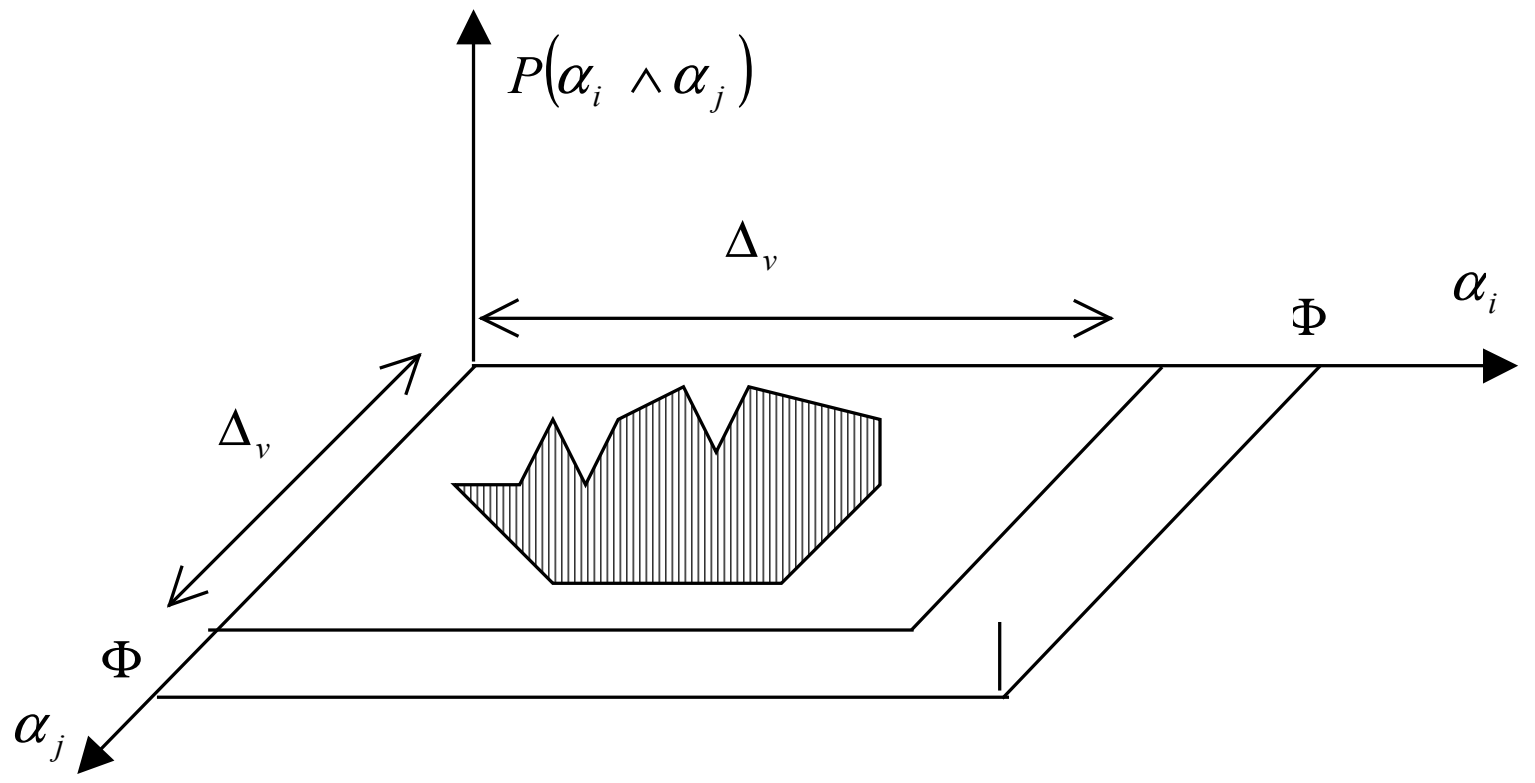

Figure 8. Example of joint probability that defines a co-occurrence relation.

A random element $\delta$ of an FDG is a *null random element* if its probability of instantiation to the null value is one, $\Pr(\delta = \Phi) = 1$. This means that all the values in the attribute tuple of every instance of a null random element are the null value $\phi$.

A *complete FDG* is an FDG with a complete graph structure *H* (but which may include null elements). The FDG extending process, needed for calculating the distance between an AG and an FDG, adds null graph elements to make an FDG complete and with a desired order (greater or equal than the original one). These null graph elements are used to signify a missing element in the "initial FDG" that represents the prototype. It is important to note that the initial FDG and the extended FDG share the same semantic and structural information, thus representing exactly the same prototype.

***Definition 6*** (*k-extension* of an FDG): A function-described graph $F=(W,B,P,R)$ of order $n$ can be *extended* to form a complete FDG $F'=(W',B',P',R')$ of order $k, k \geq n$, by adding null vertices and null arcs and extending appropriately both the set of probability density functions and the Boolean functions that relate graph elements. We call $F'$ the *k-extension* of $F$.

As a result of defining the antagonism, occurrence and existence functions of the FDGs, the value of these functions in an extended FDG is as shown in Tables 4, 5 and 6, respectively, where $\delta_i$ and $\delta_j$ are the random variables associated with elements $\gamma_i$ and $\gamma_j$.

| $A(\gamma_i,\gamma_j)$ in F' | $\gamma_i$: Null element | $\gamma_i$: Non-null element |
|---|---|---|
| $\gamma_j$: Null element | 1 | 1 |
| $\gamma_j$: Non-null element | 1 | $A(\gamma_i,\gamma_j)$ in F |

Table 4. Antagonism relation between null or non-null elements.

| $O(\gamma_i,\gamma_j)$ in F' | $\gamma_i$: Null element | $\gamma_i$: Non-null element |
|---|---|---|
| $\gamma_j$: Null element | 1 | 0 |
| $\gamma_j$: Non-null element | 1 | $O(\gamma_i,\gamma_j)$ in F |

Table 5. Occurrence relation between null or non-null elements.

| $E(\gamma_i,\gamma_j)$ in F' | $\gamma_i$ : Null element | $\gamma_i$ : Non-null element |
|---|---|---|
| $\gamma_j$ : Null element | 0 | 1 if $\Pr(\delta_i = \Phi) = 0$<br>0 otherwise |
| $\gamma_j$ : Non-null element | 1 if $\Pr(\delta_j = \Phi) = 0$<br>0 otherwise | $E(\gamma_i,\gamma_j)$ in F |

Table 6. Existence relation between null or non-null elements.

## 4.3. Relationships between functions of FDGs

The following sections discuss some interesting properties of the information represented in an FDG. These properties relate some of the aforementioned functions. Particular mention is made of the relations between first and second order probabilities the antagonism, occurrence and existence relations and the joint probability of the null value.

### 4.3.1. Non-conditional probabilities of arc attributes

Concerning the information stored in the arcs, the first-order probabilities depend only on the existence of the extreme vertices, whereas in the antagonism, occurrence and existence relations, this conditional constraint is not taken in consideration (definition 5, FDG). This is because the imposition of this constraint alone on the first-order probabilities is sufficient to guarantee that the outcome AGs of an FDG are structurally coherent.

Hence, it is interesting to observe the relation between non-conditional probabilities in the arcs and the first-order probability density functions defined in the FDGs. There are two cases: the probability of the null value and the probability of a non-null value. The relations that follow are taken into account in the study of both the second-order Boolean functions and the distance between AGs and FDGs.

Let us consider the non-conditional probability of the null value, $\Pr(\beta_i = \Phi)$, which can be calculated as the sum of the following four terms:

$$\Pr(\beta_i = \Phi) = \begin{bmatrix} \Pr(\beta_i = \Phi | \alpha_{i1} = \Phi \wedge \alpha_{i2} = \Phi) * \Pr(\alpha_{i1} = \Phi \wedge \alpha_{i2} = \Phi) + \\ \Pr(\beta_i = \Phi | \alpha_{i1} = \Phi \wedge \alpha_{i2} \neq \Phi) * \Pr(\alpha_{i1} = \Phi \wedge \alpha_{i2} \neq \Phi) + \\ \Pr(\beta_i = \Phi | \alpha_{i1} \neq \Phi \wedge \alpha_{i2} = \Phi) * \Pr(\alpha_{i1} \neq \Phi \wedge \alpha_{i2} = \Phi) + \\ \Pr(\beta_i = \Phi | \alpha_{i1} \neq \Phi \wedge \alpha_{i2} \neq \Phi) * \Pr(\alpha_{i1} \neq \Phi \wedge \alpha_{i2} \neq \Phi) \end{bmatrix} \quad (10)$$

Since the FDG describes structurally coherent AGs, the arc conditional probability of the null value when one of the extreme vertices is also null is 1, by definition. Moreover, since the attributes in the vertices are regarded as independent, the second order probabilities in the vertices can be approximated as the product of the first order probabilities. Thus,

$$\Pr(\beta_i = \Phi) = \begin{bmatrix} 1 * \Pr(\alpha_{i1} = \Phi) * \Pr(\alpha_{i2} = \Phi) + \\ 1 * \Pr(\alpha_{i1} = \Phi) * \Pr(\alpha_{i2} \neq \Phi) + \\ 1 * \Pr(\alpha_{i1} \neq \Phi) * \Pr(\alpha_{i2} = \Phi) + \\ \Pr(\beta_i = \Phi | \alpha_{i1} \neq \Phi \wedge \alpha_{i2} \neq \Phi) * \Pr(\alpha_{i1} \neq \Phi) * \Pr(\alpha_{i2} \neq \Phi) \end{bmatrix} \quad (11)$$

Therefore, the non-conditional probability of the null value, $\Pr(\beta_i = \Phi)$, can be expressed in terms of the marginal probabilities of both extreme vertices $p_{i1}(\Phi)$ and $p_{i2}(\Phi)$ and the arc conditional probability $q_i(\Phi)$, which are stored in the FDG, as follows,

$$\Pr(\beta_i = \Phi) = \begin{bmatrix} p_{i1}(\Phi) * p_{i2}(\Phi) + \\ p_{i1}(\Phi) * (1 - p_{i2}(\Phi)) + \\ (1 - p_{i1}(\Phi)) * p_{i2}(\Phi) + \\ q_i(\Phi) * (1 - p_{i1}(\Phi)) * (1 - p_{i2}(\Phi)) \end{bmatrix} \quad (12)$$

Finally, after some manipulation of the equation, we arrive at the final expression

$$\Pr(\beta_i = \Phi) = 1 - \Pr(\beta_i \neq \Phi) = 1 - (1 - q_i(\Phi)) * (1 - p_{i1}(\Phi)) * (1 - p_{i2}(\Phi)) \quad (13)$$

Let us now consider the non-conditional probability of a specific non-null value, $\Pr(\beta_i = \mathbf{b})$, where $\mathbf{b} \in \Delta_e$, which can be obtained as the sum

$$\Pr(\beta_i = \mathbf{b}) = \begin{bmatrix} \Pr(\beta_i = \mathbf{b} | \alpha_{i1} = \Phi \wedge \alpha_{i2} = \Phi) * \Pr(\alpha_{i1} = \Phi \wedge \alpha_{i2} = \Phi) + \\ \Pr(\beta_i = \mathbf{b} | \alpha_{i1} = \Phi \wedge \alpha_{i2} \neq \Phi) * \Pr(\alpha_{i1} = \Phi \wedge \alpha_{i2} \neq \Phi) + \\ \Pr(\beta_i = \mathbf{b} | \alpha_{i1} \neq \Phi \wedge \alpha_{i2} = \Phi) * \Pr(\alpha_{i1} \neq \Phi \wedge \alpha_{i2} = \Phi) + \\ \Pr(\beta_i = \mathbf{b} | \alpha_{i1} \neq \Phi \wedge \alpha_{i2} \neq \Phi) * \Pr(\alpha_{i1} \neq \Phi \wedge \alpha_{i2} \neq \Phi) \end{bmatrix} \quad (14)$$

Since the FDG describes structurally coherent AGs, the arc conditional probability of a non-null value when one of the extreme vertices is null is 0, by definition. Thus,

$$\Pr(\beta_i = \mathbf{b}) = \Pr(\beta_i = \mathbf{b} | \alpha_{i1} \neq \Phi \wedge \alpha_{i2} \neq \Phi) * \Pr(\alpha_{i1} \neq \Phi \wedge \alpha_{i2} \neq \Phi) \quad (15)$$

Due to the independence among vertices, it turns out that

$$\Pr(\beta_i = \mathbf{b}) = \Pr(\beta_i = \mathbf{b} | \alpha_{i1} \neq \Phi \wedge \alpha_{i2} \neq \Phi) * \Pr(\alpha_{i1} \neq \Phi) * \Pr(\alpha_{i2} \neq \Phi) \quad (16)$$

and, using the first-order probabilities stored in the FDG, the final equation shows again a dependence on the extreme vertices,

$$\Pr(\beta_i = \mathbf{b}) = q_i(\mathbf{b}) * (1 - p_{i1}(\Phi)) * (1 - p_{i2}(\Phi)) \quad (17)$$

### 4.3.2. Second-order functions and joint probabilities of the null value

Antagonism, occurrence and existence functions are defined independently in FDGs. However, we prove here that there is dependence among them through the first and second order probabilities of the null value. Since first order information is explicitly stored in FDGs, we will see that to find out whether two graph elements have an antagonism, occurrence or existence relation it is only necessary to know the joint probability of these two graph elements being null, $\Pr(\alpha_i = \Phi \wedge \alpha_j = \Phi)$. In a computer implementation, these relations (and the co-occurrence, if needed) can be maintained directly as relations or easily calculated, by using the equations shown below and storing the joint probability of the null value.

If we take into account that $\Pr(\alpha_i = \Phi \vee \alpha_j = \Phi) = 1 \Leftrightarrow \Pr(\alpha_i \neq \Phi \wedge \alpha_j \neq \Phi) = 0$, the antagonism in the vertices can be given by

$$A_\omega(\omega_i, \omega_j) = 1 \Leftrightarrow \Pr(\alpha_i = \Phi \vee \alpha_j = \Phi) = 1 \tag{18}$$

Using the first and second order probabilities, this is equivalent to

$$A_\omega(\omega_i, \omega_j) = 1 \Leftrightarrow \Pr(\alpha_i = \Phi) + \Pr(\alpha_j = \Phi) - \Pr(\alpha_i = \Phi \wedge \alpha_j = \Phi) = 1 \tag{19}$$

Hence, we obtain a new equation of the antagonism relation in the vertices that depends on the marginal probabilities defined in the FDG and also on the joint probabilities of the null value.

$$A_\omega(\omega_i, \omega_j) = 1 \Leftrightarrow p_i(\Phi) + p_j(\Phi) - \Pr(\alpha_i = \Phi \wedge \alpha_j = \Phi) = 1 \tag{20}$$

The antagonism function in the arcs can be expressed in a similar way. Let $\gamma_\varepsilon(\varepsilon_{ab}) = \beta_i$ and $\gamma_\varepsilon(\varepsilon_{cd}) = \beta_j$ be two random arc elements. Firstly, we note that

$$A_\varepsilon(\varepsilon_{ab}, \varepsilon_{cd}) = 1 \Leftrightarrow \Pr(\beta_i = \Phi \vee \beta_j = \Phi) = 1 \tag{21}$$

and then by using the first and second order probabilities,

$$A_\varepsilon(\varepsilon_{ab}, \varepsilon_{cd}) = 1 \Leftrightarrow \Pr(\beta_i = \Phi) + \Pr(\beta_j = \Phi) - \Pr(\beta_i = \Phi \wedge \beta_j = \Phi) = 1 \tag{22}$$

This last equation can be rewritten using the first order probabilities of vertices and arcs stored in the FDG together with the arc joint probabilities of the null value,

$$A_\varepsilon(\varepsilon_{ab},\varepsilon_{cd})=1 \Leftrightarrow \begin{pmatrix} 1-(1-q_i(\Phi))*(1-p_{i1}(\Phi))*(1-p_{i2}(\Phi))+ \\ 1-(1-q_j(\Phi))*(1-p_{j1}(\Phi))*(1-p_{j2}(\Phi))- \\ \Pr(\beta_i=\Phi\wedge\beta_j=\Phi)=1 \end{pmatrix} \tag{23}$$

Restructuring the equation, the final expression is

$$A_\varepsilon(\varepsilon_{ab},\varepsilon_{cd})=1 \Leftrightarrow \begin{pmatrix} (1-q_i(\Phi))*(1-p_{i1}(\Phi))*(1-p_{i2}(\Phi))+ \\ (1-q_j(\Phi))*(1-p_{j1}(\Phi))*(1-p_{j2}(\Phi))+ \\ \Pr(\beta_i=\Phi\wedge\beta_j=\Phi)=1 \end{pmatrix} \tag{24}$$

Now let us consider the occurrence functions between vertices. They can be rewritten using the first and second order probabilities of the null value as follows

$$O_\omega(\omega_i,\omega_j)=1 \Leftrightarrow \Pr(\alpha_j=\Phi)-\Pr(\alpha_i=\Phi\wedge\alpha_j=\Phi)=0 \tag{25}$$

Hence, replacing the first order probabilities by the associated FDG functions, we obtain

$$O_\omega(\omega_i,\omega_j)=1 \Leftrightarrow p_j(\Phi)-\Pr(\alpha_i=\Phi\wedge\alpha_j=\Phi)=0 \tag{26}$$

The occurrence relations between arcs can be restated in a similar way,

$$O_\varepsilon(\varepsilon_{ab},\varepsilon_{cd})=1 \Leftrightarrow \Pr(\beta_j=\Phi)-\Pr(\beta_i=\Phi\wedge\beta_j=\Phi)=0 \tag{27}$$

and using the FDG probability density functions we obtain

$$O_\varepsilon(\varepsilon_{ab},\varepsilon_{cd})=1 \Leftrightarrow \begin{bmatrix} 1-(1-q_j(\Phi))*(1-p_{j1}(\Phi))*(1-p_{j2}(\Phi)) \\ -\Pr(\beta_i=\Phi\wedge\beta_j=\Phi)=0 \end{bmatrix} \tag{28}$$

Finally, the equation is simplified as

$$O_\varepsilon(\varepsilon_{ab},\varepsilon_{cd})=1 \Leftrightarrow \begin{bmatrix} (1-q_j(\Phi))*(1-p_{j1}(\Phi))*(1-p_{j2}(\Phi)) \\ +\Pr(\beta_i=\Phi \wedge \beta_j=\Phi)=1 \end{bmatrix} \tag{29}$$

There is no need to mention here the existence relations on the vertices or arcs as they are directly defined through the second order probabilities of the null value (see Section 4.2).

### 4.3.3. Second-order effects of first-order information

From equations (20) and (26) it is easily derived that

$$A_\omega(\omega_i,\omega_j)=1 \wedge O_\omega(\omega_i,\omega_j)=1 \Leftrightarrow p_i(\Phi)=1 \tag{30}$$

which is to say that only null vertices are both antagonistic and occurrent to all the other vertices of the graph, and moreover, a null vertex will always be antagonistic with and occurrent to the remaining vertices. A similar equation is found for the arcs in equations (24) and (29),

$$A_\varepsilon(\varepsilon_{ab},\varepsilon_{cd})=1 \wedge O_\varepsilon(\varepsilon_{ab},\varepsilon_{cd})=1 \Leftrightarrow \Pr(\beta_i=\Phi)=1 \tag{31}$$

Some cases satisfy both antagonism and occurrence between two elements but this is contrary to the common sense. In these cases, there is one artificially inserted element. That is, it makes no sense to say that the two elements cannot appear together (antagonism) and, at the same time, that one of them must appear whenever the other does (occurrence).

On the other hand, from equation (26) and the definition of vertex existence function, we also have that

$$E_\omega(\omega_i,\omega_j)=1 \wedge O_\omega(\omega_i,\omega_j)=1 \Leftrightarrow p_j(\Phi)=0 \tag{32}$$

which is to say that only *strict non-null vertices* are both existent and occurrent from all the other vertices of the graph, and moreover, a strict non-null vertex will always be existent with and occurrent from the remaining vertices. The equation is similar for the arcs in equation (27). The final equation for the definition of arc existence function is,

$$E_{\varepsilon}(\varepsilon_{ab},\varepsilon_{cd})=1 \wedge O_{\varepsilon}(\varepsilon_{ab},\varepsilon_{cd})=1 \Leftrightarrow \Pr(\beta_j=\Phi)=0 \tag{33}$$

Equations (30) to (33) show the second-order effects of purely first-order events. For example, a single graph element is always absent or present for extended null elements and strict non-null elements, respectively, in the FDG. These effects should be taken into account when using second-order constraints to match AGs with FDGs (see Sections 5.1 and 5.2).

#### 4.3.4. Symmetry of second-order functions and the co-occurrence relation

The antagonism and existence relations are symmetric as can be seen directly from their definitions: that is $A_{\omega}(\omega_i,\omega_j)=A_{\omega}(\omega_j,\omega_i)$, $A_{\varepsilon}(\varepsilon_{ab},\varepsilon_{cd})=A_{\varepsilon}(\varepsilon_{cd},\varepsilon_{ab})$, $E_{\omega}(\omega_i,\omega_j)=E_{\omega}(\omega_j,\omega_i)$ and $E_{\varepsilon}(\varepsilon_{ab},\varepsilon_{cd})=E_{\varepsilon}(\varepsilon_{cd},\varepsilon_{ab})$.

The occurrence relation is clearly non-symmetric, but in some cases another symmetric relation, the *co-occurrence* of nodes or arcs, may be of interest. A Boolean co-occurrence function of pairs of vertices can be defined in terms of the corresponding occurrence functions as

$$C_{\omega}(\omega_i,\omega_j)=1 \Leftrightarrow O_{\omega}(\omega_i,\omega_j)=1 \wedge O_{\omega}(\omega_j,\omega_i)=1 \tag{34}$$

Similarly, we can define the co-occurrence function of pairs of arcs as

$$C_{\varepsilon}(\varepsilon_{ab},\varepsilon_{cd})=1 \Leftrightarrow O_{\varepsilon}(\varepsilon_{ab},\varepsilon_{cd})=1 \wedge O_{\varepsilon}(\varepsilon_{cd},\varepsilon_{ab})=1 \tag{35}$$

It is readily proved that these co-occurrence relations are also reflexive and transitive and therefore they are equivalence relations.

### 4.4. Synthesis of FDGs from AGs with a common labelling

Let $D = \{G^1, \ldots, G^z\}$ be a set of AGs defined over a common attribute domain $(\Delta_v, \Delta_e)$. Let $G^g = (V^g, A^g)$ where $V^g = (\Sigma_v^g, \gamma_v^g)$ and $A^g = (\Sigma_e^g, \gamma_e^g)$, for $1 \le g \le z$. Assume that there are given labelling schemes $\Psi^g = (\Psi_v^g : \Sigma_v^g \to L_v, \Psi_e^g : \Sigma_e^g \to L_e)$, $g = 1, \ldots, z$, where $\Psi_v^g$ is an injective mapping from the underlying structural vertex set of $G^g$ to a common set of vertex labels $L_v = \{1, \ldots, n\}$ and $\Psi_e^g$ similarly labels arcs with labels from $L_e = \{1, \ldots, n(n-1)\}$. The labelling schemes $\Psi^g$ can be extended to bijective mappings $\Psi'^g = (\Psi'^g_v : \Sigma'^g_v \to L_v, \Psi'^g_e : \Sigma'^g_e \to L_e)$, $g = 1, \ldots, z$, if each AG $G^g$ is previously extended to a complete graph $G'^g$ of order *n*. The arc labellings are also assumed to be consistent across all graphs in *D*, i.e. the arc from the vertex labelled *k* to the vertex labelled *l* has the same label $j = arc_number(k, l, n)$, $j \in L_e$, in all graphs. For instance, let the function $arc_number(k, l, n)$ be defined as follows:

$$arc_number(k,l,n) = \begin{cases} (k-1)(n-1)+l & \text{if } l < k \\ (k-1)(n-1)+l-1 & \text{if } l > k \end{cases}$$

Under this assumption, it is important to note that arc labellings $\Psi'^g_e$ are merely introduced for notational convenience, since all the information required is contained in the node labellings $\Psi'^g_v$.

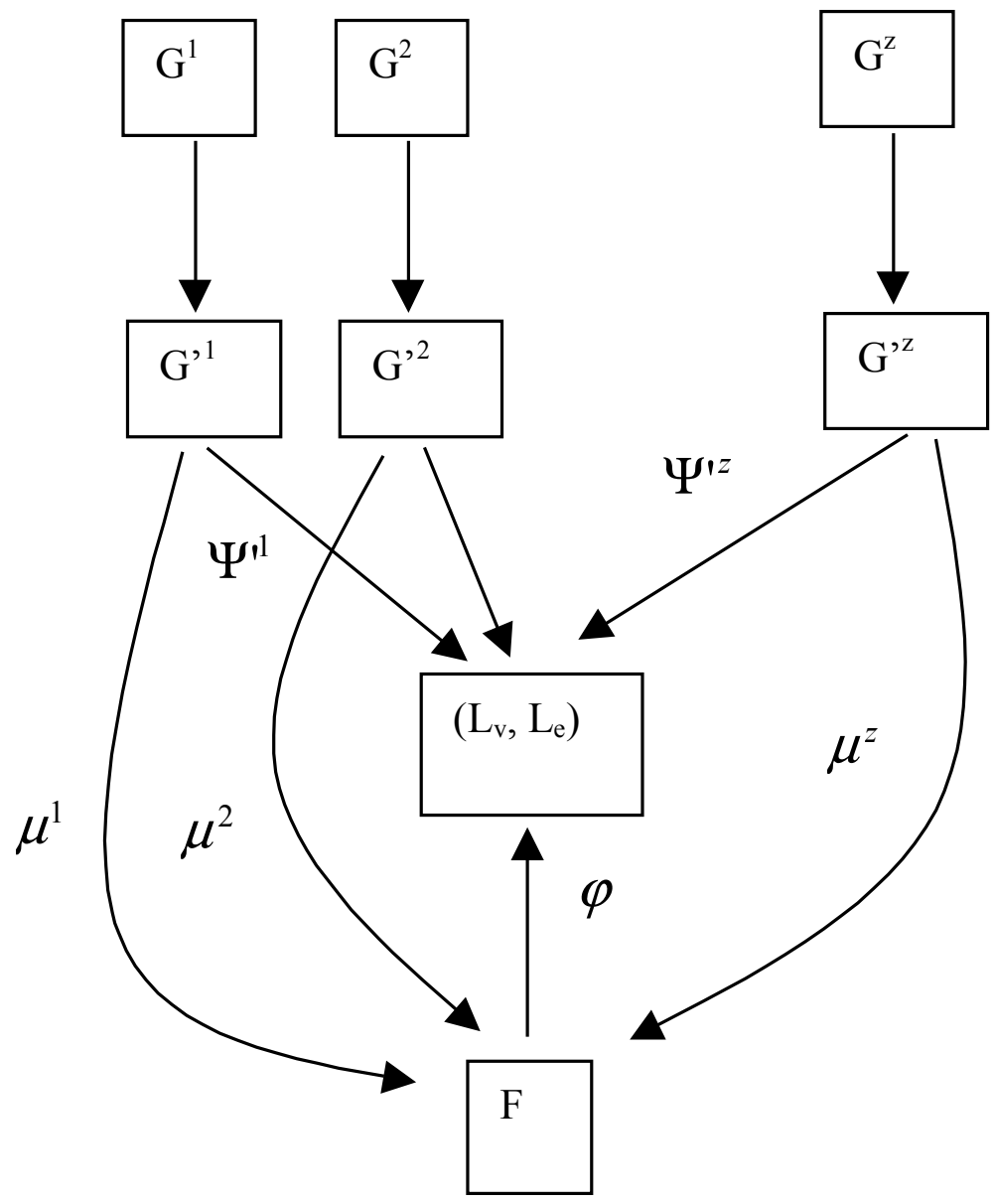

$$v_i^g = \mu_v^{g^{-1}}(\omega_i)$$
$$e_{j_1 j_2}^g = \mu_e^{g^{-1}}(\varepsilon_{j_1 j_2})$$

Figure 9. Synthesis of FDGs from AGs with a common labelling.

An FDG $F = (W, B, P, R)$ over $(\Delta_v, \Delta_e)$ can be synthesised from $D$ and $\Psi'$ in a straightforward manner (figure 9). $F$ includes a complete underlying graph structure $H = (\Sigma_\omega, \Sigma_\varepsilon)$ with a set of $n$ vertices $\Sigma_\omega = \{\omega_1, \ldots, \omega_n\}$ and a set of $n(n-1)$ arcs $\Sigma_\varepsilon = \{\varepsilon_{kl} \mid 1 \le k, l \le n,\ k \ne l\}$. The random vertex set $W = (\Sigma_\omega, \gamma_\omega)$ associates each vertex $\omega_i \in \Sigma_\omega$ with a random variable $\alpha_i = \gamma_\omega(\omega_i)$ with values in $\Delta_\omega = \Delta_v \cup \{\Phi\}$ and the random arc set $B = (\Sigma_\varepsilon, \gamma_\varepsilon)$ associates each arc $\varepsilon_{kl} \in \Sigma_\varepsilon$ with a random variable $\beta_j = \gamma_\varepsilon(\varepsilon_{kl})$ with values in $\Delta_\varepsilon = \Delta_e \cup \{\Phi\}$, where $j = arc_number(k, l, n)$.

Now, let $\varphi = (\varphi_\omega : \Sigma_\omega \to L_v,\ \varphi_\varepsilon : \Sigma_\varepsilon \to L_e)$ be a labelling scheme on $F$ defined simply by $\varphi_\omega(\omega_i) = i$ and $\varphi_\varepsilon(\varepsilon_{kl}) = arc_number(k, l, n)$. From labellings $\Psi'$ and $\varphi$, we can determine a set of bijective mappings $\{\ \mu^g = (\mu_v^g : \Sigma'^g_v \to \Sigma_\omega,\ \mu_e^g : \Sigma'^g_e \to \Sigma_\varepsilon),\ \ 1 \le g \le z\ \}$ from the AGs in $D$ to the synthesised FDG such that $\Psi'^g_v = \varphi_\omega \circ \mu_v^g$ and $\Psi'^g_e = \varphi_\varepsilon \circ \mu_e^g$, for $g = 1, \ldots, z$.

The probability density functions, $P = (P_\omega, P_\varepsilon)$, $P_\omega = \{p_i(\mathbf{a}),\ i = 1, \ldots, n\}$ of individual random vertices and $P_\varepsilon = \{q_j(\mathbf{b}),\ j = 1, \ldots, n(n-1)\}$ of individual random arcs (given non-null endpoints), can be estimated separately, in the maximum likelihood sense, using frequencies of attributes and null values in *D*. Let $v_i^g = \mu_v^{g^{-1}}(\omega_i)$ and $e_{j_1 j_2}^g = \mu_e^{g^{-1}}(\varepsilon_{j_1 j_2})$ be, respectively, the node labelled *i* and the edge labelled *j* in the attributed graph $G'^g$. Then,

$$p_i(\mathbf{a}) = \Pr(\alpha_i = \mathbf{a}) = \frac{\# g:\ 1 \le g \le z:\ \gamma_v^g(v_i^g) = \mathbf{a}}{z} \tag{36}$$

for all possible values $\mathbf{a} \in \Delta_\omega$ of $\alpha_i$, including $\Phi$, and

$$\begin{aligned} q_j(\mathbf{b}) &= \Pr(\beta_j = \mathbf{b} \mid \alpha_{j_1} \neq \Phi \wedge \alpha_{j_2} \neq \Phi) = \\ &= \frac{\# g:\ 1 \le g \le z:\ \gamma_v^g(v_{j_1}^g) \neq \Phi \wedge \gamma_v^g(v_{j_2}^g) \neq \Phi \wedge \gamma_e^g(e_{j_1 j_2}^g) = \mathbf{b}}{u_j} \end{aligned} \tag{37}$$

where $u_j$ is the number of AGs, which has both vertices $v_{j1}^g$ and $v_{j2}^g$.

$$u_j = \# g:\ 1 \le g \le z:\ \gamma_{vg}(v_{j_1}^g) \neq \Phi \wedge \gamma_{vg}(v_{j_2}^g) \neq \Phi \tag{38}$$

The binary relations $R = (A_\omega, A_\varepsilon, O_\omega, O_\varepsilon, E_\omega, E_\varepsilon)$ in the synthesised FDG can be computed as follows. The *vertex antagonism function* $A_\omega$ and the *arc antagonism function* $A_\varepsilon$ are given by

$$A_\omega(\omega_i, \omega_j) = \begin{cases} 1 & \text{if } \forall g:\ 1 \le g \le z:\ \neg\left(\gamma_{vg}(v_i^g) \neq \Phi \wedge \gamma_{vg}(v_j^g) \neq \Phi\right) \\ 0 & \text{otherwise} \end{cases} \tag{39}$$

$$A_\varepsilon(\varepsilon_{i_1 i_2}, \varepsilon_{j_1 j_2}) = \begin{cases} 1 & \text{if } \forall g:\ 1 \le g \le z:\ \neg\left(\gamma_{eg}(e_{i_1 i_2}^g) \neq \Phi \wedge \gamma_{eg}(e_{j_1 j_2}^g) \neq \Phi\right) \\ 0 & \text{otherwise} \end{cases} \tag{40}$$

The *vertex occurrence function* $O_\omega$ and the *arc occurrence function* $O_\varepsilon$ are given by

$$O_\omega(\omega_i,\omega_j)=\begin{cases}1 & \text{if } \forall g: 1\le g\le z: \neg\left(\gamma_{vg}(v_i^g)\ne\Phi \wedge \gamma_{vg}(v_j^g)=\Phi\right)\\ 0 & \text{otherwise}\end{cases} \tag{41}$$

$$O_\varepsilon(\varepsilon_{i_1i_2},\varepsilon_{j_1j_2})=\begin{cases}1 & \text{if } \forall g: 1\le g\le z: \neg\left(\gamma_{eg}(e_{i_1i_2}^g)\ne\Phi \wedge \gamma_{eg}(e_{j_1j_2}^g)=\Phi\right)\\ 0 & \text{otherwise}\end{cases} \tag{42}$$

And, finally, the *vertex existence function* $E_\omega$ and the *arc existence function* $E_\varepsilon$ are given by

$$E_\omega(\omega_i,\omega_j)=\begin{cases}1 & \text{if } \forall g: 1\le g\le z: \neg\left(\gamma_{vg}(v_i^g)=\Phi \wedge \gamma_{vg}(v_j^g)=\Phi\right)\\ 0 & \text{otherwise}\end{cases} \tag{43}$$

$$E_\varepsilon(\varepsilon_{i_1i_2},\varepsilon_{j_1j_2})=\begin{cases}1 & \text{if } \forall g: 1\le g\le z: \neg\left(\gamma_{eg}(e_{i_1i_2}^g)=\Phi \wedge \gamma_{eg}(e_{j_1j_2}^g)=\Phi\right)\\ 0 & \text{otherwise}\end{cases} \tag{44}$$

The *vertex co-occurrence function* $C_\omega$ and the *arc co-occurrence function* $C_\varepsilon$ may always be evaluated as $C_\omega(\omega_i,\omega_j)=O_\omega(\omega_i,\omega_j)\wedge O_\omega(\omega_j,\omega_i)$ and $C_\varepsilon(\varepsilon_{i_1i_2},\varepsilon_{j_1j_2})=$ $O_\varepsilon(\varepsilon_{i_1i_2},\varepsilon_{j_1j_2})\wedge O_\varepsilon(\varepsilon_{j_1j_2},\varepsilon_{i_1i_2})$, respectively.

### 4.5. Synthesis of FDGs from FDGs with a common labelling

Let $D=\{F^1,\ldots,F^h\}$ be a set of FDGs independently synthesised from disjoint subsets of a class of AGs with common homogenous domains for attributed vertices and arcs. Let $F^k=(W^k,B^k,P^k,R^k)$, for $1\le k\le h$, defined over a common attribute domain $(\Delta_v,\Delta_e)$. For each FDG $F^k$, the number of AGs from which it is formed, $z^k$, is stored. For each random arc $\beta_j^k$ of $F^k$, the number of these AGs $u_j^k$, which have both connecting vertices is also stored (See equation 38).

Assume that there are given labelling schemes $\Psi^k=(\Psi_\omega^k:\Sigma_\omega^k\to L_\omega, \Psi_\varepsilon^k:\Sigma_\varepsilon^k\to L_\varepsilon)$, $k=1,\ldots,h$, mapping the vertices and arcs of the FDGs $F^k$ into common label sets $L_\omega = \{1,\dots,n\}$ and $L_\varepsilon = \{1,\dots,n(n-1)\}$, such that all $\Psi_\omega^k$ and $\Psi_\varepsilon^k$ are injective and all arc labellings are consistent throughout the set $D$. If the order of some FDG $F^k$ is less than $n$, then $F^k$ can be extended to an isomorphic complete FDG $F'^k$ of order $n$ by adding null vertices and arcs. Therefore, the labelling schemes $\Psi^k$ can be extended to bijective mappings $\Psi'^k = \left(\Psi'^k_\omega : \Sigma'^k_\omega \to L_\omega, \Psi'^k_\varepsilon : \Sigma'^k_\varepsilon \to L_\varepsilon\right)$, $k = 1,\dots,h$, whenever each FDG $F^k$ is previously extended to a complete FDG $F'^k$ of order *n*.

As in the synthesis of FDGs from AGs, the arc labellings are also assumed to be consistent across all FDGs in *D*, i.e. the arc from the vertex labelled *t* to the vertex labelled *l* has the same label $j = arc_number(t,l,n)$, $j \in L_\varepsilon$, in all FDGs.

An FDG $F = (W,B,P,R)$ over $(\Delta_v, \Delta_e)$ can be synthesised from *D* and the common labelling $\Psi'$ as follows. $F$ includes a complete underlying graph structure $H = (\Sigma_\omega, \Sigma_\varepsilon)$ with a set of *n* vertices $\Sigma_\omega = \{\omega_1,\dots,\omega_n\}$ and a set of $n(n-1)$ arcs $\Sigma_\varepsilon = \{\varepsilon_{tl} \mid 1 \le t,l \le n,\ t \ne l\}$. The random vertex set $W = (\Sigma_\omega, \gamma_\omega)$ associates each vertex $\omega_i \in \Sigma_\omega$ with a random variable $\alpha_i = \gamma_\omega(\omega_i)$ with values in $\Delta_\omega = \Delta_v \cup \{\Phi\}$ and the random arc set $B = (\Sigma_\varepsilon, \gamma_\varepsilon)$ associates each arc $\varepsilon_{tl} \in \Sigma_\varepsilon$ with a random variable $\beta_j = \gamma_\varepsilon(\varepsilon_{tl})$ with values in $\Delta_\varepsilon = \Delta_e \cup \{\Phi\}$, where $j = arc_number(t,l,n)$.

Now, let $\varphi = (\varphi_\omega : \Sigma_\omega \to L_\omega, \varphi_\varepsilon : \Sigma_\varepsilon \to L_\varepsilon)$ be a labelling scheme on $F$ defined simply by $\varphi_\omega(\omega_i) = i$ and $\varphi_\varepsilon(\varepsilon_{tl}) = arc_number(t,l,n)$. From labellings $\Psi'$ and $\varphi$, we can determine a set of bijective mappings $\{\mu^k = (\mu^k_\omega : \Sigma'^k_\omega \to \Sigma_\omega, \mu^k_\varepsilon : \Sigma'^k_\varepsilon \to \Sigma_\varepsilon),\ 1 \le k \le h\}$ from the FDGs in $D$ to the synthesised FDG such that $\Psi'^k_\omega = \varphi_\omega \circ \mu^k_\omega$ and $\Psi'^k_\varepsilon = \varphi_\varepsilon \circ \mu^k_\varepsilon$, for $k = 1,\dots,h$. Let $\omega_i^k = \mu_\omega^{k^{-1}}(\omega_i)$ and $\varepsilon^k_{j_1 j_2} = \mu_\varepsilon^{k^{-1}}(\varepsilon_{j_1 j_2})$ be, respectively, the vertex labelled *i* and the edge labelled *j* in the FDG $F'^k$.

The probability density functions, $P = (P_\omega, P_\varepsilon)$, $P_\omega = \{p_i(\mathbf{a}),\ i = 1,\dots,n\}$ of individual random vertices and $P_\varepsilon = \{q_j(\mathbf{b}),\ j = 1,\dots,n(n-1)\}$ of individual random arcs (given non-null endpoints) can be estimated separately, using the corresponding probabilities in $P_{\omega}^{k}$ and $P_{\varepsilon}^{k}$ together with the values $z^k$ and $u_j^k$, $k = 1,...,h$.

Let $t^k$ be the normalised number of AGs used to synthesise the FDG $F^k$.

$$t^k = \frac{z^k}{\sum_{g=1}^{h} z^g}; 1 \le k \le h \tag{45}$$

And let $r_j^k$ be the normalised number of $u_j^k$ for each $F^k$ (See equation 38).

$$r_j^k = \frac{u_j^k}{\sum_{g=1}^{h} u_j^g}; 1 \le k \le h; 1 \le j \le n(n-1) \tag{46}$$

Then, for all possible values $\mathbf{a} \in \Delta_{\omega}$ of random vertex $\alpha_i$ including $\Phi$, we have that

$$p_i(\mathbf{a}) = \Pr(\alpha_i = \mathbf{a}) = \sum_{k=1}^{h} t^k * p_i^k(\mathbf{a}) \tag{47}$$

and for all possible values $\mathbf{b} \in \Delta_{\varepsilon}$ of $\beta_j$, including $\Phi$.

$$q_j(\mathbf{b}) = \Pr(\beta_j = \mathbf{b} \,|\, \alpha_{j_1} \ne \Phi \wedge \alpha_{j_2} \ne \Phi) = \sum_{k=1}^{h} r_j^k * q_j^k(\mathbf{b}) \tag{48}$$

The binary relations $R = (A_{\omega}, A_{\varepsilon}, O_{\omega}, O_{\varepsilon}, E_{\omega}, E_{\varepsilon})$ in the synthesised FDG are all readily calculated, since they are given by the logical *and* of the corresponding functions in the FDGs $F_k$. The *vertex antagonism function* $A_{\omega}$ and the *arc antagonism function* $A_{\varepsilon}$ are given by

$$A_{\omega}(\omega_i, \omega_j) = \bigwedge_{k=1}^{h} A_{\omega}^{k}(\omega_i^k, \omega_j^k) \tag{49}$$

$$A_\varepsilon\left(\varepsilon_{i_1 i_2},\varepsilon_{j_1 j_2}\right)=\overset{h}{\underset{k=1}{\Lambda}} A_\varepsilon^k\left(\varepsilon_{i_1 i}^k,\varepsilon_{j_1 j_2}^k\right) \tag{50}$$

The *vertex occurrence function* $O_\omega$ and the *arc occurrence function* $O_\varepsilon$ are given by

$$O_\omega\left(\omega_i,\omega_j\right)=\overset{h}{\underset{k=1}{\Lambda}} O_\omega^k\left(\omega_i^k,\omega_j^k\right) \tag{51}$$

$$O_\varepsilon\left(\varepsilon_{i_1 i_2},\varepsilon_{j_1 j_2}\right)=\overset{h}{\underset{k=1}{\Lambda}} O_\varepsilon^k\left(\varepsilon_{i_1 i}^k,\varepsilon_{j_1 j_2}^k\right) \tag{52}$$

And, finally, the *vertex existence function* $E_\omega$ and the *arc existence function* $E_\varepsilon$ are given by

$$E_\omega\left(\omega_i,\omega_j\right)=\overset{h}{\underset{k=1}{\Lambda}} E_\omega^k\left(\omega_i^k,\omega_j^k\right) \tag{53}$$

$$E_\varepsilon\left(\varepsilon_{i_1 i_2},\varepsilon_{j_1 j_2}\right)=\overset{h}{\underset{k=1}{\Lambda}} E_\varepsilon^k\left(\varepsilon_{i_1 i}^k,\varepsilon_{j_1 j_2}^k\right) \tag{54}$$

### 4.6. Experimental results

In order to examine the behaviour of the new representation, we performed a number of experiments with randomly generated AGs. The algorithms presented here were implemented in visual C++ and run on a Pentium II (350Mhz).

AGs were generated by a random graph generator process with the following parameters:

- *nFDG* : Number of models.
- *NT* : Number of AGs in the test set.
- *NR* : Number of AGs in the reference set for each model (FDG).
- *nv* : Number of vertices of the initial AGs.

- $ne$ : Number of arcs of the initial AGs.
- $nd$ : Number of deleted vertices of the initial AGs.
- $nl$ : Number of distorted vertices of the initial AGs.

We randomly generated $nFDG$ initial attributed graphs, one for each model, based on the parameters $nv$ and $ne$. From these graphs, the reference and test sets were derived in the following way (figure 10). For each initial AG, a reference set of $NR$ AGs was built by changing the attribute of $nd$ vertices to the null value (that is to say, they were deleted) and replacing the attribute of $nl$ vertices by another integer number (0 to 999). Also, for each initial AG, a test set of $NT/nFDG$ AGs was constructed in the same way. Thus, the whole test set was composed of $NT$ AGs and the whole reference set was composed of $nFDG$ subsets of $NR$ AGs. The FDGs were synthesised from the AGs in the corresponding reference sets using the method "FDG synthesis from AGs with a common labelling" (section 4.4).

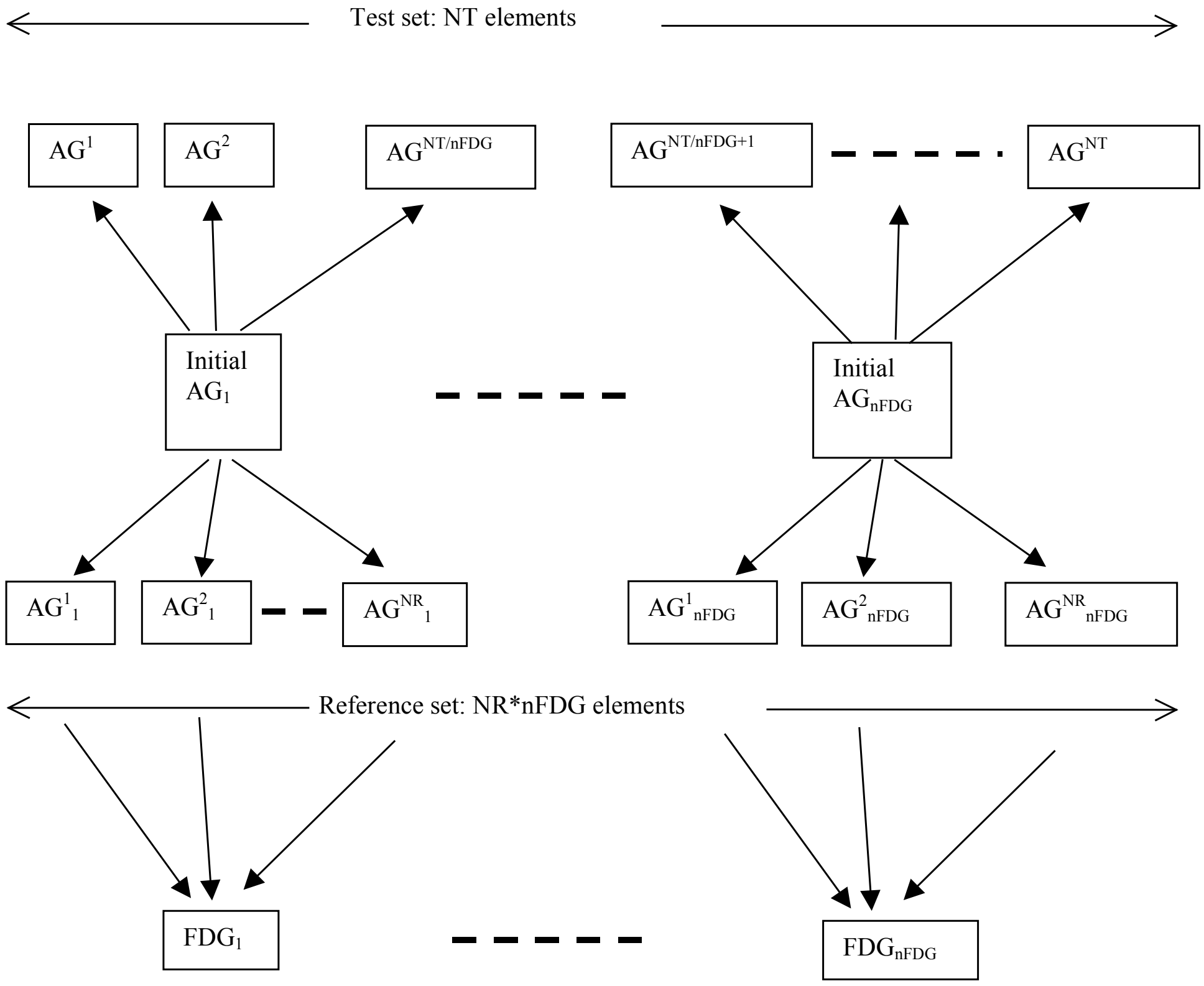

Figure 10. Random generation of the reference and test sets.

In this experiment, we were interested in the relation between the number of AGs used to synthesise the FDGs ($NR$), the number of vertices in these AGs ($nv$') and four features of the synthesised FDG structure (the number of vertices, antagonisms, existences and occurrences).

Table 7 gives an overview of the experiments and indicates the different parameter values of the graph generator, the figures where the results are shown and the features of the AGs generated. $nv$' is the number of vertices in the generated AGs and *Non-modified* denotes the number of vertices that have the same attribute value as in the initial graph. The number of FDGs was set to 1, $nFDG = 1$, and the number of AGs in the test set to 0, $NT = 0$ (there is no recognition process in the first experiment).

| Figures | Initial Graph | | | | Generated AGs | |
|---|---|---|---|---|---|---|
| | *nv* | *ne* | *nd* | *nl* | *nv'* | *Non-modified* |
| 11.a 13.a 15.a | 9 | 36 | 6,3 | 1,2 | 3,6 | 2,4 |
| 11.b 13.b 15.b | 18 | 72 | 15,12,9,6 | 1,2,3,4 | 3,6,9,12 | 2,4,6,8 |
| 11.c 13.c 15.c | 27 | 108 | 24,21,18,15,9,3 | 1,2,3,4,6,8 | 3,6,9,12,18,24 | 2,4,6,8,12,16 |
| 11.d 13.d 15.d | 36 | 144 | 33,30,27,24,18,12 | 1,2,3,4,6,8 | 3,6,9,12,18,24 | 2,4,6,8,12,16 |

Table 7. Parameters of the first experiments.

Figure 11 shows the number of vertices of the synthesised FDG when the number of vertices of the AGs ($nv$') and the number of AGs in the reference set ($NR$) are varied. Figures 11.a to 11.d show the results from different number of vertices of the initial graphs, $nv$. Figure 11.e shows three different combinations of the values $nv$ and $nv$'. The number of vertices of the FDG increases while the number of AGs remains small until it reaches the correct value. Furthermore, the bigger the AGs are, the faster the number of FDG vertices reaches the maximum value. When $NR = 1$ the number of FDG vertices is the same than the number of AG vertices, $nv$'.

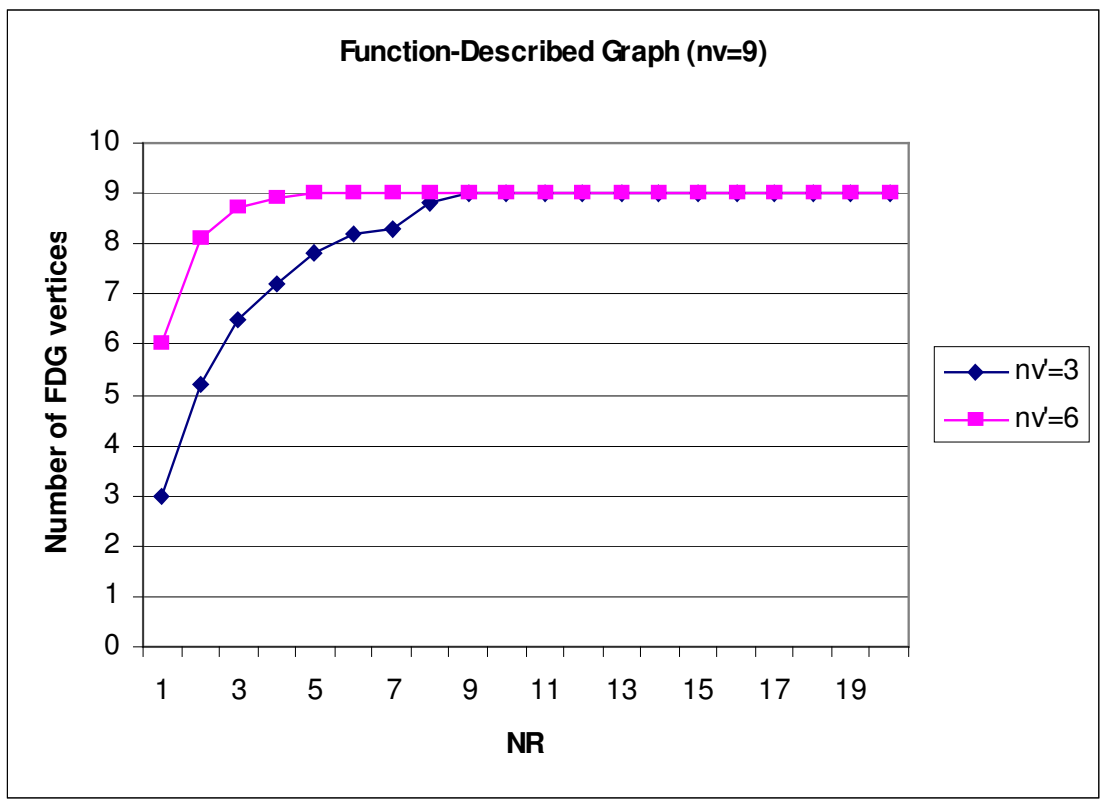

(a) nv=9

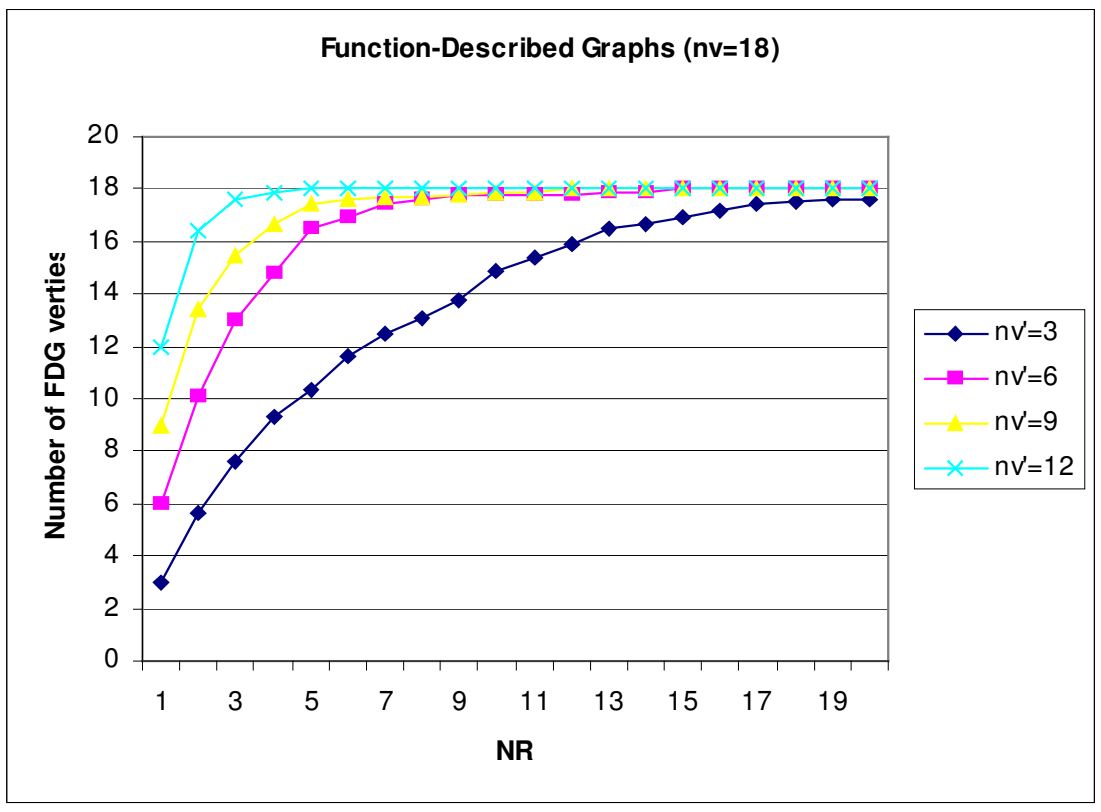

(b) nv=18

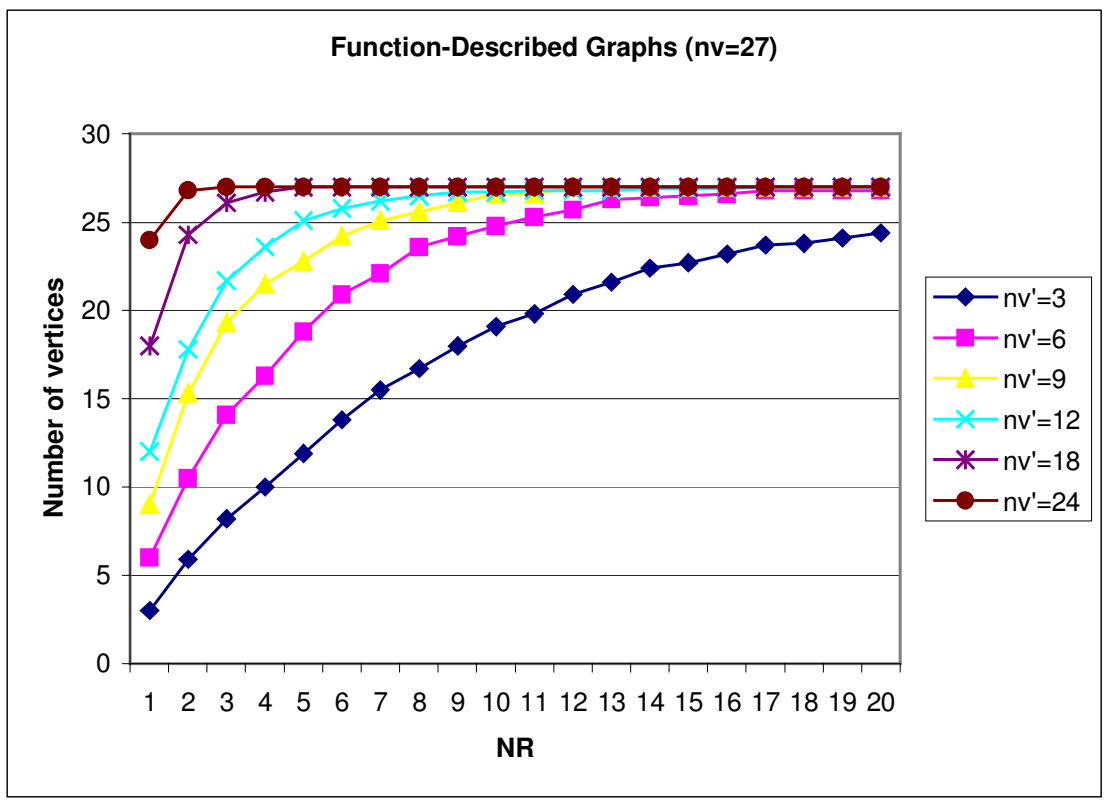

(c) nv=27

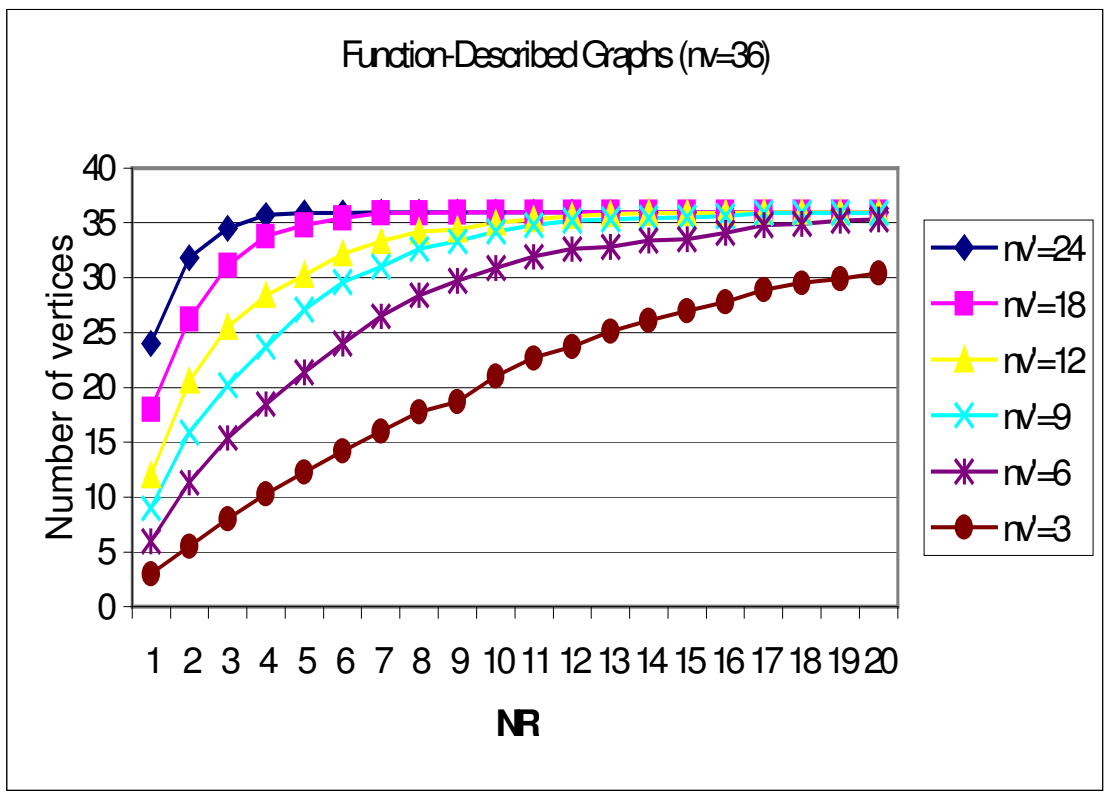

(d) nv=36

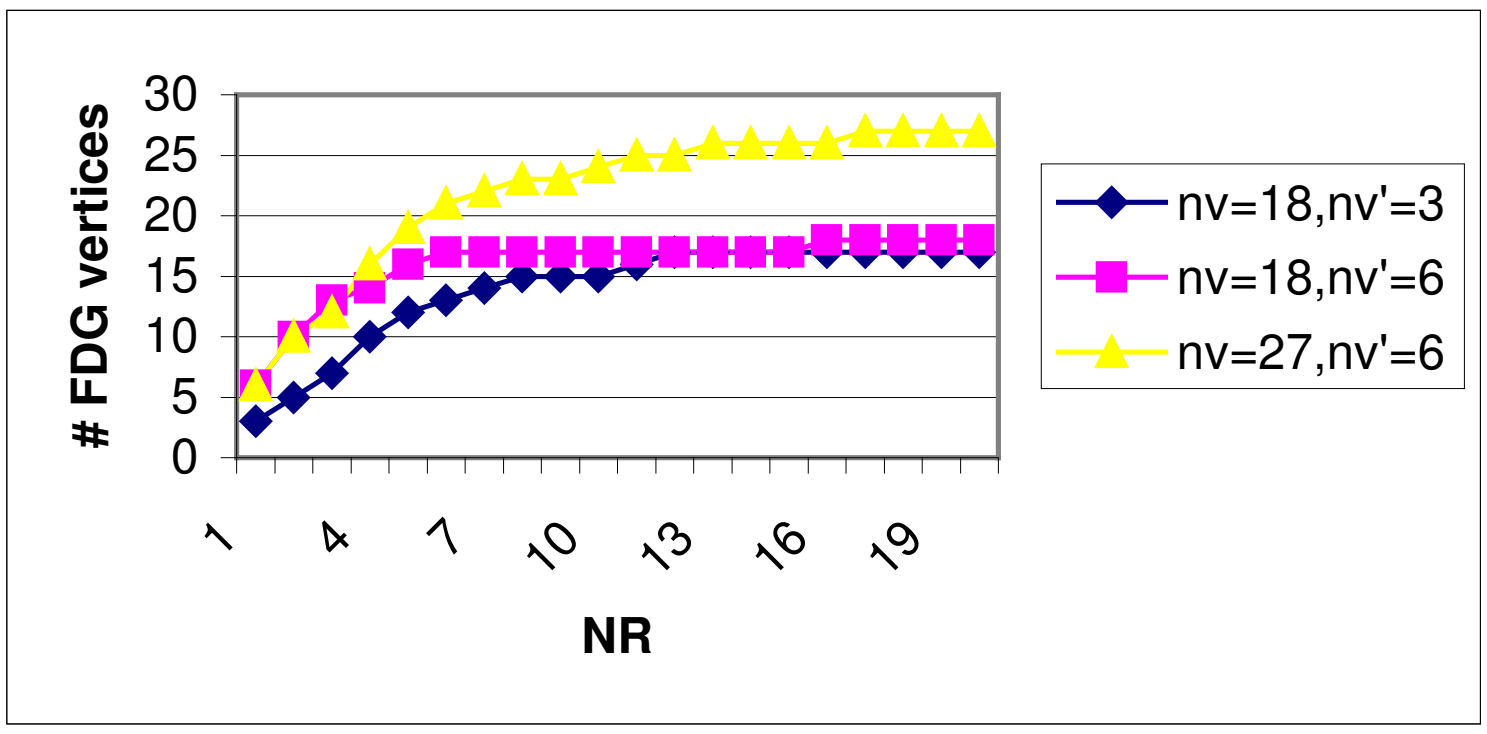

(e) nv=18, 27 and nv'=3, 6

Figure 11. Number of vertices of the FDG.

Figure 11.d shows the number of vertices when the initial graph have 36 vertices. In the cases that the AGs are very small $nv'=3$ and $nv'=6$, the maximum value is not reached which means that the FDG is not totally covered by the AGs. For instance, in the case that the number of AG vertices is 3, and 20 AGs are used to synthesise the FDG, only 30 vertices are generated instead of 36.

Figure 12 shows the number of antagonisms in the synthesised FDG. We observe that the maximum value of the number of antagonisms increases when the number of vertices in the AGs ($nv'$) decreases. Moreover, for a given $nv'$, the number of antagonisms is zero when the FDG has been synthesised with only one AG, that is $NR = 1$. This is because, the whole vertices of the FDG appear in the same AG and so there are not antagonistic. Furthermore, the number of antagonisms increases when $NR$ increases until reaching a maximum value and then it decreases monotonically if $NR$ is further increased. In all the tests, the maximum number of antagonisms is reached once the maximum number of FDG vertices ($nv$) is reached (see figure 11). This is due to the fact that when new vertices are added to the FDG structure new antagonisms between them and the other vertices are added. Whereas, for a fixed graph structure (the number of FDG vertices is not modified), second-order relations can only be removed but not added when more outcome AGs are introduced in the reference set or update the FDG.

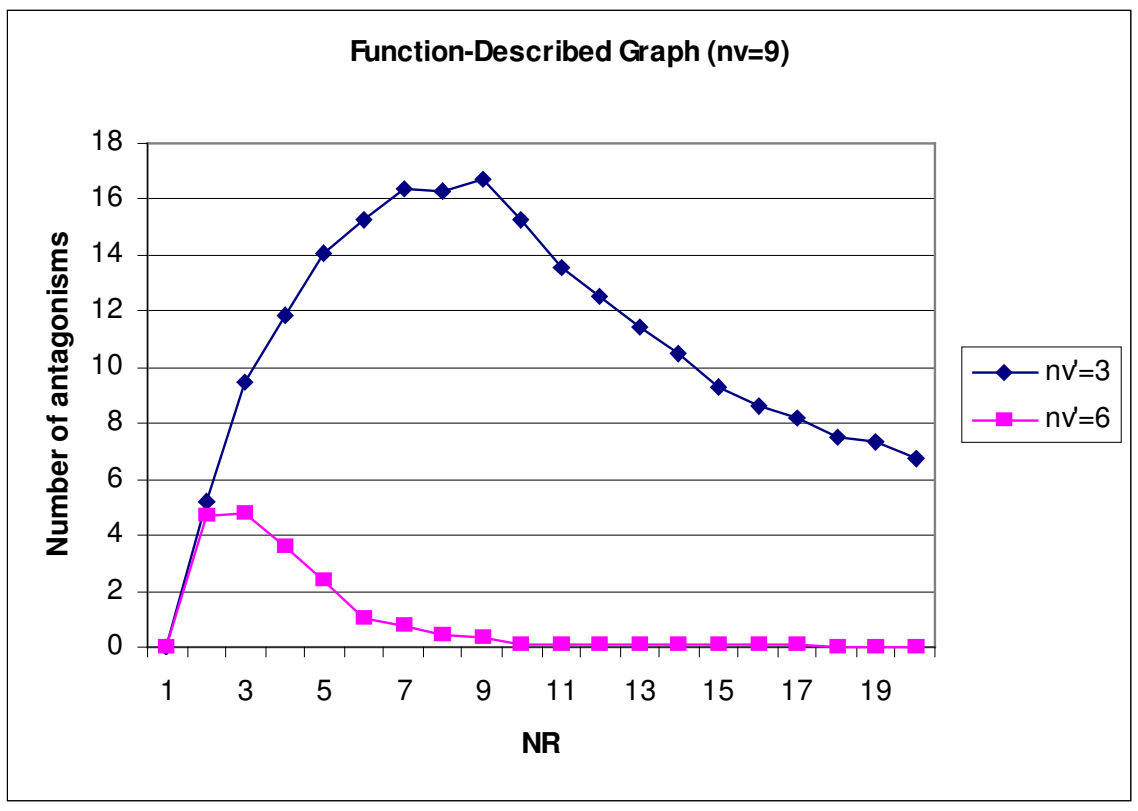

(a) nv=9

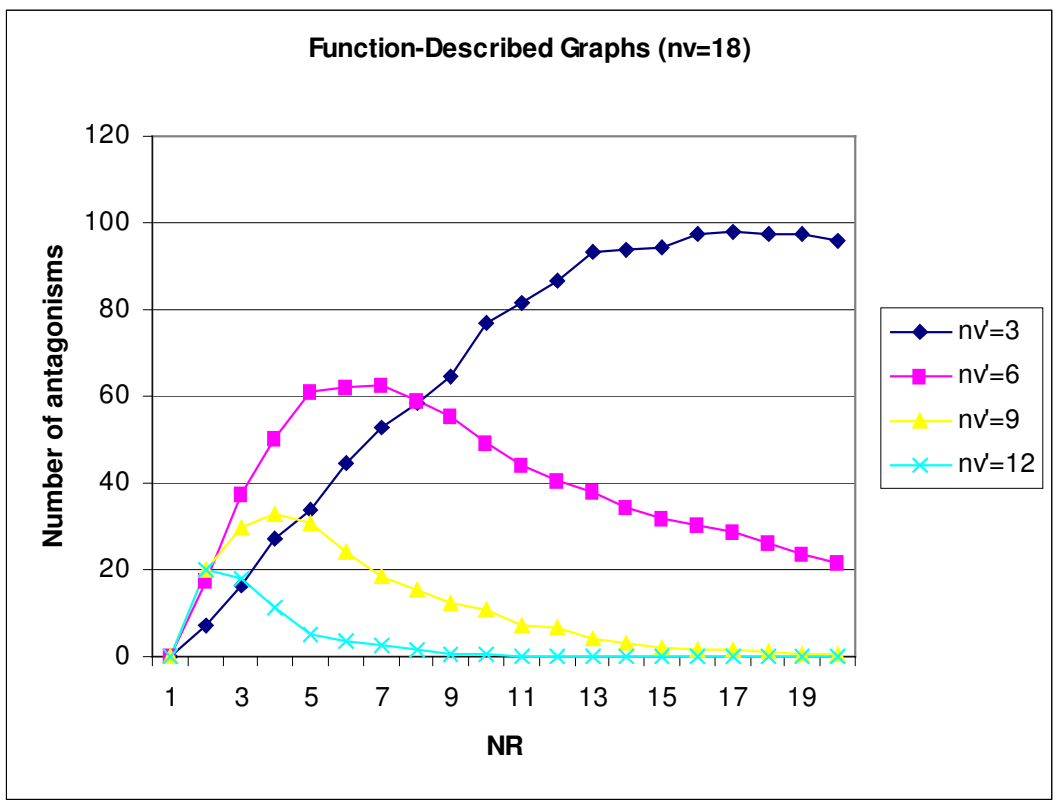

(b) nv=18

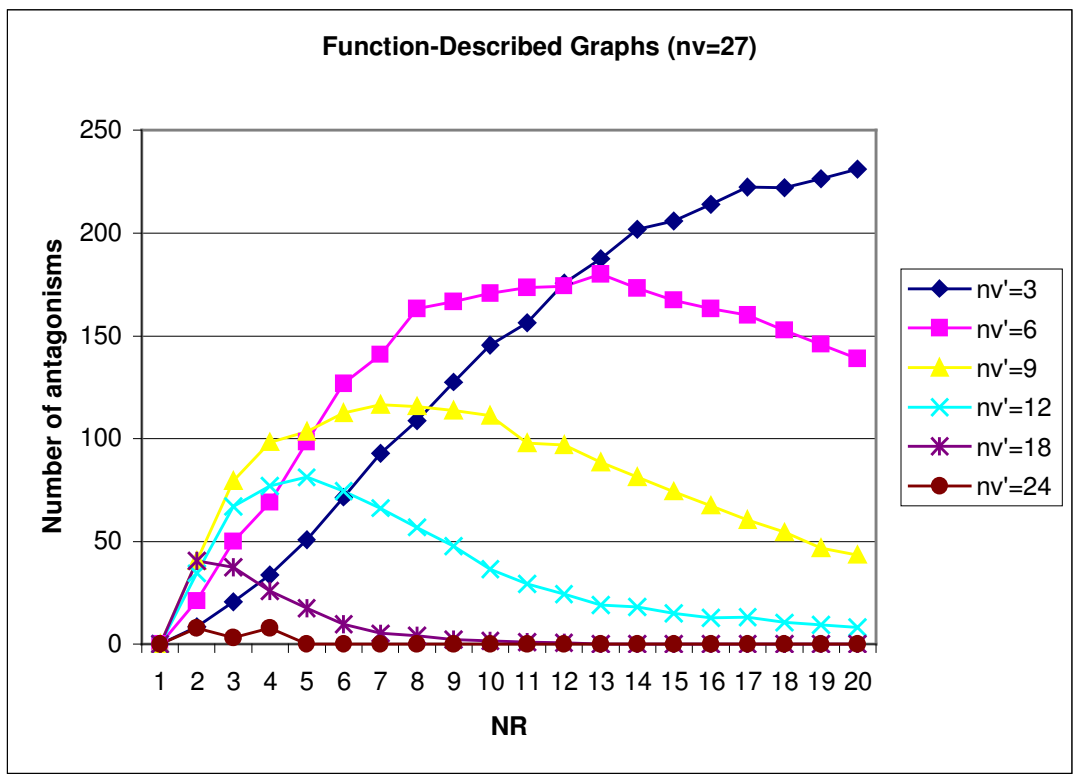

(c) nv=27

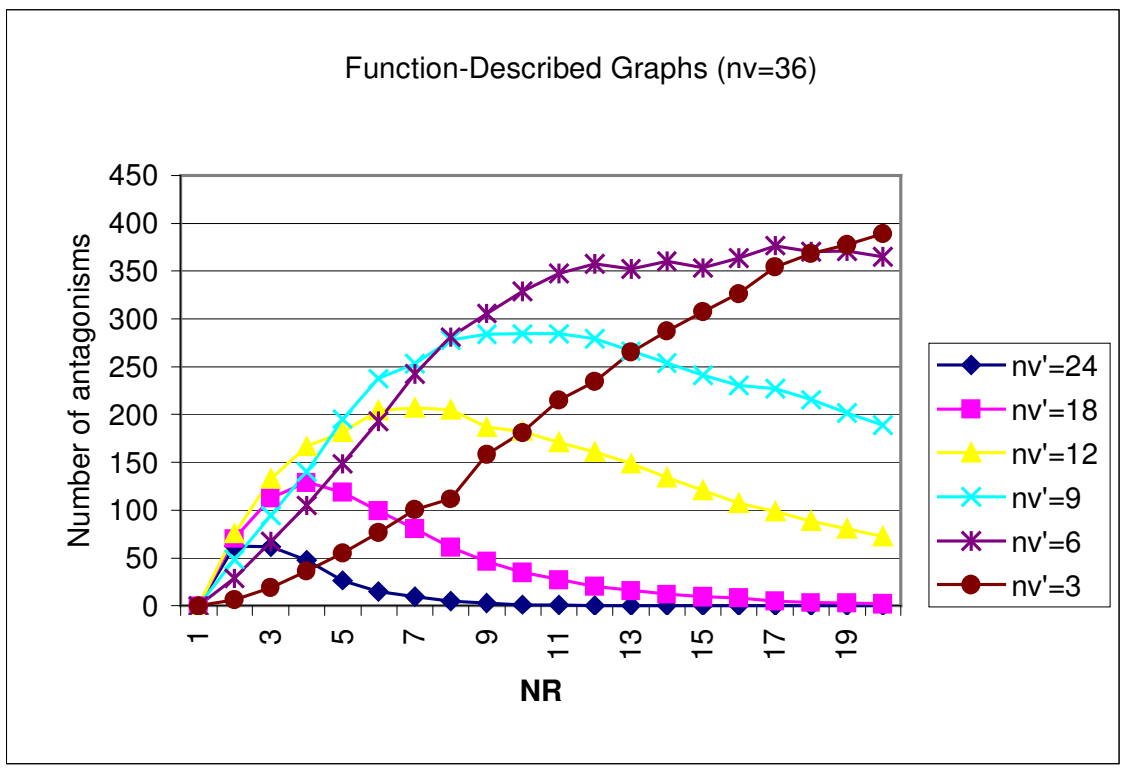

(d) nv=36

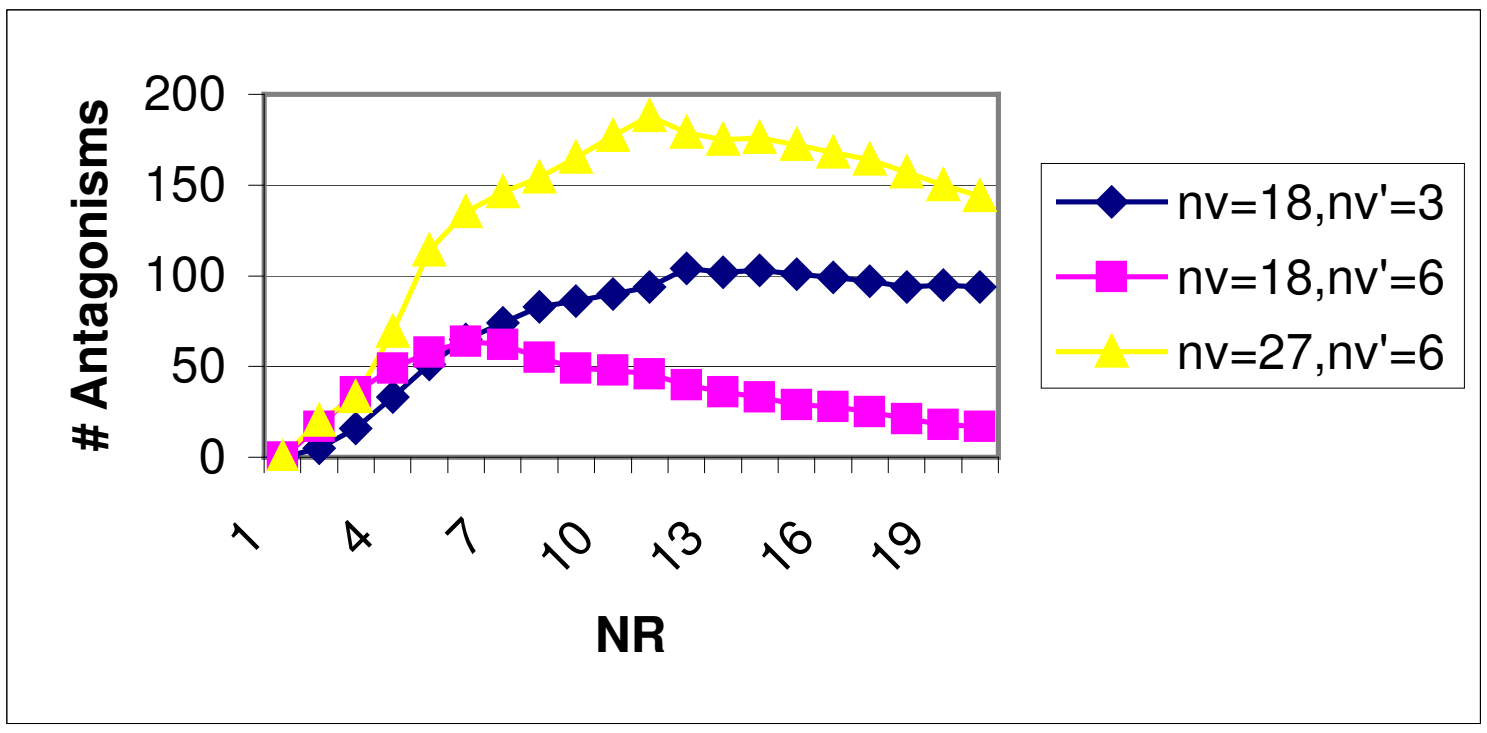

(e) nv=18, 27 and nv'=3, 6

Figure 12. Number of Antagonisms.

Figure 12.d shows the case in which $nv = 36$. In the cases that $nv'$ is greater than 6, the maximum value is reached when the number of vertices also reaches the maximum value (see figure 11). In the cases $nv' = 3$ and $nv' = 6$, the maximum number of vertices is not reached and neither the maximum number of antagonisms.

Figure 13 shows the number of occurrences of the synthesised FDG. Unlike the antagonism function, for a fixed $NR$, the number of occurrences increases when the number of vertices in the AGs ($nv'$) also increases. Bigger are the AGs, more occurrence functions appear between the FDG vertices. When the number of vertices of the AGs is very small the FDG is partially covered by “small pieces” and so, there are few occurrences. And also, when $NR = 1$ the whole FDG vertices are mutually occurrent. As in the antagonism case, the number of occurrences increases when $NR$ increases until reaching a maximum value and then decreases when $NR$ is further increased. The explanation of these results is similar than of the results on the antagonisms. Nevertheless, in the occurrence case, the maximum value is reached before the number of FDG vertices reaches the maximum value.

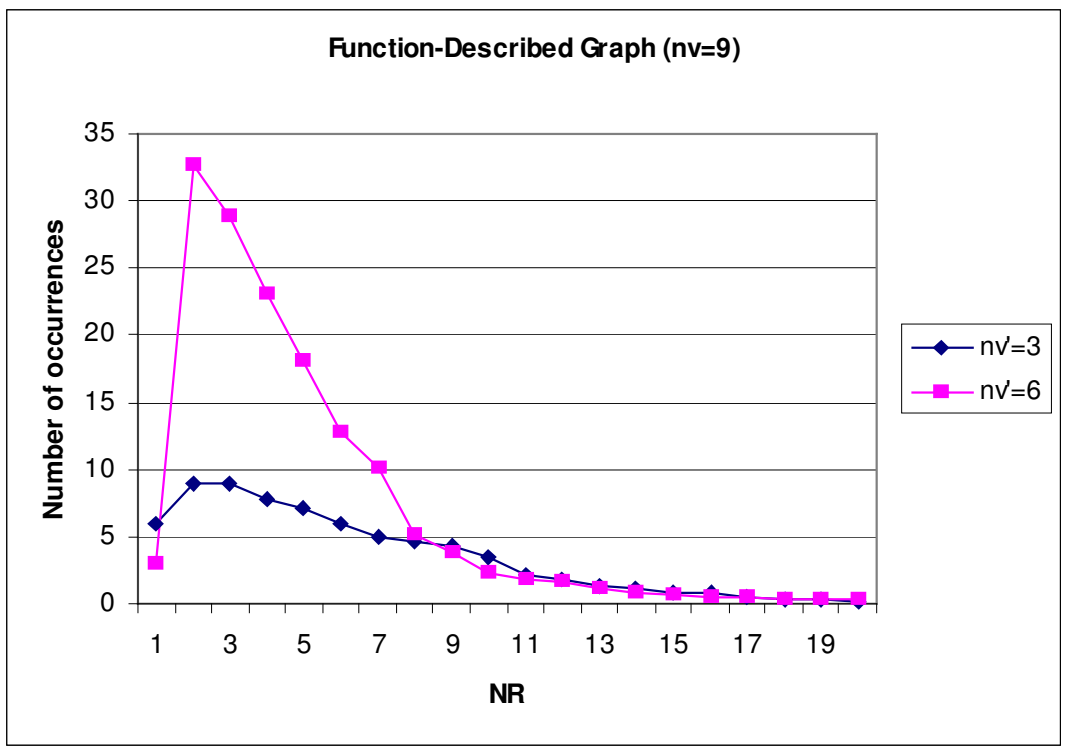

(a) nv=9

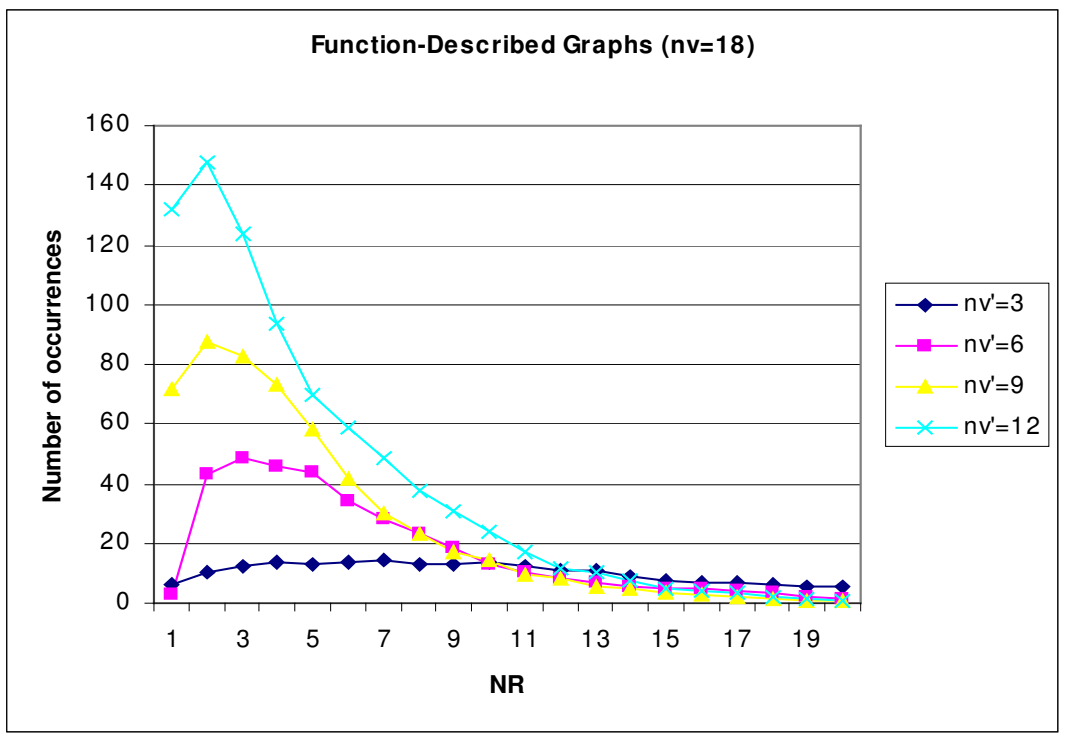

(b) nv=18

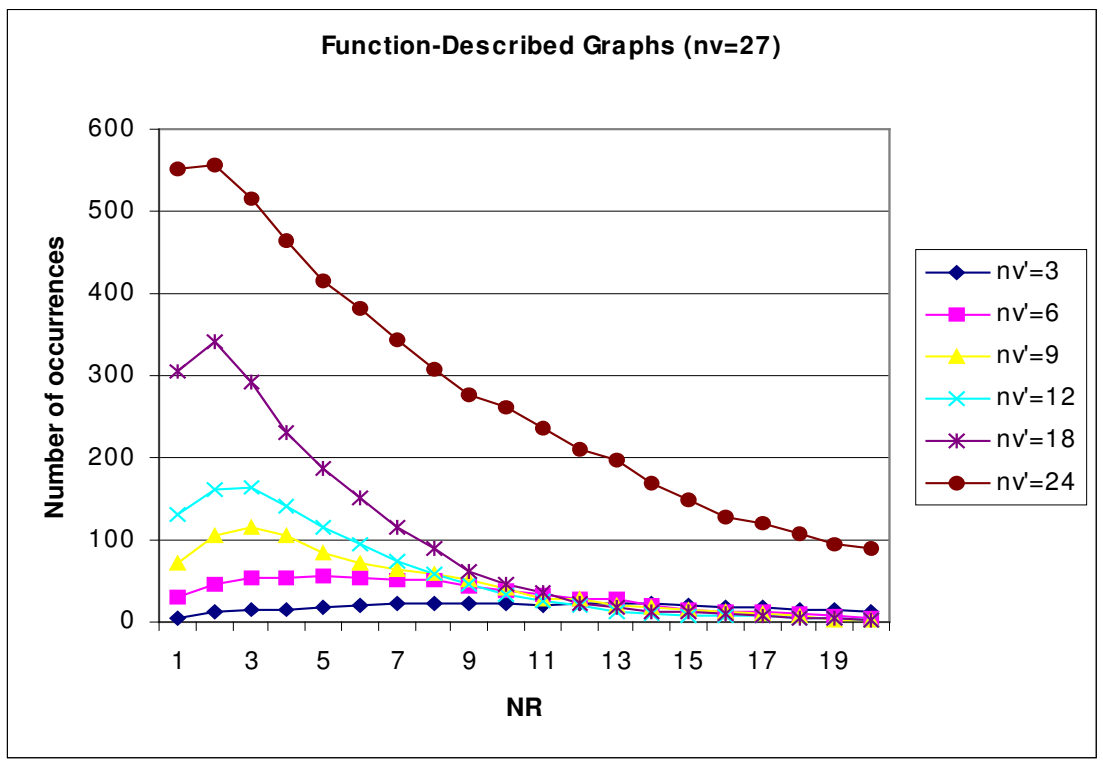

(c) nv=27

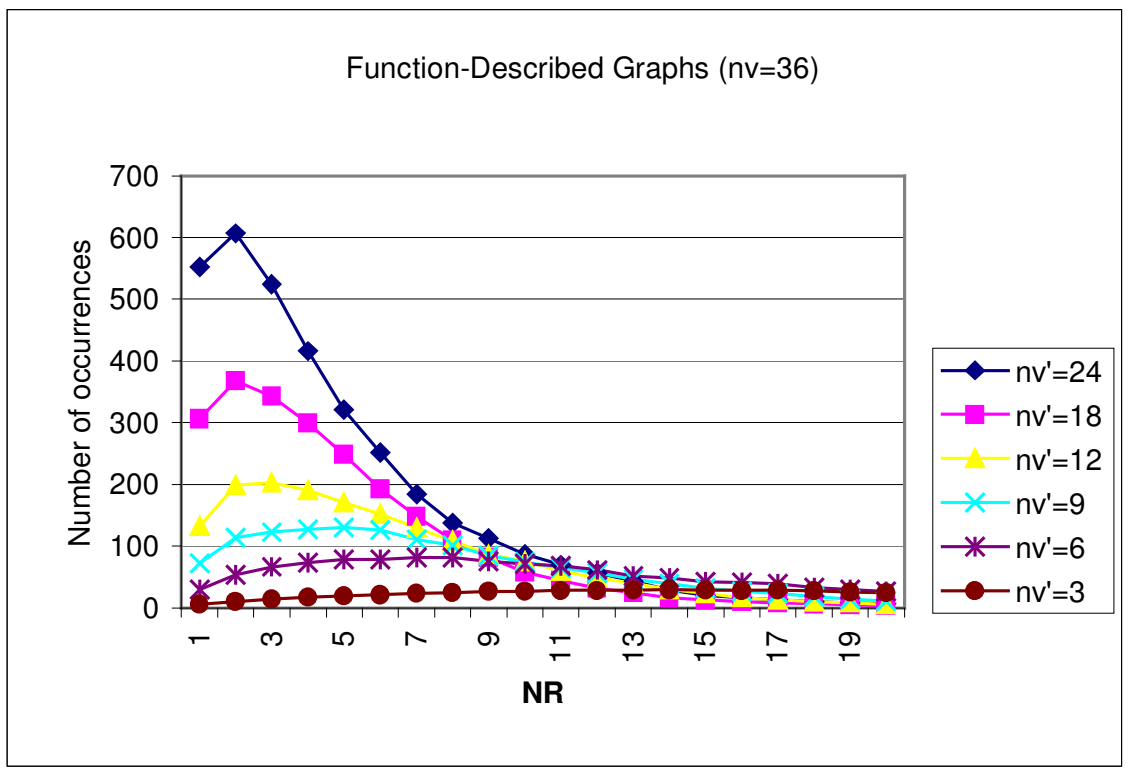

(d) nv=36

Figure 13. Number of Occurrences.

Figure 14 shows the number of existences of the synthesised FDG. The average of the number of existences is a monotonically decreasing function due to when new AGs are included the existence relations only can be removed. As in the occurrence case, when $NR = 1$ the whole FDG vertices have existence relations.

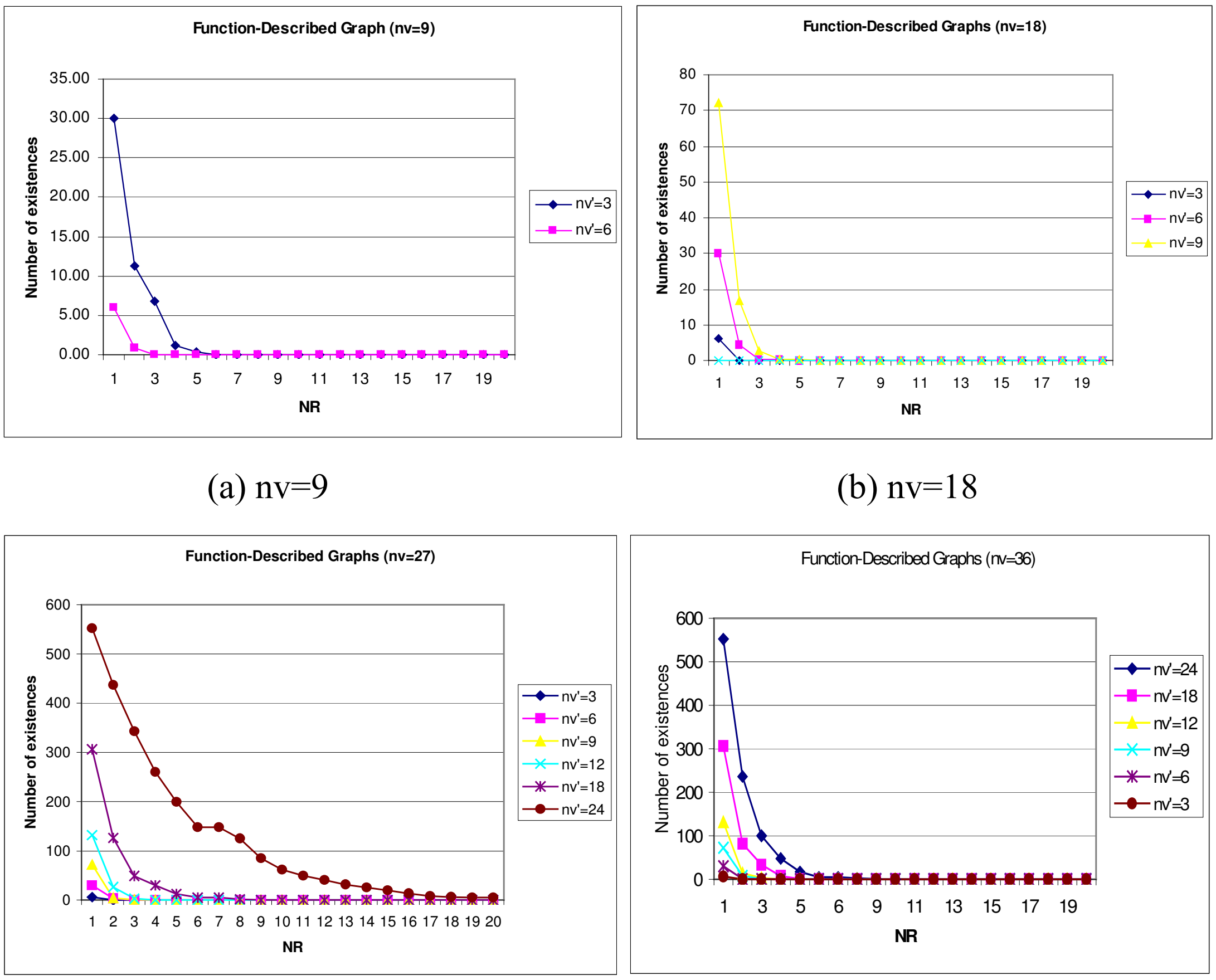

(a) nv=9 (b) nv=18

(c) nv=27 (d) nv=36

Figure 14. Number of Existences.

# 5. Distance measures for matching AGs to FDGs

This chapter presents the distance measure between AGs and FDGs. In the first section, the distance is introduced by means of Bayesian theory. Then, the distance measure between AGs and FDGs is presented using restrictions and it is compared with the Sanfeliu distance measure (Sanfeliu and Fu, 1983) in section 5.2. However, spurious elements mean that this distance becomes too coarse in real applications. Therefore section 5.3 proposes another distance in which the second order constraints (constraints on the antagonism, occurrence and existence relations) are relaxed. Finally, in section 5.4, a simple example of both distances is outlined.

## 5.1. Distance measure and the Bayesian theory framework

To study the matching between an AG *G* (data graph) and an FDG *F* (prototype graph) we use the Bayesian theory framework, as in (Wilson, 1997). The distance measures that will be proposed to match AGs with FDGs are somehow related to the maximum *a posteriori* probability $P(h|G)$ of a labelling function $h: G \to F$ given the measurements on the data graph *G*. The type of labelling function that is considered matches graph elements (vertices and arcs) of the AG with those of the FDG. The particular definition of the distance measure $d$ and how it is obtained from the most probable (or least costly) labelling function $h_d$ will be specified later depending on the type of constraints.

In any case, we should attempt to minimise the cost $C_h$ of the matching assignment with respect to a set *H* of valid labelling functions *h*. Theoretically, the optimal approach would be to compute the cost $C_h$ as a monotonic decreasing function of the *a posteriori* probability, $C_h = func(P(h|G))$. But, in practice, a well-founded approach, which also leads to optimal matching if uniform *a priori* probabilities (within *H*) are assumed, is to take the cost as a monotonic decreasing function of the conditional probability of the data graph given the labelling function, $C_h = func(P(G|h))$.

To illustrate this, let us apply the Bayes theorem to the *a posteriori* probability, which gives

$$P(h|G) = \frac{P(G|h) * P(h)}{P(G)} \tag{55}$$

where the three probabilities in the right hand side of the equation deserve some comments. First of all, the joint overall probability of the graph *G*, $P(G)$, is fixed independently of the match $h$ and therefore it is not taken into account in the definition of the cost $C_h$. The joint conditional probability $P(G|h)$ models the probability of the known measurements given a match or configuration. Since these measurements (attribute values) on the vertices and arcs are considered independent to each other we can factorise the joint conditional probability as

$$P(G|h) = \prod_{\forall x} \Pr(\gamma(y) = \gamma(x) \mid h(x) = y) \tag{56}$$

where *x* and *y* are graph elements in the AG and the FDG respectively, $\gamma(y)$ is the random variable associated with *y*, $\gamma(x)$ is the attribute value in *x*, and all the elements of both graphs have to appear in the productory (possibly by extending the domain and range of the mapping with null elements). Finally, the *a priori* probability $P(h)$ represents the configuration probability, which, unlike the previous one, models the structural aspects of the AG and the FDG being matched, and defines the allowable configurations in $H$. In some approaches, the structure that is inconsistent is immediately rejected because the configuration probability is defined as a binary distribution: $P(h) = c$, $c > 0$, if $h \in H$, i.e. when the labelling represents a function in which all structural constraints are fully satisfied; and $P(h) = 0$ when $h \notin H$. The positive constant *c* is merely introduced here for normalisation purposes.

In summary, the Bayes theorem maximises the product $P(G|h) * P(h)$ not the posterior probability $P(h|G)$ and reaches the optimal matching $h_d$. Assuming a uniform probability distribution among the valid labelling functions, the problem is reduced to that of filtering the invalid mappings and maximising $P(G|h)$ within the remaining set $H$. This approach is at the heart of graph search algorithms, such as the subgraph isomorphism and maximal clique finding methods.

The above framework, however, fails to tackle adequately one of the fundamental problems in computer vision and other pattern recognition fields: namely, that data extracted from the information sources (e.g. images) is noisy, incomplete or uncertain. For this reason, both the posterior probability and the conditional probability mentioned above might sometimes provide a coarse measure of the quality of a match. For example, if an extraneous element is added to a perfectly matched data graph the instantiation probability is zero, i.e. $P(G|h)=0$. We must therefore admit that there may be both extraneous and missing elements in the data graphs. As a consequence, inconsistent matches should no longer be discarded as incorrect since they could be the result of graph corruption (Wilson, 1997).

For our purposes, we require a fine but robust matching cost that not only makes powerful use of the measurement information in the data graphs (attribute values) and in the prototypes (random variable distributions) but is also effective way of constraining the possible matches, if we want the system to be able to discern between prototypes. The matching measure must be soft for two reasons: first, because it is assumed that in real applications the patterns are distorted by noise, and second, because a prototype has to represent not only the objects in the reference set but also the ones that are "near" them.

First of all, and for the sake of robustness, the mapping $h$ is not defined from the initial AG that represents the pattern to the initial FDG that represents the class, but from the *k*-extended AG to the *k*-extended FDG. In this way, it accepts that there may be some missing graph elements or some extraneous graph elements introduced by noisy effects. A missing element in the AG will be represented by a null element in the extended AG, and an extraneous element in the AG should be mapped to a null element in the extended FDG. Since it is desired to allow a priori all the isomorphisms, the number of vertices $k$ in the extended graphs is set to the sum of the number of vertices in both initial graphs. Hence, the limit situations in which all the graph elements in the FDG are missing in the AG or all the graph elements in the AG are extraneous are covered.

Let $G'$ be a *k*-extension of the AG $G$ and $F'$ be a *k*-extension of the FDG $F$. Then, $G'$ and $F'$ are structurally isomorphic and complete with the same number of vertices $k$. They also share a common attribute domain $(\Delta_v, \Delta_e)$. Now, the labelling function is defined as a mapping $h: G' \to F'$, and the *a priori* configuration probability of $h$ is assumed to be $P(h) = c$ if $h \in H$, and $P(h) = 0$ if $h \notin H$, where the set $H$ is to be determined and $c$ is a positive constant (in theory, $c = 1/|H|$). Since graphs do not have any predetermined orientation and each orientation is given by a morphism $h$, a *global cost* $C_h$ is associated with each $h$ in a set of valid mappings $H$, and the measure of similarity is defined as the minimum of all such costs,

$$d = \min_{h \in H} \{C_h\} \tag{57}$$

In addition, an optimal labelling $h_d$ is given by

$$h_d = \arg \min_{h \in H} \{C_h\} \tag{58}$$

We wish the global cost $C_h$ to provide a quantitative idea of the match quality through the mapping $h$. This cost is based on the joint conditional probability that the AG is generated from the FDG given labelling $h$, that is, $C_h = func(P(G|h))$. For instance, $C_h = -\ln(P(G|h))$ would be a possible choice, but it is not the most appropriate because of its high sensitivity to noise. If only one of the probabilities were to be zero, then the distance obtained would be $\infty$. Note that the joint probability $P(G|h)$ cannot be estimated directly and has to be approximated by the product of the first-order probabilities of the elements. In this case, the previous choice is equivalent to

$$C_h = -\sum_{\forall x} \ln(\Pr(\gamma(y) = \gamma(x) | \, h(x) = y)) \tag{59}$$

However, if only one graph element were to have a probability of zero, the joint probability would be zero and $C_h$ would be infinite. Since this may happen due to the noisy presence of an unexpected element (insertion) or the absence of a prototype element (deletion), if only one graph element were not properly mapped due to clutter, the involved graphs would be wrongly considered to be completely different.

## 5.2. Distance measure between AGs and FDGs using restrictions

We have shown that it is better to decompose the global cost $C_h$ into the sum of bounded individual costs associated with the element matches. Although this cost has the major flaw that the joint probability is not considered as a whole, it has the advantage that clutter affects the global cost only locally. An *individual cost* $C(x,y)$ represents the dissimilarity between two mapped elements *x* and *y*, and it could even be based on the first-order probabilities of the elements, $C(x,y) = func(\Pr(\gamma(y) = \gamma(x) | h(x) = y))$, even though it is bounded by some fixed constant, $C(x,y) \leq \text{Max}$, for instance $C(x,y) \leq 1$.

The global cost is therefore computed as the sum of the individual costs of all the matches between graph elements,

$$C_h = \sum_{\forall x} C(x, h(x)) \tag{60}$$

The main concepts underlying the definition of the distance measures between AGs and FDGs have been introduced above. To define the different specific measures, it is only needed to define the set of valid mappings *H* and the individual costs $C(x,y)$.

### 5.2.1. Definition of restrictions on the valid morphisms

The basic elements in the distance which uses graph elements $d_f$ are vertices and arcs. For this reason, the configuration probability is defined on a morphism $f$ from G' with an underlying structure $(\Sigma_v, \Sigma_e)$ to F' with an underlying structure $(\Sigma_\omega, \Sigma_\varepsilon)$. This morphism labels graph elements. It is assumed that the extended AG G' and the extended FDG F' are complete and contain the same number of vertices (same order). Since vertices and arcs represent two different kinds of information, the global morphism is actually defined as a pair of morphisms $f=(f_v,f_e)$, where $f_v:\Sigma_v\rightarrow\Sigma_w$ is a mapping defined on the vertices (the basic parts of the object) and $f_e:\Sigma_e\rightarrow\Sigma_\varepsilon$ is a mapping defined on the arcs (the relations between these basic parts).

The set $F_0$ is composed of all the possible morphisms $f=(f_v,f_e)$.

However, if the structural information of the graphs is to be used, some restrictions have to be imposed to attain a valid morphism, which will depend on the particular application. Let $R_i(f)$ denote that $f$ fulfils a certain constraint $R_i$ and let $F_i$ denote the set of configurations $f$ such that $R_i(f)$. Next, five types of constraints are defined together with the set of functions that satisfy them. Thus, depending on the nature of the application, the set of valid mappings $H$ is defined as a mapping of the $F_i$ or as the intersection of two or more mappings.

**Constraint** $R_1$: The morphisms $f_v$ and $f_e$ are both defined as bijective functions.

$$(f_v(v_i)=f_v(v_j)\Leftrightarrow v_i=v_j)\wedge(f_e(e_{ij})=f_e(e_{pq})\Leftrightarrow e_{ij}=e_{pq})\tag{61}$$

When this constraint is applied, the basic elements and their relations in the pattern appear once and only once in the prototype (monomorphism). For instance, if the pattern is a chair, different feet of the chair have to be mapped with different feet of the prototype chair. In addition, since the domain and the range of each mapping have the same cardinality, the above constraint guarantees a structural isomorphism.

**Constraint** $R_2$: The morphism $f=(f_v,f_e)$ has to be coherent structurally.

$$(f_v(v_i)=\omega_p\wedge f_v(v_j)=\omega_q)\Rightarrow(f_e(e_{ij})=\varepsilon_{pq})\tag{62}$$

If a pair of vertices of G' are mapped with a pair of vertices of F', then the arcs in G' and F' that connect these vertices have to be mapped together. An arc in the graph is a relation between two basic elements in the object or in the prototype, therefore, since a relation depends directly on their linking elements, $f_e$ is regarded to be dependent on $f_v$. Hence, the global morphism $f$ is totally determined by the morphism between vertices $f_v$.

**Constraint** $R_3$: The second-order constraints in the vertices have to be fulfilled (except for second-order constraints induced by FDG extended vertices).

The satisfaction of constraint $R_3$ means satisfying the antagonism, occurrence and existence relations between vertices. However, the relations that include FDG null vertices should not be taken into account, since they are artificially introduced in the extension of the FDG (see tables 4 - 6).

If two vertices are antagonistic, they do not appear together in any of the AGs used to synthesise the FDG. The joint probability of these two vertices both having a non-null value is considered to be zero. The mapping from two vertices of G' to two antagonistic vertices of F' is allowed only if at least one of the vertices involved is null. This means that what has been added to be matched is an extended vertex. Thus, the first condition of constraint $R_3$ can be expressed as the rule

$$\begin{aligned}&\left(A_\omega(\omega_p,\omega_q)=1 \wedge f_v(v_i)=\omega_p \wedge f_v(v_j)=\omega_q\right)\Rightarrow\\&\left(\mathbf{a}_i=\Phi \vee \mathbf{a}_j=\Phi \vee \Pr(\alpha_p=\Phi)=1 \vee \Pr(\alpha_q=\Phi)=1\right)\end{aligned} \tag{63}$$

which, using the first-order probability density functions included in the FDGs, is

$$\begin{aligned}&\left(A_\omega(\omega_p,\omega_q)=1 \wedge f_v(v_i)=\omega_p \wedge f_v(v_j)=\omega_q\right)\Rightarrow\\&\left(\mathbf{a}_i=\Phi \vee \mathbf{a}_j=\Phi \vee p_p(\Phi)=1 \vee p_q(\Phi)=1\right)\end{aligned} \tag{64}$$

The two rightmost terms in the consequent of the above rule make it possible to match a non-null vertex of the AG to an extended (null) vertex of the FDG. Otherwise, due to the second-order effect shown in table 4, the FDG null vertices would never be matched to actual vertices in the AG and, consequently, they would be of no help in bringing some flexibility to the matching process.

Now, if a vertex is occurrent to another, and provided that the former has appeared in an AG used to synthesise the FDG, the latter will also have appeared. This means that the joint probability of these two vertices is zero when the second vertex is a null element and the first one is not. If it is considered desirable to keep the same dependence in the matched AG, when two vertices of the AG are mapped to two occurrent vertices of the FDG, the morphism is allowed in two situations. The first one is when the vertex of the AG mapped to the first element in the occurrence relation is null or the vertex of the AG mapped to the second element is non-null (occurrence satisfaction). The second one is when at least one of the FDG vertices is null, i.e. an extended vertex. This also allows a non-null vertex of the AG to be matched to an extended vertex of the FDG (insertion). Thus, the second condition of constraint $R_3$ is given by

$$\begin{aligned}&\left(O_\omega\left(\omega_p,\omega_q\right)=1\wedge f_v(v_i)=\omega_p\wedge f_v\left(v_j\right)=\omega_q\right)\Rightarrow\\&\left(\mathbf{a}_i=\Phi\vee\mathbf{a}_j\neq\Phi\vee\Pr\left(\alpha_p=\Phi\right)=1\right)\end{aligned}\tag{65}$$

which, using the probability density functions defined in the FDGs, is

$$\begin{aligned}&\left(O_\omega\left(\omega_p,\omega_q\right)=1\wedge f_v(v_i)=\omega_p\wedge f_v\left(v_j\right)=\omega_q\right)\Rightarrow\\&\left(\mathbf{a}_i=\Phi\vee\mathbf{a}_j\neq\Phi\vee p_p(\Phi)=1\right)\end{aligned}\tag{66}$$

Note that the term $p_q(\Phi)=1$ is not included in the consequent of the above rule because it would be totally redundant, since $O_\omega\left(\omega_p,\omega_q\right)=1\wedge p_q(\Phi)=1\Rightarrow p_p(\Phi)=1$. In other words, we could restate the rule as "either the occurrence relation is satisfied or at least one of the FDG vertices is null".

Finally, if an existence relation between two FDG vertices holds, at least one of the AG vertices mapped to them should be non-null, i.e. included in the original AG. However, as in the cases of the previous relations, the existence constraint is defined to be fulfilled as well when one of the FDG vertices is null (this may only happen if the other FDG vertex is a strict non-null vertex). Thus, the third condition of constraint $R_3$ is given by

$$\left(E_{\omega}\left(\omega_p,\omega_q\right)=1\wedge f_v\left(v_i\right)=\omega_p\wedge f_v\left(v_j\right)=\omega_q\right)\Rightarrow$$
$$\left(\mathbf{a}_i\neq\Phi\vee\mathbf{a}_j\neq\Phi\vee\Pr\left(\alpha_p=\Phi\right)=1\vee\Pr\left(\alpha_q=\Phi\right)=1\right) \tag{67}$$

which, using the probability density functions defined in the FDGs, is

$$\left(E_{\omega}\left(\omega_p,\omega_q\right)=1\wedge f_v\left(v_i\right)=\omega_p\wedge f_v\left(v_j\right)=\omega_q\right)\Rightarrow$$
$$\left(\mathbf{a}_i\neq\Phi\vee\mathbf{a}_j\neq\Phi\vee p_p(\Phi)=1\vee p_q(\Phi)=1\right). \tag{68}$$

**Constraint** $R_4$: The second-order constraints in the arcs have to be fulfilled (except for second-order constraints induced by FDG null arcs).

The reasons and explanations are similar to the vertex case, but the use of the arc conditional probabilities stored in the FDG has to be considered. For notational purposes, let $\gamma_e\left(e_{ij}\right)=\mathbf{b}_m$ and $\gamma_e\left(e_{kt}\right)=\mathbf{b}_n$ be, respectively, the attribute values of the arcs $e_{ij}$ and $e_{kt}$ in the AG, and let $\gamma_{\varepsilon}\left(\varepsilon_{ab}\right)=\beta_g$ and $\gamma_{\varepsilon}\left(\varepsilon_{cd}\right)=\beta_f$ be, respectively, the random variables associated with arcs $\varepsilon_{ab}$ and $\varepsilon_{cd}$ in the FDG.

The antagonism constraint on the arcs can be expressed as the rule

$$\left(A_{\varepsilon}\left(\varepsilon_{ab},\varepsilon_{cd}\right)=1\wedge f_e\left(e_{ij}\right)=\varepsilon_{ab}\wedge f_e\left(e_{kt}\right)=\varepsilon_{cd}\right)\Rightarrow$$
$$\left(\mathbf{b}_m=\Phi\vee\mathbf{b}_n=\Phi\vee\Pr\left(\beta_g=\Phi\right)=1\vee\Pr\left(\beta_f=\Phi\right)=1\right) \tag{69}$$

which, taking into account equation (13) given in Section 4.3.1 and using the probability density functions represented in the FDG, is equivalent to the rule

$$\left(A_{\varepsilon}\left(\varepsilon_{ab},\varepsilon_{cd}\right)=1\wedge f_e\left(e_{ij}\right)=\varepsilon_{ab}\wedge f_e\left(e_{kt}\right)=\varepsilon_{cd}\right)\Rightarrow$$
$$\begin{pmatrix}\mathbf{b}_m=\Phi\vee\mathbf{b}_n=\Phi\vee\left(1-q_g(\Phi)\right)*\left(1-p_a(\Phi)\right)*\left(1-p_b(\Phi)\right)=0\vee\\ \left(1-q_f(\Phi)\right)*\left(1-p_c(\Phi)\right)*\left(1-p_d(\Phi)\right)=0\end{pmatrix} \tag{70}$$

The occurrence constraint on the arcs is given by

$$\left(O_{\varepsilon}\left(\varepsilon_{ab},\varepsilon_{cd}\right)=1\wedge f_e\left(e_{ij}\right)=\varepsilon_{ab}\wedge f_e\left(e_{kt}\right)=\varepsilon_{cd}\right)\Rightarrow$$
$$\left(\mathbf{b}_m=\Phi\vee\mathbf{b}_n\neq\Phi\vee\Pr\left(\beta_g=\Phi\right)=1\right) \tag{71}$$

As in the antagonism case, equation (13), the arc occurrence constraint can be expressed using the FDG probability density functions as

$$\begin{aligned}&\left(O_\varepsilon(\varepsilon_{ab},\varepsilon_{cd})=1\wedge f_e(e_{ij})=\varepsilon_{ab}\wedge f_e(e_{kt})=\varepsilon_{cd}\right)\Rightarrow\\&\left(\mathbf{b}_m=\Phi\vee\mathbf{b}_n\neq\Phi\vee(1-q_g(\Phi))*(1-p_a(\Phi))*(1-p_b(\Phi))=0\right)\end{aligned}\qquad(72)$$

Finally, the existence constraint on the arcs is given by

$$\begin{aligned}&\left(E_\varepsilon(\varepsilon_{ab},\varepsilon_{cd})=1\wedge f_e(e_{ij})=\varepsilon_{ab}\wedge f_e(e_{kt})=\varepsilon_{cd}\right)\Rightarrow\\&\left(\mathbf{b}_m\neq\Phi\vee\mathbf{b}_n\neq\Phi\vee\Pr(\beta_g=\Phi)=1\vee\Pr(\beta_f=\Phi)=1\right)\end{aligned}\qquad(73)$$

or, with the FDG probability density functions, by

$$\begin{aligned}&\left(E_\varepsilon(\varepsilon_{ab},\varepsilon_{cd})=1\wedge f_e(e_{ij})=\varepsilon_{ab}\wedge f_e(e_{kt})=\varepsilon_{cd}\right)\Rightarrow\\&\begin{pmatrix}\mathbf{b}_m\neq\Phi\vee\mathbf{b}_n\neq\Phi\vee(1-q_g(\Phi))*(1-p_a(\Phi))*(1-p_b(\Phi))=0\vee\\(1-q_f(\Phi))*(1-p_c(\Phi))*(1-p_d(\Phi))=0\end{pmatrix}\end{aligned}\qquad(74)$$

**Constraint** $R_5$: The relative order between non-null arcs in planar graphs has to be preserved when a structural isomorphism is applied.

Let $r_t$ be the number of arcs departing from a node $t$ in G and let $n_t:\{1,\ldots,r_t\}\rightarrow\Sigma_v$ be a function that represents an ordering of these arcs. Similarly, let $r_d$ be the number of arcs departing from a node $d$ in F and let $n_d:\{1,\ldots,r_d\}\rightarrow\Sigma_\omega$ be an ordering function of these arcs. Then, the planar graph constraint is given by

$$\begin{aligned}&\forall t,i,j,k:1\leq i<j<k\leq r_t:\\&\begin{bmatrix}\left(f_v(v_t)=\omega_d\wedge f_e(e_{tn_t(i)})=\varepsilon_{dn_d(a)}\wedge f_e(e_{tn_t(j)})=\varepsilon_{dn_d(b)}\wedge f_e(e_{tn_t(k)})=\varepsilon_{dn_d(c)}\right)\Rightarrow\\((a<b<c)\vee(c<a\wedge(b<c\vee a<b)))\end{bmatrix}\end{aligned}\qquad(75)$$

Note that a rotational shift is permitted in the above expression.

### 5.2.2. Individual costs of matching elements

We now turn our attention to the individual cost of matching a pair of elements, one from an AG and one from an FDG. The cost is defined as a normalised function which depends on the dissimilarity between the two mapped elements, as given by the negative logarithm of the probability of instantiating the random element of the FDG to the corresponding attribute value in the AG. That is

$$C(x,y)=\begin{cases}\dfrac{-\ln\left(\Pr\left(\gamma(y)=\gamma(x)\mid f(x)=y\right)\right)}{-\ln\left(K_{\Pr}\right)} & \text{if } \Pr\left(\gamma(y)=\gamma(x)\mid f(x)=y\right)\geq K_{\Pr}\\ 1 & \text{otherwise}\end{cases}\tag{76}$$

where the cost $C(x,y)$ is bounded by $[0,1]$, and the positive constant $K_{\Pr}\in[0,1]$ is a threshold on low probabilities that is introduced to prevent the case $\ln(0)$, which gives negative infinity. Hence, $C(x,y)=1$ will be the cost of matching a null element of the FDG to a non-null element of the AG or of matching an FDG element to an AG element whose attribute value has a very low probability of instantiation. That is to say, $\Pr\left(\gamma(y)=\gamma(x)\mid f(x)=y\right)\leq K_{\Pr}$.

The individual cost of the vertices is defined using the probabilities stored in the FDG as

$$C_{f_v}\left(v_i,\omega_q\right)=\begin{cases}\dfrac{-\ln\left(p_q(\mathbf{a}_i)\right)}{-\ln\left(K_{\Pr}\right)} & \text{if } p_q(\mathbf{a}_i)\geq K_{\Pr}\\ 1 & \text{otherwise}\end{cases}\tag{77}$$

The individual cost of the arcs is defined using the arc conditional probabilities as follows. Let $\gamma_e\left(e_{ij}\right)=\mathbf{b}_m$ in the AG arc and let $\gamma_\varepsilon\left(\varepsilon_{ab}\right)=\beta_n$ in the matched FDG arc. Then, in general,

$$C_{f_e}\left(e_{ij},\varepsilon_{ab}\right)=\begin{cases}\dfrac{-\ln\left(q_n\left(\mathbf{b}_m\right)\right)}{-\ln\left(K_{\Pr}\right)} & \text{if } q_n\left(\mathbf{b}_m\right)\geq K_{\Pr}\\ 1 & \text{otherwise}\end{cases} \tag{78}$$

However, if either $v_i$ or $v_j$ is a null extended vertex in the AG, then the conditional probability $q_n(\mathbf{b}_m)$ is not applicable, since it depends on the existence of the two extreme vertices, and must be replaced by the conditional probability $\Pr(\beta_n = \mathbf{b}_m \mid \alpha_a = \Phi \vee \alpha_b = \Phi)$, which is 1 if $\mathbf{b}_m = \Phi$ and 0 otherwise.

Finally, the total cost of a given mapping $f$ is calculated as

$$C_f = K_1 * \sum_{\forall v_i \in \Sigma_v \text{ of } G'} C_{f_v}\left(v_i, f_v\left(v_i\right)\right) + K_2 * \sum_{\forall e_{ij} \in \Sigma_e \text{ of } G'} C_{f_e}\left(e_{ij}, f_e\left(e_{ij}\right)\right) \tag{79}$$

The two terms are weighted by non-negative constants $K_1$ and $K_2$, to compensate for the different number of elements in the additions. Thus, the distance measure between an AG and an FDG is defined as the minimum cost achieved by a valid morphism *f*,

$$d_f = \min_{f \in H}\left\{C_f\right\} \tag{80}$$

### 5.2.3. Comparison with edit-operation distances

Sanfeliu and Fu's classical framework for comparing graphs (Sanfeliu and Fu, 1983) is based on the idea that the distance between two structures is the lowest global cost of transforming one into the other using edit operations (see section 2.2). Six edit operations are used in graphs: deletion, insertion and substitution of vertices and arcs. There is a certain cost associated with each edit operation, and thus, the global cost is the sum of the costs of the edit operations involved.

However, in our approach, all these edit operations can be viewed as substitutions and, more importantly, the individual costs depend on the attribute values and the probability density functions of the random elements and are therefore variable. There are four type of matches on the vertices and arcs. They depend on whether the elements belong to the initial graphs G and F or whether they have been added while extending the graphs to G' and F', respectively.

Let us first study the match between vertices $v_i$ and $\omega_q$ by analysing the four possible cases:

1) Both vertices belong to the initial graphs. The cost, therefore, depends on the probability of the attribute value in the vertex of G, $p_q(\mathbf{a}_i)$, and it can be seen as a *substitution cost* which depends on the semantic information,

$$C_{f_v}(v_i, \omega_q) = \begin{cases} \dfrac{-\ln(p_q(\mathbf{a}_i))}{-\ln(K_{\mathrm{Pr}})} & \text{if } p_q(\mathbf{a}_i) \geq K_{\mathrm{Pr}} \\ 1 & \text{otherwise} \end{cases} \tag{81}$$

2) The vertex of the AG belongs to the initial graph G and the vertex of the FDG is a null element (added during the extension process). The probability in the null elements of the FDG F' is 0 for any actual attribute value in the vertex of G, that is to say, $p_q(\mathbf{a}_i) = 0$, and therefore,

$$C_{f_v}(v_i, \omega_q) = 1 \tag{82}$$

This case can be regarded as an *insertion* operation with a constant cost.

3) The vertex of the AG G' is a null element (added during the extension process) and the vertex of the FDG belongs to the initial graph F. The cost depends on the probability of the null value in the random vertex of the FDG,

$$C_{f_v}(v_i, \omega_q) = \begin{cases} \dfrac{-\ln(p_q(\Phi))}{-\ln(K_{\mathrm{Pr}})} & \text{if } p_q(\Phi) \geq K_{\mathrm{Pr}} \\ 1 & \text{otherwise} \end{cases} \tag{83}$$

This case can be considered as a *deletion* operation, but it is important to emphasise that the cost is not constant but variable and dependent on the probability of the null value. Suppose this probability is high, as most of the AGs used to synthesise this FDG do not contain this vertex. Then, the cost will be low and, therefore, this individual match will not substantially increase the distance value. On the contrary, if most of the AGs used to synthesise the FDG include this vertex, then the probability of the null value will be low and the cost high.

4) Both vertices have been added during the extension process. It is not desirable that this match should influence the global cost, as an arbitrary number of null elements can be generated independently of the initial graphs. Bearing in mind that in this situation $p_q(\Phi)=1$ (by definition) then this case can be regarded as a *substitution* operation with zero cost.

$$C_{f_v}(v_i, \omega_q) = \frac{-\ln(1)}{-\ln(K_{\mathrm{Pr}})} = 0 \tag{84}$$

We move on to study the match between the arcs $e_{ij}$ and $\varepsilon_{ab}$. For convenience we assume that $\gamma_e(e_{ij}) = \mathbf{b}_m$ and that $\gamma_\varepsilon(\varepsilon_{ab}) = \beta_n$. We should point out that the first-order probability functions stored in the arcs of F' are conditioned to the existence of the corresponding extreme vertices in G. As before, the following four cases must be analysed:

1) Both arcs belong to the initial graphs. It is a *substitution* operation but we have to distinguish whether the vertices connected by $\varepsilon_{ab}$ in F have been matched to vertices that belong to the initial graph G or to null vertices. In the first situation, the probability $q_e(\mathbf{b}_m)$ is applicable and the cost depends on the semantic knowledge,

$$C_{f_e}(e_{ij}, \varepsilon_{ab}) = \begin{cases} \dfrac{-\ln(q_n(\mathbf{b}_m))}{-\ln(K_{\mathrm{Pr}})} & \text{if } q_n(\mathbf{b}_m) \geq K_{\mathrm{Pr}} \\ 1 & \text{otherwise} \end{cases} \tag{85}$$

In the second situation, which cannot occur if the morphism belongs to $F_2$ since it is a non-coherent match, one of the endpoints of $\varepsilon_{ab}$ is matched to a null vertex. Bearing in mind that $\Pr(\beta_n \neq \Phi | \alpha_a = \Phi \vee \alpha_b = \Phi) = 0$, the substitution cost applied is the highest one,

$$C_{f_e}(e_{ij}, \varepsilon_{ab}) = 1 \tag{86}$$

2) The arc of the AG belongs to the initial graph (it is non-null) but the arc of the FDG is a null element. In this case we have an *insertion cost* that is constant and maximum. Nevertheless, it can be reached in two different ways. If $q_n$ is applicable because the extreme vertices of $\varepsilon_{ab}$ are matched to non-null vertices of G', then $q_n(\mathbf{b}_m) = 0$ for all $\mathbf{b}_m \neq \Phi$. On the contrary, if $q_n$ is not applicable because the extreme vertices of $\varepsilon_{ab}$ are matched to null vertices of G', then $\Pr(\beta_n \neq \Phi | \alpha_a = \Phi \vee \alpha_b = \Phi) = 0$ , and thus,

$$C_{f_e}(e_{ij}, \varepsilon_{ab}) = 1 \tag{87}$$

3) The AG arc is a null element and the FDG arc belongs to the initial F (*deletion* operation). Here, as in the first case, we have to distinguish whether the vertices connected by the arc of F have been matched to vertices of G or to null vertices. In the first situation, $q_n(\mathbf{b}_m)$ is applicable and the cost depends on the probability of the null value,

$$C_{f_e}(e_{ij}, \varepsilon_{ab}) = \begin{cases} \dfrac{-\ln(q_n(\Phi))}{-\ln(K_{\Pr})} & \text{if } q_n(\Phi) \geq K_{\Pr} \\ 1 & \text{otherwise} \end{cases} \tag{88}$$

In the second situation, which is also a perfectly coherent match, because an extreme vertex of a null arc can be a null vertex in a morphism belonging to $F_2$, $q_n(\mathbf{b}_m)$ does not apply but $\Pr(\beta_n = \Phi | \alpha_a = \Phi \vee \alpha_b = \Phi) = 1$, and therefore

$$C_{f_e}(e_{ij}, \varepsilon_{ab}) = \frac{-\ln(1)}{-\ln(K_{\Pr})} = 0 \tag{89}$$

4) Finally, the case in which both arcs are null elements in their respective graphs. As in the vertex case, the total cost should not be influenced by this match, so the individual cost is zero in the two possible cases. The first, when the extreme vertices of $\varepsilon_{ab}$ have been matched to null elements, $q_n$ is not defined but $\Pr(\beta_n = \Phi | \alpha_a = \Phi \vee \alpha_b = \Phi) = 1$. The second, when the extreme vertices of $\varepsilon_{ab}$ have been matched to non-null elements, $q_n$ is defined as $q_n(\Phi) = 1$. In both cases it turns out that

$$C_{f_e}(e_{ij}, \varepsilon_{ab}) = \frac{-\ln(1)}{-\ln(K_{\Pr})} = 0 \tag{90}$$

### 5.3. Distance between AGs and FDGs relaxing 2nd order constraints

Second order relations of are useful for constraining the set of possible labellings between AGs and FDGs through restrictions $R_3$ and $R_4$. The aimed of this constraint is to obtain the best labelling function *f*, taking into account, as much as possible, the structure of the cluster of AGs that was used to synthesise the FDG. Nevertheless, in real applications, AGs can be distorted by external noise and, therefore, the constraints $R_3$ and $R_4$ associated to the second order relations have to be relaxed to prevent a noisy AG being misclassified due to non-fulfilment of any of these constraints. For instance, due to the second-order effect shown in equation (32) (section 4.3.3), the deletion of a strict non-null vertex of the FDG will almost always involve the non-fulfilment of some of the existence or occurrence constraints induced by that strict non-null vertex.

To gain more flexibility and robustness, instead of applying hard binary constraints, some local non-negative costs may be added to the global cost of the labelling depending on the second-order probabilities of the graph elements. Equations (91) to (93) show how to define these costs for the vertices. They assume that $f_v(v_i)=\omega_p$ and $f_v(v_j)=\omega_q$. These equations cover the three following qualitative cases: $C_{A_\omega}$ represents the presence of both vertices in the AG, $C_{O_\omega}$ represents the presence of only one of them, and $C_{E_\omega}$ the absence of both vertices. Note that, as in the definition of restriction $R_3$, the second-order costs induced artificially by FDG null vertices are not taken into account.

$$C_{A_\omega}(v_i,v_j,\omega_p,\omega_q)=\begin{cases}1-\Pr(\alpha_p\neq\Phi\wedge\alpha_q\neq\Phi) & if\begin{pmatrix}\mathbf{a}_i\neq\Phi\wedge\mathbf{a}_j\neq\Phi\wedge\\ p_p(\Phi)\neq 1\wedge p_q(\Phi)\neq 1\end{pmatrix}\\ 0 & \text{otherwise}\end{cases} \tag{91}$$

$$C_{O_\omega}(v_i,v_j,\omega_p,\omega_q)=\begin{cases}1-\Pr(\alpha_p\neq\Phi\wedge\alpha_q=\Phi) & if\begin{pmatrix}\mathbf{a}_i\neq\Phi\wedge\mathbf{a}_j=\Phi\wedge\\ p_p(\Phi)\neq 1\end{pmatrix}\\ 0 & \text{otherwise}\end{cases} \tag{92}$$

$$C_{E_\omega}(v_i,v_j,\omega_p,\omega_q)=\begin{cases}1-\Pr(\alpha_p=\Phi\wedge\alpha_q=\Phi) & if\begin{pmatrix}\mathbf{a}_i=\Phi\wedge\mathbf{a}_j=\Phi\wedge\\ p_p(\Phi)\neq 1\wedge p_q(\Phi)\neq 1\end{pmatrix}\\ 0 & \text{otherwise}\end{cases} \tag{93}$$

Equations (94) to (96) show the corresponding three types of second-order costs in the case of the arcs, assuming $\gamma_e(e_{ij})=\mathbf{b}_m$ and $\gamma_e(e_{kt})=\mathbf{b}_n$ as well as $\gamma_\varepsilon(\varepsilon_{ab})=\beta_e$ and $\gamma_\varepsilon(\varepsilon_{cd})=\beta_f$ for convenience. As in the definition of restriction $R_4$, the second-order costs induced by FDG null arcs are not taken into account.

$$C_{A_\varepsilon}\left(e_{ij},e_{kt},\varepsilon_{ab},\varepsilon_{cd}\right)=\begin{cases}1-\Pr\left(\beta_e\neq\Phi\wedge\beta_f\neq\Phi\right) & if\begin{pmatrix}\mathbf{b}_m\neq\Phi\wedge\mathbf{b}_n\neq\Phi\wedge\\ \Pr\left(\beta_e=\Phi\right)\neq1\wedge\\ \Pr\left(\beta_f=\Phi\right)\neq1\end{pmatrix}\\ 0 & \text{otherwise}\end{cases} \quad (94)$$

$$C_{O_\varepsilon}\left(e_{ij},e_{kt},\varepsilon_{ab},\varepsilon_{cd}\right)=\begin{cases}1-\Pr\left(\beta_e\neq\Phi\wedge\beta_f=\Phi\right) & if\begin{pmatrix}\mathbf{b}_m\neq\Phi\wedge\mathbf{b}_n=\Phi\wedge\\ \Pr\left(\beta_e=\Phi\right)\neq1\end{pmatrix}\\ 0 & \text{otherwise}\end{cases} \quad (95)$$

$$C_{E_\varepsilon}\left(e_{ij},e_{kt},\varepsilon_{ab},\varepsilon_{cd}\right)=\begin{cases}1-\Pr\left(\beta_e=\Phi\wedge\beta_f=\Phi\right) & if\begin{pmatrix}\mathbf{b}_m=\Phi\wedge\mathbf{b}_n=\Phi\wedge\\ \Pr\left(\beta_e=\Phi\right)\neq1\wedge\\ \Pr\left(\beta_f=\Phi\right)\neq1\end{pmatrix}\\ 0 & \text{otherwise}\end{cases} \quad (96)$$

Since the second-order probabilities are not actually stored in the FDGs, they are replaced by the second-order relations, and the costs are therefore coarser. That is to say, some second-order non-negative costs are added to the global cost of the labelling when second-order constraints (antagonism, occurrence, existence) are broken. Equations (97) to (102) show the final second-order costs, which can only be 1 or 0, associated to the three relations of antagonism, occurrence and existence between pairs of vertices and pairs of arcs, respectively.

The cost of the antagonism on the vertices and arcs,

$$C_{A_\omega}\left(v_i,v_j,\omega_p,\omega_q\right)=\begin{cases}A_\omega\left(\omega_p,\omega_q\right) & if\begin{pmatrix}\mathbf{a}_i\neq\Phi\wedge\mathbf{a}_j\neq\Phi\wedge\\ p_p(\Phi)\neq1\wedge p_q(\Phi)\neq1\end{pmatrix}\\ 0 & \text{otherwise}\end{cases} \quad (97)$$

$$C_{A_\varepsilon}\left(e_{ij},e_{kt},\varepsilon_{ab},\varepsilon_{cd}\right)=\begin{cases}A_\varepsilon\left(\varepsilon_{ab},\varepsilon_{cd}\right) & if\begin{pmatrix}\mathbf{b}_m\neq\Phi\wedge\mathbf{b}_n\neq\Phi\wedge\\ (1-q_e(\Phi))*(1-p_a(\Phi))*(1-p_b(\Phi))\neq0\wedge\\ (1-q_f(\Phi))*(1-p_c(\Phi))*(1-p_d(\Phi))\neq0\end{pmatrix}\\ 0 & \text{otherwise}\end{cases} \quad (98)$$

The cost of the occurrences on the vertices and arcs,

$$C_{O_\omega}\left(v_i, v_j, \omega_p, \omega_q\right) = \begin{cases} O_\omega\left(\omega_p, \omega_q\right) & if\ \mathbf{a}_i \neq \Phi \wedge \mathbf{a}_j = \Phi \wedge p_p(\Phi) \neq 1 \\ 0 & \text{otherwise} \end{cases} \tag{99}$$

$$C_{O_\varepsilon}\left(e_{ij}, e_{kt}, \varepsilon_{ab}, \varepsilon_{cd}\right) = \begin{cases} O_\varepsilon\left(\varepsilon_{ab}, \varepsilon_{cd}\right) & if \begin{pmatrix} \mathbf{b}_m \neq \Phi \wedge \mathbf{b}_n = \Phi \wedge \\ (1 - q_e(\Phi)) * (1 - p_a(\Phi)) * \\ (1 - p_b(\Phi)) \neq 0 \end{pmatrix} \\ 0 & \text{otherwise} \end{cases} \tag{100}$$

And the cost of the existence on the vertices and arcs,

$$C_{E_\omega}\left(v_i, v_j, \omega_p, \omega_q\right) = \begin{cases} E_\omega\left(\omega_p, \omega_q\right) & if \begin{pmatrix} \mathbf{a}_i = \Phi \wedge \mathbf{a}_j = \Phi \wedge \\ p_p(\Phi) \neq 1 \wedge p_q(\Phi) \neq 1 \end{pmatrix} \\ 0 & \text{otherwise} \end{cases} \tag{101}$$

$$C_{E_\varepsilon}\left(e_{ij}, e_{kt}, \varepsilon_{ab}, \varepsilon_{cd}\right) = \begin{cases} E_\varepsilon\left(\varepsilon_{ab}, \varepsilon_{cd}\right) & if \begin{pmatrix} \mathbf{b}_m = \Phi \wedge \mathbf{b}_n = \Phi \wedge \\ (1 - q_e(\Phi)) * (1 - p_a(\Phi)) * (1 - p_b(\Phi)) \neq 0 \wedge \\ (1 - q_f(\Phi)) * (1 - p_c(\Phi)) * (1 - p_d(\Phi)) \neq 0 \end{pmatrix} \\ 0 & \text{otherwise} \end{cases} \tag{102}$$

Now, the global cost on the labelling function $C_f^R$ is redefined with the two original terms that depend on the first-order probability information, and six more terms that depend on the second-order constraints:

$$C_f^R = \begin{cases} K_1 * \sum_{\forall v_i \in \Sigma_v \text{ of } G'} C_{f_v}\left(v_i, f_v(v_i)\right) + K_2 * \sum_{\forall e_{ij} \in \Sigma_e \text{ of } G'} C_{f_e}\left(e_{ij}, f_e(e_{ij})\right) + \\ K_3 * \sum_{\forall v_i, v_j \in \Sigma_v \text{ of } G'} C_{A_\omega}\left(v_i, v_j, f_v(v_i), f_v(v_j)\right) + K_4 * \sum_{\forall e_{ij}, e_{kt} \in \Sigma_e \text{ of } G'} C_{A_\varepsilon}\left(e_{ij}, e_{kt}, f_e(e_{ij}), f_e(e_{kt})\right) + \\ K_5 * \sum_{\forall v_i, v_j \in \Sigma_v \text{ of } G'} C_{O_\omega}\left(v_i, v_j, f_v(v_i), f_v(v_j)\right) + K_6 * \sum_{\forall e_{ij}, e_{kt} \in \Sigma_e \text{ of } G'} C_{O_\varepsilon}\left(e_{ij}, e_{kt}, f_e(e_{ij}), f_e(e_{kt})\right) + \\ K_7 * \sum_{\forall v_i, v_j \in \Sigma_v \text{ of } G'} C_{E_\omega}\left(v_i, v_j, f_v(v_i), f_v(v_j)\right) + K_8 * \sum_{\forall e_{ij}, e_{kt} \in \Sigma_e \text{ of } G'} C_{E_\varepsilon}\left(e_{ij}, e_{kt}, f_e(e_{ij}), f_e(e_{kt})\right) + \end{cases} \tag{103}$$

The eight terms are weighted by non-negative constants $K_1$ to $K_8$, to compensate for the different number of elements in the additions and to balance the influence of second-order costs with respect to first-order costs in the overall value. Finally, the distance measure between an AG and an FDG using both first-order and second-order costs $d_f^R$ is defined as the minimum cost achieved by a valid morphism $f$:

$$d_f^R = \min_{f \in H} \left\{ C_f^R \right\} \tag{104}$$

Note that $d_f^R = d_f$ if $K_3 = K_4 = K_5 = K_6 = K_7 = K_8 = \infty$ and $H \subset F_3 \cap F_4$.

### 5.4. Example of the distances between AGs and FDGs

Section 4.1 proposes that the domain of the joint probability of vertices $\omega_1$ and $\omega_2$ be split into four regions (Figure 3). Assuming that the probability of each region can be zero or greater than zero, there are 16 different combinations of the joint probability. Nevertheless, since the sum of the joint probability throughout the four regions equals 1 (Equation 9), the probabilities of the four regions cannot all be zero (the FDG would be incorrect). Figure 15.b shows the 16 combinations. The joint probability domain is on the left. An X is written in one of the four regions if and only if the sum of the probabilities in that region is greater than zero. The only possible structure obtained from the corresponding joint probability composed of the vertices $\omega_1$ and $\omega_2$ and some second order relations is on the right (Figure 15.a).

Note that it is not "logical" for both *occurrence* and *antagonism* relations to be satisfied at the same time between two elements (Equation 31, Section 4.3.3.). If two elements cannot exist in the same AG (antagonism) one of them cannot always exists when the other exists (occurrence). This combination only appears in cases in which one of the elements is null (cases 1 to 5); that is, it is synthetically created element. Moreover, the 16$^{th}$ combination is impossible in a correct FDG and, therefore, the four second-order relations cannot appear between two graph elements at the same time.

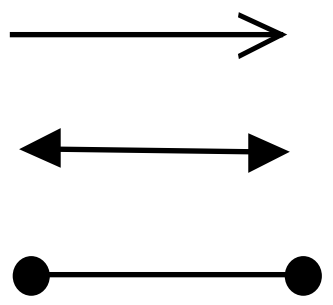

$O_\omega(\omega_1, \omega_2) = 1$

$A_\omega(\omega_1, \omega_2) = 1$

$E_\omega(\omega_1, \omega_2) = 1$

Figure 15.a.

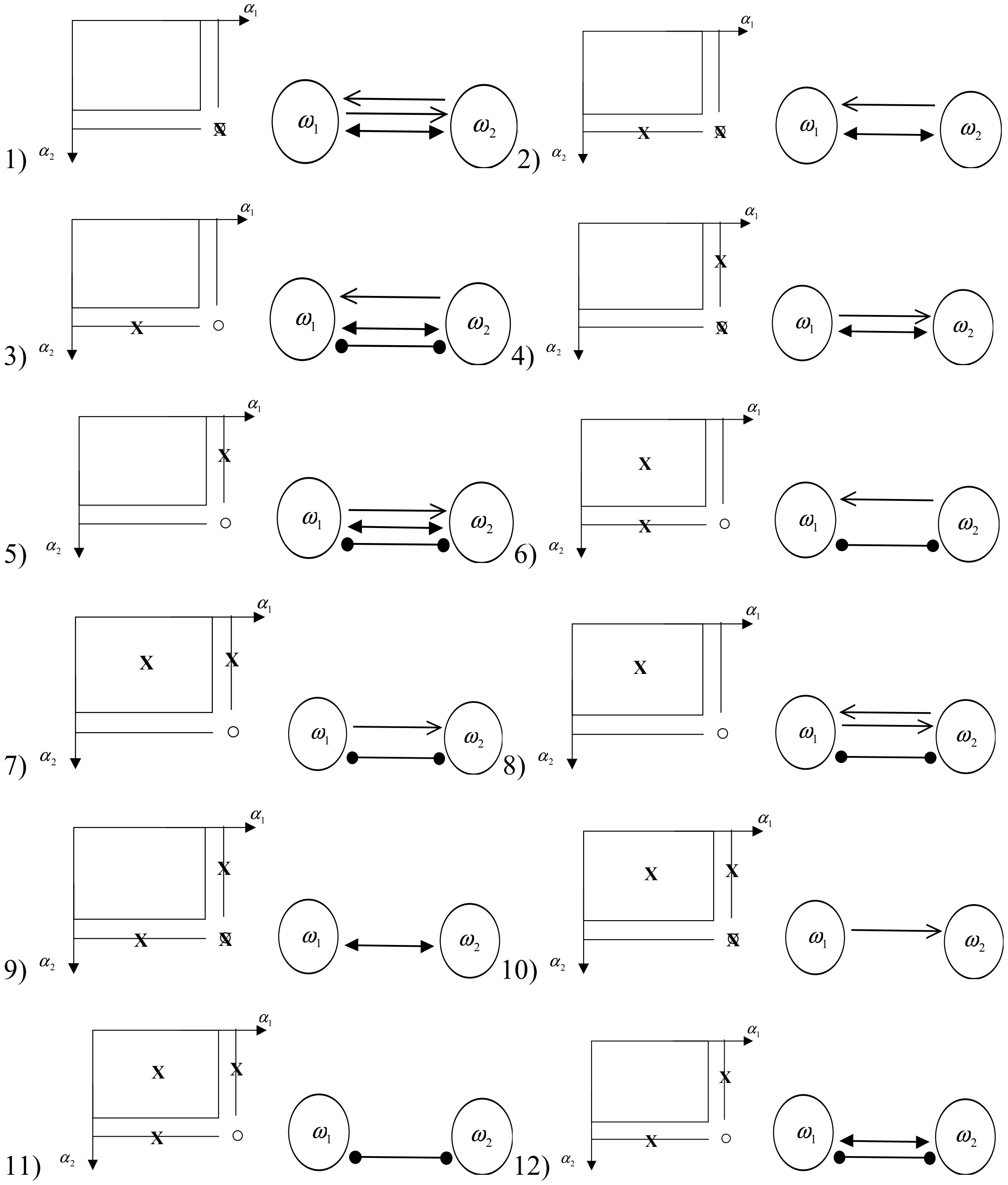

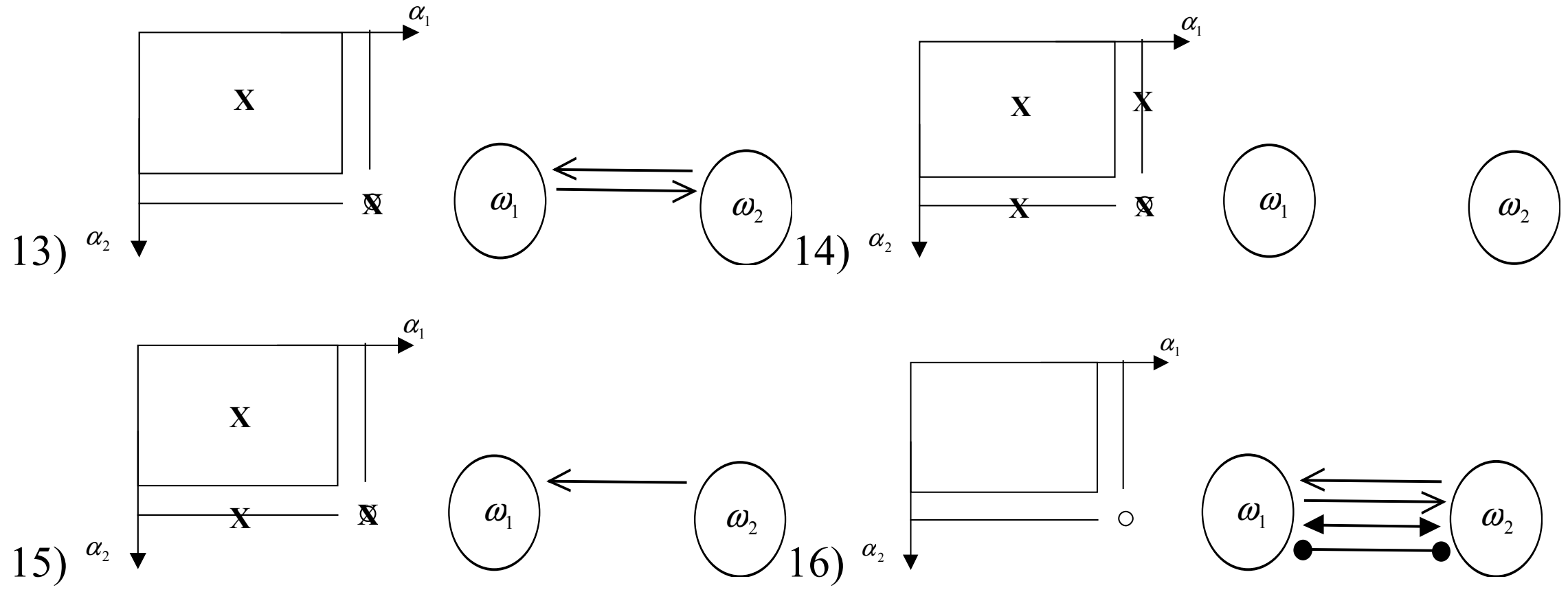

Figure 15. Sixteen combinations of the joint probability.

Tables 8 and 9 show the cost of matching the AG vertices $v_1$ and $v_2$ to the FDG vertices $\omega_1$ and $\omega_2$, respectively, in the 15 possible combinations of the FDG vertices. The costs in Table 8 are computed by applying the distance measure with restrictions $C_f$ (Section 5.2), whereas the costs in Table 9 are computed by applying the distance measure in which the second order constraints have been relaxed $C_f^R$ (Section 5.3).

The first order probability costs $C_1$, $C_1 \in [0,1]$, and $C_2$, $C_2 \in [0,1]$, depend on the attribute values and are not specified. Unlike edit distances (Sanfeliu and Fu, 1983), a *strict deletion* (d1 for the mapping $(v_1,\omega_1)$ or d2 for the mapping $(v_2,\omega_2)$) appears when a null AG vertex is matched to an FDG vertex with zero probability of being null. An *insertion* (i1 for the mapping $(v_1,\omega_1)$ or i2 for the mapping $(v_2,\omega_2)$) appears when a non-null AG vertex is matched to a null FDG vertex. Non-allowed labellings in the distance with restrictions are represented by the symbol $\infty$ in Table 8. The costs associated with the second order relations in the distance relaxing constraints are represented by $C_A$ (antagonism), $C_E$ (existence) and $C_O$ (occurrence) in Table 9.

| $a_1$ | $\neq \Phi$ | $\neq \Phi$ | $\Phi$ | $\Phi$ |
|---|---|---|---|---|
| $a_2$ | $\neq \Phi$ | $\Phi$ | $\neq \Phi$ | $\Phi$ |
| Case 1 | $1+1$ (i1,i2) | $1+0$ (i1) | $0+1$ (i2) | $0+0$ |
| Case 2 | $C_1+1$ (i2) | $C_1+1$ | $C_1+1$ (i2) | $C_1+0$ |
| Case 3 | $C_1+1$ (i2) | $C_1+0$ | $1+1$ (i2, d1) | $1+0$ (d1) |
| Case 4 | $1+C_2$ (i1) | $1+C_2$ (i1) | $0+C_2$ | $0+C_2$ |
| Case 5 | $1+C_2$ (i1) | $1+1$ (i1,d2) | $0+C_2$ | $0+1$ (d2) |
| Case 6 | $C_1+C_2$ | $C_1+C_2$ | $\infty$ (d1) | $\infty$ (d1) |
| Case 7 | $C_1+C_2$ | $\infty$ (d2) | $C_1+C_2$ | $\infty$ (d2) |
| Case 8 | $C_1+C_2$ | $\infty$ (d2) | $\infty$ (d1) | $\infty$ (d1,d2) |
| Case 9 | $\infty$ | $C_1+C_2$ | $C_1+C_2$ | $C_1+C_2$ |
| Case 10 | $C_1+C_2$ | $\infty$ | $\infty$ | $C_1+C_2$ |
| Case 11 | $C_1+C_2$ | $C_1+C_2$ | $C_1+C_2$ | $\infty$ |
| Case 12 | $\infty$ | $C_1+C_2$ | $C_1+C_2$ | $\infty$ |
| Case 13 | $C_1+C_2$ | $\infty$ | $\infty$ | $C_1+C_2$ |
| Case 14 | $C_1+C_2$ | $C_1+C_2$ | $C_1+C_2$ | $C_1+C_2$ |
| Case 15 | $C_1+C_2$ | $C_1+C_2$ | $C_1+C_2$ | $C_1+C_2$ |

Table 8. Cost with restrictions on the 15 combinations of two vertices

The distance in which the 2nd order constraints have been relaxed has been introduced to prevent the labelling from being rejected due to a strict deletion or to the non-fulfilment of a second-order constraint. For instance, in case 8 (both vertices appear in all the AGs which are used to synthesise the FDG), when both labelled AG vertices are non-null, the cost is $C_1+C_2$ in both distances. Nevertheless, if one of the AG vertices is null, then the labelling is not allowed in the distance with second order restrictions but only one cost is added in the distance in which the 2nd order constraints have been relaxed, $C_1+1+C_O$.

| $a_1$ | $\neq \Phi$ | $\neq \Phi$ | $\Phi$ | $\Phi$ |
|---|---|---|---|---|
| $a_2$ | $\neq \Phi$ | $\Phi$ | $\neq \Phi$ | $\Phi$ |
| Case 1 | $1+1$ (i1,i2) | $1+0$ (i1) | $0+1$ (i2) | $0+0$ |
| Case 2 | $C_1+1$ (i2) | $C_1+1$ | $C_1+1$ (i2) | $C_1+0$ |
| Case 3 | $C_1+1$ (i2) | $C_1+0$ | $1+1$ (i2, d1) | $1+0$ (d1) |
| Case 4 | $1+C_2$ (i1) | $1+C_2$ (i1) | $0+C_2$ | $0+C_2$ |
| Case 5 | $1+C_2$ (i1) | $1+1$ (i1,d2) | $0+C_2$ | $0+1$ (d2) |
| Case 6 | $C_1+C_2$ | $C_1+C_2$ | $1+C_2+C_O$ (d1) | $C_1+C_2+C_E$ (d1) |
| Case 7 | $C_1+C_2$ | $C_1+1+C_O$ (d2) | $C_1+C_2$ | $C_1+C_2+C_E$ (d2) |
| Case 8 | $C_1+C_2$ | $C_1+1+C_O$ (d2) | $1+C_2+C_O$ (d1) | $C_1+C_2+C_E$ (d1,d2) |
| Case 9 | $C_A$ | $C_1+C_2$ | $C_1+C_2$ | $C_1+C_2$ |
| Case 10 | $C_1+C_2$ | $C_1+C_2+C_O$ | $C_1+C_2+C_O$ | $C_1+C_2$ |
| Case 11 | $C_1+C_2$ | $C_1+C_2$ | $C_1+C_2$ | $C_1+C_2+C_E$ |
| Case 12 | $C_1+C_2+C_A$ | $C_1+C_2$ | $C_1+C_2$ | $C_1+C_2+C_E$ |
| Case 13 | $C_1+C_2$ | $C_1+C_2+C_O$ | $C_1+C_2+C_O$ | $C_1+C_2$ |
| Case 14 | $C_1+C_2$ | $C_1+C_2$ | $C_1+C_2$ | $C_1+C_2$ |
| Case 15 | $C_1+C_2$ | $C_1+C_2$ | $C_1+C_2$ | $C_1+C_2$ |

Table 9. Cost relaxing restrictions on the 15 combinations of two vertices.

# 6. Algorithms for computing the distance measures

A reasonable choice for the set of valid morphisms is to take $H = F_1 \cap F_2$ , i.e. a one-to-one mapping is required between the vertices of the extended graphs, and the arc mapping is determined from the vertex mapping assuming structural coherence. In addition, constraint $R_5$ (the relative order between arcs has to be preserved) should be incorporated if the application deals with planar graphs; in this case, $H = F_1 \cap F_2 \cap F_5$. The second-order constraints $R_3$ and $R_4$ may be applied as hard restrictions, i.e. $H = F_1 \cap F_2 \cap F_3 \cap F_4 \cap F_5$, or relaxed into second-order costs, as discussed before.

The branch and bound technique applied on the FDG problem is first presented in section 6.1. The computation of the distance with second order restrictions and the distance relaxing these restrictions is studied in Sections 6.2 and 6.3, respectively. The complexity of the distance computation is presented in section 6.4. Section 6.5 shows some experiments with random graphs in which the matching algorithms are used.

## *6.1. Branch and bound technique*

The distance and the optimal morphism between an AG and an FDG are calculated by an algorithm for error-tolerant graph matching. Our approach is based on a tree search by the A* algorithm, where the search space is reduced by a *branch and bound* technique. The algorithm searches a tree where the nodes represent possible mappings between vertices of both graphs and branches represent combinations of pairs of graph vertices that satisfy the labelling constraints. Hence, the path from the root to the leaves represent allowed labellings *f*, which belong to *H*.

The distance measure has been theoretically defined so that both graphs are extended to have the same number of elements and to be complete. Nevertheless, in practice, our algorithm only needs the FDG to be extended with one null vertex, because the different permutations of the null vertices are regarded as equivalent labellings. Thus, the AG spurious vertices are possibly matched (through an actually non-injective mapping) to this unique null FDG vertex. On the other hand, the FDG graph elements that remain unmatched when arriving at a leaf are considered to be matched with null AG vertices $v_\Phi$ or null AG arcs $e_\Phi$. Consequently, a final cost of deleting these elements is added to the cost of the labelling in the leaves of the search tree.

The constraint $R_1$ ($f$ has to be bijective) is fulfilled by constructing the search tree for all the vertices except for the null vertex of the FDG which has been introduced. The null vertex is permitted to be the image of several AG vertices. The restriction $R_2$ ($f$ has to be coherent structurally) is intrinsically fulfilled since $f_e$ is determined by $f_v$, which is given by the current path. The constraint $R_5$ (the relative order between arcs has to be preserved) is easily verified by applying its definition at each node of the search tree. Finally, the verification of the second-order constraints $R_3$ and $R_4$ is discussed in Section 6.2.

In general, solving a branch and bound problem requires a *branch evaluation function* and a *global evaluation function*. The former assigns a cost to the branch incident to a node *N* of the tree, which is the cost of the new match (or pair) appended. The latter is used to guide the search at a node *N* and refers to the cost of the best complete path through *N* (i.e. including the pairs of vertices already matched when arriving at *N*). In our case, we also must define the *deleting function*, which evaluates the cost of deleting the unmatched graph elements of the FDG when a leaf is reached.

The cost of a labelling *f* is given by the value of the *FDG global evaluation function* in the corresponding leaf, $T_f$, of the search tree. For the distance measure which uses restrictions,

$$C_{f\in H} = l^*\left(T_f\right) \tag{105}$$

and, for the distance measure which relaxes second-order constraints,

$$C^R_{f\in H} = l^{R*}\left(T_f\right) \tag{106}$$

The global evaluation function $l^*$ (see section 6.2), is a function which is reminiscent of the one presented in (Wong, You and Chan, 1990), and in which some terms have been redefined using our notation and others have been updated to solve the FDG matching problem. The global evaluation function $l^{R*}$, which is described in Section 6.3, also takes into account the second-order costs coming from the non-fulfilment of the FDG Boolean functions of antagonism, occurrence and existence.

## *6.2. Computation of the distance measure using restrictions*

Each node *N* of the search tree at level $p > 0$ is described by a collection of pairs of vertices of the graphs, $N = \{(v_i, \omega_{q_i})\}$, where $i = 1,2,...,p$ corresponds to the vertex indices of the vertices in the AG $v_i$ and $q_i$ are the various vertex indices of the vertices in the FDG $\omega_{q_i}$ such that $f_v(v_i) = \omega_{q_i}$. We also define the sets $N_v = \{v_1, v_2, ..., v_p\}$ and $N_\omega = \{\omega_{q_1}, \omega_{q_2}, ..., \omega_{q_p}\}$ of vertices that have already been matched between both graphs, and the sets $M_v = \{v_{p+1}, v_{p+2}, ..., v_n\}$ and $M_\omega = \{\omega_j \mid \omega_j \notin N_\omega\}$ of vertices that have not been matched yet. Assume that $N = \{(v_1, \omega_{q_1}), (v_2, \omega_{q_2}), ..., (v_p, \omega_{q_p})\}$ indicates the only path from the root to a tree node *N* and $T = \{(v_1, \omega_{q_1}), (v_2, \omega_{q_2}), ..., (v_n, \omega_{q_n})\}$ indicates the only path from the root to a leaf *T*.

The *branch evaluation function K* is defined as the cost of the new match between vertices plus the cost of all the arcs related to these two vertices. This match involves vertices from $N_v$ and $N_\omega$. Thus, the cost assigned to the branch incident to *N* is given by

$$K(v_p, \omega_{q_p}) = K_1 * C_{f_v}(v_p, \omega_{q_p}) + K_2 * \sum_{s=1}^{p-1} \left( C_e(e_{ps}, \varepsilon_{q_p q_s}) + C_e(e_{sp}, \varepsilon_{q_s q_p}) \right) \tag{107}$$

where $C_e(e, \varepsilon)$ is the cost of matching the arc $e$ in the AG to the arc $\varepsilon$ in the FDG, which is computed depending on the existence of both arcs as follows

$$C_e(e,\varepsilon)=\begin{cases} C_{f_e}(e,\varepsilon) & if\ e\in\Sigma_e \wedge \varepsilon\in\Sigma_\varepsilon \\ C_{f_e}(e,\varepsilon_\Phi)=1 & if\ e\in\Sigma_e \wedge \varepsilon\notin\Sigma_\varepsilon \\ C_{f_e}(e_\Phi,\varepsilon) & if\ e\notin\Sigma_e \wedge \varepsilon\in\Sigma_\varepsilon \\ C_{f_e}(e_\Phi,\varepsilon_\Phi)=0 & if\ e\notin\Sigma_e \wedge \varepsilon\notin\Sigma_\varepsilon \end{cases} \tag{108}$$

The *global evaluation function* $l^*(N)$ at a node *N* of level *p* is defined as the cost $g^*(N)$ of an optimal path from the root to the node *N* plus the cost $h^*(N)$ of an optimal path from the node *N* to a leaf $T=\{(v_i,\omega_{q_i}) \mid i=1,2,...,n\}$ constrained to be reached through the node *N*:

$$l^*(N)=g^*(N)+h^*(N) \tag{109}$$

$$g^*(N)=\sum_{i=1}^{p} K(v_i,\omega_{q_i}) \tag{110}$$

$$h^*(N)=\min_t\left\{\sum_{i=p+1}^{n} K(v_i,\omega_{q_i})+d(T)\right\} \tag{111}$$

where *t* denotes a feasible path from *N* to *T*, and $d(T)$ is the *deleting function*, which computes the cost of deleting the FDG vertices that have not been matched in a leaf,

$$d(T)=K_1 * \sum_{\forall \omega_j \in M_\omega} C_{f_v}(v_\Phi,\omega_j) \tag{112}$$

Note that the arcs $\varepsilon_{ij}$ in which one of the extreme vertices is not matched, $\omega_i \in M_\omega$ or $\omega_j \in M_\omega$, are not considered in the deleting function, since the cost of deleting these arcs is always zero.

The global evaluation function is then defined in a leaf $T$ as the cost $g^*(T)$ of an optimal path from the root to the leaf plus the cost of deleting the unmatched graph elements $d(T)$,

$$l^*(T) = g^*(T) + d(T) \tag{113}$$

Moreover, the global evaluation function $l^*(N)$ is unknown in an inner node $N$, since $h^*(N)$ is also unknown, and can only be approximated by a consistent lower-bounded estimate.

For that purpose, let $K'(v_i, \omega_j)$ be the cost of adding a pair of vertices to $N$, where $v_i \in M_v$ and $\omega_j \in M_\omega$, defined as

$$K'(v_i, \omega_j) = K_1 * C_{f_v}(v_i, \omega_j) + K_2 * \sum_{s=1}^{p} \left( C_e(e_{is}, \varepsilon_{jq_s}) + C_e(e_{si}, \varepsilon_{q_s j}) \right) \tag{114}$$

Then, for each unmatched vertex $v_i \in M_v$, a corresponding vertex $\omega_j \in M_\omega$ can be associated so that the cost $K'(v_i, \omega_j)$ is minimised. Next, from the sum of the minimal $K'$ for all the unmatched vertices, a consistent lower bounded estimate $h(N)$ of $h^*(N)$ is yielded,

$$h(N) = \sum_{i=p+1}^{n} \min_{\forall \omega_j \in M_\omega} \{ K'(v_i, \omega_j) \} \tag{115}$$

Finally, the *heuristic function* $l(N)$ that estimates $l^*(N)$ in a node $N$ is given by

$$l(N) = g^*(N) + h(N) \tag{116}$$

Note that, in a leaf $T$, $h(T) = 0$, and therefore, $l(T) \le l^*(T)$. It is also interesting to observe that the function $h(N)$ does not include a term for estimating the value of the deleting function $d(T)$, since it is hard to give a consistent lower bounded estimate of it (other than zero).

We move on to study the satisfaction of the second-order restrictions $R_3$ and $R_4$ by a labelling function. When the algorithm arrives at any leaf *T*, if $\omega_i \in N_\omega$ then it is sure that $\omega_i$ has been matched to a non-null AG vertex with a certain attribute value $a_i \neq \Phi$; however, if $\omega_i \in M_\omega$ then it is considered to be matched to a null AG vertex $v_\Phi$ (with $a_\Phi = \Phi$). After these considerations, the labelling $f$ does not belong to $F_3$ (it does not satisfy $R_3$, that is, it does not satisfy equation (64)) if the following rule is fulfilled at the corresponding leaf $T_f$:

$$\forall \omega_i, \omega_j \in \Sigma_\omega : \begin{Bmatrix} (\omega_i \in N_w \wedge \omega_j \in N_w \wedge A_\omega(\omega_i, \omega_j) = 1) \vee \\ (\omega_i \in N_w \wedge \omega_j \in M_w \wedge O_\omega(\omega_i, \omega_j) = 1) \vee \\ (\omega_i \in M_w \wedge \omega_j \in M_w \wedge E_\omega(\omega_i, \omega_j) = 1) \end{Bmatrix} \Rightarrow f \notin F_3 \tag{117}$$

Likewise, the labelling *f* does not belong to $F_4$ (it does not satisfy $R_4$, that is, it does not satisfy equation (70)) if the following rule is fulfilled at $T_f$:

$$\forall \varepsilon_{ij}, \varepsilon_{pq} \in \Sigma_\varepsilon : \begin{Bmatrix} (e_{ab} \in \Sigma_e \wedge e_{cd} \in \Sigma_e \wedge A_\varepsilon(\varepsilon_{ij}, \varepsilon_{pq}) = 1) \vee \\ (e_{ab} \in \Sigma_e \wedge e_{cd} \in \Sigma'_e - \Sigma_e \wedge O_\varepsilon(\varepsilon_{ij}, \varepsilon_{pq}) = 1) \vee \\ (e_{ab} \in \Sigma'_e - \Sigma_e \wedge e_{cd} \in \Sigma'_e - \Sigma_e \wedge E_\varepsilon(\varepsilon_{ij}, \varepsilon_{pq}) = 1) \end{Bmatrix} \Rightarrow f \notin F_4 \tag{118}$$

where $f_e(e_{ab}) = \varepsilon_{ij}$, $f_e(e_{cd}) = \varepsilon_{pq}$, and $\Sigma'_e - \Sigma_e$ is the set of null arcs in the extended AG. If $\omega \in N_\omega$ in a tree node then $\omega \in N_\omega$ in a leaf of the same path but, if $\omega \in M_\omega$ in a tree node, then it is not possible to know if $\omega \in M_\omega$ or $\omega \in N_\omega$ in a leaf of the same path. For this reason only the conditions that involve vertices that belong to $N_\omega$ can be evaluated in the tree nodes while the other conditions have to be evaluated in the leaves. Hence, some antagonism constraints can be evaluated in the inner nodes and therefore may be useful for pruning the search tree. The occurrence and existence constraints can only be evaluated in the leaves, and therefore are not helpful for pruning purposes. Therefore, at any inner node *N*, the following decision is taken about the satisfaction of the second-order constraints on the vertices by the partially defined function *f*,

$$\begin{array}{ll} f \notin F_3 & \textit{if } \omega_i \in N_w \wedge \omega_j \in N_w \wedge A_\omega(\omega_i, \omega_j) = 1 \\ f \in F_3 & \textit{otherwise} \end{array} \tag{119}$$

For the second-order restrictions on the arcs, however, the decision rule is

$$\begin{array}{ll} f \notin F_4 & \textit{if } \begin{pmatrix} \omega_i \in N_w \wedge \omega_j \in N_w \wedge \omega_p \in N_w \wedge \omega_q \in N_w \wedge \\ e_{ab} \in \Sigma_e \wedge e_{cd} \in \Sigma_e \wedge A_\varepsilon(\varepsilon_{ij}, \varepsilon_{pq}) = 1 \end{pmatrix} \\ f \in F_4 & \textit{otherwise} \end{array} \tag{120}$$

where $f_e(e_{ab}) = \varepsilon_{ij}$ and $f_e(e_{cd}) = \varepsilon_{pq}$ are already included in the partially defined function $f$ when the four vertices $\omega_i, \omega_j, \omega_p, \omega_q$ belong to $N_w$.

Finally, at any leaf *T* reached, the conditions (66), (68), (72), (74) are checked to establish whether the complete function $f$ satisfies both $R_3$ and $R_4$ or not. It is only at that moment that the occurrence and existence constraints are verified.

Algorithm 1 computes the distance measure $d_f$ and the corresponding optimal labelling $f_{opt} \in H$ between a given AG and a given FDG. It only invokes the recursive procedure TreeSearch at the root node.

---

**Algorithm 1**: Distance-measure-between-AG-and-FDG

**Inputs**: $G$ and $F$ : A given AG and a given FDG.

**Outputs**: The distance measure $d_f$ and the optimal labelling $f_{opt} \in H$ .

**Begin**

$d_f := \text{BigNumber}$ ; $f_{opt} := \varnothing$ ;

$\text{TreeSearch}(G, F, \varnothing, 0, v_1, d_f, f_{opt})$;

**End-algorithm**

---

**Procedure** $\text{TreeSearch}(G, F, f, g^*, v_i, d_f, f_{opt})$

**Input parameters:** $G$ and $F$ : An AG and an FDG;

$f$ : Optimal path (or labelling) from the root to the current node;

$g^*$ : Minimum value from the root to the current node;

$v_i$ : Current AG vertex to be matched;

**Input/Output parameters**: The best measure $d_f$ and the corresponding labelling $f_{opt} \in H$ obtained so far from the leaves already visited during the tree search.

**Begin**

**For** each vertex $\omega_j$ of the FDG $F$ not used yet in $f$ or $\omega_\Phi$ **do**

$K := \text{Branch-evaluation-function}(G, F, f, v_i, \omega_j)$

$h := \text{Bound-estimate-function}(G, F, f \cup \{f(v_i) = \omega_j\})$;

$l := g^* + K + h$ ; {Heuristic function of $l^*$}

**If** $f \cup \{f(v_i) = \omega_j\} \in H$ **and** $l < d_f$ **then**

**If** $i < n$ **then** {there is another vertex of the AG still not matched}

$\text{TreeSearch}(G, F, f \cup \{f(v_i) = \omega_j\}, g^* + K, v_{i+1}, d_f, f_{opt})$;

**Else** {all the vertices of the AG have been matched}

$d := \text{Deleting-function}(G, F, f \cup \{f(v_i) = \omega_j\})$;

**If** $d_f > g^* + K + d$ **then**

$d_f := g^* + K + d$ ; $f_{opt} := f \cup \{f(v_i) = \omega_j\}$;

**End-if**

**End-if**

**End-if**

**End-for**

**End-procedure**

---

### *6.3. Computation of the distance relaxing second-order constraints*

The FDG second-order functions affect the value of this distance measure, not by constraining the labelling but by increasing its cost. For this reason, three functions used in the *branch-and-bound* approach must be redefined: the branch evaluation function, $K^R(N)$; the cost of adding a pair of vertices in a tree node, $K'^R(N)$; and the cost of deleting the unmatched FDG vertices at a leaf, $d^R(T)$. When they are redefined, the set of valid labellings is $H = F_1 \cap F_2 \cap F_5$ or $H = F_1 \cap F_2$ depending on whether the planar graph condition is imposed or not.

The branch evaluation function is redefined by adding two more terms, $C_3$ and $C_4$, which refer to the second-order costs on the vertices and the arcs, respectively,

$$K^R\left(v_p, \omega_{q_p}\right) = K\left(v_p, \omega_{q_p}\right) + C_3\left(v_p, \omega_{q_p}\right) + C_4\left(v_p, \omega_{q_p}\right) \tag{121}$$

The term $C_3$ takes into account the possible second-order relations between the FDG vertex $w_{q_p}$ mapped in the tree node $N$ and all the FDG vertices mapped in the tree nodes from the root to $N$,

$$C_3\left(v_p, \omega_{q_p}\right) = \sum_{s=1}^{p-1} \begin{pmatrix} K_3 * C_{A_\omega}\left(v_p, v_s, \omega_{q_p}, \omega_{q_s}\right) + K_7 * C_{E_\omega}\left(v_p, v_s, \omega_{q_p}, \omega_{q_s}\right) + \\ K_5 * C_{O_\omega}\left(v_p, v_s, \omega_{q_p}, \omega_{q_s}\right) + K_5 * C_{O_\omega}\left(v_s, v_p, \omega_{q_s}, \omega_{q_p}\right) \end{pmatrix} \tag{122}$$

Considering that $a_p \neq \Phi$ and $a_s \neq \Phi : 1 \leq s \leq p-1$, since the AG has not been extended, then the costs associated with the occurrence relations (equation 99) and existence relations (equation 101) are always zero, and therefore,

$$C_3\left(v_p, \omega_{q_p}\right) = K_3 * \sum_{s=1}^{p-1} C_{A_\omega}\left(v_p, v_s, \omega_{q_p}, \omega_{q_s}\right) \tag{123}$$

Similarly, the term $C_4$ takes into account the possible second-order relations between those FDG arcs that connect $w_{q_p}$ and mapped vertices in the tree nodes from the root to $N$ and those FDG arcs in which both extremes have also been mapped in the tree nodes from the root to $N$,

$$C_4\left(v_p,\omega_{q_p}\right)=\sum_{s,t,r=1}^{p-1}\begin{pmatrix} K_4 * C_{A_\varepsilon}\left(e_{ps},e_{tr},\varepsilon_{q_pq_s},\varepsilon_{q_tq_r}\right)+K_4 * C_{A_\varepsilon}\left(e_{sp},e_{tr},\varepsilon_{q_sq_p},\varepsilon_{q_tq_r}\right)+ \\ K_8 * C_{E_\varepsilon}\left(e_{ps},e_{tr},\varepsilon_{q_pq_s},\varepsilon_{q_tq_r}\right)+K_8 * C_{E_\varepsilon}\left(e_{sp},e_{tr},\varepsilon_{q_sq_p},\varepsilon_{q_tq_r}\right)+ \\ K_6 * C_{O_\varepsilon}\left(e_{ps},e_{tr},\varepsilon_{q_pq_s},\varepsilon_{q_tq_r}\right)+K_6 * C_{O_\varepsilon}\left(e_{sp},e_{rt},\varepsilon_{q_sq_p},\varepsilon_{q_rq_t}\right)+ \\ K_6 * C_{O_\varepsilon}\left(e_{tr},e_{ps},\varepsilon_{q_tq_r},\varepsilon_{q_pq_s}\right)+K_6 * C_{O_\varepsilon}\left(e_{tr},e_{sp},\varepsilon_{q_tq_r},\varepsilon_{q_sq_p}\right) \end{pmatrix} \tag{124}$$

which can be simplified by eliminating the null terms (equations 100 and 102) as

$$C_4\left(v_p,\omega_{q_p}\right)=K_4 * \sum_{s,t,r=1}^{p-1} C_{A_\varepsilon}\left(e_{ps},e_{tr},\varepsilon_{q_pq_s},\varepsilon_{q_tq_r}\right)+C_{A_\varepsilon}\left(e_{sp},e_{tr},\varepsilon_{q_sq_p},\varepsilon_{q_tq_r}\right) \tag{125}$$

The cost of adding a pair of vertices to $N$ is redefined by adding two more terms, $C'_3$ and $C'_4$,

$$K'^R\left(v_i,\omega_j\right)=K'\left(v_i,\omega_j\right)+C'_3\left(v_i,\omega_j\right)+C'_4\left(v_i,\omega_j\right) \tag{126}$$

The term involving the second-order costs on the vertices, $C'_3$, is defined as

$$C'_3\left(v_i,\omega_j\right)=\sum_{s=1}^{p}\begin{pmatrix} K_3 * C_{A_\omega}\left(v_i,v_s,\omega_j,\omega_{q_s}\right)+K_7 * C_{E_\omega}\left(v_i,v_s,\omega_j,\omega_{q_s}\right)+ \\ K_5 * C_{O_\omega}\left(v_i,v_s,\omega_j,\omega_{q_s}\right)+K_5 * C_{O_\omega}\left(v_s,v_i,\omega_{q_s},\omega_j\right) \end{pmatrix} \tag{127}$$

and, considering that the AG vertices involved (non-null elements) have been matched to null or non-null FDG vertices and equations 99 and 101, this can be simplified as

$$C'_3\left(v_i,\omega_j\right)=K_3 * \sum_{s=1}^{p} C_{A_\omega}\left(v_i,v_s,\omega_j,\omega_{q_s}\right) \tag{128}$$

Similarly, the term involving the second-order costs on the arcs, $C'_4$, adds the costs between arcs whose external vertices have already been matched. After simplification, this term is given by

$$C'_4\left(v_i,\omega_j\right)=K_4 * \sum_{s,t,r=1}^{p} C_{A_\varepsilon}\left(e_{is},e_{tr},\varepsilon_{jq_s},\varepsilon_{q_tq_r}\right)+C_{A_\varepsilon}\left(e_{si},e_{tr},\varepsilon_{q_sj},\varepsilon_{q_tq_r}\right) \tag{129}$$

With $K'^R$ defined, the consistent lower bounded estimate is computed as

$$h^R(N)=\sum_{i=p+1}^{n}\min_{\forall \omega_j\in M_\omega}\left\{K'^R\left(v_i,\omega_j\right)\right\} \tag{130}$$

Finally, the second order costs of the relations between the unmatched elements in a leaf (that is to say, being $\omega_i \in M_\omega$ in *T*), have to be added to the cost of deleting the unmatched elements of the FDG. Recall that if $\omega_i \in M_\omega$ in a leaf then it is considered that it is matched to a null AG vertex $v_\Phi$ (with $a_\Phi = \Phi$ ).

$$d^R(T)=d(T)+C''_3(T)+C''_4(T) \tag{131}$$

where $C''_3(T)$ is the cost on the vertices defined as

$$C''_3(T)=\begin{cases}\displaystyle\sum_{\substack{\forall \omega_i,\omega_j\in M\omega\\ i<j}}\begin{pmatrix}K_3*C_{A_\omega}\left(v_\Phi,v_\Phi,\omega_i,\omega_j\right)+K_7*C_{E_\omega}\left(v_\Phi,v_\Phi,\omega_i,\omega_j\right)+\\ K_5*C_{O_\omega}\left(v_\Phi,v_\Phi,\omega_i,\omega_j\right)+K_5*C_{O_\omega}\left(v_\Phi,v_\Phi,\omega_j,\omega_i\right)\end{pmatrix}+\\ \displaystyle\sum_{\substack{\forall \omega_i\in M\omega\\ \forall \omega_{q_j}\in N\omega\\ i<qj}}\begin{pmatrix}K_3*C_{A_\omega}\left(v_\Phi,v_j,\omega_i,\omega_{q_j}\right)+K_7*C_{E_\omega}\left(v_\Phi,v_j,\omega_i,\omega_{q_j}\right)+\\ K_5*C_{O_\omega}\left(v_\Phi,v_j,\omega_i,\omega_{q_j}\right)+K_5*C_{O_\omega}\left(v_j,v_\Phi,\omega_{q_j},\omega_i\right)\end{pmatrix}+\\ \displaystyle\sum_{\substack{\forall \omega_j\in M\omega\\ \forall \omega_{q_i}\in N\omega\\ qi<j}}\begin{pmatrix}K_3*C_{A_\omega}\left(v_i,v_\Phi,\omega_{q_i},\omega_j\right)+K_7*C_{E_\omega}\left(v_i,v_\Phi,\omega_{q_i},\omega_j\right)+\\ K_5*C_{O_\omega}\left(v_i,v_\Phi,\omega_{q_i},\omega_j\right)+K_5*C_{O_\omega}\left(v_\Phi,v_i,\omega_j,\omega_{q_i}\right)\end{pmatrix}\end{cases} \tag{132}$$

This can be reduced to the following equation by considering equations (97), (99) and (101),

$$
C''_3(T)=\begin{cases} K_7 * \sum\limits_{\substack{\forall \omega_i,\omega_j \in M\omega \\ i<j}} C_{E_\omega}\left(v_\Phi, v_\Phi, \omega_i, \omega_j\right) + K_5 * \sum\limits_{\substack{\forall \omega_i \in M\omega \\ \forall \omega_{q_j} \in N\omega \\ i<qj}} C_{O_\omega}\left(v_j, v_\Phi, \omega_{q_j}, \omega_i\right) \\ + K_5 * \sum\limits_{\substack{\forall \omega_j \in M\omega \\ \forall \omega_{q_i} \in N\omega \\ qi<j}} C_{O_\omega}\left(v_i, v_\Phi, \omega_{q_i}, \omega_j\right) \end{cases} \tag{133}
$$

The cost on the arcs, $C''_4(T)$, has to consider all the combinations between matched and non-matched vertices that the arcs connect. After a simplification step, in which equations (98), (100) and (102) are considered, the cost is given by

$$
C''_4(T)=\begin{cases} K_8 * \sum\limits_{\substack{\forall \left\{\begin{smallmatrix}\omega_i,\omega_j \\ \omega_k,\omega_t\end{smallmatrix}\right\} \in M\omega \\ i<k \vee (i=k \wedge j<t)}} C_{E_\varepsilon}\left(e_\Phi, e_\Phi, \varepsilon_{ij}, \varepsilon_{kt}\right) + K_6 * \sum\limits_{\substack{\forall \omega_i,\omega_j \in M\omega \\ \forall \omega_{qk},\omega_{q_t} \in N\omega \\ i<k \vee (i=k \wedge j<t)}} C_{O_\varepsilon}\left(e_{kt}, e_\Phi, \varepsilon_{q_k q_t}, \varepsilon_{ij}\right) \\ + K_6 * \sum\limits_{\substack{\forall \omega_k,\omega_t \in M\omega \\ \forall \omega_{qi},\omega_{q_j} \in N\omega \\ i<k \vee (i=k \wedge j<t)}} C_{O_\varepsilon}\left(e_{ij}, e_\Phi, \varepsilon_{q_i q_j}, \varepsilon_{kt}\right) \end{cases} \tag{134}
$$

Algorithm 1 can be used to compute the distance measure $d_f^R$ and the corresponding optimal labelling $f_{opt}^R \in H$. It does so by modifying the three functions *Branch-evaluation-function*, *Bound-estimate-function* and *Deleting-function* in the procedure TreeSearch and removing the verification of $f \in F_3$ and $f \in F_4$ when $f \in H$ is evaluated. The above three functions have to be modified to compute $K^R$, $h^R(N)$, and $d^R(T)$, respectively.

### *6.4. Complexity of distance computation*

Taking into account the set of all possible labelling combinations of the vertices within the set of allowable mappings $H = F_1 \cap F_2$, and regarding the different permutations of the null vertices as equivalent labellings, the number of possible matches between the extended graphs increases to

$$\binom{n}{i} V_{n-i}^{m} = \frac{n!}{(n-i)!\,i!} \frac{m!}{(m-n+i)!} \tag{135}$$

where $m$ and $n$ are the number of vertices in the original graphs and $i$ is the number of AG vertices that are matched to the null FDG vertex. The extended graphs are assumed to have $m+n$ vertices. $\binom{n}{i}$ denotes the combinations of the $n-i$ matched vertices and $V_{n-i}^{m}$ denotes the combinations of the $i$ vertices matched to the null vertex, that is, the variations of $m$ elements taken in groups of *n-i* elements (see figure 16).

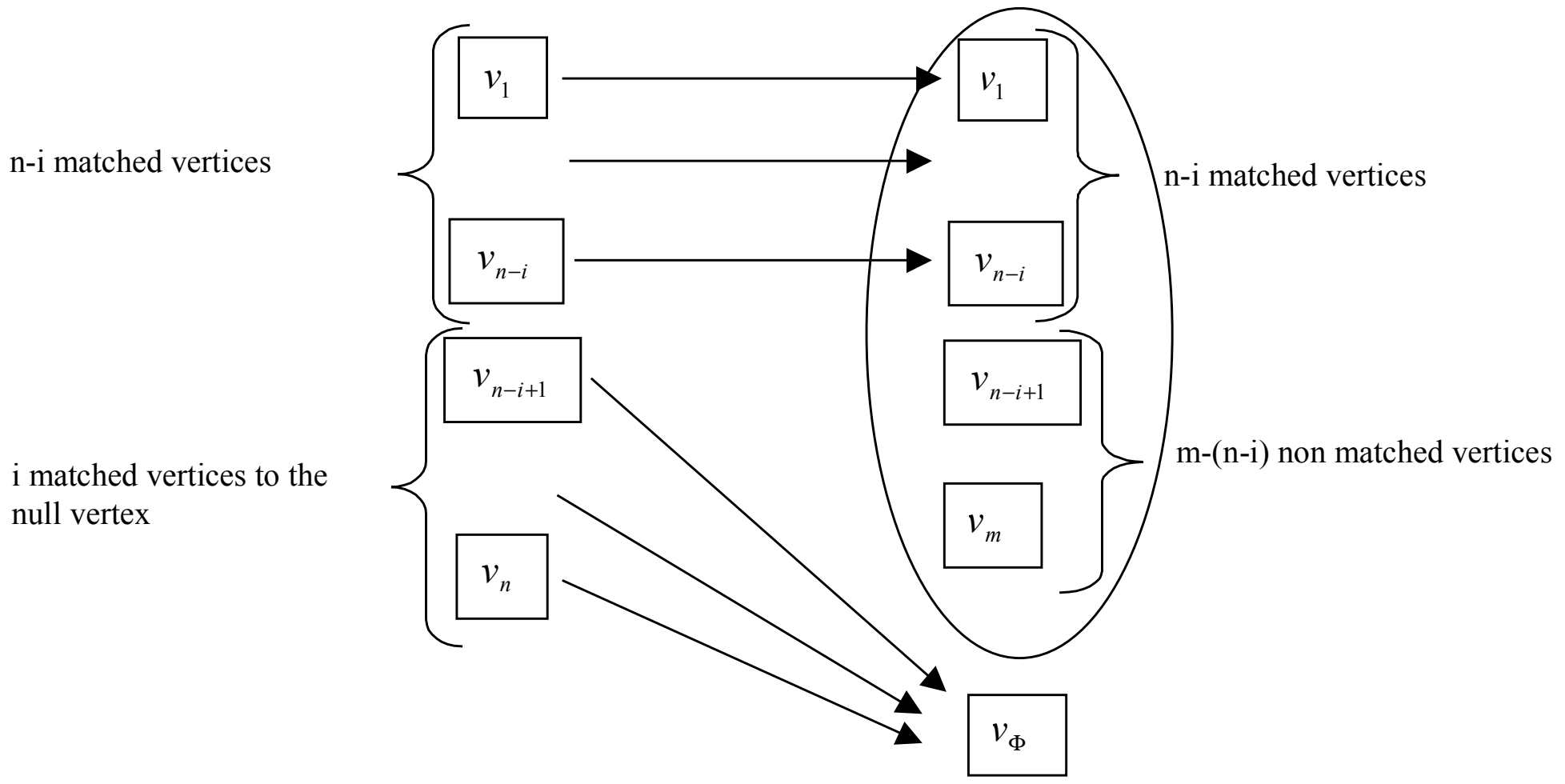

Figure 16. Labelling between two graphs with *n* and *m* vertices and *i* vertices mapped to the null vertex.

If we first consider the case $n \geq m$, the minimum number of vertices matched to the null one is $n-m$ and the maximum is $n$, so, the number of labellings is

$$Q_{n,m} = \sum_{i=n-m}^{n} \binom{n}{i} V_{n-i}^{m} \quad if \quad n \geq m \tag{136}$$

and considering equation (135),

$$Q_{n,m} = \sum_{i=n-m}^{n} \frac{n!\,m!}{(n-i)!\,i!\,(m-n+i)!} \quad if \quad n \geq m \tag{137}$$

which can be rewritten by,

$$Q_{n,m} = n!m!\sum_{k=0}^{m}\frac{1}{(k+n-m)!\,(m-k)!k!} \quad if \quad n \geq m \tag{138}$$

where $k = i - n + m$.

If the case $n < m$ is now considered, the minimum number of vertices matched to the null one is 0 and the maximum is $n$, so, the number of labellings is

$$Q_{n,m} = \sum_{k=0}^{n}\binom{n}{k} V_{n-k}^{m} \quad if \quad n < m \tag{139}$$

Considering equation (135) and reorganising the above one, the final expression is,

$$Q_{n,m} = n!m!\sum_{k=0}^{n}\frac{1}{(k+m-n)!\,(n-k)!k!} \quad if \quad n < m \tag{140}$$

Therefore, if an exhaustive search algorithm such as *backtracking* were used to compute the distance by evaluating the cost of each possible match and finding the minimum, the number of possible labellings would be $Q_{n,m}$. Figure 17 shows a possible search tree. Note that the number of siblings is reduced when the father is a vertex from the graph but it is kept unmodified if the father is the null vertex and also, that the number of labellings is the number of leaves.

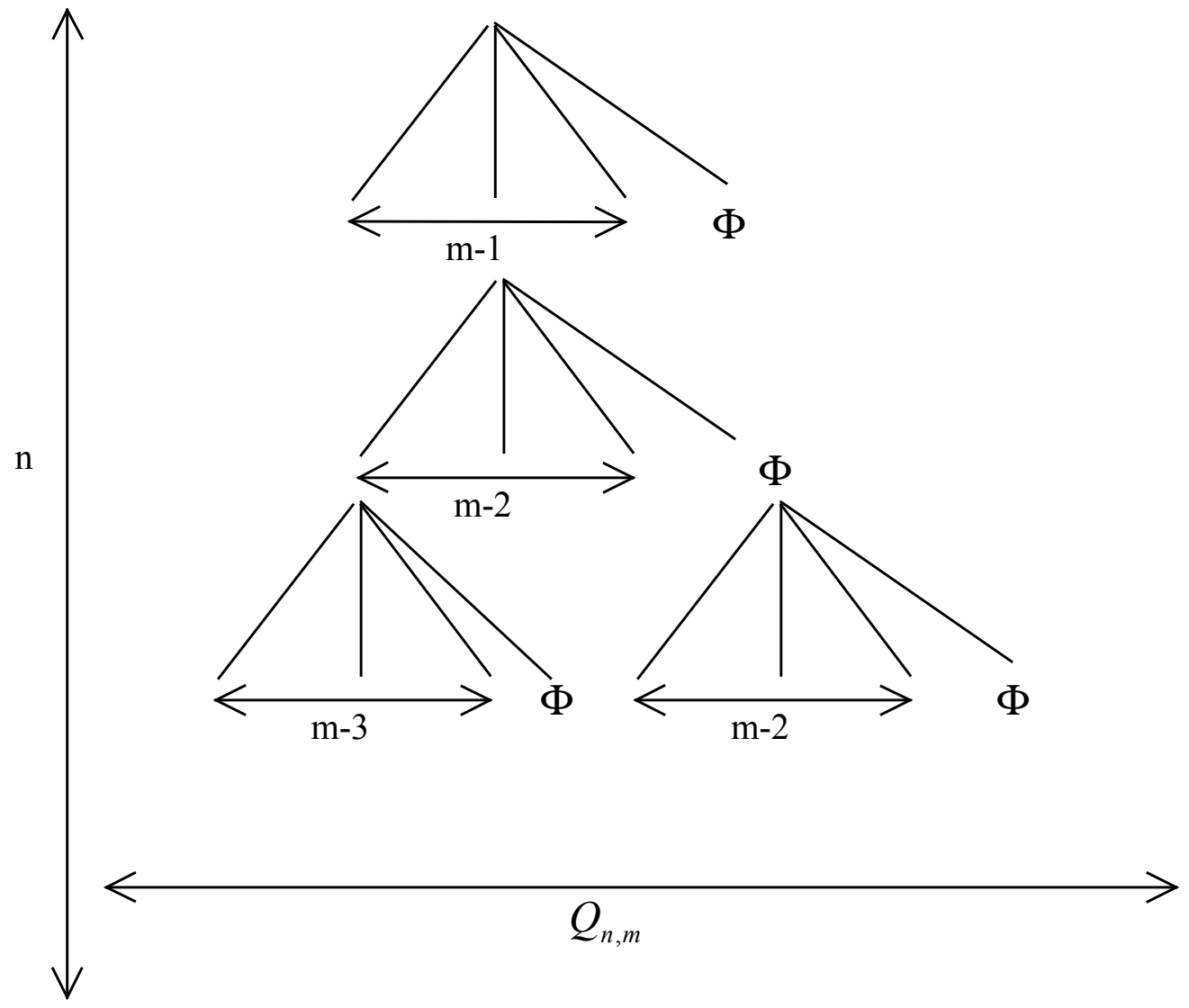

Figure 17. Search tree for two graphs of *n* and *m* number of vertices where $n \geq m$.

Nevertheless, the time complexity of the computation of the distance does not depend on the number of labellings but on the number of nodes in the tree search. Therefore, it is defined by

$$T_{n,m} = O_v \ A_{n,m} \tag{141}$$

where $O_v$ is the time complexity of processing one node and $A_{n,m}$ is the number of nodes of the tree search, which is computed by,

$$A_{n,m} = \sum_{i=0}^{n} Q_{i,m} \tag{142}$$

By incorporating additional constraints, such as the planar graph constraint $R_5$ and the second-order constraints $R_3$ and $R_4$ (i.e. $H = F_1 \cap F_2 \cap F_3 \cap F_4 \cap F_5$), the search tree can be somewhat pruned, although the combinatorial complexity remains. In real applications, the best way of pruning the search tree and still calculating the distance measure and the optimal labelling is by means of some sort of *branch and bound* algorithm (Wong, You and Chan, 1990) such as the one presented in Section 6.2. A good heuristic function in the branch and bound algorithm may prune the search tree quite considerably and speed up the process of finding the optimal solution impressively (Larrosa *et al.*, 1999). However, even in this case, the worst-case complexity remains non-polynomial. Efficient algorithms that compute sub-optimal approximations of the given distance measure are presented in the next chapter.

### 6.5. Experimental results

In order to examine the performance of the new matching algorithm in practice, we carried out a number of experiments with randomly generated graphs. The random graph generator was the same as the one explained in section 4.6 with the following parameter values:

| Initial | Graph | | | Generated | AGs |
|---|---|---|---|---|---|
| *nv* | *ne* | *nd* | *nl* | *nv'* | *Non-modified* |
| 27 | 108 | 21 | 2 | 6 | 4 |

Table 10. Parameters of the second experiments.

The number of FDGs was set to 10, $nFDG = 10$, and the number of AGs in the test set to 100, $NT = 100$. The number of AGs in the reference set for each FDG, $NR$, was set to 1, 3, 6, 9, 12, 15 and 18.

We were interested in measuring the ability of the branch and bound algorithm to prune the decision tree while applying different cost values to the antagonisms on the vertices. The reason why we did not test occurrence and existence relations is because they are not useful for pruning the search tree (Section 6.2). Hence, the weights on the costs were $K_1 = K_2 = 1$, $K_4 = K_5 = K_6 = K_7 = K_8 = 0$ and $K_3$ varied for different tests. In the first tests, no antagonisms were considered, $K_3 = 0$; in the second tests, the distance relaxing second order constraints were applied with $K_3 = 1$; and in the last tests, the distance with constraints were applied with $K_3 = Big_number$. Figure 18 shows the average of the number of vertices and antagonisms of the synthesised FDGs throughout the number of AGs. These results have been extracted from figures 11 and 12.

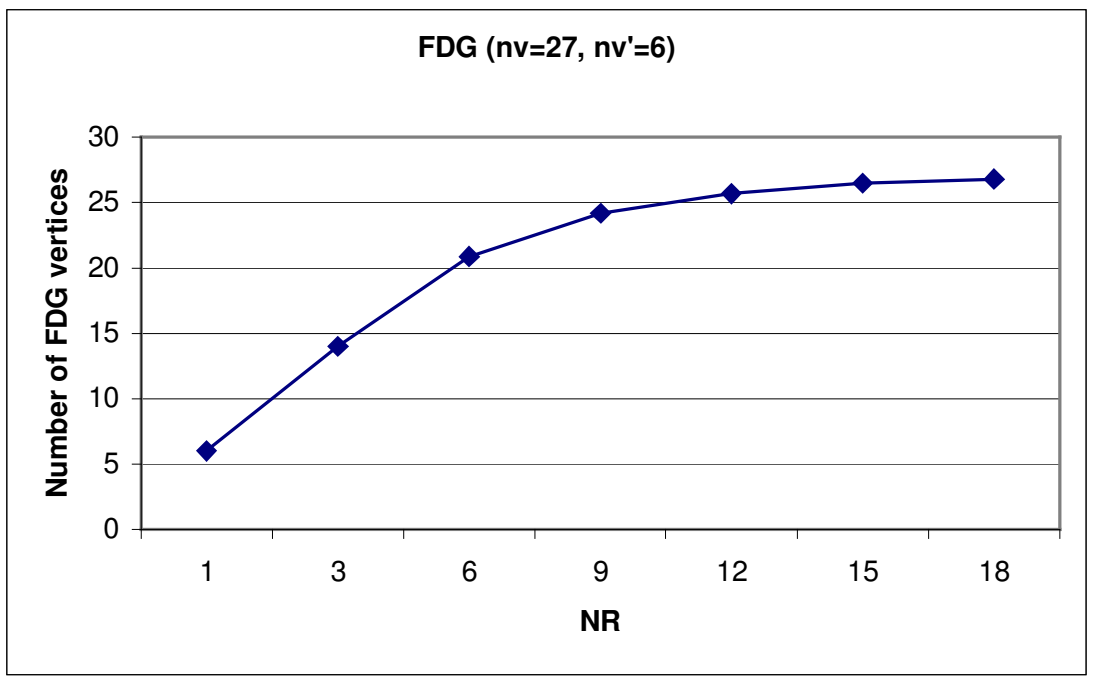

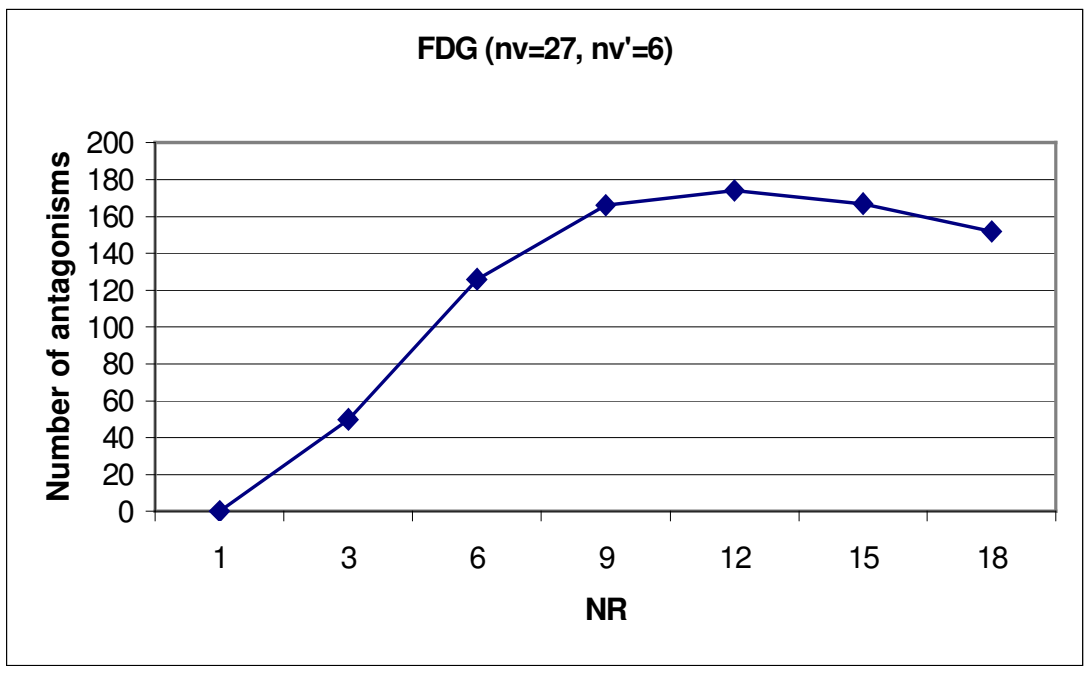

Figure 18. Average of the number of vertices and antagonisms in the FDGs.

Figure 19 shows the ratio of correctness. We observe that the correctness increases when the number of AGs used to synthesise the FDGs also increases. The best results appear when the antagonisms are considered with a cost. When there are few antagonisms (figure 18.b), the distance that considers the antagonisms as strict relations obtains better results than without considering the antagonisms. Nevertheless, when the number of antagonisms increases, the strict second-order relations discards too many labellings and so it is better not to consider them.

Moreover, the ratio is always better when the antagonisms are applied with a cost $K_3 = 1$ than when they are not applied ($K_3 = 0$) or strictly applied ($K_3 = Big_number$). In the extreme values of $NR$ (low or high) there are few antagonisms (figure 18) and so there is less difference between using the antagonisms or not but, in the values of $NR$ that the maximum number of antagonisms is obtained, the difference on the correctness applying or not the antagonisms is also maximum.

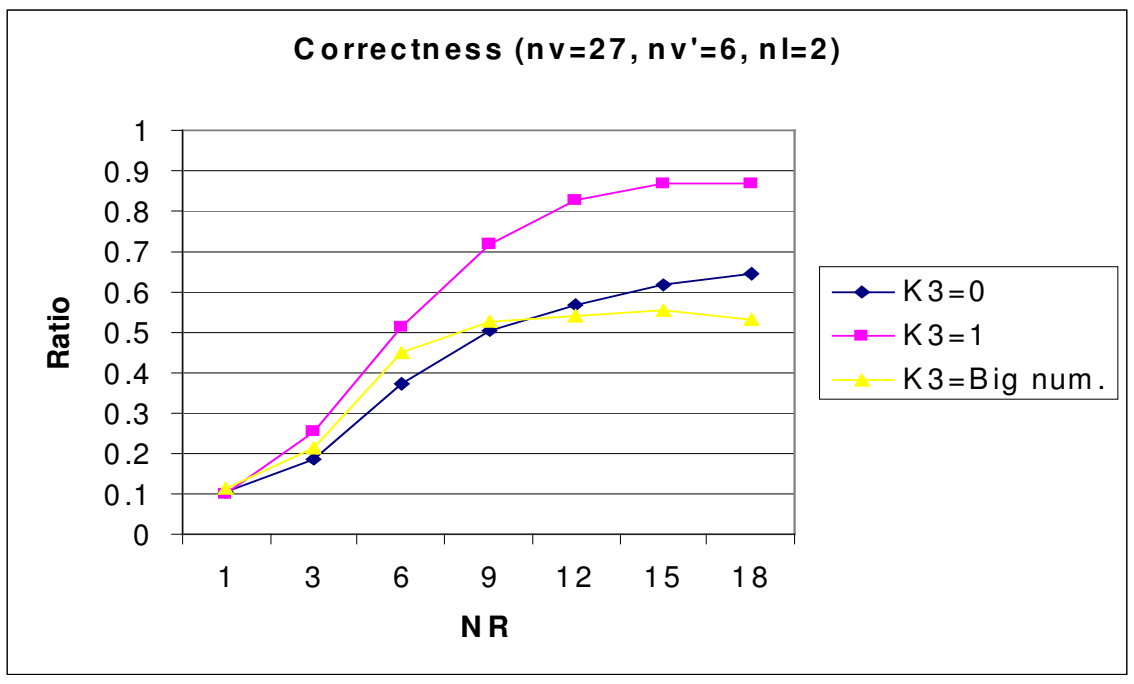

Figure 19. Correctness in the branch and bound algorithm.

Figure 20 shows the run time of the above experiments. The antagonism relations are useful to prune the search tree and decrease the run time. The maximum difference between the case in which the antagonisms are not used, $K3 = 0$, and the other two appear when the maximum number of antagonisms is obtained ($NR = 12$). The fastest tests are the ones in which $K_3 = Big_number$. This is because the branch and bound prunes more paths in the tree search. Nevertheless, since the correctness is lower, some of these paths were the optimal labellings.

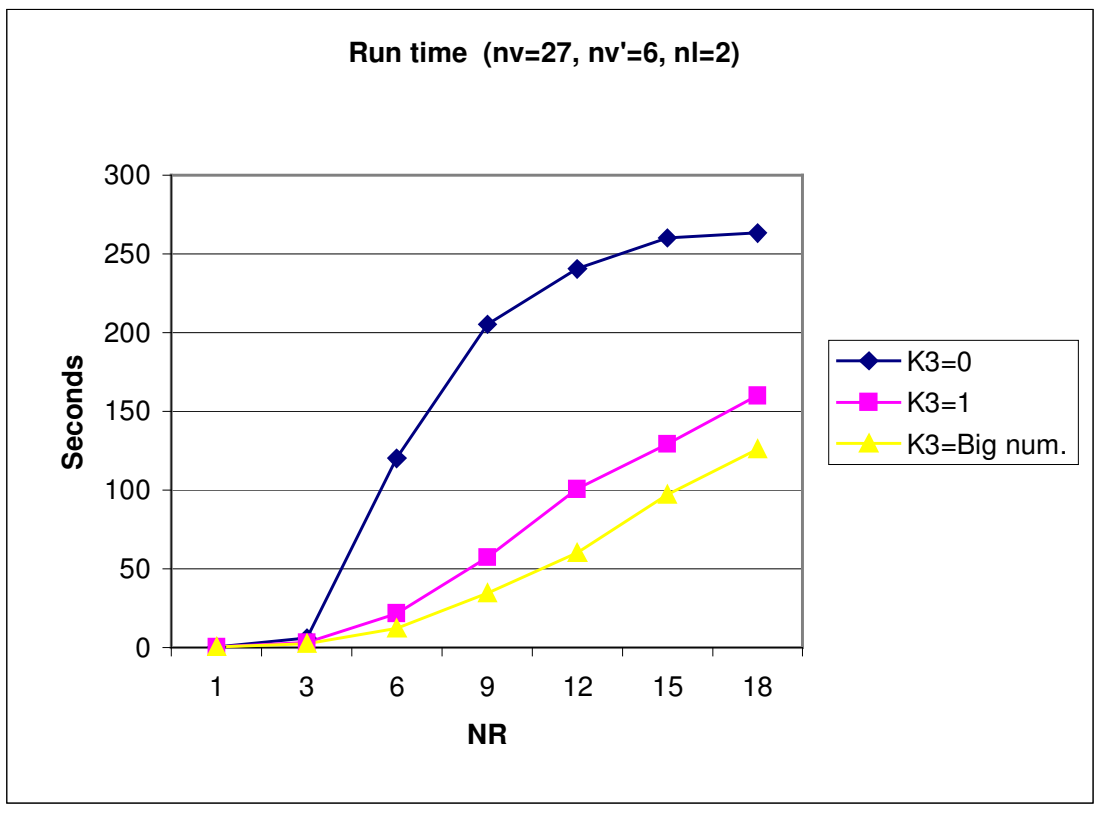

Figure 20. Run time in the branch and bound algorithm.

# 7. Efficient algorithms for computing sub-optimal distances

We propose a method which further reduces the search space of our branch and bound algorithm by initially discarding mappings between vertices. The match of a pair of vertices is discarded if the obtained degree of similarity between their related sub-units is lower than a threshold. Thus, the graphs are broken down into manageable sub-units that overlap, so that two adjacent sub-units contain information about each other through the mutual graph elements they contain.

A classical probabilistic relaxation scheme is presented in section 7.1. Then, the above commented sub-units are presented and defined in section 7.2 and 7.3, respectively. Moreover, a distance measure between sub-units and a matching algorithm to compute this distance are proposed in sections 7.4 and 7.5. We propose two techniques to compute a sub-optimal distance between AGs and FDGs. In the first, section 7.6, a non-iterative method is proposed to discard non-probable matches. In the second, section 7.7, two relaxation schemes are presented. The difference between them is the initialisation of the probabilities. Section 7.8 presents some results with random graphs.

## *7.1. Probabilistic relaxation schemes*

Relaxation schemes are optimisation techniques in which the variables of the scheme are iteratively updated in order to approach a stationary point of the update equations. They can be used to optimise a matching criterion which has a maximum at the stationary point. A classic probabilistic relaxation scheme is due to Rosenfeld, Hummel and Zucker (Rosenfeld et al., 1976), and was conceived as an object labelling algorithm. Since it's conception, the approach has been widely used for image processing tasks including graph matching.

We will denote the probability that the AG vertex $v_i$ matches to the FDG vertex $\omega_a$ at the iteration $t$ as $P_i^t(a)$. The Rosenfeld *et al.* Scheme specifies that the probabilities at iteration $t+1$ should be given by

$$P_i^{t+1}(a) = \frac{P_i^t(a)\left(1+Q_i^t(a)\right)}{\sum_{\omega_{a'} \in \Sigma_\omega} P_i^t(a')\left(1+Q_i^t(a')\right)} \tag{143}$$

where $Q_i^t(a)$ is called the support function. The numerator in the update formula can be viewed as the product of two evidential factors; the probability $P_i^t(a)$ just described which represents the local information and the support function $Q_i^t(a)$ which represents the probability of the surrounding matches given that $v_i$ matches $\omega_a$. The denominator simply ensures the normalisation of the probabilities, i.e. that $\sum_{\omega_a \in \Sigma_\omega} P_i^t(a) = 1$. This step is necessary since the support functions are not true probabilities. Rosenfeld *et al.* define the support function as

$$Q_i^t(a) = \sum_{v_j \in N(v_i)} d_{i,j} \sum_{\omega_b \in N(\omega_a)} r_{i,j}(a,b) P_j^t(b) \tag{144}$$

where $r_{i,j}(a,b)$ is called the compatibility function and $N(v_i)$ and $N(\omega_a)$ are the set of vertices which are connected to $v_i$ and $\omega_a$, respectively. The coefficients $d_{i,j}$ are used to make $Q_i^t(a)$ be in the range $[-1,1]$, provided that $\sum_{v_j \in N(v_i)} d_{i,j} = 1$.

In the original form, the $r_{i,j}(a,b)$ are purely arbitrary, application dependent and no method for their specification is offered. The initial probabilities, $P_i^0(a)$, are also application dependent.

### *7.2. Splitting the graphs into sub-units*

The larger the topological structure of the sub-units is, the more effective the scheme is at discarding unacceptable labellings. In these terms, small structural units perform badly and the contextual information of the matching process is impoverished. For instance, in the relaxation methods described in (Rosenfeld *et al.*, 1976; Feng *et al.*, 1994; O'leary and S. Peleg , 1983; Christmas *et al.*, 1995), the sub-units are composed of a pair of vertices and their connecting arcs, and the support function does not consider that the labelling has to be bijective (constraint $R_1$), or that the relative order between arcs has to be preserved (constraint $R_5$). If, on the other hand, the structural sub-units are too large, then those constraints can be fulfilled but the computational requirements of the matching process become excessively cumbersome; the limitation stems from the need to explore the space of mappings between these sub-units. As a compromise between representational power and computational requirements, Wilson and Hancock proposed splitting the graphs into sub-units which are structured by a central vertex and all the adjacent vertices are connected to it by an arc (Wilson and Hancock, 1997). For convenience we refer to these sub-units as *expanded vertices* (Figure 21). Their probabilistic approach has the main drawback that the semantic information is used only in the initialisation step while the support function is only based on symbolic information.

We present here three different approaches. The first approach takes the distance between expanded vertices with semantic knowledge as an informative heuristic for selecting a reduced set of configurations prior to the computation of an *approximation to the distance measure* through the branch and bound algorithm. In order to provide a good selection (which hopefully includes the optimal labelling) and reduce the intrinsically used heuristics, the distance measure between expanded vertices is taken to be the same as the one defined between the whole graphs. The main drawback of this approach is that the selection of a vertex is based only on local information. To solve this problem, two other approaches are presented which are based on the Rosenfeld relaxation method (Rosenfeld *et al.*). The advantage of this relaxation method is that the global information flows through the probabilities in each iteration of the algorithm. Nevertheless, the structural sub-units are too small to keep the structural knowledge. The difference between both approaches is how they initialise the probabilities. In the first, it is used the distance between expanded vertices, thus, the initial probabilities are located using some structural information. In the second, it is only used the distance between vertices.

### *7.3. Expanded Vertices*

An AG expanded vertex (Figure 21.a) or an FDG expanded vertex (Figure 21.b) are an AG or an FDG, respectively, with a specific structure formed by a *central vertex* and all the adjacent vertices, called *external vertices*, which are connected to it by an outgoing arc. More formally, an *AG expanded vertex* $EV_i$ over $(\Delta_v, \Delta_e)$ with an underlying graph structure $H = (\Sigma_v, \Sigma_e)$ is defined as an AG $EV_i = (V, A)$ where the vertices belong to a finite set $\Sigma_v = \{v_1, ..., v_i, ... v_n\}$ that contains a given vertex $v_i$ and the arcs belong to a finite set $\Sigma_e = \{e_{i,1}, ..., e_{i,i-1}, e_{i,i+1}, ... e_{i,n}\}$ formed by all the arcs departing from $v_i$ in the AG. The vertex $v_i$ is called the *central vertex* and the vertices $v_j : i \neq j, 1 \leq j \leq n$ are called the *external vertices*.

An AG *G* of order *k* over the domain $(\Delta_v, \Delta_e)$ with structure $H = (\Sigma_v, \Sigma_e)$ can be represented as a set of expanded vertices $\{EV_i\}$, such that $G = \bigcup_{i=1..k} EV_i$, where $v_i$ is the central vertex of $EV_i, \forall v_i \in \Sigma_v$.

It should be noted that when the whole underlying structure *H* of the AG is described by a set of expanded vertices, each arc appears just once (each arc belongs to just one expanded vertex) but vertices are possibly included more than once (each vertex is a central vertex of an expanded vertex and also an external vertex of as many expanded vertices as input arcs it receives).

An *FDG expanded vertex* $EW_i$ over $(\Delta_v, \Delta_e)$ with an underlying graph structure $H = (\Sigma_\omega, \Sigma_\varepsilon)$ is defined as an FDG $EW_i = (W, B, P, R)$ with a finite set of vertices $\Sigma_\omega = \{\omega_1, ..., \omega_i, ... \omega_m\}$ that contains a given vertex $\omega_i$ and a finite set of arcs $\Sigma_\varepsilon = \{\varepsilon_{i,1}, ..., \varepsilon_{i,i-1}, \varepsilon_{i,i+1}, ... \varepsilon_{i,n}\}$ formed by all the arcs departing from $\omega_i$ in the FDG. Again, the vertex $\omega_i$ is called the *central vertex* and the vertices $\omega_j : i \neq j, 1 \leq j \leq m$ are called the *external vertices*.

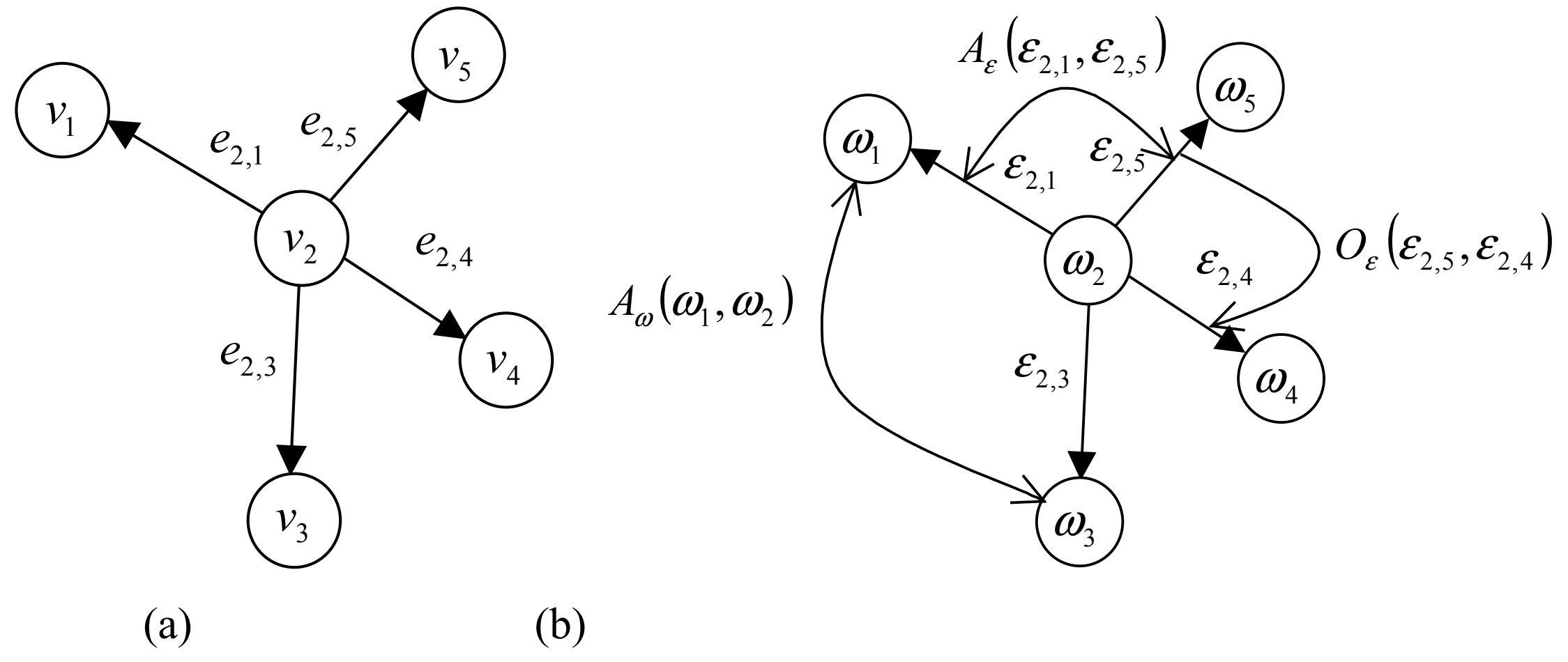

Figure 21. Expanded vertex of an AG (a) and an FDG (b).

An FDG $F$ can be partially represented in distributed form as the set of its expanded vertices $\{EW_i\}$, where $\omega_i$ is the central vertex of $EW_i, \forall \omega_i \in \Sigma_\omega$.

We note that the whole underlying structure of an FDG is included in this new representation; the arcs appear once but vertices may be included more than once. However, the antagonism, occurrence and existence relations of the original FDG cannot be represented entirely: those that involve two graph elements belonging to the same expanded vertex are included, but this is not possible when the related graph elements belong to different expanded vertices.

### 7.4. Distance measure between expanded vertices

The distance between an AG expanded vertex and an FDG expanded vertex is the same as the distance defined between an AG and an FDG. This is possible because expanded vertices can be seen as graphs with a specific structure. However, when both expanded vertices are non-null, the mapping $f$ between graph elements of expanded vertices $EV_i$ and $EW_j$ is constrained to satisfy the obvious requirement $f(v_i)=\omega_j$, i.e. the central vertices are mapped to one another.

Bearing this in mind, and not including the second-order costs, the maximum distance $d_{max}$ obtained between two non-null expanded vertices is

$$d_{max} = \begin{cases} 2n-1 & \text{if } n \geq m \\ n+m-1 & \text{if } n < m \end{cases} \tag{145}$$

where *n* and *m* are the number of vertices of the AG expanded vertex and the FDG expanded vertex, respectively. If $n \geq m$ then the maximum distance value is the cost of substituting $m-1$ arcs and *m* vertices plus the cost of inserting $n-m$ arcs and vertices. However, if $n < m$, the maximum value is the cost of substituting $n-1$ arcs and *n* vertices plus the cost of deleting $m-n$ AG vertices. The cost of deleting the AG arcs is zero when the external vertices have been deleted.

This maximum value is used in section 7.6 to normalise the distances between vertices.

## 7.5. A fast algorithm for computing the distance between expanded vertices

Algorithm 2 calculates the distance between non-null expanded vertices with a computational cost $O(n^2 \cdot m)$ where *n* and *m* are the number of vertices of the AG and FDG expanded vertices, respectively. It is based on the idea that if the labelling function has to be structurally coherent and the order between arcs has to be maintained, then the external vertices and their arcs can be regarded as a cyclic string where each element is a pair (arc, external vertex). Thus, the distance between expanded vertices is computed as the cost of matching the central vertices plus the distance between both cyclic strings. Gregor and Thomason showed that the distance between two cyclic strings can be obtained as the minimum of the distances between one orientation of the first string and all possible orientations of the second one (Gregor and Thomason, 1996).

In the case of an AG expanded vertex $EV_i$, the string is defined as $S_i = \{(e_{i(1)}, v_{i(1)}), \ldots, (e_{i(n')}, v_{i(n')})\}$ where $\{v_{i(1)}, \ldots, v_{i(n')}\}$ is an ordered set of external vertices of $EV_i$, $e_{i(j)} = e_{ik} \Leftrightarrow v_{i(j)} = v_k$ and *n'* is the number of external vertices, *n'*=*n*-1. Similarly, the string related to an FDG expanded vertex $EW_j$ is defined as $S_j = \{(\varepsilon_{i(1)}, \omega_{i(1)}), \ldots, (\varepsilon_{i(m')}, \omega_{i(m')})\}$ where $\{\omega_{i(1)}, \ldots, \omega_{i(m')}\}$ is an ordered set of external vertices of $EW_j$, $\varepsilon_{i(j)} = \varepsilon_{ik} \Leftrightarrow \omega_{i(j)} = \omega_k$ and *m'*=*m*-1.

**Algorithm 2**: Distance-between-expanded-vertices

**Inputs**: $EV_i$ : AG expanded vertex.

$EW_j$ : FDG expanded vertex.

**Output**: The distance $d_f$ between $EV_i$ and $EW_j$ .

**Begin**

Let $S_i$ and $S_j$ be the cyclic strings of $EV_i$ and $EW_j$ , respectively.

$d_f := \infty$

$D(0,0) := K_1 * C_{f_v}\left(v_i, \omega_j\right)$. { $D(0,0) := 0$ in the Levenshtein algorithm}

**For** s:=0 **to** n'-1 **do** {orientation of the AG expanded vertex}

**For** l:=1 **to** n' **do**

$D(l,0) := D(l-1,0) + K_1 * C_{f_v}\left(v_{i(l+s)}, \omega_\Phi\right) + K_2 * C_{f_e}\left(e_{i,i(l+s)}, \varepsilon_\Phi\right)$

**end-for**

**For** k:=1 **to** m' **do**

$D(0,k) := D(0,k-1) + K_1 * C_{f_v}\left(v_\Phi, \omega_{j(k)}\right) + K_2 * C_{f_e}\left(e_\Phi, \varepsilon_{j,j(k)}\right)$

**end-for**

**For** l:=1 **to** n' **do**

**For k:**=1 **to** m' **do**

$m_1 := D(l-1,k-1) + K_1 * C_{f_v}\left(v_{i(l+s)}, \omega_{j(k)}\right) + K_2 * C_{f_e}\left(e_{i,i(l+s)}, \varepsilon_{j,j(k)}\right)$ {Subs. in Levenshtein alg.}

$m_2 := D(l-1,k) + K_1 * C_{f_v}\left(v_{i(l+s)}, \omega_\Phi\right) + K_2 * C_{f_e}\left(e_{i,i(l+s)}, \varepsilon_\Phi\right)$ {Insertion in Levenshtein alg.}

$m_3 := D(l,k-1) + K_1 * C_{f_v}\left(v_\Phi, \omega_{j(k)}\right) + K_2 * C_{f_e}\left(e_\Phi, \varepsilon_{j,j(k)}\right)$ {Deletion in Levenshtein alg.}

$D(l,k) := \min\left(m_1, m_2, m_3\right)$

**end-for**

**end-for**

**if** $D(n',m') < d_f$ **then**

$d_f := D(n',m')$

**end-if**

**end-for**

**end-algorithm**

For each possible orientation of one of the expanded vertices (subscript $s$ in the algorithm), the distance between the strings $S_i$ and $S_j$ is computed by a method reminiscent of the Levenshtein algorithm (Levenshtein, 1966; Tanaka and Kasai, 1976; Bunke and Sanfeliu (Eds.), 1990), with the difference that the cost of inserting or deleting string elements is obtained by substituting these elements with null vertices and arcs. Moreover, since the minimal expanded vertex is formed by only one central vertex, the cost of substituting the central vertices is initially applied independently of the string alignment cost caused by arcs and external vertices. An algorithm to search the distance between two cyclic strings is proposed in (Maes, 1990) with a computational cost $O(n \cdot m \cdot \log(m))$. Nevertheless we decided to use the Levenshtein algorithm applied $n$ times for its simplicity and considering that the average number of output arcs for each vertex is not usually very big. The Maes algorithm is going to be applied in a future work.

## 7.6. A non-iterative sub-optimal method for computing the distance between AGs and FDGs

Figure 22 shows the basic scheme of algorithm 3, proposed to compute a sub-optimal approximation of $d_f^R(G,F)$ within a set of labelling functions *H*, denoted $d_f^{R\wedge}(G,F)$, and also, the best labelling reached $f\wedge \in H$. Obviously, $d_f^{R\wedge}(G,F) \geq d_f^R(G,F)$, since some labelling functions, which belong to *H*, are discarded due to the constraints derived from the distances between expanded vertices. The scheme of the algorithm consists of three main modules. The first one computes all the distances between expanded vertices using the distance measure without second order costs, $d_f(EV_i, EW_j)$. The thresholding module decides which vertex matches are discarded using an externally imposed threshold $\tau$, $\tau \in [0,1]$. If the normalised distance between expanded vertices is higher than $\tau$ the mapping is considered unlikely and it is forbidden.

$$Forbid(v_i,\omega_j)=\begin{cases} True & \text{if } \dfrac{d_f(EV_i,EW_j)}{2n_i-1} > \tau \wedge n_i \geq m_j \\ True & \text{if } \dfrac{d_f(EV_i,EW_j)}{n_i+m_j-1} > \tau \wedge n_i < m_j \\ False & \text{otherwise} \end{cases} \tag{146}$$

where $n_i$ and $m_j$ represent the number of vertices of the expanded vertices $EV_i$ and $EW_j$, respectively.

The last module computes the sub-optimal distance $d_f^{R\wedge}(G,F)$ using the branch and bound algorithm presented in chapter 6 but only on the set of vertex mappings allowed by the thresholding module, the ones for which $Forbid(v_i^G,\omega_j^F)=False$.

Note that, given a set of distances between expanded vertices, the lower $\tau$ is, the more restrictions are imposed. Therefore, the branch and bound algorithm has to explore fewer possible morphisms and therefore, the measure is more approximated and the process is faster. If $\tau=1$, the combinatorial algorithm explores the whole set of possible morphisms and the output is exactly $d_f^R(G,F)$. However, if $\tau=0$, only the matches of totally isomorphic subunits are not discarded, and the combinatorial algorithm explores few (or no) morphisms, in which case $d_f^{R\wedge}(G,F)$ is "very sub-optimal".

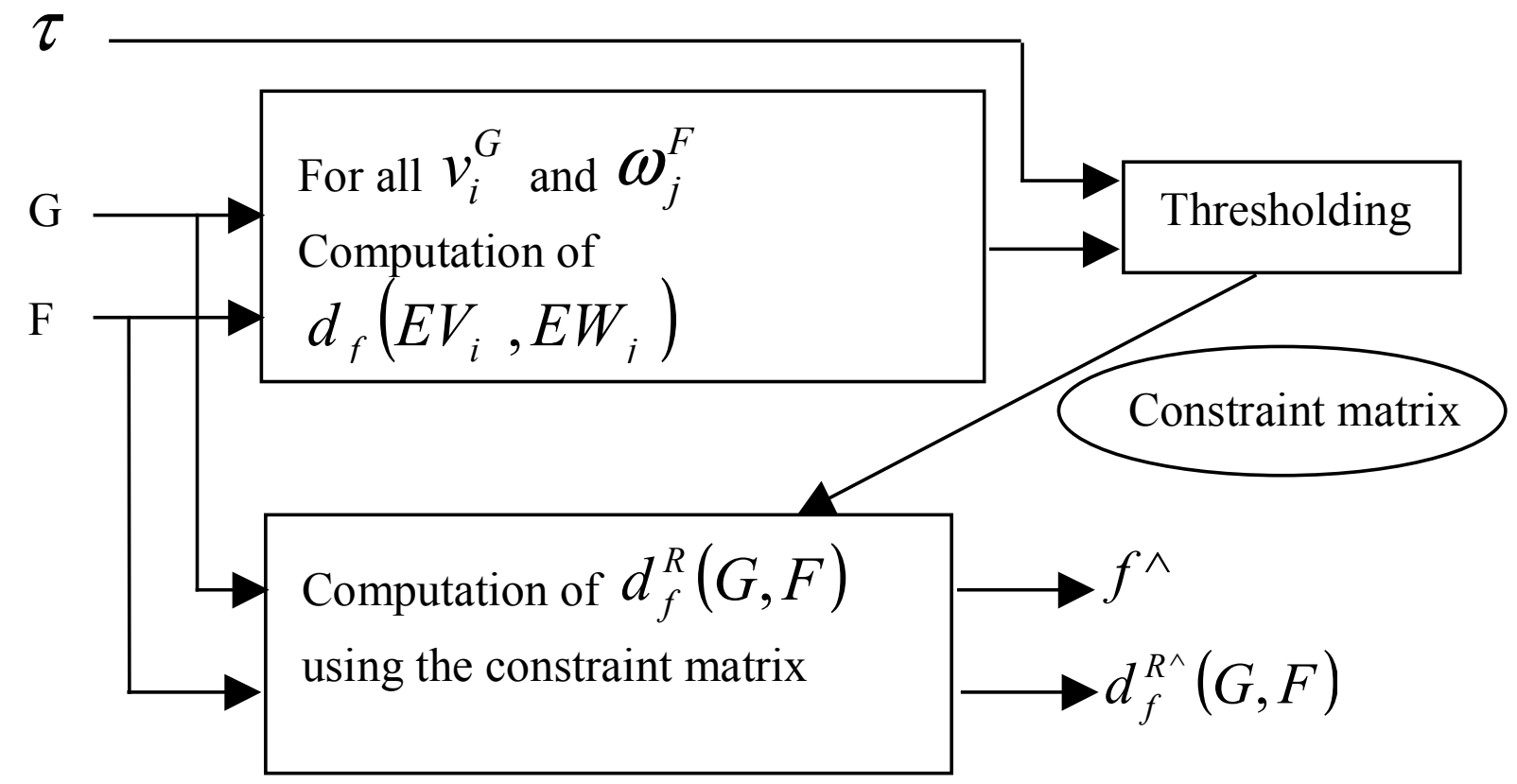

Figure 22. Basic scheme of algorithm 3 for sub-optimal distance computation.

## 7.7. Probabilistic relaxation methods to compute the distance between AGs and FDGs

We apply the Rosenfeld *et al.* approach to compute two sub-optimal distances between AGs and FDGs in this section. The efficient algorithms are composed by three main modules similar as the one proposed in section 7.6 (figure 22). The first module computes the probability matrix and the second module decides which mappings are accepted depending on these probabilities. The third module is similar as the one in section 7.6. The compatibility function is defined as follows,

$$r_{i,j}(a,b) = e^{-\left(C_{f_v}(v_i,\omega_a)+C_{f_v}(v_j,\omega_b)+C_{f_e}(e_{ij},\varepsilon_{ab})\right)} \tag{147}$$

In the first relaxed algorithm, the initial probabilities are calculated depending on the distance between vertices,

$$P_i^0(a) = \frac{e^{-C_{f_v}(v_i,\omega_a)}}{\sum_{\omega_{a'}\in\Sigma_\omega} e^{-C_{f_v}(v_i,\omega_{a'})}} \tag{148}$$

In the second relaxed algorithm, they are defined depending on the distance between expanded vertices,

$$P_i^0(a) = \frac{e^{-d_f(EV_i,EW_a)}}{\sum_{\omega_{a'}\in\Sigma_\omega} e^{-d_f(EV_i,EW_{a'})}} \tag{149}$$

The aim of the second option is to establish the initial probabilities nearer to the stationary point than the first option although the computational cost is higher. This is achieved by taking into account the plannar graph restriction, $R_5$.

## 7.8. Experimental results

In order to examine the performance of the efficient matching algorithms in practice, we carried out a number of experiments with randomly generated graphs. The parameter values of the random graph generator were:

| Initial Graph | | | | Generated AGs | |
|---|---|---|---|---|---|
| *nv* | *ne* | *nd* | *nl* | *nv'* | *Non-modified* |
| 27 | 108 | 21 | 2 | 6 | 4 |

Table 11. Parameters of the third experiments.

Similarly than the second experiments, the number of FDGs was set to 10 and synthesised in a supervised manner and the number of AGs in the test set to 100. The number of AGs in the reference set for each FDG to 1, 3, 6, 9, 12, 15 and 18.

The aim of the experiments in this section is to compare the three proposed efficient algorithms. As in the experiments in section 6.5, the branch and bound module computed the distance with the following weights on the costs: $K_1 = K_2 = 1$, $K_5 = K_7 = K_4 = K_6 = K_8 = 0$. The weight on the vertex antagonism was $K_3 = 1$ as it was the value in which the best results were obtained. Figures 23, 24 and 25 show the correctness and run time in the recognition process obtained by the three efficient algorithms. Figure 23 shows the results of the non-iterative algorithm. In the case $T = 1$, the optimal distance is obtained and the first module of the algorithm is not used. Figure 24 shows the results of the relaxed algorithm in which the probabilities were initialised by the distance between expanded vertices. Figure 25 shows the results of the relaxed algorithm in which the probabilities were initialised by the distance between vertices. In the case $T_p = 0$, the optimal distance is obtained and the first module of the algorithm is not used.

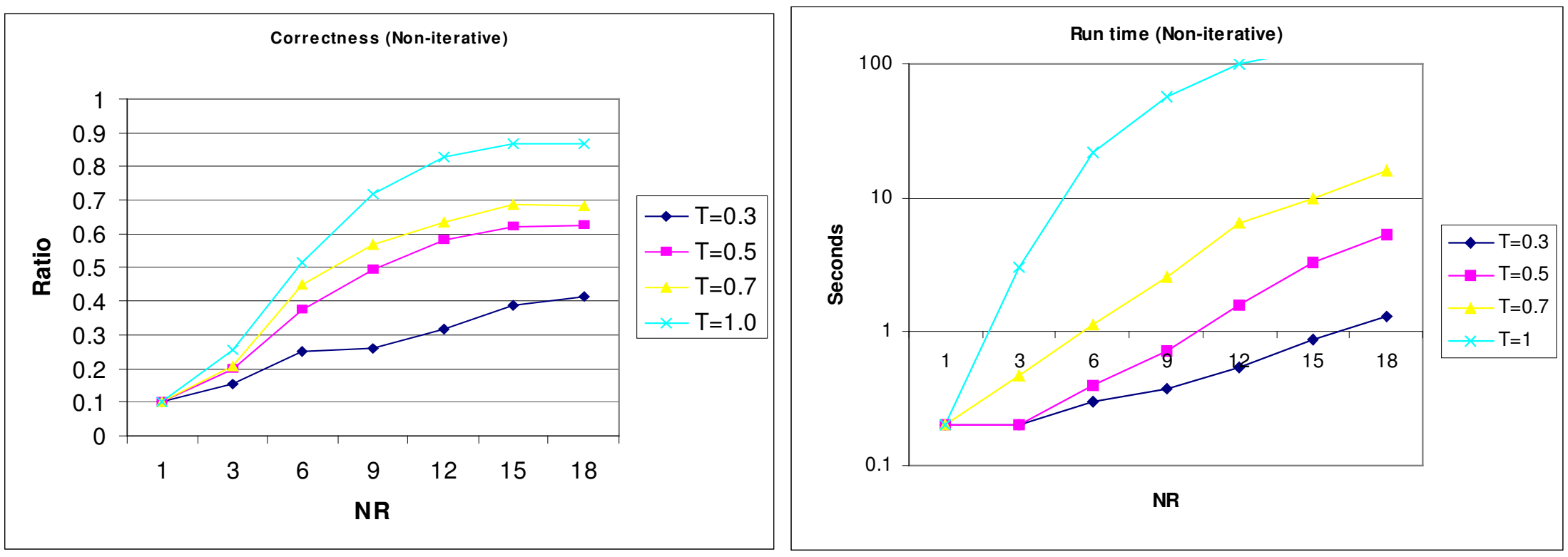

Figure 23. Correctness and run time obtained by the non-iterative algorithm

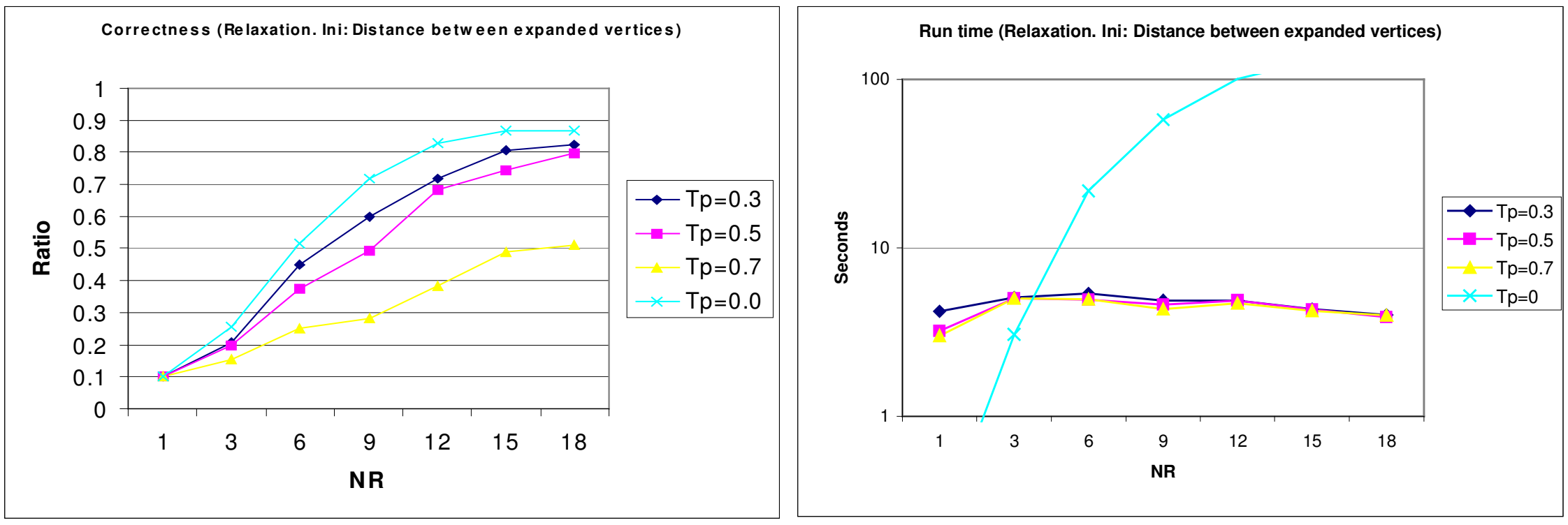

Figure 24. Correctness and run time obtained by the relaxation algorithm. The probabilities were initialised by the distance between expanded vertices.

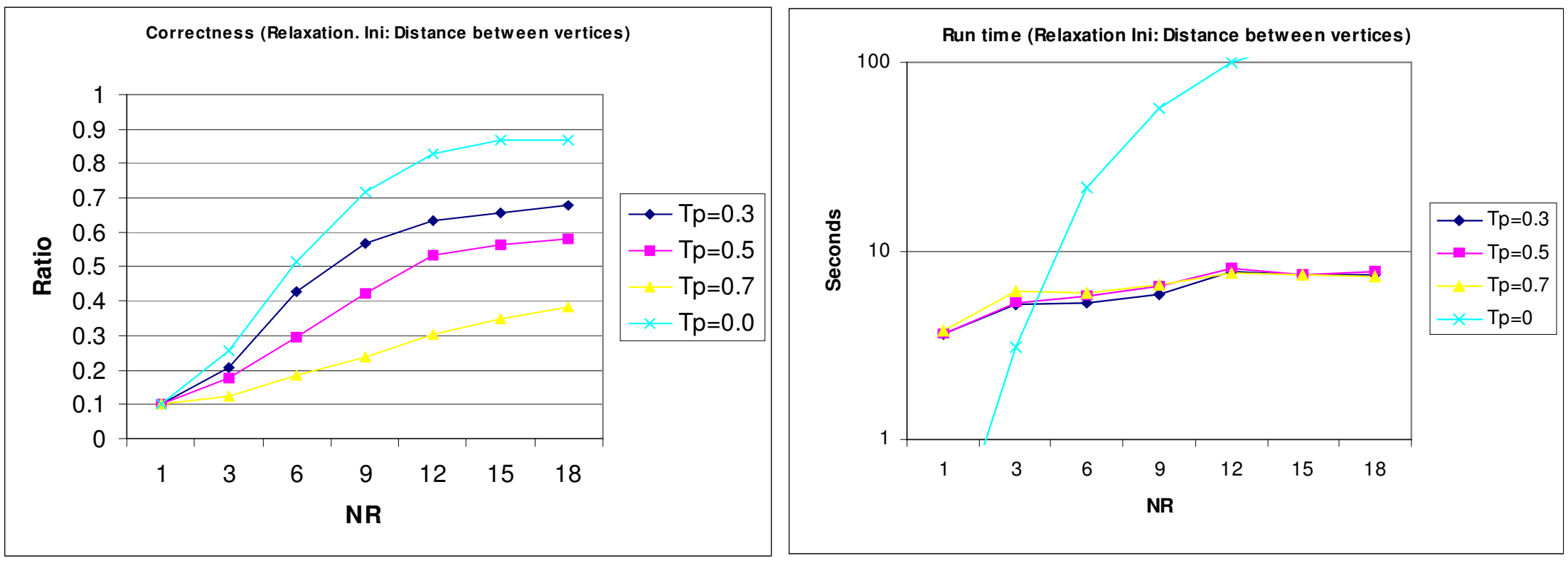

Figure 25. Correctness and run time obtained by the relaxation algorithm. The probabilities were initialised by the distance between vertices.

The relaxation algorithm in which the probabilities are initialised by the distance between expanded vertices obtains better correctness and less run time than the algorithm in which the probabilities are initialised by the distance between vertices. It is because the initial probabilities are set closer to the stationary point and the time spent to compute the distance between the expanded vertices compensates the efficiency in pruning of the search tree. The run time is almost constant throughout the *NR* axis in the tests in which the probabilities are initialised with the distance between vertices but there is slight increase when the probabilities are initialised with the distance between vertices. This is not the case of the non-iterative algorithm in which the run time increases when the number of AGs increases.

# 8. Clustering of AGs using FDGs

Given a set of AGs, which are initially supposed to belong to the same class, we do not, in general, have any way of synthesising an FDG that represents the ensemble unless we can first establish a common labelling of their vertices. However, given a set of isomorphisms applied to a common underlying structure, an FDG can be generated from the ensemble using the process *FDG-synthesis-from-labelled-AGs* presented in section 4.4. We want to choose the common labelling which will minimise the measures of dissimilarity between the given AGs and the resulting FDGs. This global optimisation problem does not lead to a computationally practical method for choosing the labelling, because there are too many possible orientations to consider, especially when the number and order of the AGs is high. Therefore, we propose two sub-optimal methods for synthesising FDGs from a set of unlabelled AGs.

In one of them, the FDGs are updated by the AGs, which are sequentially introduced as in (Seong *et al*., 1993). The advantage of this method is that the learning and recognition processes can be interleaved, i.e. the recognition does not need to wait for all the input instances, but is available after each AG has been processed. The main drawback of this incremental approach is that different FDGs can be synthesised from a set of unlabelled AGs depending on the order of presentation of the AGs.

To infer some unique FDGs, we propose a hierarchical method, which is carried out by successively merging pairs of FDGs with minimal distance, as in (Wong and You, 1985). The drawback here is that the full ensemble of AGs is needed to generate the FDGs. Since a distance measure between FDGs has not been presented, the original distances between the AGs of the ensemble are used in this second approach. The hierarchy of AGs (dendogram) can be computed by a *complete method* or by a *single method*, depending on whether the similarity of an object to a cluster is taken as its similarity to the farthest or to the closest member within the cluster, respectively (Gordon, 1987; Fisher, 1987; Wallance and Kanade, 1990).

When the ensemble of patterns is composed of several classes and the assignment of patterns to classes is unknown (unclassified AGs), we need a clustering process to synthesise an FDG for each class. Thus, a distance measure threshold $d_\alpha$ is introduced to determine the splitting condition. Nevertheless, if the given AGs are considered to belong to the same class and a single FDG is desired, the clustering process can be applied using a large distance threshold such that the splitting condition is never fulfilled and, therefore, only one FDG is synthesised.

Below, we present the *Incremental-clustering-of-AGs* (section 8.1) and the *Hierarchical-clustering-of-AGs* (section 8.2) methods to cluster a set of AGs and synthesise the corresponding FDG for each class. Moreover, some experiments are included in section 8.3.

## 8.1. Incremental (dynamic) clustering of AGs

**Algorithm 4** computes the *Incremental-clustering-of-AGs.* It generates some clusters from a sequence of AGs and synthesises a set of FDGs, one for each cluster. The algorithm uses two previously described procedures: *FDG-synthesis-from-labelled-AGs* (section 4.4), to transform an AG into an equivalent FDG, and *FDG-synthesis-from-labelled-FDGs* (section 4.5) to build an FDG from two FDGs with a given labelling. Moreover, the procedure *Extend-labelling-AG-FDG* extends the AG and the FDG with null elements to make them structurally isomorphic and also extends the given labelling accordingly. And also, the procedure *Update-an-FDG-with-an-AG* synthesises an FDG from a previous FDG, a new AG, and a labelling between them. The clustering method relies on (and is parameterised by) a distance threshold $d_\alpha$ and a matching algorithm $M(G,F)$ that is supposed to return an optimal (or a "good" suboptimal) labelling between an AG *G* and an FDG *F*, according to an appropriate distance measure. This distance measure is parameterised by the weights on the costs on the first and second-order relations (section 5.3). The weights on the first-order relations, $K_1$ for the vertices and $K_2$ for the arcs, are application dependent and therefore they also are parameters in the clustering algorithm. The other weights, from $K_3$ to $K_8$, for the antagonisms, occurrences and existences on the vertices and arcs, are useful to keep the structural information through the labelling. Since this FDG structure has to be updated, and more graph elements have to be included into it, we do not consider this knowledge and so they are set to zero.

---

**Algorithm 4:** Incremental-clustering-of-AGs

**Inputs**: A sequence of AGs $G_1,...G_m$, $m \geq 1$, over a common domain.

A matching algorithm $M(G,F)$ between an AG and an FDG that finds an optimal or sub-optimal labelling according to the weights $K_1$ and $K_2$.

A threshold $d_\alpha$ on the distance between an AG and an FDG.

**Output**: A set of FDGs $C^{F'}$ that represents the clusters of $G_1,...G_m$.

**Begin**

$n := 1$

$F_n$ := FDG-synthesis-from-labelled-AGs( $G_1$ ) {build the first FDG from $G_1$ using the method in sec. 4.4}

**for** $i := 2$ **to** $m$ **do**

  **for** $j := 1$ **to** $n$ **do**

    let $d_j : G_i, F_j \to \Re$ and $\mu_j : G_i \to F_j$ be the distance and labelling found by $M(G_i, F_j)$

    let $d_x = min\{d_k : 1 \leq k \leq j\}$

  **end-for**

  **if** $d_x \leq d_\alpha$ **then**

    $(G'_i, F'_x, \mu'_x)$:=Extend-labelling-AG-FDG $(G_i, F_x, \mu_x)$ {described next}

    $F_x$ := Update-an-FDG-with-an-AG $(G'_i, F'_x, \mu'_x)$ {described next}

  **else**

    $F_{n+1}$ := FDG-synthesis-from-labelled-AGs( $G_i$ ) {build $F_{n+1}$ from $G_i$ using the synthesis sec. 4.4}

    $n := n + 1$

  **end-if**

**end-for**

Let $C^{F'} = \{F_1, F_2, ..., F_n\}$

**end-algorithm**

---

The synthesis process given a common labelling (section 4.4) needs the AGs and the common labelling be structurally isomorphics. For this reason, the procedure *Extend-labelling-AG-FDG* extends the labelling and both graphs in the following way. First, both graphs are extended with null vertices, as many as the number of non-labelled vertices of the other graph. Thus, the updated graphs have the same number of vertices. And second, the original vertices that had not have been labelled, are labelled now with the null vertices created in the extended graphs.

---

**Procedure** Extend-labelling-AG-FDG $(G, F, \mu)$ **returns** $(G', F', \mu')$

**if** $\mu$ is not bijective **then**

Let $\mu' : G' \to F'$ be a bijective mapping that extends $\mu$ by extending $G$ to $G'$ and $F$ to $F'$

with null vertices and arcs appropriately

**else**

$\mu' := \mu$ ; $G' := G$ ; $F' := F$

**end-if**
**end-procedure**

---

And the last procedure updates an FDG using two procedures described in previous sections. This procedure is carried out in two steps since we have not formally defined the synthesis of an FDG using another FDG and an AG. For this reason, the AG is first transformed into an FDG and then the returned FDG is built by the synthesis from two FDGs.

---

**Procedure** Update-an-FDG-with-an-AG $(G', F', \mu')$ **returns** $F$

{build $H$ from $G'$ using the method described in section 4.4}

$H$ := FDG-synthesis-from-labelled-AGs( $G'$ )

let $\gamma : G' \to H$ be a bijective mapping used in the previous synthesis.

let $\varphi : H \to F'$ be the bijective mapping determined by the composition $\mu' = \varphi \circ \gamma$

$F$ := FDG-synthesis-from-labelled-FDGs $(H, F', \varphi)$ {build $F$ using synthesis in sec. 4.5 with 2 FDGs}

**end-procedure**

---

## 8.2. Hierarchical (agglomerative) clustering of AGs

**Algorithm 5** computes the *Hierarchical-clustering-of-AGs.* It generates some clusters from a set of AGs by a single or complete agglomerative method and synthesises the FDGs that describe the classes obtained. The clustering method is parameterised by a matching algorithm between AGs, $M(G,G')$, a distance threshold $d_\alpha$, and a Boolean value, which indicates the type of agglomerative method. The algorithm uses four main procedures, which are described below.

---

**Algorithm 5:** Hierarchical-clustering-of-AGs

**Inputs**: A set of AGs $C^G = \{G_1,...,G_m\}$ , $m \geq 1$, over a common domain.

A matching algorithm $M(G,G')$ between AGs that finds an optimal or good sub-optimal labelling according to a distance between AGs.

A threshold $d_\alpha$ on the distance between AGs.

A Boolean value *complete* which is true or false depending on whether a complete or a single agglomerative method is desired.

**Output**: A set of FDGs $C^{F'}$ that represents the clusters of $G_1,...,G_m$.

**Begin**

{Find the set of distances between AGs $C^d = \{d_{i,j} : 1 \leq i,j \leq m : i < j\}$

and the associated labellings $C^f = \{f_{i,j} : 1 \leq i,j \leq m : i < j\}\}$

$(C^f, C^d)$:=Labelling-and-distance-between-AGs$(C^G, M)$

{Define the set of FDGs consisting of one AG $C^F = \{F_i : 1 \leq i \leq m\}$

and the labellings between FDGs $C^\varphi = \{\varphi_{i,j} : 1 \leq i,j \leq m : i < j\}\}$

$(C^F, C^\varphi)$:=Initial-FDGs-and-labellings$(C^G, C^f)$

{Build the FDGs $C^{F'}$ by a single or complete hierarchical method}

$C^{F'}$:=Agglomerative-clustering-of-FDGs$(C^F, C^\varphi, C^d, d_\alpha, \text{complete})$

**end-algorithm**

---

In the first procedure, called *Labelling-and-distance-between-AGs*, the distance measures and labellings between AGs are obtained. Note that the whole set of AGs has not to be structurally isomorphic and so, the computed labellings may not be bijective. This is because, a global common labelling between AGs is not obtained.

---

**Procedure** Labelling-and-distance-between-AGs$(C^G, M)$ **returns** $(C^f, C^d)$

**for all** $G_i, G_j \in C^G$

Let $f_{i,j}: G_i \rightarrow G_j$ and $d_{i,j}: G_i, G_j \rightarrow \Re$ found by $M(G_i, G_j)$

**end-for-all**

Let $C^f = \{f_{i,j} : 1 \le i, j \le m; i < j\}$;

Let $C^d = \{d_{i,j} : 1 \le i, j \le m; i < j\}$

**end-procedure**

---

In the second procedure, called *Initial-FDGs-and-labellings*, the AGs and labellings are transformed into equivalent FDGs and their corresponding labellings. At this point, the initial patterns are described in FDGs, which represent only one AG. The distance between them is the one obtained between the AGs. Their labellings are obtained as a composition of the labellings between their equivalent AGs.

---

**Procedure** Initial-FDGs-and-labellings$(C^G, C^f)$ **returns** $(C^F, C^\varphi)$

**for all** $G_i \in C^G$

$F_i$ := FDG-synthesis-from-labelled-AGs($G_i$) {build the FDG $F_i$ from $G_i$ using the synth. in sec. 4.4}

let $\gamma_i : G_i \rightarrow F_i$ be the bijective mapping used in the previous synthesis

**end-for-all**

**for all** $F_i, F_j : 1 \le i, j \le m : i < j$

let $\varphi_{i,j}: F_i \rightarrow F_j$ be a mapping defined by $\varphi_{i,j} = \gamma_j \circ f'_{i,j} \circ \gamma_i^{-1}$

**end-for-all**

Let $C^\varphi = \{\varphi_{i,j} : 1 \le i, j \le m; i < j\}$

Let $C^F = \{F_i : 1 \le i \le m\}$

**end-procedure**

---

In the third procedure, called *Agglomerative-clustering-of-FDGs*, a set of FDGs is generated by a complete or single method. The new feature of this procedure is that the merging of two clusters redefines not only the distance measures but also the labelling between the clusters. Note that it is not possible to use an *average clustering method* (Gordon, 1987) since an “average labelling” between two FDGs cannot be defined. As in the incremental approach, the procedures *FDG-synthesis-from-labelled-AGs* and *FDG-synthesis-from-labelled-FDGs* are used as well.

**Procedure** Agglomerative-clustering-of-FDGs $\left(C^F, C^\varphi, C^d, d_\alpha, \text{complete}\right)$ **returns** $C^{F'}$

Let $C^F = \{F_i : 1 \le i \le m\}$; Let $C^{F'} = C^F$

Let $C^\varphi = \{\varphi_{i,j} : 1 \le i, j \le m; i < j\}$

Let $C^d = \{d_{i,j} : 1 \le i, j \le m; i < j\}$

**for** $i := 1$ **to** $m$ **do**

  $d_{i,i} := \infty$

**end-for**

Let $d_{x,y}$ be the minimum distance in $C^d$

**while** $d_{x,y} \le d_\alpha$ **do**

  $\left(F'_x, F'_y, \varphi'_{x,y}\right)$:=Extend-labelling-FDG-FDG $\left(F_x, F_y, \varphi_{x,y}\right)$ {described next}

  $F_y$ := FDG-synthesis-from-labelled-FDGs $\left(F'_x, F'_y, \varphi'_{x,y}\right)$ {described in section 4.5}

  Remove $F_x$ from $C^{F'}$

  **for all** $d_{x,i}, d_{i,x} \in C^d$ **do**

    **if** complete **then {** complete-method **}**

      **if** $d_{i,x} > d_{i,y}$ **then**

        $d_{i,y} := d_{i,x}$; $\varphi_{i,y} := \varphi_{x,y} \circ \varphi_{i,x}$; $d_{i,x} := \infty$

      **end-if**

      **if** $d_{x,i} > d_{y,i}$ **then**

        $d_{y,i} := d_{x,i}$; $\varphi_{y,i} := \varphi_{x,i} \circ \varphi_{y,x}$; $d_{x,i} := \infty$

      **end-if**

    **else {** single-method **}**

      **if** $d_{i,x} < d_{i,y}$ **then**

        $d_{i,y} := d_{i,x}$; $\varphi_{i,y} := \varphi_{x,y} \circ \varphi_{i,x}$; $d_{i,x} := \infty$

      **end-if**

      **if** $d_{x,i} < d_{y,i}$ **then**

        $d_{y,i} := d_{x,i}$; $\varphi_{y,i} := \varphi_{x,i} \circ \varphi_{y,x}$; $d_{x,i} := \infty$

      **end-if**

    **end-if**

  **end-for-all**

**end-while**

**end-procedure**

The synthesis process from FDGs given a common labelling (section 4.5) needs the FDGs and the common labelling be structurally isomorphics as it was commented in the incremental clustering. The procedure *Extend-labelling-FDG-FDG* extends the labelling and both FDGs similarly than the procedure *Extend-labelling-AG-FDG*.

---

**Procedure** Extend-labelling-FDG-FDG $(F, H, \varphi)$ **returns** $(F', H', \varphi')$

**if** $\varphi'$ is not bijective **then**

Let $\varphi : F \to H$ be a bijective mapping that extends $\varphi'$ by extending $F'$ to $F$ and $H'$ to $H$

with null vertices and arcs appropriately

**else**

$\varphi := \varphi'$; $F := F'$; $H := H'$

**end-if**

**end-procedure**

---

### 8.3. Experimental results

In order to examine the ability of these algorithms to synthesise the FDGs in practice, we performed a number of experiments with randomly generated AGs. The random AGs were generated as in section 6.5 or 7.8, that is, $nFDG = 10$, $NT = 100$ and $NR = \{1,3,6,9,12,18\}$.

| Initial Graph | | | | Generated AGs | |
|---|---|---|---|---|---|
| *nv* | *ne* | *nd* | *nl* | *nv'* | *Non-modified* |
| 27 | 108 | 21 | 2 | 6 | 4 |

Table 12. Parameters of the fourth experiments.

From each set of AGs representing one cluster in the reference set, an FDG was synthesised by one of the four different methods proposed to generate the FDGs. The three clustering methods with unsupervised labelling commented above are the incremental or also called dynamic method, the hierarchical complete (or agglomerative complete) and the hierarchical single (or agglomerative single). These methods have an input parameter that is the distance threshold. We have given to this parameter a big number to obtain only one FDG from each cluster. The fourth method is the synthesis method with a given labelling described in section 4.4.

In the classification process, the branch-and-bound algorithm was used to compute the optimal distance. The weights on the costs were $K_1 = K_2 = 1$, $K_4 = K_5 = K_6 = K_7 = K_8 = 0$ and $K_3$ had the values $K_3 = 0$ or $K_3 = 1$. Each experimental run was repeated ten times and the average of the results was recorded as a result.

Figure 26 shows the average of the number of vertices and antagonism in the FDGs in which the four methods were applied. We observe that in small and medium number of AGs, the non-supervised methods obtain less number of vertices. It is because vertices in different AGs, which represent different basic parts in the original object, are merged in only one FDG vertex when the labelling was not given. We also note that when there are many AGs, the number of generated vertices using the dynamic clustering is bigger than the number of vertices of the original AGs, $nv$. We think that the spurious elements are not matched to other vertices and so, they remain as different ones. The hierarchical methods obtain less vertices than the dynamic method. This is because some AGs that represent different parts of the original AG have nearly similar structural and semantic information and so, the distance between them is small. In the hierarchical methods, these AGs are merged in the first steps of the clustering algorithms and so their vertices and arcs are mapped although representing different parts of the original AG. In the incremental (dynamic) method, the order of merging the AGs is imposed. If two AGs, which represent different parts of the original AG, have to be merged in the first iterations of the algorithm then the synthesised FDG does not have the same structure than the "original AG" since vertices and arcs that represent different parts may be merged in only one FDG vertex or arc. When new AGs are introduced in the synthesis process, the matching algorithm cannot completely recognise the structure and so, it generates more vertices.

We also observe that the dynamic method generates more second-order relations than the other methods when the number of AGs is big. We have checked that the most of these second-order relations appear between few elements and also, that some of these elements have similar semantic information than another FDG vertex. Therefore these elements represent the same element in the original AG but, due to the order of presenting the AGs, they have been generated as different elements. In a post-processing step used to reduce the FDGs or delete the spurious elements, the second-order relations could be useful to discern which vertices delete from the FDGs.

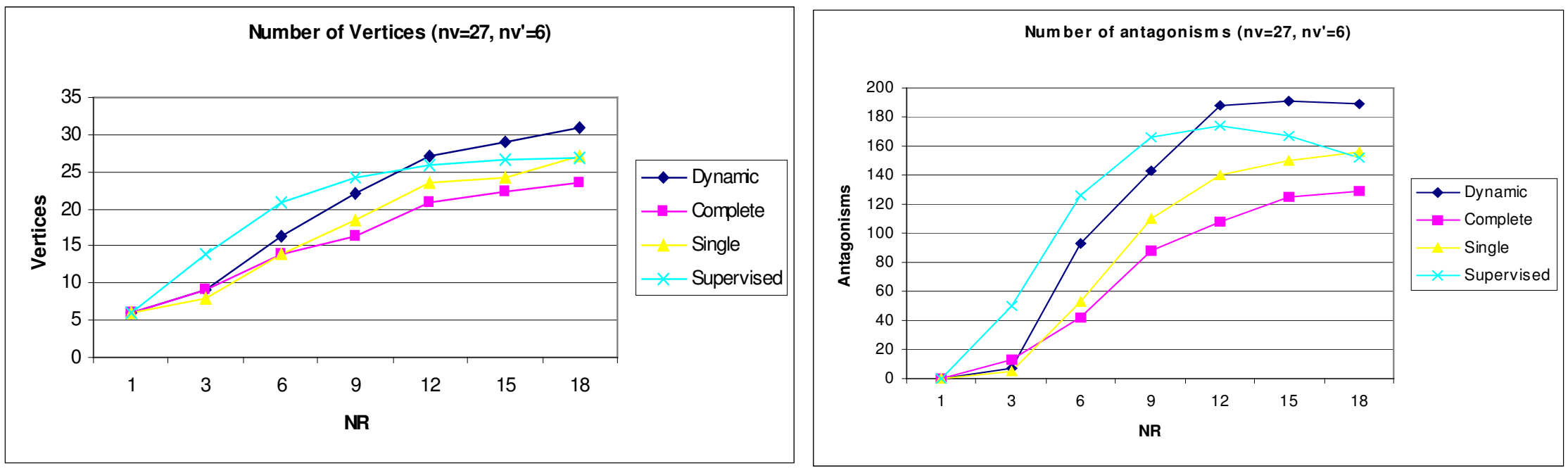

Figure 26. Number of elements in the FDGs applying different clustering methods.

Figure 27 shows the ratio of correctness when the FDGs were built by the four different clustering methods. In the left column, the antagonisms were not used and in the right column, the antagonism were applied with a cost 1. We observe that in the four methods, results are better using the antagonisms than without the antagonisms. The complete method obtains the most important difference. It is the authors believe that it is because the FDGs are smaller (figure 26.a), and so, the structure of the AGs has spread between less FDG vertices and so more useful are these relations to discern the AG vertices.

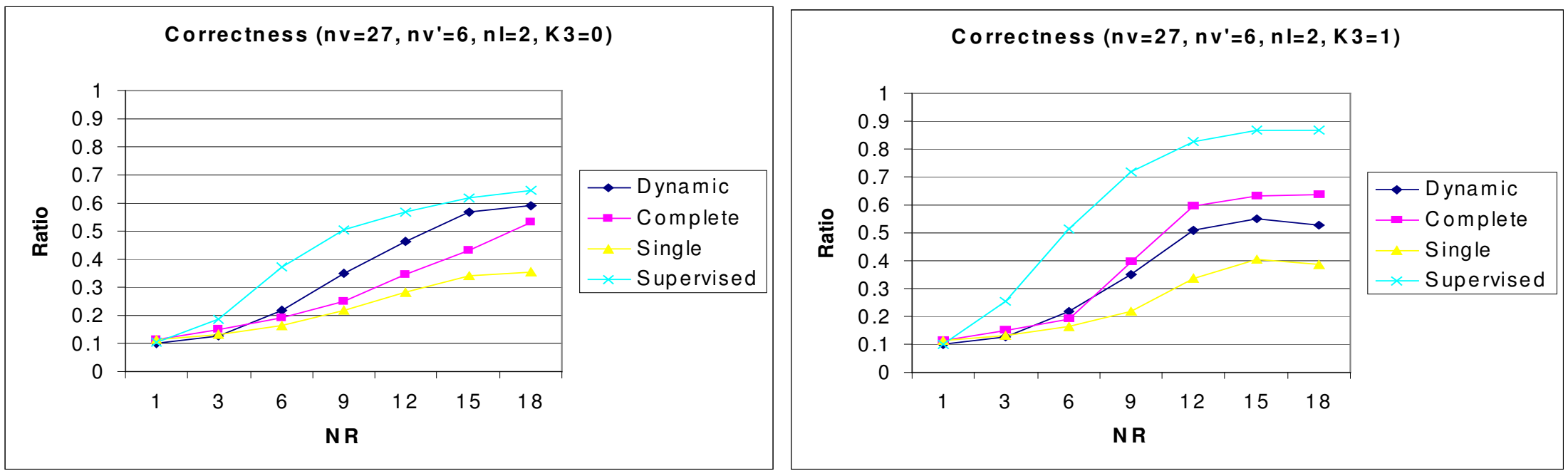

Figure 27. Ratio of correctness applying different clustering methods.

Figure 28 shows the average run time of the classification process in logarithmic scale. In all the cases the time spent to classify the AGs decreases when the antagonisms are used. We can also observe that the biggest difference appear when the number of antagonism is big, for instance, in the supervised method and $NR = 9$.

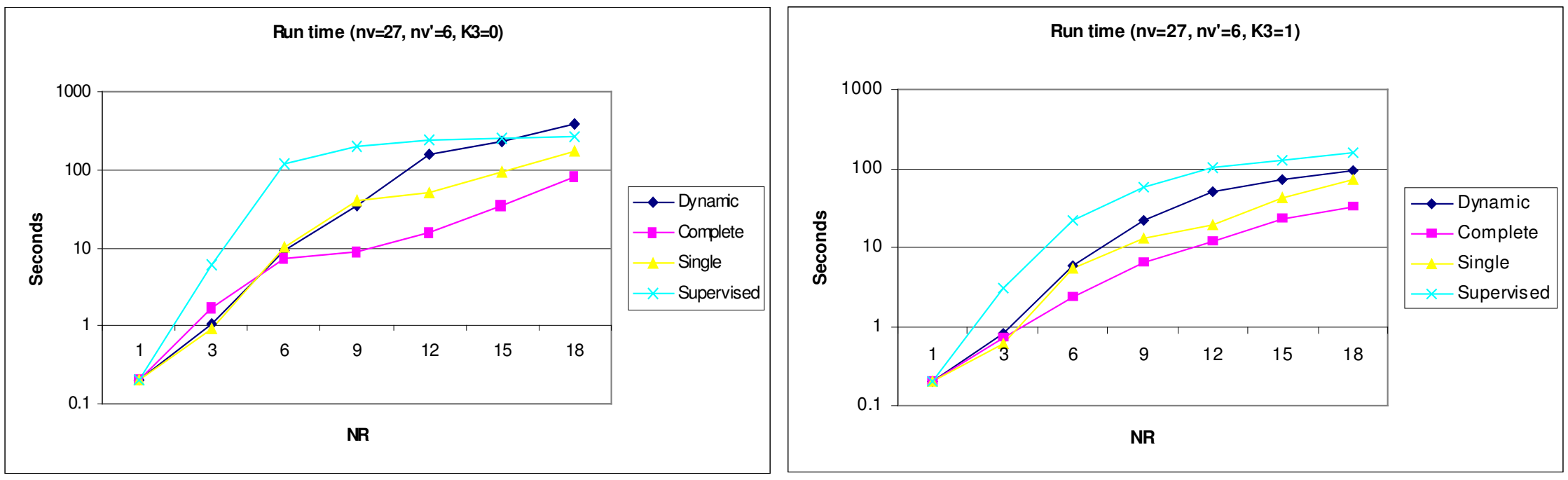

Figure 28. Run time applying different clustering methods.

# 9. Experimental validation of FDGs using artificial 3D objects

We present an illustrative example of FDGs, which is based on artificially created 3D objects with planar faces. We designed five objects by a CAD program and then, we defined five sets of views from these objects. Furthermore, we built an attributed graph from each view in which the vertices represent the planar faces and the arcs represent the edges between faces. The exact value of the area of the face and the length of the edges are the only attributes of the vertices and the arcs. Instead of applying the projective distortion on the area of the faces and the distance of the edges, we have modified the attribute values of the vertices and arcs by Gausian noise. Moreover, some vertices and arcs are deleted and inserted. We wanted the attributes on the vertices and arcs be invariant but with some degree of noise. For this reason, the initial values of the attributes were set to the area and the distance instead of a random value although we are aware of they are not invariant to the projective distortion. Moreover, the Gausian noise and the insertion and deletion of some graph elements represent the noise on the measurement of the data.

This experiment is only useful to make a first validation and test of the algorithms and methods presented in this thesis since it does not use real images and the extraction of the attributed graphs is not made automatically. A real application is presented in the next chapter. The advantage of this experiment is that we can use the synthesis of FDG given a common labelling (section 4.4) since we know which vertices from the AGs represent the same planar face of the object. Thus, we can discern the effects of computing the optimal and sub-optimal distance between AGs and FDGs using different costs on the second-order relations in the recognition process and the clustering algorithms in the classification process. Section 9.1 explains the set of samples. In sections 9.2 and 9.3, FDGs are synthesised with a given common labelling. The former assesses the effects of applying second-order relations in the computation of both distances between AGs, i.e. with hard restrictions and with relaxed second-order costs. The latter studies the balance between effectiveness and efficiency in the classification process when the efficient algorithm was applied. Finally, section 9.4 shows the capability of FDGs to represent an ensemble of AGs when a supervised and unsupervised clustering was applied.

### *9.1. The set of samples*

The original data was composed of 101 AGs, which represent the semantic and structural information of the views taken of five objects (appendix 1.1, views 101 to 121 of object 1, views 201 to 221 of object 2, views 301 to 312 of object 3, views 401 to 424 of object 4 and views 501 to 523 of object 5).

Vertices in the AGs represent the faces with one attribute, which is the area of the face (average of the areas 11.0). Arcs represent the edges between faces, with one attribute, which is the length of the edge (average of the lengths 3.0). appendix 1.2 shows the visible faces for each 3D-object. Appendix 1.3 shows the semantic and structural information of the AGs with the following structure,

Number of vertices (n)
Attribute vertex 1 Attribute vertex 2 ... Attribute vertex n
Attribute edge 1,1 Attribute edge 1,2 ... Attribute edge 1,n
Attribute edge 2,1 Attribute edge 2,2 ...
...
Attribute edge n,1 Attribute edge n,2 ... Attribute edge n,n

Note that some AGs are similar although the views are different, for instance AGs 203 and 206 in figure 29. This particularity of the reference set has an important effect on the recognition and clustering process. The AGs are planar graphs that have between one and nine vertices. When the visible faces are not touching in the object, the graph can be disjoint (Figure 29).

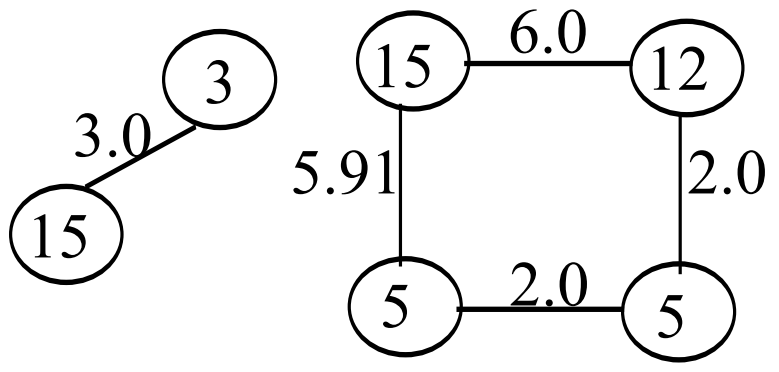

Figure 29. The AG that represents views 203 and 206 with the attributes in the vertices and arcs.

The AGs in the test set and the reference set are the AGs obtained from the views and modified by some structural or semantic noise. The results in the following tables are the average of computing the test 20 times. The semantic noise, which is added to the attribute values of the vertices and arcs, is obtained by a random number generation with a median of 0.0 and a standard deviation: 0.0, 2.0, 4.0, 8.0 and 12.0. The structural noise, also obtained by a random number generator, deletes or includes 0, 1 or 2 vertices, which represent 0%, 20% and 40% of the average structure, respectively.

## 9.2. Effects of second-order relations

The aim of the first test is to assess how the antagonism and occurrence relations between vertices affect the computation of the distance with second-order restrictions or relaxing them. Results are compared with the nearest neighbour classifier in which the Sanfeliu distance (section 2.2) is used as a dissimilarity measure between elements.

### 9.2.1. Recognition by FDG classifier

Five FDGs were built from the AGs that represent their views using the supervised method *synthesis-of-FDGs-from-AGs-with-a-common-labelling* (see section 4.4). In this method the labelling function between graph elements is imposed using the information taken from the views and tables in appendix 1.2. Vertices and arcs in the FDGs represent planar faces and edges between faces, respectively, as in the AG case. Moreover, the number of vertices of the FDGs is similar to the number of faces of the objects. There is an antagonism relation between two graph elements when these elements are never seen together in the same view. Moreover, an occurrence relation appears when a face is visible whenever another face is also visible in all the views. For instance, in a concave object, there is an occurrence between the faces around a hole and the faces on the bottom of it. There is no existence relation because there is no pair of faces such that at least one of the two faces is visible in all views. Appendix 1.4 shows the structure of these FDGs.

Valid mappings must be coherent structurally and consistent with the planar graph restriction; that is, they must belong to $H = F_1 \cap F_2 \cap F_5$ when both distances are applied, in the recognition task. The second-order relations on the vertices, $R_3$, are used in some tests. Furthermore, the second-order relations on the arcs, $R_4$, are not used since there is a geometrical relationship between the antagonisms and occurrences in the vertices and arcs. The weights in the distance relaxing second-order relations were set as follows: $K_1 = 1$ (vertices), $K_2 = 1/2$ (arcs), $K_3 = 1$ (second-order relations on the vertices), and $K_4 = 0$ (second order relations on the arcs).

When one attempts to recognise a three-dimensional object from one view, the occluded parts of the object should not have an influence on the distance value. In our method, this is carried out by considering the cost of mapping a null element of the AG to an element of the FDG to be zero.

### 9.2.2. Recognition by the nearest neighbour classifier

In this test, the nearest neighbour classifier was used with a neighbourhood of 5 and the Sanfeliu distance, described in section 2.2, was used as the dissimilarity measure between elements. The labellings between graph elements were constrained to be bijective, structurally coherent and consistent with the planar graph restriction, $H = F_1 \cap F_2 \cap F_5$. The weight of the insertion and deletion of nodes and arcs were defined as a constant: $W_{vi} = 1$, $W_{ei} = 1$, $W_{vd} = 1$ and $W_{ed} = 1$. The weights of the substitution operations were defined according to the attribute values of the graph elements of both graphs. For the vertices, $W_{vs}(v_i, v'_j) = 1$ if $(a_i - a'_j)^2 > K_a$ and $W_{vs}(v_i, v'_j) = 0$ otherwise, where $v_i$ and $v'_j$ are two vertices of both graphs and $K_a = 4.0$ is a threshold on the noise of the attribute value. For the arcs, assuming that $b_m$ and $b'_n$ are the attribute values of the arcs $e_{i,j}$ and $e'_{k,t}$, $W_{es}(e_{i,j}, e'_{k,t}) = 1$ if $(b_m - b'_n)^2 > K_b$ and $W_{es}(e_{i,j}, e'_{k,t}) = 0$ otherwise, where $K_b = 4.0$ is a noise threshold.

### 9.2.3. Results

Table 13 shows the recognition ratio throughout different levels of semantic (standard deviation) and structural noise (number of vertices deleted or inserted) to the test and reference set. The FDG classifier was used with the distance with hard second-order constraints and the distance relaxing second-order constraints and the Nearest neighbour classifier. When the distance with second-order restrictions is used and the semantic or structural noise is low, the classification correctness is higher when the antagonism and occurrence relations are applied. Nevertheless, when the noise increases, best results appear when no relations are applied. However, if the distance relaxing second-order restrictions is used, the classification correctness is always higher when the antagonisms and occurrences are taken into account. The FDG classifier gives worse results than the nearest-neighbour classifier only when the structural or semantic noise is very low.

| | **# spurious vertices** | **0** | **0** | **0** | **0** | **0** | **1** | **2** | **1** |
|---|---|---|---|---|---|---|---|---|---|
| | **Standard Deviation** | **0.0** | **2.0** | **4.0** | **8.0** | **12.0** | **0.0** | **0.0** | **8.0** |
| **FDGs distance with 2nd order constraints** | Without relations | 100 | 90.1 | 89.7 | 88.6 | 86.3 | 70.8 | 67.7 | 68.7 |
| | With antagonism | 100 | 92.5 | 89.3 | 87.0 | 84.9 | 61.6 | 54.4 | 57.4 |
| | With occurrence | 100 | 91.9 | 89.9 | 88.2 | 85.2 | 62.5 | 59.5 | 59.5 |
| | With antag. & occurr. | 100 | 95.1 | 90.2 | 86.6 | 80.8 | 60.7 | 53.2 | 56.2 |
| **FDGs distance relaxing 2nd order Relations** | Without relations | 100 | 90.1 | 89.7 | 88.6 | 86.3 | 70.8 | 67.7 | 68.7 |
| | With antagonism | 100 | 92.3 | 91.5 | 91.3 | 87.2 | 80.5 | 75.3 | 75.5 |
| | With occurrence | 100 | 95.6 | 92.4 | 91.5 | 86.4 | 81.2 | 77.2 | 76.4 |
| | With antag. & occurr. | 100 | 98.7 | 97.1 | 95.0 | 92.5 | 89.2 | 85.2 | 83.6 |
| **Nearest Neighbour** | | 100 | 98.9 | 82.6 | 62.6 | 52.4 | 90.0 | 58.6 | 58.6 |

Table 13. Correctness obtained by the FDG classifier (using the optimal distances with second-order relations and relaxing them) and by the nearest-neighbour classifier (using the Sanfeliu distance).

## 9.3. Efficient algorithms

| **# spurious vertices** | **0** | **0** | **0** | **0** | **1** | **2** | **1** |
|---|---|---|---|---|---|---|---|
| **Standard Deviation** | **0.0** | **4.0** | **8.0** | **12.0** | **0.0** | **0.0** | **8.0** |
| $\tau = 1$ | 100 | 97.1 | 95.0 | 92.5 | 89.2 | 85.2 | 83.6 |
| $\tau = 0.4$ | 100 | 96.5 | 94.9 | 92.6 | 87.5 | 82.4 | 80.3 |
| $\tau = 0.3$ | 100 | 80.3 | 75.6 | 71.0 | 81.5 | 77.7 | 72.1 |
| $\tau = 0.2$ | 100 | 73.7 | 70.0 | 56.2 | 72.3 | 65.3 | 63.2 |
| $\tau = 0.1$ | 100 | 46.3 | 26.9 | 25.2 | 56.2 | 50.1 | 45.3 |

Table 14. Correctness obtained by the FDG classifier using the sub-optimal distance relaxing second-order relations with different thresholds.

The aim of the following tests was to study the balance between effectiveness and efficiency in the classification process in which the non-iterative efficient algorithm is used. The distance relaxing second-order relations is shown to be the one that gives best results with different levels of noise in the above section. So, this distance is used in the efficiency tests. Furthermore, the AGs and FDGs are obtained as mentioned in sections 9.1 and 9.2.1, respectively. Table 14 shows the recognition ratio applying different levels of semantic (standard deviation) and structural noise (number of vertices deleted or inserted) in the test set and different thresholds in the sub-optimal distance relaxing 2nd order relations.

Table 15 shows the computational costs associated with the results of Table 15. All the values are normalised by the time spent to compute the classification according to the optimal distance, that is $\tau = 1$, and without noise.

| **# spurious vertices** | **0** | **0** | **0** | **0** | **1** | **2** | **1** |
|---|---|---|---|---|---|---|---|
| **Standard Deviation** | **0.0** | **4.0** | **8.0** | **12.0** | **0.0** | **0.0** | **8.0** |
| $\tau = 1$ | **1.00** | 1.51 | 3.03 | 3.75 | 2.07 | 3.13 | 4.21 |
| $\tau = 0.4$ | 0.99 | 1.21 | 1.59 | 2.79 | 1.95 | 2.97 | 3.54 |
| $\tau = 0.3$ | 0.84 | 1.01 | 1.00 | 1.08 | 1.12 | 1.03 | 2.06 |
| $\tau = 0.2$ | 0.57 | 0.77 | 0.86 | 0.75 | 1.15 | 0.78 | 0.89 |
| $\tau = 0.1$ | 0.09 | 0.56 | 0.49 | 0.35 | 0.89 | 0.35 | 0.47 |

Table 15. Computational costs (normalised run times) obtained by the FDG classifier using the sub-optimal distance relaxing second-order relations with different thresholds.

In the cases where $\tau = 1$, all the mappings are allowed and the first and second steps are not computed, so their run times are saved. For this reason if $\tau < 1$ the run time may be higher than when $\tau = 1$.

The efficiency of this application does not improve without a corresponding loss in classification correctness. We believe that there are two basic reasons for this. One is that some combinations that the first step forbids may be removed by the branch and bound technique. The other is that, the AGs are small (five vertices on average), and so the cost of computing the first step is equivalent to the cost of the third, considering the branch and bound technique.

## 9.4. Clustering

The last tests were carried out to show how capable FDGs are at representing an ensemble of AGs when a supervised or unsupervised clustering process is applied. First, we used a supervised clustering; that is, the FDGs were synthesised using the AGs that belong to the same 3D object (tables 16 and 17). And second, we applied an unsupervised clustering; the FDGs were automatically generated using the whole reference set of AGs. That is, independently of the view or of the 3D object from which were extracted (table 18). In this case, we considered that the FDG belongs to the 3D object that the most of the AGs used to synthesise it belong to.

### 9.4.1. Supervised clustering

In the first experiments, we used four different methods. In the first one, the labellings were manually determined from the original data and the FDGs were synthesised as explained in section 9.2.1. The number of vertices of the FDGs, shown in the first column of table 16, is similar to the number of faces of the objects. In the rest of the methods, there is no a-priori information of the labellings. The distance threshold $d_\alpha$ was set to a big number, such that, only one FDG was synthesised for each reference set. In the second one, the FDGs were obtained by the procedure *Incremental-clustering-of-AGs* (section 8.1). The number of FDG vertices is shown in the fourth column of table 16. The AGs were presented to the algorithm as the point of view rotated around the object. Therefore, the new faces are gradually introduced to the system, which minimises the effect of merging different faces in only one FDG vertex. The third and fourth methods use the *Hierarchical-clustering-of-FDGs* algorithm (section 8.2), with a complete and single agglomerative technique, respectively. The Sanfeliu distance measure (section 2.2) was used to find the distance and the optimal labelling between AGs.

Note that in the clustering methods such that the labelling is not given the number of vertices of the FDGs were reduced (2$^{nd}$, 3$^{rd}$ and 4$^{th}$ column of Table 16). This means that more than one face of the object is represented by one vertex in the FDG. This is partly due to the fact that some faces of these objects are similar and thus, the matching algorithm between AGs matches their corresponding vertices. So, the clustering process considers these faces as the same vertex in the FDG.

| | Synthesis from labelled AGs | Hierarchical-complete clustering | Hierarchical-single clustering | Incremental clustering |
|---|---|---|---|---|
| **FDG 1** | 14 | 9 | 9 | 11 |
| **FDG 2** | 15 | 10 | 9 | 10 |
| **FDG 3** | 9 | 7 | 7 | 11 |
| **FDG 4** | 10 | 8 | 7 | 9 |
| **FDG 5** | 9 | 8 | 8 | 11 |

Table 16. Number of vertices of the FDGs generated by different clustering methods

Table 17 shows the ratio of correctness of the FDG classifier and the nearest neighbour classifier that applies the Sanfeliu distance measure between AGs as a distance between elements. The test set was obtained from the reference set by applying a zero-mean Gaussian noise that modifies the semantic information with a standard deviation of 8.0. The structure was also modified by inserting or deleting one of the vertices. The results of the nearest neighbour classifier were only better than the FDG classifier when the single method was used. On the contrary of the results obtained in the experimental results, figure 27 right, the complete clustering gave lower results than the incremental clustering. It is the author's believe that it is because the order of presenting the new graphs to the incremental synthesis is of crucial importance. In the random graph experiments, the order was also random, without considering the similarity between the introduced graphs.

| Supervised clustering | | Correctness |
|---|---|---|
| **FDG:** | Labelled AGs | 83.6% |
| | Incremental | 80.3% |
| | Complete | 74.6% |
| | Single | 40.3% |
| **Nearest Neighbour** | | 58.6%. |

Table 17. Correctness obtained by the FDG classifier using the supervised clustering and the nearest-neighbour classifier.

### 9.4.2. Unsupervised clustering

Table 18 shows the clustering and classifying results using the FDG unsupervised clustering methods and the Nearest-neighbour classifier. The synthesis with labelled AGs cannot be used since in these experiments the FDGs can be synthesised with AGs from different 3D objects and so the FDG vertices can represent faces from different objects. Moreover, the order of presenting the AGs to the incremental method was defined randomly to try not to influence on the synthesised FDGs. The distance threshold $d_\alpha$ was set to obtain five classes in all the methods. There is a decrease in the correctness in all the methods except the complete clustering. We also show the average of the number of vertices of the FDGs. The classification using the complete method obtains the best results and also it is the only method that the correctness increases respect the supervised clustering (table 17). In these tests, we show again that the order of introducing the AGs in the incremental method is very important. Again, the nearest neighbour classifier only obtains better results than the single method.

| Unsupervised Clustering | | # FDGs | $d_\alpha$ | # FDG vertices | Correctness |
|---|---|---|---|---|---|
| **FDG:** | Incremental | 5 | 15 | 11 | 69.2% |
| | Complete | 5 | 25 | 9 | 85.5% |
| | Single | 5 | 25 | 9 | 33.1% |
| **Nearest Neighbour** | | --- | --- | --- | 48.1%. |

Table 18. Correctness obtained by the FDG classifier using the unsupervised clustering.

# 10. Application of FDGs to 3D objects

We present a real application of FDGs to recognise coloured objects using 2D images. The tests presented here are the first ones obtained in the Spanish project TAP98-0473. The aim of this project is to design a robot that has to recognise objects in a given environment, for instance, in an office. The recognition process is implemented by the FDG approach with a high number of objects and views. During the learning process, the robot is guided to move around the 3D objects and to take images from different perspectives. These images are segmented by the method outlined in the next section, which extracts the AGs. An FDG is synthesised from the set of AGs providing from each object.

In this test, we took sixteen views from each object. In each view, the angle was incremented 22.5 degrees (appendix 2.2). The views taken with the angles 0, 45, 90, 135, 180, 225, 270 and 315 were used to synthesise the FDGs in the learning process. The other views, 22.5, 67.5, 115.5, 157.5, 202.5, 247.5, 292.5 and 337.5 compose the test set.

In the validation of the FDGs, chapter 9, the incremental synthesis obtains better results than the hierarchical synthesis in the supervised clustering when a proper sequence of presenting the AGs is given. Moreover, the incremental synthesis has the advantage that the learning and recognition process can be interleaved and also that the synthesis complexity is only lineal on the number of AGs. For these reasons, in this project we prefer the incremental synthesis to generate the FDGs that represent the objects to be recognised.

The main problems in real applications provide from the segmentation module. From the structural point of view, if a region is not detected (their pixels are merged with a near region) then there is a missing vertex in the AGs. Furthermore, if a region is split into two, then there is a spurious vertex. And also, from the semantic point of view, there is always some degree of uncertainty in obtaining the attribute values. To solve, as much as possible these problems, first, it is useful to use redundancy data and for this reason, some of the visible faces in the image are also visible in other images. Second, it is needed a calibration process to extract the kind of regions that the user desires.
The remainder of this chapter is as follows, section 10.1 outlines the method used to segment the images, section 10.2 describes the calibration process and section 10.3 presents some of the obtained results.

### *10.1. Segmentation of the images*

Segmentation is the first essential and important step of low-level vision. Segmentation is a process of partitioning the image into some regions such that each region is homogeneous and the union of two adjacent regions is not homogeneous. Hundreds of segmentation techniques are present in the literature (Pal and Pal, 1993). In this application, we have selected the graph-theoretical approach to cope with image segmentation because it has good mathematical basements and some segmentation problems are easily translated into graph-related problems by analogy. Moreover, in the graph theoretical approach, image region extraction and finding region edges are dual problems. The worst disadvantage of this approach, as can be seen in (Wu and Leahy, 1993; Vlachos and Constantinides, 1993; Xu and Uberbacher, 1997), is that these sort of algorithms are very time consuming, which makes their implementation prohibitive in some real applications. For this reason, we have chosen the greedy algorithm for graph partition proposed in (Felzenswalb and Huttenlocher, 1998). This algorithm makes decisions based on local properties of image, as could be pixel differences, and also global properties of the image such as not over-segmentation and not sub-segmentation. It was implemented and compared to other approaches in (Vergés-Llahí and Sanfeliu, 2000).
The input of our segmentation algorithm is a view of the object and the output is an attributed graph. The vertices of the AG represent regions which have similar colour and there is an arc between adjacent regions. The attribute on the vertices is the hue of the colour and the attribute on the arcs is the difference between the hues of the connecting vertices. The hue is not well defined when the colour tends to the grey, that is, the colour is not saturated. For this reason, the neighbourhoods of non-saturated pixels are merged into regions independently of their colour. The attribute on the node is represented by the average of the hue. Figure 30 (page 162) shows the four objects, the segmented views and their corresponding AGs. Appendix 2.3 and 2.4 show the segmented images of the objects and the computed AGs.

The input parameters of the algorithm are:

*Threshold on the colour difference*: If the colour between two adjacent pixels is lower than this threshold, then they belong to the same region.

*Threshold on the number of pixels*: The regions that have less number of pixels than this threshold are merged into their adjacent regions.

*Threshold on the compactness*: The thin regions are deleted if their compactness is greater than this threshold. It is useful the delete the regions that appear in the edges of the faces.

*Threshold on the saturation*: If the saturation of the pixel colour is lower than this threshold, then the pixel belongs to a non-saturated region.

*Threshold on the average size*: This is a global feature used to control the average size of the regions.

### 10.2. Calibration process

To calibrate the segmentation module we used four monochrome objects. From each object we took four images varying the angle (appendix 2.1.1). We wanted that the segmented images had only one region (appendix 2.1.2) and that the AGs computed from these views had only one vertex (appendix 2.1.3). The parameters were set as follows:

| | |
|---|---|
| Threshold on the colour difference | 8 |
| Threshold on the number of pixels | 100 |
| Threshold on the compactness | 0.1 |
| Threshold on the saturation | 20 |
| Threshold on the average size | 200 |

Table 19. Calibration parameters

Moreover, an FDG was synthesised using the four AGs that belong to the same object. These FDGs had to have two vertices. One vertex represents the object and the other one represents the background. In the recognition process, the 16 AGs had to be correctly classified. As it was commented in section 4.1, the probability density functions in the FDGs are represented non-parametrically and we use a discretisation process to represent them computationally. The discretisation of the hue plays an important rule on the number of graph elements in the FDGs. If it is discretised in few intervals and these intervals are large, then only one FDG vertex is synthesised but the AGs are not properly classified. On the contrary, if the hue is discretised in many intervals and these intervals are small, then more than two vertices are generated in the FDGs. The values of the hue obtained in the segmentation are:

| | View 1 | View 2 | View 3 | View 4 |
|---|---|---|---|---|
| Blue object | 43 | 42 | 43 | 43 |
| Green object | 153 | 147 | 150 | 149 |
| Yellow object | 3 | 4 | 4 | 4 |
| Red object | 96 | 92 | 95 | 96 |

Table 20. Value of the vertex attributes in the calibration AGs

We discretised the hue in intervals of 10. Moreover, the probabilities were computed as follows,

$$p(\mathbf{a}) = \frac{1}{2}\Pr(R(a)) + \frac{1}{4}\Pr(R^{+}(a)) + \frac{1}{4}\Pr(R^{-}(a)) \tag{150}$$

where $R(\mathbf{a})$ represents the interval that $\mathbf{a}$ belongs to and $R^{+}(\mathbf{a})$ and $R^{-}(\mathbf{a})$ are the previous and the next intervals.

### *10.3. Practical results*

Table 21 shows the number of graph elements of the synthesised FDGs using the incremental method. The AGs were presented as the observer was rotating around the objects and the optimal distance was computed between the AGs and the FDGs (any relaxation method was applied).

| Synthesis | FDG 1 | FDG 2 | FDG 3 | FDG 4 |
|---|---|---|---|---|
| Number of vertices | 8 | 7 | 10 | 9 |
| Number of arcs | 36 | 34 | 52 | 30 |
| Number of antagonisms | 8 | 0 | 6 | 4 |
| Number of occurrences | 15 | 16 | 18 | 24 |
| Number of existences | 0 | 2 | 0 | 2 |

Table 21. Number of graph elements in the FDGs (supervised clustering & dynamic synthesis).

Table 23 shows the classification of each view (or AG) in one of the four objects (or FDGs) in the first column. The incorrectly classified objects are marked in bold. The distance between the AG and the FDG is shown in the second column. The optimal distance was computed. The average time to classify one AG (that is to compare the AG with the 4 FDGs) was 1.05 seconds and the correctness is 78%.

Table 22 shows the ratio of correctness and the average run time of classifying one AG using different thresholds. When $\tau = 1$, the module that computes the distance between expanded vertices is not used (see section 7.6), for this reason, it is possible to spent more time in the recognition process when $\tau$ is lower than 1 (for instance $\tau = 0.9$).

| $\tau$ | Correctness | Run time |
|---|---|---|
| 1.0 | 0.78 | 1.05 |
| 0.9 | 0.78 | 1.15 |
| 0.7 | 0.78 | 0.85 |
| 0.5 | 0.78 | 0.5 |
| 0.3 | 0.68 | 0.4 |
| 0.1 | 0.62 | 0.2 |

Table 22. Correctness and run time applying different thresholds.

| **Recognition correctness** | Classification | Distance |
|---|---|---|
| Object 1 View 22.5 | FDG 1 | 1.00 |
| Object 1 View 67.5 | FDG 1 | 1.31 |
| Object 1 View 115.5 | FDG 1 | 0.89 |
| Object 1 View 157.5 | FDG 1 | 1.91 |
| Object 1 View 202.5 | FDG 1 | 1.31 |
| Object 1 View 247.5 | FDG 1 | 1.36 |
| Object 1 View 292.5 | FDG 1 | 0.21 |
| Object 1 View 337.5 | **FDG 2** | 1.39 |
| Object 2 View 22.5 | FDG 2 | 4.03 |
| Object 2 View 67.5 | FDG 2 | 2.97 |
| Object 2 View 115.5 | **FDG 3** | 2.80 |
| Object 2 View 157.5 | FDG 2 | 2.45 |
| Object 2 View 202.5 | **FDG 3** | 0.45 |
| Object 2 View 247.5 | FDG 2 | 5.28 |
| Object 2 View 292.5 | **FDG 3** | 3.87 |
| Object 2 View 337.5 | FDG 2 | 4.31 |
| Object 3 View 22.5 | FDG 3 | 3.87 |
| Object 3 View 67.5 | FDG 3 | 9.16 |
| Object 3 View 115.5 | FDG 3 | 4.99 |
| Object 3 View 157.5 | FDG 3 | 3.78 |
| Object 3 View 202.5 | FDG 3 | 2.20 |
| Object 3 View 247.5 | FDG 3 | 2.73 |
| Object 3 View 292.5 | FDG 3 | 1.36 |
| Object 3 View 337.5 | FDG 3 | 1.51 |
| Object 4 View 22.5 | **FDG 3** | 0.33 |
| Object 4 View 67.5 | FDG 4 | 2.15 |
| Object 4 View 115.5 | FDG 4 | 0.23 |
| Object 4 View 157.5 | **FDG 3** | 1.41 |
| Object 4 View 202.5 | FDG 4 | 1.69 |
| Object 4 View 247.5 | FDG 4 | 1.65 |
| Object 4 View 292.5 | FDG 4 | 1.87 |
| Object 4 View 337.5 | **FDG 3** | 0.63 |

Table 23. Classification and distance of the AGs.

Tables 24.a and 24.b show the distances between AGs in the reference set and the AGs in the test set using the Sanfeliu and Fu algorithm (section 2.2). The minimum distance is shown in the column headed by *md* and also in bold in the table. There is a "+" or a "*" if the AG was correctly classified using the 1 nearest neighbours and the 3 nearest neighbours, respectively. The costs on the insertions, deletions and substitutions of vertices and arcs have been set as follows:

The costs on the insertion and deletion take all the same value: $C_{vi} = C_{ei} = C_{vd} = C_{ed} = 1$. The cost on the substitution is:

$$C_s = \begin{cases} 1 & if \quad abs(a,b) > 10 \\ 1/2 & if \quad 5 \geq abs(a,b) \geq 10 \\ 0 & if \quad abs(a,b) < 5 \end{cases} \qquad (151)$$

where $C_s$ represents $C_{vs}$ if *a* and *b* are attribute values of two vertices and $C_{es}$ if they are attribute values of two arcs.

The average time spent to classify each AG in the test set was 1.93 seconds. The correctness using the 1-NN is 56% and using the 3-NN is 59%. FDGs obtain better results than the 1 nearest neighbour or the 3 nearest neighbours. Given a test view, the most similar views in the reference set have to be the previous and next views in the rotating sequence of the same object. Nevertheless, the distance is usually greater than zero because the different topology and the variation on the colour. In the FDG classifier, these variations are compensated by the structural and probabilistic knowledge of the whole object. Moreover, the FDG method classifies the views faster than the nearest neighbours. Although the FDGs are bigger than the AGs, the second-order relations prune the search tree. In addition, only one comparison for class is needed in the FDGs and sixteen in the nearest neighbours.

| | Object 1 of reference set | | | | | | | | Object 2 of reference set | | | | | | | |
|---|---|---|---|---|---|---|---|---|---|---|---|---|---|---|---|---|
| View: | 22.5 | 067.5 | 112.5 | 157.5 | 202.5 | 247.5 | 292.5 | 315.5 | 022.5 | 067.5 | 112.5 | 157.5 | 202.5 | 247.5 | 292.5 | 315.5 |
| Object 1 of test set | | | | | | | | | | | | | | | | |
| View 000 | 07.0 | 09.0 | 16.0 | 12.0 | 14.0 | **05.0** | 07.5 | 12.0 | 14.5 | 15.0 | 11.5 | 12.0 | 14.5 | 19.0 | 25.0 | 21.0 |
| View 045 | **01.0** | 02.0 | 18.0 | 14.0 | 16.0 | 12.0 | 13.5 | 17.0 | 15.0 | 16.0 | 13.0 | 15.5 | 16.0 | 18.0 | 26.5 | 18.5 |
| View 090 | 09.5 | 12.0 | 16.0 | 17.0 | 09.5 | **09.0** | 10.0 | 12.0 | **09.0** | 14.0 | 14.0 | 10.0 | 14.0 | 21.0 | 29.5 | 21.5 |
| View 135 | 18.0 | 18.0 | 16.5 | 21.0 | 13.0 | 16.0 | 14.5 | 14.5 | 17.0 | 14.5 | 14.0 | 17.5 | 18.0 | 26.0 | 28.0 | 23.0 |
| View 180 | 14.5 | 14.0 | 19.5 | 14.5 | 14.0 | 16.5 | 16.0 | 17.5 | 16.0 | 14.0 | 15.0 | 18.5 | 16.5 | 22.0 | 27.5 | 22.0 |
| View 225 | 20.0 | 20.5 | 24.0 | 19.5 | 21.0 | **16.0** | 19.5 | 23.0 | 24.0 | 22.0 | **16.0** | 19.5 | 21.0 | 18.5 | 23.0 | 22.5 |
| View 270 | 17.0 | 19.0 | 17.5 | 21.5 | 10.0 | **05.0** | 08.0 | 14.0 | 13.0 | 16.5 | 15.0 | **05.0** | 13.0 | 20.0 | 26.0 | 22.0 |
| View 315 | 17.0 | 19.0 | 16.0 | 20.5 | 12.0 | 08.0 | **05.0** | 09.0 | 16.0 | 14.5 | 14.0 | 10.0 | 14.5 | 20.0 | 26.0 | 21.0 |
| Object 2 of test set | | | | | | | | | | | | | | | | |
| View 000 | 16.0 | 14.0 | 23.5 | 17.5 | 18.0 | 17.0 | 18.5 | 21.0 | 12.5 | 14.5 | 14.5 | 20.0 | 15.5 | 20.5 | 26.5 | 16.0 |
| View 045 | 18.5 | 18.5 | 17.0 | 20.0 | 12.0 | 14.0 | 14.5 | 11.0 | 13.0 | **06.5** | 09.5 | 10.0 | 11.5 | 22.0 | 23.0 | 20.0 |
| View 090 | 14.0 | 14.0 | 17.5 | 21.0 | 11.0 | 14.0 | 14.0 | 12.0 | 13.5 | **05.0** | 10.5 | 13.0 | 12.0 | 22.0 | 24.5 | 18.5 |
| View 135 | 14.0 | 13.0 | 19.0 | 19.0 | 13.0 | 15.5 | 15.5 | 18.0 | 13.5 | **09.0** | 14.5 | 13.5 | 15.0 | 23.0 | 28.0 | 23.0 |
| View 180 | 18.0 | 20.0 | 14.0 | 24.0 | 11.5 | 11.0 | 13.0 | 17.0 | 12.5 | 15.5 | 16.5 | 10.0 | 13.0 | 21.0 | 26.0 | 23.0 |
| View 225 | **08.0** | 12.0 | 22.5 | 14.0 | 19.0 | 17.0 | 19.0 | 19.0 | 20.0 | 20.0 | 20.0 | 20.0 | 15.5 | 13.0 | 22.0 | 13.5 |
| View 270 | 17.0 | 19.0 | 21.5 | 13.5 | 17.0 | 13.5 | 14.5 | 20.0 | 15.0 | 17.0 | 12.0 | 17.0 | 13.0 | 16.5 | 24.0 | 12.0 |
| View 315 | 31.0 | 32.5 | 32.5 | 30.0 | 31.5 | 30.0 | 31.0 | 30.0 | 31.0 | 27.5 | 29.5 | 27.0 | 23.0 | 25.5 | 25.0 | **19.5** |
| Object 3 of test set | | | | | | | | | | | | | | | | |
| View 000 | 16.0 | 15.0 | 24.0 | 17.5 | 19.0 | 14.0 | 16.5 | 21.0 | 21.0 | 17.5 | 19.0 | 18.0 | 20.0 | 18.5 | 19.0 | 15.5 |
| View 045 | 19.5 | **15.0** | 28.5 | 22.5 | 27.0 | 24.0 | 26.5 | 28.0 | 26.0 | 23.0 | 23.0 | 29.0 | 26.5 | 24.0 | 30.0 | 21.0 |
| View 090 | 20.0 | **19.0** | 27.5 | 22.5 | 28.0 | 24.0 | 23.5 | 27.0 | 24.0 | 24.5 | 26.5 | 26.5 | 25.0 | 25.0 | 32.0 | 24.0 |
| View 135 | 20.0 | **16.5** | 26.0 | 23.0 | 25.0 | 26.0 | 26.0 | 25.0 | 22.0 | 25.0 | 18.0 | 26.0 | 25.0 | 24.0 | 28.5 | 19.5 |
| View 180 | 11.0 | 12.5 | 16.0 | 17.5 | 12.0 | 08.5 | 08.5 | 12.0 | 15.0 | 12.0 | 12.0 | 08.5 | 11.5 | 15.0 | 22.0 | 17.0 |
| View 225 | 14.5 | 14.0 | 21.5 | 17.5 | 13.0 | 12.0 | 14.5 | 15.0 | 15.0 | 14.5 | 15.0 | 17.0 | 14.5 | 17.5 | 26.0 | 15.5 |
| View 270 | 21.0 | 20.5 | 22.0 | 21.5 | 21.5 | 16.5 | 17.5 | 21.5 | 22.0 | 16.0 | 21.5 | 17.5 | 19.0 | 19.5 | 18.0 | 17.0 |
| View 315 | 23.0 | 21.0 | 19.5 | 20.0 | 19.0 | 21.0 | 21.0 | 20.0 | 20.5 | 18.5 | **16.0** | 21.5 | 21.5 | 22.0 | 24.0 | 19.5 |
| Object 4 of test set | | | | | | | | | | | | | | | | |
| View 000 | 17.0 | 19.0 | 18.5 | 21.0 | 07.5 | 10.0 | 14.5 | 16.0 | 10.0 | 10.0 | 13.5 | 06.5 | 06.5 | 18.0 | 26.0 | 17.5 |
| View 045 | 15.5 | **11.0** | 21.0 | **11.0** | 18.0 | 17.5 | 17.5 | 19.5 | 18.5 | 16.0 | 18.0 | 21.5 | 17.0 | 22.5 | 31.0 | 22.0 |
| View 090 | 17.0 | 19.0 | 20.5 | 21.0 | 10.0 | 10.0 | 13.5 | 17.0 | 10.0 | 15.0 | 16.0 | 12.0 | 13.0 | 21.0 | 32.0 | 22.0 |
| View 135 | 17.0 | 19.0 | 21.0 | 23.0 | 10.0 | 10.0 | 14.5 | 18.0 | 07.0 | 15.0 | 12.0 | 10.0 | 13.5 | 23.0 | 31.0 | 21.0 |
| View 180 | 21.0 | 20.5 | 26.0 | 20.0 | 20.0 | 22.0 | 20.5 | 23.5 | 23.5 | 17.5 | 20.0 | 24.5 | 22.0 | 20.5 | 24.0 | 18.5 |
| View 225 | 20.0 | 20.0 | 24.5 | 25.0 | 23.0 | 23.0 | 23.0 | 24.5 | 13.0 | 19.0 | 17.0 | 23.0 | 23.0 | 25.0 | 33.0 | 19.0 |
| View 270 | 21.0 | 23.0 | 26.0 | 26.0 | 21.0 | 21.0 | 22.5 | 25.0 | 14.0 | 20.5 | 18.0 | 21.5 | 20.5 | 23.0 | 33.0 | 17.0 |
| View 315 | 13.5 | 14.0 | 18.0 | 15.0 | 08.0 | 10.0 | 10.0 | 13.5 | 14.5 | 10.0 | 12.0 | 11.5 | 10.0 | 18.5 | 26.0 | 15.0 |

Table 24.a. Distances between the whole test set and the object 1 and 2 of the reference set.

| View: | Object 3 of reference set 22.5 | 067.5 | 112.5 | 157.5 | 202.5 | 247.5 | 292.5 | 315.5 | Object 4 of reference set 22.5 | 067.5 | 112.5 | 157.5 | 202.5 | 247.5 | 292.5 | 315.5 | md |
|---|---|---|---|---|---|---|---|---|---|---|---|---|---|---|---|---|---|
| Object 1 of test set | | | | | | | | | | | | | | | | | |
| View000 | 12.5 | 24.0 | 24.5 | 11.0 | 08.0 | 17.0 | 19.5 | 17.0 | 15.0 | 13.0 | 12.0 | 15.0 | 17.0 | 24.5 | 22.0 | 15.0 | 05.0+* |
| View045 | 17.0 | 24.5 | 24.0 | 12.0 | 14.5 | 18.5 | 20.0 | 19.0 | 17.0 | 15.0 | 14.0 | 17.0 | 17.5 | 25.5 | 21.0 | 16.0 | 01.0+* |
| View090 | 15.0 | 25.0 | 24.0 | 12.5 | 10.0 | 17.5 | 20.5 | 21.5 | 15.0 | 14.0 | 10.5 | 10.0 | 18.5 | 24.0 | 21.0 | 10.0 | 09.0+* |
| View135 | 19.0 | 29.0 | 26.0 | 23.5 | 20.5 | 15.0 | 25.5 | 23.0 | **12.0** | 15.5 | 19.0 | 19.0 | 17.0 | 30.0 | 23.0 | 15.0 | 12.0 |
| View180 | 18.5 | 28.0 | 28.0 | 19.0 | 17.0 | 18.0 | 19.0 | 21.0 | 16.0 | 17.5 | 15.0 | 17.0 | 14.5 | 24.0 | 23.0 | **12.0** | 12.0 * |
| View225 | 19.5 | 28.0 | 25.5 | 22.0 | 18.0 | 16.5 | 24.0 | 21.5 | 23.0 | 24.5 | 19.0 | 25.0 | 18.0 | 28.5 | 26.5 | 20.5 | 16.0+* |
| View270 | 10.0 | 27.0 | 27.0 | 20.0 | 11.0 | 16.0 | 28.0 | 22.5 | 16.0 | 14.0 | 11.0 | 09.0 | 20.0 | 29.0 | 24.0 | 09.5 | 05.0+* |
| View315 | 12.0 | 27.0 | 26.0 | 19.0 | 11.0 | 17.0 | 27.0 | 22.0 | 10.5 | 16.0 | 15.0 | 16.0 | 20.0 | 27.0 | 23.0 | 13.0 | 05.0+* |
| Object 2 of test set | | | | | | | | | | | | | | | | | |
| View000 | 16.0 | 22.5 | 21.5 | **10.0** | 15.5 | 18.5 | 16.0 | 16.5 | 16.0 | 18.0 | 13.0 | 18.5 | 15.5 | 18.0 | 21.5 | 19.0 | 10.0 |
| View045 | 09.5 | 24.0 | 25.5 | 20.0 | 15.0 | 10.0 | 23.0 | 18.0 | 11.5 | 15.0 | 15.0 | 12.0 | 10.0 | 20.0 | 20.0 | 10.0 | 06.5+* |
| View090 | 13.0 | 25.5 | 22.5 | 15.0 | 14.0 | 11.5 | 15.0 | 22.0 | 10.0 | 14.5 | 13.0 | 15.0 | 15.0 | 24.0 | 20.5 | 11.5 | 05.0+* |
| View135 | 09.5 | 27.0 | 24.0 | 18.5 | 18.5 | 10.0 | 21.0 | 17.0 | 12.0 | 16.0 | 16.5 | 17.0 | 13.0 | 23.5 | 23.0 | 12.0 | 09.0+ |
| View180 | 13.5 | 28.5 | 25.5 | 21.0 | 17.0 | 17.5 | 30.0 | 23.0 | 14.5 | 13.0 | 12.0 | **08.5** | 23.0 | 27.5 | 22.5 | 10.5 | 08.5 |
| View225 | 18.5 | 21.5 | 22.0 | 18.0 | 15.0 | 16.0 | 22.0 | 18.0 | 20.5 | 20.0 | 16.0 | 19.0 | 20.5 | 25.5 | 22.0 | 16.5 | 08.0 |
| View270 | 13.5 | 25.0 | 17.5 | 14.0 | 11.5 | 13.0 | 20.0 | **11.0** | 14.0 | 16.5 | 16.0 | 14.5 | 14.0 | 18.5 | 17.0 | 15.0 | 11.0 |
| View315 | 27.0 | 31.0 | 23.0 | 29.5 | 28.5 | 25.0 | 25.0 | 29.0 | 26.5 | 30.0 | 27.0 | 32.0 | 27.5 | 29.5 | 30.0 | 28.5 | 19.5+* |
| Object 3 of test set | | | | | | | | | | | | | | | | | |
| View000 | 16.0 | 24.0 | 24.5 | 16.0 | **12.5** | 19.0 | 22.0 | 19.5 | 22.5 | 22.0 | 19.0 | 24.0 | 14.0 | 27.0 | 27.0 | 19.0 | 12.5+* |
| View045 | 25.5 | 22.5 | 27.0 | 19.5 | 22.5 | 26.0 | 22.0 | 26.0 | 25.5 | 24.0 | 24.0 | 29.0 | 19.5 | 25.5 | 28.0 | 27.0 | 15.0 |
| View090 | 25.0 | 22.5 | 20.5 | 20.5 | 22.5 | 23.5 | 20.5 | 25.0 | 26.0 | 22.5 | 22.0 | 27.0 | 22.5 | 24.0 | 24.5 | 25.5 | 19.0 |
| View135 | 23.0 | **16.5** | 17.0 | 24.0 | 24.0 | 21.0 | 22.5 | 19.0 | 21.5 | 21.5 | 24.0 | 23.5 | 19.0 | 25.0 | 19.0 | 23.0 | 16.5 * |
| View180 | 10.5 | 24.5 | 19.0 | 15.0 | 11.5 | 15.0 | 22.0 | 19.5 | 12.0 | 12.5 | **08.0** | 13.0 | 17.0 | 25.0 | 23.0 | 12.0 | 08.0 |
| View225 | 16.0 | 23.5 | 19.0 | 14.0 | 11.5 | 12.5 | 16.0 | 20.0 | 15.5 | 15.5 | 12.0 | 17.0 | **11.0** | 22.0 | 20.0 | 12.5 | 11.0 |
| View270 | 16.0 | 26.5 | 20.5 | 17.0 | **13.0** | 14.0 | 22.5 | 23.5 | 21.5 | 22.0 | 21.5 | 24.5 | 15.0 | 24.0 | 23.5 | 18.0 | 13.0+* |
| View315 | 18.5 | 23.5 | 23.5 | 23.5 | 19.0 | 18.5 | 20.0 | 19.0 | 19.0 | 22.5 | 23.0 | 21.0 | 17.5 | 25.0 | 21.0 | 19.0 | 16.0 |
| Object 4 of test set | | | | | | | | | | | | | | | | | |
| View000 | 11.5 | 29.0 | 23.0 | 20.0 | 16.5 | 08.0 | 24.0 | 21.5 | 05.0 | 08.0 | 08.0 | 07.5 | 14.0 | 20.5 | 18.0 | **01.0** | 01.0+* |
| View045 | 21.5 | 26.0 | 21.0 | 15.5 | 16.0 | 14.0 | 19.0 | 21.5 | 16.0 | 11.5 | 18.0 | 19.5 | 15.0 | 21.0 | 17.0 | 16.0 | 11.0 |
| View090 | 17.5 | 28.0 | 24.0 | 20.0 | 16.0 | 14.5 | 26.0 | 25.0 | 12.5 | 12.0 | 08.0 | **04.0** | 17.0 | 23.0 | 18.0 | 05.5 | 04.0+* |
| View135 | 13.5 | 24.0 | 24.5 | 20.0 | 16.0 | 15.0 | 26.5 | 21.0 | 12.5 | 12.0 | 10.0 | **00.0** | 16.0 | 22.0 | 17.0 | 07.5 | 00.0+* |
| View180 | 25.5 | 26.0 | 24.0 | **16.0** | 18.5 | 20.0 | 18.0 | 25.5 | 20.5 | 19.5 | 20.5 | 25.0 | 17.0 | 23.0 | 24.5 | 19.0 | 16.0 |
| View225 | 22.5 | 22.0 | 18.0 | 22.5 | 20.0 | 20.0 | 21.0 | 22.0 | 19.0 | 18.5 | 21.0 | 19.0 | 16.0 | 12.0 | **05.0** | 21.0 | 05.0+* |
| View270 | 19.5 | 26.0 | 16.0 | 22.0 | 22.5 | 18.0 | 20.0 | 24.0 | 17.0 | 16.5 | 19.0 | 17.0 | 15.0 | 14.5 | **00.0** | 18.5 | 00.0+* |
| View315 | 12.0 | 26.0 | 23.0 | 18.0 | 14.5 | 08.5 | 20.0 | 19.5 | **05.0** | **05.0** | 08.5 | 12.5 | 14.5 | 21.5 | 18.0 | 06.0 | 05.0+* |

Table 24.b. Distances between the whole test set and the object 3 and 4 of the reference set.

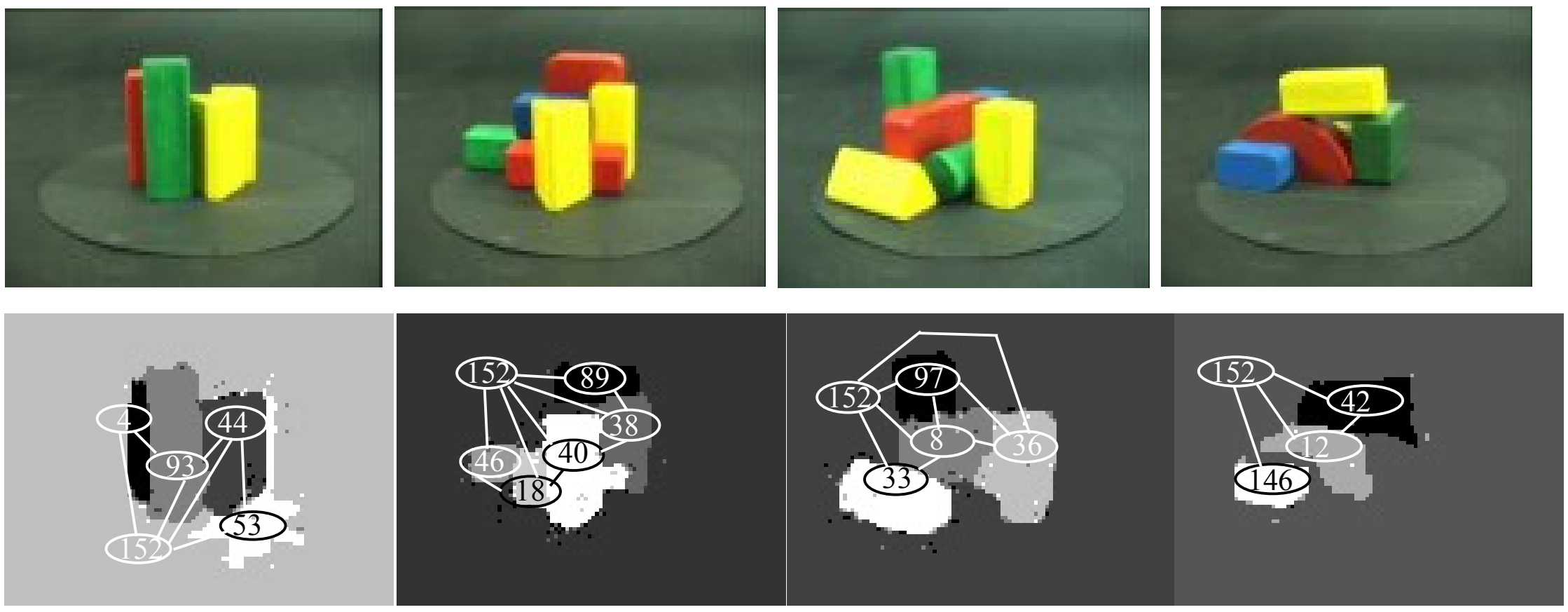

Figure 30. The 4 objects, their segmented views and their corresponding AGs.

# 11. Conclusions and future work

Function-described graphs (FDGs) are a type of compact representation of a set of attributed graphs (AGs) that borrow from random graphs the capability of probabilistic modelling of structural and attribute information, while improving the capacity of first-order random graphs to record structural relationships that consistently appear through the data. This is achieved by incorporating qualitative knowledge of the second-order probabilities of the elements that are expressed as relations (Boolean functions) between pairs of vertices and pairs of arcs in the FDG.

If only the relations between vertices are considered, the space complexity of the FDG representation is not greater than that of a first-order random graph, $n^2$, since all the relations defined (antagonism, occurrence, existence) can be obtained easily from the first-order marginal probabilities and just one second-order probability for each pair of vertices (namely, the probability of both vertices being null at the same time). However, if arc relations are considered as well, the space complexity obviously increases to $a^2$, where $a$ is the number of arcs. For dense graphs, this involves a storage (and a time complexity for creation, maintenance and verification of the arc relations) of order $n^4$ instead of $n^2$, which can be a severe drawback in practice.

In this doctoral thesis, we have studied the problem of matching an AG to an FDG for recognition or classification purposes from a Bayesian perspective. A distance measure between AGs and FDGs was derived from the principle of maximum likelihood, but robustness was assured because the effects of extraneous and missing elements were considered only locally. In this first measure, the set of valid mappings which have to be explored can be reduced by imposing the second-order relations as hard restrictions to be fulfilled. A second and more robust distance measure has also been given, in which second-order constraints are relaxed into local costs, at the expense of losing the theoretical link with the maximum likelihood source.

A branch-and-bound algorithm, adapted from (Wong, You and Chan, 1990), has been proposed for computing both distance measures together with their corresponding optimal labellings. We have seen that the antagonism relations are useful for pruning the search tree and also for increasing the correctness in the recognition process. However, the tree's computational complexity is non-polynomial, and therefore, alternative efficient methods which can provide good sub-optimal measures and labellings have also been presented. The aim of these methods is to reduce the number of possible vertex mappings before the branch and bound algorithm is run to compute the distance measure.

The synthesis of an FDG from a set of commonly labelled AGs has been described in detail. This synthesis can be regarded as a supervised learning method because all the AGs in the sample require a given common labelling. We have also reported two methods for synthesising FDGs from a set of unlabelled AGs and for clustering, in a non-supervised manner, a set of AGs into a set of FDGs representing subclasses. The former is based on an incremental process, which has the advantage that the learning and classification process can be interleaved. The latter is based on a hierarchical process. The whole ensemble of samples is needed to build the FDGs but the process does not depend on the order of presentation of the AGs as in the first method.

Three different experimental tests were carried out to asses the usefulness of our new representation and the algorithms presented here.

The first type of tests is presented in the last section in chapters 4, 6, 7 and 8. These tests are performed with randomly generated attributed graphs with the aim of validating the approaches presented in each chapter independently, as much as possible, of the other approaches. In section 4.6 we presented the structure of the FDGs (the number of vertices, antagonisms, occurrences and existences) depending on the number of attributed graphs used to generate it and the number of elements of these graphs. We realise that the number of FDG elements depend on the nature of the attributed graphs and we obtained some results of the number of antagonisms and vertices useful to understand other results presented in the following chapters. In section 6.5, we show the correctness and run time of the recognition process varying the cost on the antagonism. The best correctness appears when the antagonisms are relaxed (the cost of antagonisms is 1) although the fastest comparison is performed when the antagonisms are considered as strict restrictions (the cost of antagonism is a big number). The run time is much bigger when the antagonisms are not considered, therefore we conclude that the antagonisms are useful, not only for increasing the correctness but also for decreasing the run time. In section 7.8, we studied the balance between the run time and the correctness of the efficient algorithms and the optimal algorithm. We show that it is worth to initialise the probabilities by the distance between expanded vertices although this process is more expensive than initialising them by the distance between vertices. The non-iterative algorithm is not useful when the number of AGs used to synthesise the FDGs is medium or big since the run time is higher and the correctness lower than the iterative algorithms. Finally, in section 8.3, we compared the clustering algorithms. The best results are obtained using the complete method. We observed that this is not the case in the experimental validation that is commented in the previous section. We observed that spurious vertices in the FDGs are connected to the other vertices by antagonisms. Therefore, antagonisms not always are useful for decreasing the run time and increasing the correctness in the recognition process, but also they are useful for locating the spurious elements and deleting them from the FDGs in the learning process.
The second type of tests is an experimental validation of the FDGs. The attributed graphs are obtained from some artificially created 3D objects. The aim of these tests is to enforce the usefulness of our method by studying the algorithms and methods as a whole. The advantage of using artificial data is that we can perform the synthesis given a labelling described in section 4.4. There is any surprising result except for the clustering algorithms. We observe that, on the contrary of the results obtained in the above experiments, in the supervised clustering, the incremental (or dynamic) method obtains better results than the complete method. Nevertheless, when the unsupervised clustering is used, the complete method again obtains the best results.

The last tests are performed using real data. Some images were taken from four objects and segmented by analysing the colour. The problems providing from the segmentation step, such as the non-recognition of a region, the extraction of the attributes or the appearance of a spurious graph element are solved satisfactorily by the use of Function-Described Graphs. Therefore, we conclude that they are a useful tool for structural pattern recognition applied to computer vision.

In a practical point of view, further investigation is needed to compare the performance of FDGs with the other approaches in different applications and also to study the performance of the segmentation module. We are working in a middle and long-term project. Results presented in chapter 10 are only the first experiments of this project using real data. Now, we will work with office objects (tables, columns, chairs, telephones, ...).

In a theoretical point of view, second order relations can be substituted by second order probabilities since they are a coarse approximation of them. Thus, the distance relaxing second order costs would have a finer value. Nevertheless, the relation between the increase in correctness in the classification process and the increase in computational and storage cost would have to be considered.

The application of relaxation techniques in the first module of our efficient algorithm needs to be studied in depth since the run time and the correctness is of capital importance in our present project. A new support function needs to be defined that includes structural and first and second order information.

Finally, it would be interesting to study a learning method with positive and negative samples. The negative samples would influence the synthesis algorithms (learning process). Moreover, a process of reduction of the FDG could be studied. The main idea is to delete the graph elements that are considered not important in the representation of the model or to merge, by a clustering process, some of these graph elements.